\documentclass{article}

\usepackage{iclr2027_conference,times}
\iclrfinalcopy
\usepackage{flafter}

\usepackage[utf8]{inputenc}
\usepackage[T1]{fontenc}
\usepackage{hyperref}
\usepackage{placeins}
\usepackage{float}
\usepackage{wrapfig}
\usepackage{capt-of}
\usepackage{url}
\usepackage{booktabs}
\usepackage{amsfonts}
\usepackage{nicefrac}
\usepackage{microtype}
\usepackage{xcolor}
\definecolor{citationblue}{RGB}{58,142,193}
\hypersetup{
  colorlinks=true,
  citecolor=citationblue,
  linkcolor=citationblue,
  urlcolor=citationblue,
  filecolor=citationblue
}
\usepackage{colortbl}
\definecolor{ourrow}{rgb}{1.00,0.87,1.00}
\usepackage{algorithm}
\usepackage[noend]{algpseudocode}
\usepackage{tikz}
\usepackage{graphicx}
\usepackage{multirow}
\usetikzlibrary{positioning,calc,fit,arrows.meta,backgrounds,shapes.geometric,
                decorations.pathmorphing}
\usepackage{pgfplots}
\pgfplotsset{compat=1.18}
\usepgfplotslibrary{groupplots}

\definecolor{cnumfill}{rgb}{1.00,0.79,1.00}
\definecolor{cnumtext}{rgb}{0.00,0.00,0.00}
\newcommand{\cnum}[1]{%
  \tikz[baseline=(cn.base)]{%
    \node[shape=circle, fill=cnumfill, text=cnumtext, inner sep=1.1pt,
          minimum size=1.05em, font=\footnotesize\bfseries] (cn) {#1};}%
}

\definecolor{stepfill}{rgb}{0.18,0.49,0.20}
\DeclareRobustCommand{\stepnum}[1]{%
  \tikz[baseline=(sn.base)]{%
    \node[shape=circle, fill=stepfill, text=white, inner sep=1.1pt,
          minimum size=1.05em, font=\footnotesize\bfseries] (sn) {#1};}%
}

\newcommand{\mch}{\check{\vmu}_z}
\definecolor{paseoLatent}{HTML}{2E7D32}
\definecolor{paseoMeas}{HTML}{E06010}
\newcommand{\cstate}[1]{\textcolor{red}{#1}}
\newcommand{\cproxy}[1]{\textcolor{blue}{#1}}
\newcommand{\clat}[1]{\textcolor{paseoLatent}{#1}}
\newcommand{\cmeas}[1]{\textcolor{paseoMeas}{#1}}
\newcommand{\ech}{\check{\eta}_z}

\newcommand{\appendixpart}[2]{%
  \par\phantomsection
  \addcontentsline{toc}{part}{Part~#1: #2}%
  {\centering\Large\scshape Part~#1: #2\par}\vspace{1.2em}}
\makeatletter
\renewcommand*\l@part[2]{%
  \addpenalty{-\@highpenalty}\addvspace{0.6em plus 2pt}%
  {\parindent\z@\leavevmode\large\scshape #1\par}\nobreak}
\makeatother

\DeclareRobustCommand{\paseo}{\textsc{PASEO}}

\usepackage{amsmath,amsfonts,bm}

\def\eqref#1{equation~\ref{#1}}

\def\1{\bm{1}}

\def\rvepsilon{{\mathbf{\epsilon}}}

\def\rvb{{\mathbf{b}}}
\def\rvc{{\mathbf{c}}}
\def\rvd{{\mathbf{d}}}
\def\rve{{\mathbf{e}}}

\def\rvh{{\mathbf{h}}}
\def\rvu{{\mathbf{i}}}

\def\rvm{{\mathbf{m}}}
\def\rvn{{\mathbf{n}}}

\def\rvu{{\mathbf{u}}}
\def\rvv{{\mathbf{v}}}
\def\rvw{{\mathbf{w}}}
\def\rvx{{\mathbf{x}}}
\def\rvy{{\mathbf{y}}}
\def\rvz{{\mathbf{z}}}

\def\rmH{{\mathbf{H}}}
\def\rmI{{\mathbf{I}}}
\def\rmJ{{\mathbf{J}}}

\def\rmM{{\mathbf{M}}}

\def\rmU{{\mathbf{U}}}
\def\rmV{{\mathbf{V}}}
\def\rmW{{\mathbf{W}}}

\def\vzero{{\bm{0}}}

\def\vmu{{\bm{\mu}}}

\DeclareMathAlphabet{\mathsfit}{\encodingdefault}{\sfdefault}{m}{sl}
\SetMathAlphabet{\mathsfit}{bold}{\encodingdefault}{\sfdefault}{bx}{n}

\def\gA{{\mathcal{A}}}

\def\gD{{\mathcal{D}}}
\def\gE{{\mathcal{E}}}

\def\emLambda{{\Lambda}}
\def\emA{{A}}

\def\emC{{C}}

\def\emH{{H}}

\def\emK{{K}}

\def\emS{{S}}

\def\emU{{U}}
\def\emV{{V}}

\newcommand{\R}{\mathbb{R}}

\DeclareMathOperator*{\argmin}{arg\,min}

\numberwithin{equation}{section}

\title{\raggedright Perturb-and-Solve: Efficient \mbox{Learned-Operator} Conditioning for Latent Diffusion Inverse Problems}

\author{%
  \begin{tabular}[t]{@{}l@{\qquad}l@{\qquad}l@{}}
    Abduragim Shtanchaev & Arip Asadulaev & Luiza Labazanova \\[3pt]
    Aidar Alimbayev      & Karim Salta    & Eric Moulines \\[5pt]
  \end{tabular} \\
  Mohamed bin Zayed University of Artificial Intelligence \\
  \texttt{abduragim.shtanchaev@mbzuai.ac.ae}
}

\begin{document}
\addtocontents{toc}{\protect\setcounter{tocdepth}{-1}}

\maketitle
\lhead{Preprint. Under review.}

\begin{abstract}

Latent diffusion models serve as powerful priors for solving inverse problems
in image restoration, such as deblurring, inpainting, and super-resolution.
\textbf{Current methods have a trade-off between generality and efficiency.} Solvers
that are restricted to a fixed set of degradation operators are fast and
efficient. Methods that support arbitrary degradation operators are slow and
require gradients through the diffusion network.
To break this bottleneck, we introduce \paseo\
(\textbf{P}erturb-\textbf{A}nd-\textbf{S}olve for
\textbf{E}fficient \textbf{O}perator conditioning), a method that uses a small
($\sim$1M parameters) learned network to degrade diffusion model predictions in latent space.
\textbf{\paseo\ supports learned degradation operators without back-propagating
through the diffusion network.} We efficiently sample reconstructions from
an approximate posterior by combining the diffusion model's prediction with the
observed image. We do this by adding noise and solving linear equations based
on a local linear approximation of the learned network, without building or
inverting large covariance matrices.
Across super-resolution, deblurring, and inpainting on FFHQ and COCO, \paseo\
achieves strong perceptual quality while running up to $\textcolor{citationblue}{19\times}$ faster and
using up to $\textcolor{citationblue}{34\%}$ less peak memory than the tested baselines, with the same or fewer
model evaluations.

\end{abstract}

\newlength{\introtextfloatsep}
\setlength{\introtextfloatsep}{\textfloatsep}
\setlength{\textfloatsep}{8pt}
\vspace{-5pt}
\vspace{-5pt}
\section{Introduction}
\label{sec:intro}

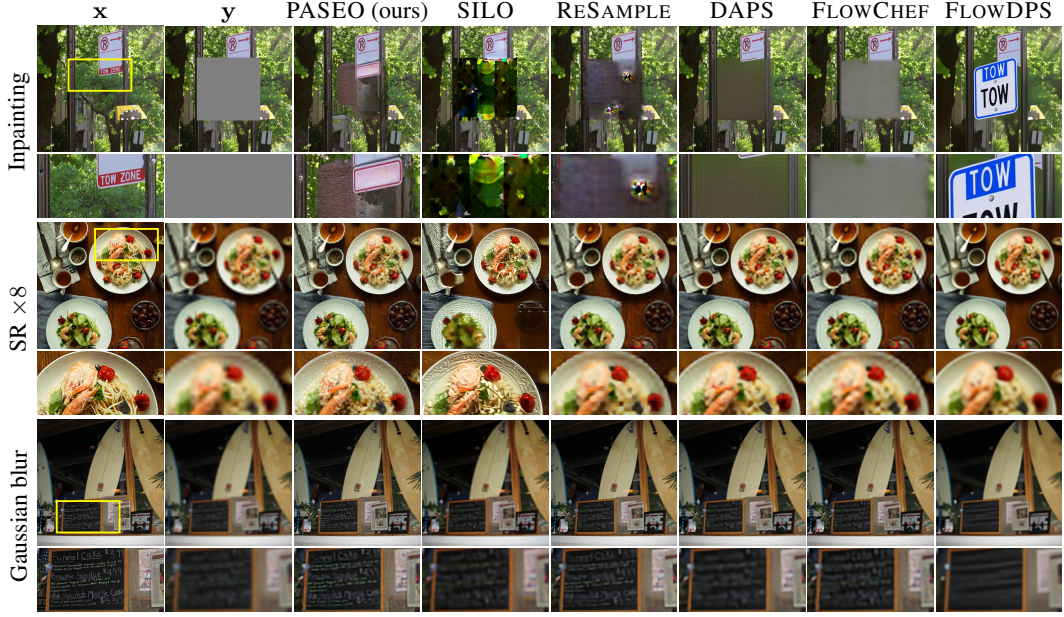
\begin{figure}[t]
\centering
\definecolor{zoombox}{rgb}{1.00,0.98,0.01}
\ifdefined\qcw\else\newlength{\qcw}\fi
\setlength{\qcw}{\dimexpr(\linewidth-19pt)/8\relax}
\newcommand*{\qcimg}[1]{\includegraphics[width=\qcw]{figures/qualitative_coco/#1}}
\newcommand*{\qcbox}[5]{%
  \begin{tikzpicture}[baseline=0pt]
    \node[inner sep=0pt, anchor=south west] at (0,0) {\qcimg{#1}};
    \draw[zoombox, line width=0.7pt]
      ({#2/512*\qcw}, {\qcw-#3/512*\qcw}) rectangle ({#4/512*\qcw}, {\qcw-#5/512*\qcw});
  \end{tikzpicture}}
\newcommand*{\qcstack}[2]{\begin{tabular}[b]{@{}c@{}}#1\\[1pt]#2\end{tabular}}
\newcommand*{\qccol}[2]{\qcstack{\qcimg{#1_#2}}{\qcimg{#1_#2_zoom}}}
\newcommand*{\qclabel}[1]{\rotatebox{90}{\makebox[\dimexpr1.5\qcw+1pt\relax][c]{\small #1}}}
\newcommand*{\qcrow}[6]{%
  \qclabel{#2} &
  \qcstack{\qcbox{#1_x}{#3}{#4}{#5}{#6}}{\qcimg{#1_x_zoom}} &
  \qccol{#1}{y} & \qccol{#1}{paseo} & \qccol{#1}{silo} & \qccol{#1}{resample} &
  \qccol{#1}{daps} & \qccol{#1}{flowchef} & \qccol{#1}{flowdps}}
\setlength{\tabcolsep}{0pt}
\renewcommand{\arraystretch}{0}
\begin{tabular}{@{}c@{\hspace{3pt}}c@{\hspace{1pt}}c@{\hspace{1pt}}c@{\hspace{1pt}}c@{\hspace{1pt}}c@{\hspace{1pt}}c@{\hspace{1pt}}c@{\hspace{1pt}}c@{}}
 & \small $\rvx$ & \small $\rvy$ & \makebox[\qcw][c]{\small \paseo\ (ours)} & \small \textsc{SILO} & \small \textsc{ReSample}
 & \small \textsc{DAPS} & \small \textsc{FlowChef} & \small \textsc{FlowDPS} \\[3pt]
\qcrow{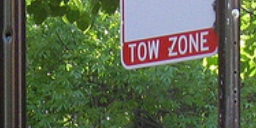}{Inpainting}{128}{136}{384}{264} \\[2pt]
\qcrow{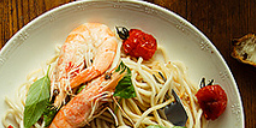}{SR $\times8$}{236}{24}{492}{152} \\[2pt]
\qcrow{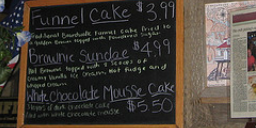}{Gaussian blur}{80}{330}{336}{458} \\
\end{tabular}
\caption{Visual comparison on COCO with SD3.5-medium prior.}
\label{fig:coco-qualitative}
\end{figure}

In image and signal processing, a signal of interest often cannot be observed directly and therefore must instead be recovered from indirect, incomplete or noisy measurements.
This is known as an \textit{inverse problem}, where we seek to recover an
\textit{unknown signal} $\rvx$ from an \textit{indirect measurement} $\rvy$ of that signal. The same measurement can come from different signals: for example, different sharp images
can produce similar blurred observations. We therefore need additional
information about what the original signal could be. The Bayesian formulation expresses
this knowledge through a prior $p(\rvx)$ and uses a likelihood
$p(\rvy\mid\rvx)$ to describe how a candidate signal could produce the
observed measurement. Together, they define the posterior distribution
$p(\rvx\mid\rvy)\propto p(\rvy\mid\rvx)p(\rvx)$, which describes plausible
signals given the measurement.
Inverse problems with diffusion models aim at 
generating plausible reconstructions through posterior sampling. For image reconstruction, the prior must capture the structure and detail of natural images, motivating the use of
expressive generative models.

Latent diffusion models (LDMs)~\citep{rombach2022ldm} provide
effective image priors for inverse problems~\citep{rout2023psld,song2024resample,
raphaeli2025silo}. They learn the structure and detail of natural images and
generate images by working in a compact latent space. A pretrained
autoencoder connects this space to images: its encoder maps an image to a
latent representation, and its decoder maps that representation back to an
image. Working in this smaller space reduces the cost of image generation.

For image reconstruction, generating realistic images is not enough. The reconstructed image must also be similar to the measurement $\rvy$. 
In the literature, this similarity is often called \textit{data consistency}. Enforcing data consistency is harder
with LDMs because the measurement $\rvy$ is in pixel space, while the model works in latent space.
Many solvers therefore enforce data consistency by repeatedly
running the encoder and decoder and back-propagating~\citep{rout2023psld,
song2024resample,rout2024stsl,chung2024p2l,raphaeli2025silo}, increasing
runtime and memory.

Recent methods aim to improve reconstruction
quality and reduce these costs.
LD-SMC~\citep{achituve2025ldsmc} and
DDSMC~\citep{ekstrom2025ddsmc} focus on more accurate posterior sampling,
while SITCOM~\citep{alkhouri2025sitcom}, learnable linear
extrapolation~\citep{zhang2025lle}, SPIN~\citep{shtanchaev2026spin}, and
CWGF~\citep{spagnoletti2026cwgf} create faster samplers.
Other approaches use the structure of linear measurements, as in
DMPS~\citep{meng2024dmps} and pDDIM~\citep{jiao2026pddim}; modify the sampling procedure,
as in \textsc{DAPS}~\citep{zhang2025daps}; or extend reconstruction to
flow-based generative models, as in \textsc{PnP-Flow}~\citep{martin2025pnpflow}
and \textsc{FlowDPS}~\citep{kim2025flowdps}.
We build on the following two methods for reconstruction in latent space.
\textsc{DING}~\citep{moufad2026ding} samples exactly from approximate
posterior transitions without back-propagating through the denoising network.
Its latent-space formulation is limited to inpainting, where a mask
separates observed and missing regions and enables a simple exact sampling update. Other degradations generally require a different update.
On the other hand, \textsc{SILO}~\citep{raphaeli2025silo} uses a small
trained network to describe more general degradation processes in latent
space.
We denote this learned operator by $\rmH_\psi$, where $\psi$
denotes its learned parameters. From a clean latent, the network
predicts the encoding of the corresponding degraded image.
This avoids repeatedly running the encoder and decoder
during sampling, but \textsc{SILO} still back-propagates through the learned
operator and the denoising network at every step to guide the reconstruction process.
Our aim is to combine these benefits to support learned
degradation operators in latent space while avoiding propagating gradients through the denoiser network.

\paseo\ efficiently updates its estimate toward an approximate posterior by first forming a local linear approximation of the learned network $\rmH_\psi$ around the current clean-latent prior estimate obtained from the diffusion step. To avoid forming the posterior covariance matrix and computing matrix inverses, we add independent Gaussian noise to the encoded measurement $\rvy_{z}$ and the mean of the clean-latent prior. We then solve the resulting linear equations using conjugate gradients (CG) to obtain a \textit{corrected} clean latent. For a linear operator, this basic construction gives an exact posterior sample when the equations are solved exactly. With the nonlinear learned network, the result is an approximate posterior sample, because we linearize the network locally and solve the equations with only a few CG iterations. The solver requires derivatives only through $\rmH_\psi$, without back-propagating through the denoiser or decoder. The encoder and decoder are each used only once, to encode the degraded image $\rvy$ and decode the reconstruction. We make three \textcolor{magenta}{\textbf{\textit{contributions:}}}

\cnum{1}~\textbf{A back-propagation-free sampler for learned operators.}
We turn each denoising step into a Gaussian inference problem in latent
space and sample it by perturbing and solving. For a linear operator and an
exact solve, this sample is exact; for a nonlinear one, we linearize the
operator locally, and the sample is approximate. This takes
back-propagation-free latent updates beyond inpainting, to any degradation
with a learned latent operator.

\cnum{2}~\textbf{Fast and memory-efficient by design.}
The update needs only a few CG iterations through the small
learned network ($\sim$1M parameters). It never back-propagates through the
denoiser, never forms a Jacobian or covariance matrix, and runs the
autoencoder only twice per reconstruction.

\cnum{3}~\textbf{Strong reconstruction quality at lower cost.}
Across five restoration tasks on FFHQ and COCO, with a diffusion prior
(RV-v5.1) and a flow prior (SD3.5), \paseo\ ranks first or second in LPIPS on
almost every task. With SD3.5, it runs up to \textcolor{citationblue}{$19\times$}
faster and uses up to $\textcolor{citationblue}{34\%}$ less peak memory than the
baselines, with the same or fewer denoiser evaluations.

\par
\setlength{\textfloatsep}{\introtextfloatsep}
\vspace{-5pt}
\section{Background}
\label{sec:background}
\providecommand{\restored}[1]{{\color{orange}#1}}

\textbf{Diffusion and flow-matching priors.}
Denoising diffusion models~\citep{song2019ncsn,ho2020ddpm,song2021sde}
construct a probability path $(p_t)_{t\in[0,1]}$ between the data distribution
$p_0$ and a reference distribution
$p_1:=\mathbf{N}(0,\rmI_{d_{\rvx}})$. A conditional path
$p_{t|0}(\rvx_t\mid\rvx_0)$ induces the marginal $p_t(\rvx_t)=\int p_{t|0}(\rvx_t\mid\rvx_0) p_0(\rvx_0)\,d\rvx_0$.
Although this density is generally intractable, samples are obtained by
drawing $\rvx_0\sim p_0$ followed by
$\rvx_t\sim p_{t|0}(\cdot\mid\rvx_0)$. For the Gaussian conditional path
$p_{t|0}(\cdot\mid\rvx_0)
=\mathbf{N}(\alpha_t\rvx_0,\sigma_t^2\rmI_{d_{\rvx}})$, this becomes
\begin{equation}
  \rvx_t=\alpha_t\rvx_0+\sigma_t\rvx_1,\qquad
  \rvx_0\sim p_0,\quad \rvx_1\sim p_1,\quad
  \rvx_0\perp\!\!\!\perp\rvx_1.
  \label{eq:foward_noise}
\end{equation}
where schedules $\alpha_t, \sigma_t$ satisfy $(\alpha_0,\sigma_0)=(1,0)$ and
$(\alpha_1,\sigma_1)=(0,1)$, with $\alpha_t$ non-increasing and $\sigma_t$
non-decreasing. Variance-preserving diffusion uses
$\alpha_t^2+\sigma_t^2=1$~\citep{ho2020ddpm}, whereas flow matching commonly
uses $(\alpha_t,\sigma_t)=(1-t,t)$~\citep{lipman2023flowmatching}.

The conditional path describes how a clean image produces a noisy state. Conversely, the distribution of the clean image given that noisy state follows from Bayes' rule: $p_{0|t}(\rvx_0\mid\rvx_t)
\propto p_{t|0}(\rvx_t\mid\rvx_0)p_0(\rvx_0)$. A learned denoiser approximates the mean of this reverse conditional distribution: $\hat{\rvx}^\theta_0(\rvx_t,t)
\approx \hat{\rvx}_0(\rvx_t,t) := \mathbb{E}[\rvx_0\mid\rvx_t]
=\int \rvx_0p_{0|t}(\rvx_0\mid\rvx_t)d\rvx_0.$

To generate samples, we follow this probability path in reverse,
starting from a noise sample $\rvx_1\sim p_1$ and moving toward the data
distribution $p_0$. Learned transitions approximate this reverse sampling
process. DDIM~\citep{song2021ddim} defines one such transition:
\begin{equation}
  \rvx_{t}
  =\alpha_{t}\hat{\rvx}^{\theta}_0(\rvx_{t+1},t+1)
  +(\sigma_{t}^2-\eta_{t}^2)^{1/2}
    \hat{\rvx}^{\theta}_1(\rvx_{t+1},t+1)
  +\eta_{t}\rvw,
  \qquad \rvw\sim\mathbf{N}(0,\rmI_{d_{\rvx}}).
\label{eq:b-ddim}
\end{equation}
Here $0\leq\eta_{t}\leq\sigma_{t}$,
and
$ \hat{\rvx}^{\theta}_1(\rvx_t,t) := (\rvx_t-\alpha_t\hat{\rvx}^{\theta}_0(\rvx_t,t))/\sigma_t$
estimates the noise coordinate. This follows from
$\mathbb{E}[\rvx_1\mid\rvx_t]
=(\rvx_t-\alpha_t\hat{\rvx}_0(\rvx_t,t))/\sigma_t$; equivalently, the clean
estimate is recovered from the noise one as
$\hat{\rvx}^{\theta}_0(\rvx_t,t)
=(\rvx_t-\sigma_t\hat{\rvx}^{\theta}_1(\rvx_t,t))/\alpha_t$, an identity used
in \eqref{eq:b-ding} and again in Section~\ref{sec:method}. The update is
deterministic when $\eta_{t}=0$ and stochastic otherwise.
We write $p^{\eta,\theta}_{t\mid t+1}(\rvx_{t}\mid\rvx_{t+1})$ for the law
of \eqref{eq:b-ddim}.

\textbf{Inverse problems.}
We consider the forward model
\begin{equation}
  \rvy=\gA(\rvx)+\rvn,\qquad
  \rvn\sim\mathbf{N}(0,\sigma_{\rvy}^2\rmI_{d_{\rvy}}).
\label{eq:i-forward}
\end{equation}
where $\rvx\in\R^{d_{\rvx}}$ is the \textit{unknown signal},
$\rvy\in\R^{d_{\rvy}}$ is the \textit{measurement}, and the known degradation operator
$\gA:\R^{d_{\rvx}}\to\R^{d_{\rvy}}$ may represent masking, blur,
downsampling, or compression.
The posterior introduced earlier combines the image prior $p_0$
with a likelihood, denoted here by $\ell_0$, that describes how a candidate
image could produce the observation.
The Gaussian likelihood is
$\ell_0(\rvy\mid\rvx)
=\mathbf{N}(\rvy;\gA(\rvx),\sigma_{\rvy}^2\rmI_{d_{\rvy}})$, and the
resulting posterior distribution is
\begin{equation}
  p_0(\rvx_0\mid\rvy)
  \propto \ell_0(\rvy\mid\rvx_0)p_0(\rvx_0).
\end{equation}

To draw approximate samples from this posterior, we use the observed
image $\rvy$ to guide the pretrained diffusion sampler, whose transitions are
$p^{\eta,\theta}_{t\mid t+1}(\rvx_{t}\mid\rvx_{t+1})$. This requires the
likelihood of $\rvy$ given the current noisy state $\rvx_t$, denoted by
$\ell_t(\rvy\mid\rvx_t)$. Computing this likelihood requires averaging over
the possible clean images given that noisy state:
$\ell_t(\rvy\mid\rvx_t):=\int\ell_0(\rvy\mid\rvx_0)\,
p_{0|t}(\rvx_0\mid\rvx_t)\,d\rvx_0$~\citep[Equation~2.20]{daras2024survey}.
Evaluating this integral over the high-dimensional space of clean images
is generally intractable.

To avoid computing this integral exactly, diffusion solvers use an approximate correction
to the denoiser's clean-image prediction.
The posterior denoiser averages the possible clean images given both the
noisy state $\rvx_t$ and the observed image $\rvy$:
$\hat{\rvx}_0(\rvx_t,t\mid\rvy):=\mathbb{E}[\rvx_0\mid\rvx_t,\rvy]$.
For the Gaussian noising model above, this average equals the estimate
based only on $\rvx_t$, plus a correction using
$\rvy$~\citep[Equations~2.15 and~2.17]{daras2024survey}:
\begin{equation}
  \hat{\rvx}_0(\rvx_t,t\mid\rvy)
  =\hat{\rvx}_0(\rvx_t,t)
  +\alpha_t^{-1}\sigma_t^2
    \nabla_{\rvx_t}\log\ell_t(\rvy\mid\rvx_t).
\label{eq:b-posterior-denoiser}
\end{equation}
The pretrained denoiser $\hat{\rvx}_0^\theta(\rvx_t,t)$ approximates the first term,
$\hat{\rvx}_0(\rvx_t,t)$. The gradient
$\nabla_{\rvx_t}\log\ell_t(\rvy\mid\rvx_t)$ is called the
\emph{guidance term}. To approximate it, a standard approach evaluates the
likelihood at the denoiser's clean-image prediction instead of averaging
over the possible clean images~\citep{ho2022video,chung2023dps}:
\begin{equation}
  \ell_t^\theta(\rvy\mid\rvx_t)
  :=\ell_0\!\left(
  \rvy\mid\hat{\rvx}_0^\theta(\rvx_t,t)
  \right).
\label{eq:b-dirac}
\end{equation}
This approximation ignores uncertainty about the clean image but lets solvers
guide sampling with the observation, without further training the denoiser.

\textbf{Guidance in latent space.}
We use a pretrained latent diffusion model with encoder $\gE$ and decoder
$\gD$~\citep{rombach2022ldm}. From this point on, the \textit{$\rvx$-family} denotes \textit{latent
diffusion states} and their clean and noise estimates in $\R^d$, so
\eqref{eq:b-ddim} is read as a latent transition. We color the diffusion state
\cstate{\textbf{red} $\rvx_t$}, the proxy \cproxy{\textbf{blue} $\rvx_t^{\mathrm{pxy}}$}, and the
encoded measurement \cmeas{\textbf{orange} $\gE(\rvy)=\rvy_{z}$}. In this setting, the two solvers we
build on use different approximations to the intermediate likelihood
$\ell_t(\rvy\mid\rvx_t)$. \textsc{DING}~\citep{moufad2026ding} avoids differentiating through the denoiser.
It rewrites the likelihood \eqref{eq:b-dirac} in terms of
the noise prediction and evaluates that prediction at an independent proxy
$\textcolor{blue}{\rvx_t^{\mathrm{pxy}}} := \rvm_t + \eta_t \rvw', \quad \rvw' \sim \mathbf{N}(0, \rmI)$ with $\rvm_t$ the mean of the prior transition in
\eqref{eq:b-ddim}, giving
\begin{equation}
  \begin{aligned}
  \ell_t^{\theta}(\rvy\mid\textcolor{red}{\rvx_t})
  &=\ell_0\!\left(\rvy\mid
    \hat{\rvx}^{\theta}_0(\textcolor{red}{\rvx_t},t)\right)
   \overset{\mathrm{(i)}}{=}\ell_0\!\left(\rvy\,\middle|\,
    (\textcolor{red}{\rvx_t}-\sigma_t
    \hat{\rvx}^{\theta}_1(\textcolor{red}{\rvx_t},t))/\alpha_t\right)
    \\[3pt]
  &\approx\ell_0\!\left(\rvy\,\middle|\,
    (\textcolor{red}{\rvx_t}-\sigma_t
    \hat{\rvx}^{\theta}_1(\textcolor{blue}{\rvx_t^{\mathrm{pxy}}},t))/\alpha_t\right)
   =:\ell_t^{\theta}
    (\rvy\mid\textcolor{red}{\rvx_t},\textcolor{blue}{\rvx_t^{\mathrm{pxy}}}).
  \end{aligned}
\label{eq:b-ding}
\end{equation}
Here, (i) uses the relation between the clean and noise predictions stated
after \eqref{eq:b-ddim}.
The proxy noise $\textcolor{blue}{\rvx_t^{\mathrm{pxy}}}$ is fixed with respect to $\textcolor{red}{\rvx_t}$, making the likelihood
argument \textit{affine}. Therefore, evaluating the gradient
$\nabla_{\textcolor{red}{\rvx_t}} \log \ell_t^{\theta}
(\rvy\mid\textcolor{red}{\rvx_t},\textcolor{blue}{\rvx_t^{\mathrm{pxy}}})$
does not require differentiation through the denoiser
$\hat{\rvx}^{\theta}_1(\textcolor{blue}{\rvx_t^{\mathrm{pxy}}},t)$.
Because $\textcolor{blue}{\rvx_t^{\mathrm{pxy}}}$ is fixed and, for
inpainting, the mask only selects the observed coordinates, combining the Gaussian likelihood
$\textcolor{black}{\ell_t^\theta(\rvy\mid\textcolor{red}{\rvx_t},
\textcolor{blue}{\rvx_t^{\mathrm{pxy}}})}$ with the Gaussian prior transition
$\textcolor{black}{p^{\eta,\theta}_{t\mid t+1}
(\textcolor{red}{\rvx_t}\mid\rvx_{t+1})}$ gives another Gaussian.
\textsc{DING} uses this construction to sample exactly from the
resulting surrogate posterior transition
$p^\theta_{t\mid t+1}
(\textcolor{red}{\rvx_t}\mid\rvx_{t+1},
\textcolor{blue}{\rvx_t^{\mathrm{pxy}}},\rvy)
\propto\allowbreak
\ell_t^\theta(\rvy\mid\textcolor{red}{\rvx_t},
\textcolor{blue}{\rvx_t^{\mathrm{pxy}}})\,\allowbreak
p^{\eta,\theta}_{t\mid t+1}(\textcolor{red}{\rvx_t}\mid\rvx_{t+1})$.

\textsc{SILO}~\citep{raphaeli2025silo} uses the same likelihood
approximation in \eqref{eq:b-dirac} but replaces decoding, applying the
degradation $\gA$, and re-encoding with a learned latent operator
$\rmH_\psi\approx\gE\circ\gA\circ\gD$. With the encoded measurement
$\cmeas{\gE(\rvy)}\in\R^d$, it uses
\begin{equation}
  \ell_t^{\theta,\psi}(\rvy\mid\textcolor{red}{\rvx_t})
  :=\mathbf{N}\!\left(
  \cmeas{\gE(\rvy)};
  \rmH_\psi(\hat{\rvx}^{\theta}_0(\textcolor{red}{\rvx_t},t)),
  r^2\rmI_d\right).
\label{eq:b-silo}
\end{equation}
Here, $r$ is the assumed noise scale of the latent
forward degradation model.
This permits any degradation in the latent space for which an accurate $\rmH_\psi$ can be
trained, while evaluating $\gE$ and $\gD$ only once per trajectory. Its
guidance $\nabla_{\textcolor{red}{\rvx_t}}\log\ell_t^{\theta,\psi}(\cdot)$ nevertheless
backpropagates through both $\rmH_\psi$ and
$\hat{\rvx}^{\theta}_0$ at every step.

\vspace{-5pt}
\section{Method}
\label{sec:method}
\providecommand{\hlz}{}
\providecommand{\yz}{\rvy_{z}}
\providecommand{\tyz}{\tilde{\rvy}_{z}}
\providecommand{\bxi}{\bm{\xi}}
\providecommand{\xzero}[2]{\hat{\rvx}_0^{\theta}(#1,#2)}
\providecommand{\xone}[2]{\hat{\rvx}_1^{\theta}(#1,#2)}

\begingroup
\setlength{\columnsep}{10pt}
\setlength{\intextsep}{8pt}

\begin{figure}[!t]
  \setlength{\abovecaptionskip}{4pt}
  \centering
  \resizebox{\textwidth}{!}{\input{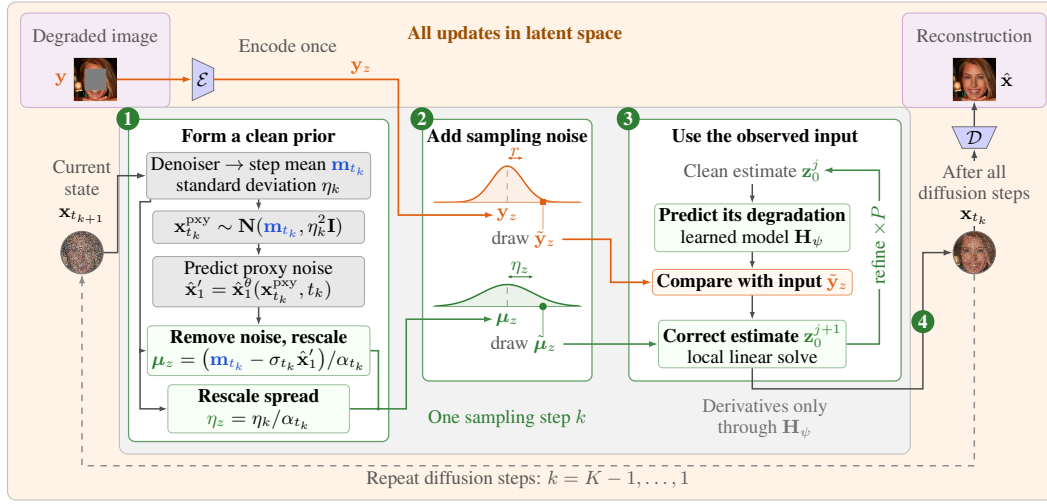}}
  \caption{\textbf{Overview of PASEO.} Each diffusion step is conditioned on the
  measurement through a small learned operator $\rmH_\psi$ in latent space.
  Only $\rmH_\psi$ is differentiated, \textit{never the denoiser}, and the autoencoder
  runs only once. Numbers match Steps~1--4 (Algorithm~\ref{alg:silo-jding}).}
  \label{fig:method}
\end{figure}

We begin with the observation that \eqref{eq:b-ding} does not produce a Gaussian surrogate posterior transition that can be sampled in closed form when the degradation operator is a learned network $\rmH_\psi$. \paseo\ uses the proxy construction from \eqref{eq:b-ding}, which avoids gradients through the denoiser, and extends it to \eqref{eq:b-silo}. This yields a Gaussian approximation to the surrogate posterior transition. We achieve this by (i) locally approximating the learned operator with a linear map, and (ii) using perturbations to sample the surrogate transition efficiently without forming or inverting its covariance matrix.  Figure~\ref{fig:method} gives an overview and Algorithm~\ref{alg:silo-jding} the pseudocode.  

\textbf{Training the latent operator.} Before describing
the sampler, we briefly explain how $\rmH_\psi$ is trained. Following
\citet{raphaeli2025silo}, we train a small network offline for each
degradation and latent space to predict the encoded degraded measurement
from the encoded clean image, minimizing
an $\ell_1$ prediction loss. The diffusion model
and autoencoder remain fixed, and $\rmH_\psi$ is kept fixed during
sampling. Full training details are given in
Appendix~\ref{sec:operator-training-details}.

\textbf{\stepnum{1}~The denoiser only sets the prior.}
We now turn to sampling with the trained operator, which acts on clean latents; we denote them $\clat{\rvz_0}$ and color them \clat{\textbf{green}}. Diffusion sampling, however, starts from a noisy state $\cstate{\rvx_t}$, so we express the diffusion state at $t$ as a distribution over the clean latent. From the current state $\rvx_{t+1}$, the
DDIM transition \eqref{eq:b-ddim} proposes the next state
$\cstate{\rvx_t}=\rvm_t+\eta_t\rvw$ with $\rvw\sim\mathbf{N}(0,\rmI_d)$, where
$\rvm_t:=\alpha_t\xzero{\rvx_{t+1}}{t+1}
+(\sigma_t^2-\eta_t^2)^{1/2}\xone{\rvx_{t+1}}{t+1}$.
As in \eqref{eq:b-ding}, \paseo\ also draws an independent proxy state
$\cproxy{\rvx_t^{\mathrm{pxy}}}=\rvm_t+\eta_t\rvw'$, evaluates the
denoiser once on it, and keeps its noise prediction
$\cproxy{\hat{\rvx}_1'}:=\xone{\cproxy{\rvx_t^{\mathrm{pxy}}}}{t}$ fixed for
the rest of the step. With $\cproxy{\hat{\rvx}_1'}$ fixed, a state and its
clean latent are linked by the simple relation
$\cstate{\rvx_t}=\alpha_t\clat{\rvz_0}+\sigma_t\cproxy{\hat{\rvx}_1'}$, as in
\eqref{eq:foward_noise}. Solving it for $\clat{\rvz_0}$ and substituting the
next-state draw gives the \textit{clean-latent prior}:
\begin{equation}
\clat{\rvz_0}=\frac{\cstate{\rvx_t}-\sigma_t\cproxy{\hat{\rvx}_1'}}{\alpha_t}
=\frac{\rvm_t-\sigma_t\cproxy{\hat{\rvx}_1'}}{\alpha_t}
+\frac{\eta_t}{\alpha_t}\rvw
=\clat{\vmu_z}+\eta_z\rvw
\quad\Rightarrow\quad
\clat{\rvz_0}\sim\mathbf{N}(\clat{\vmu_z},\eta_z^2\rmI_d).
\label{eq:clean-latent-prior}
\end{equation}
This is where the denoiser's role ends. Its two forward passes set
$\clat{\vmu_z}$ (the schedule sets $\eta_z$), and nothing below
differentiates through it.

\textbf{\stepnum{2}~Perturb and solve: a posterior sample from
one linear solve.}
Before turning to the learned network, it helps to look at a linear latent
degradation $\emH\in\R^{d\times d}$. Here the posterior has a closed form, which
makes it easy to see how the measurement corrects the prior and why sampling
from it is expensive. Step~3 then carries this over to $\rmH_\psi$. The forward degradation model is
$\cmeas{\yz}=\emH\clat{\rvz_0}+\rvepsilon$, with
$\cmeas{\yz}:=\gE(\cmeas{\rvy})$ and
$\rvepsilon\sim\mathbf{N}(0,r^2\rmI_d)$, where $r$ is the latent measurement
scale in \eqref{eq:b-silo}. It tells how likely a measurement
$\cmeas{\yz}$ is for a given clean latent $\clat{\rvz_0}$:
$\mathbf{N}(\cmeas{\yz};\emH\clat{\rvz_0},r^2\rmI_d)$. Multiplying this by
the clean-latent prior \eqref{eq:clean-latent-prior} gives the posterior,
$p(\clat{\rvz_0}\mid\cmeas{\yz})\propto
\mathbf{N}(\cmeas{\yz};\emH\clat{\rvz_0},r^2\rmI_d)\,
\mathbf{N}(\clat{\rvz_0};\clat{\vmu_z},\eta_z^2\rmI_d)$. Because $\emH$ is
linear, this product is again a Gaussian, $\mathbf{N}(\rvm_p,\mathbf{\Sigma}_p)$
\citep[Eq.~2.116]{bishop2006prml}. Written with the Woodbury
identity~\citep{henderson1981inverse}, its mean is
\begin{equation}
\rvm_p=\clat{\vmu_z}+\emK\big(\cmeas{\yz}-\emH\clat{\vmu_z}\big),\qquad
\emK:=\eta_z^2\emH^T\emS^{-1},\qquad
\emS:=r^2\rmI_d+\eta_z^2\emH\emH^T
\label{eq:method-posterior-gain}
\end{equation}
(Appendix~\ref{sec:app-gain-form}). The formula starts from the prior mean,
compares the measurement with what the operator predicts there, and corrects
by the mismatch. The gain $\emK$ decides how much to trust the measurement:
a broad prior (large $\eta_z$) or a precise measurement (small $r$) moves
the result further toward the measurement.

Computing this mean does not require inverting $\emS$: we solve the linear
system $\emS\rvv=\cmeas{\yz}-\emH\clat{\vmu_z}$ for $\rvv$ and set
$\rvm_p=\clat{\vmu_z}+\eta_z^2\emH^T\rvv$. Drawing a sample is harder. For a mask, $\emS$ is diagonal and the sample follows coordinate
by coordinate. For a general
$\emH$, $\emS$ couples all coordinates and the covariance $\mathbf{\Sigma}_p$
is a dense $d\times d$ matrix. A direct sample would take the form
$\clat{\rvz_0}=\rvm_p+\mathbf{\Sigma}_p^{1/2}\bxi$, with
$\bxi\sim\mathbf{N}(0,\rmI_d)$, which needs this matrix and its square root;
both are out of reach at latent sizes (with SD3.5, $d=65{,}536$, so
$\mathbf{\Sigma}_p$ has about $4.3$ billion entries).

\paseo\ gets a sample without ever building the covariance. It adds noise to
the two inputs of the mean formula, each at its own level: the prior mean
gets the prior's spread $\eta_z$, and the measurement gets the measurement
noise $r$ (Step~2 in Figure~\ref{fig:method}). It then applies the same
formula to the noisy inputs (Step~3). With independent
$\bxi_z,\bxi_y\sim\mathbf{N}(0,\rmI_d)$,
\begin{equation}
\boxed{\;
\clat{\rvz_0^+}=\clat{\tilde{\vmu}_z}
+\emK\big(\cmeas{\tyz}-\emH\clat{\tilde{\vmu}_z}\big),\quad
\clat{\tilde{\vmu}_z}=\clat{\vmu_z}+\eta_z\bxi_z,\quad
\cmeas{\tyz}=\cmeas{\yz}+r\bxi_y\;}
\label{eq:perturb-and-solve}
\end{equation}
\par
\algrenewcommand\algorithmicrequire{\textbf{Input:}}
\definecolor{algcomment}{gray}{0.64}
\definecolor{algline}{RGB}{150,150,150}
\definecolor{algorithmbackground}{rgb}{1.00,0.97,1.00}
\newcommand{\LineComment}[2]{\Statex
  \hspace{\algorithmicindent}%
  \tikz[baseline=(algcommenttext.base)]{%
    \node[anchor=west,inner sep=0pt,outer sep=0pt,align=left,
          text width=\dimexpr\linewidth-\algorithmicindent-12pt\relax,
          text=algcomment,font=\ttfamily] (algcommenttext) at (12pt,0)
          {/* #2~*/};
    \node[circle,fill=stepfill,text=white,inner sep=0pt,outer sep=0pt,
          minimum size=8pt,font=\fontsize{6}{6}\selectfont\bfseries]
          at (4pt,0) {#1};}}
\newcommand{\RightStepComment}[2]{\unskip\hfill
  \tikz[baseline=(algrightcomment.base)]{%
    \node[inner sep=0pt,outer sep=0pt,text=algcomment,font=\rmfamily]
          (algrightcomment) {\(\triangleright\) #2};
    \node[circle,fill=stepfill,text=white,inner sep=0pt,outer sep=0pt,
          minimum size=8pt,font=\fontsize{6}{6}\selectfont\bfseries]
          at ([xshift=6pt]algrightcomment.east) {#1};}}
\newsavebox{\paseoAlgorithmBox}
\makeatletter
\let\paseo@plainfloatbox\float@makebox
\renewcommand{\float@makebox}[1]{%
  \vbox{%
    \setbox\paseoAlgorithmBox=\paseo@plainfloatbox{#1}%
    \edef\paseo@bgwidth{\the\dimexpr\wd\paseoAlgorithmBox+4pt\relax}%
    \edef\paseo@bgdepth{\the\dimexpr\ht\paseoAlgorithmBox+\dp\paseoAlgorithmBox+4pt\relax}%
    \offinterlineskip
    \hbox{\tikz[overlay]{%
      \fill[algorithmbackground,rounded corners=4pt]
        (-4pt,4pt) rectangle (\paseo@bgwidth,-\paseo@bgdepth);}}%
    \nointerlineskip\unvbox\paseoAlgorithmBox}}
\begin{wrapalgorithm}[30]{L}{0.52\textwidth}
\fontsize{8}{11}\selectfont
\algrenewcommand\algorithmicindent{0.8em}
\algrenewcommand\alglinenumber[1]{{\color{algline}\fontsize{6.5}{8}\selectfont #1:}}
\caption{\textbf{\paseo\ } - \texttt{Pseudocode}}
\label{alg:silo-jding}
\begin{algorithmic}[1]
\Require decreasing timesteps $(t_k)_{k=K}^{0}$; latent
measurement $\cmeas{\yz}$; latent operator $\rmH_{\psi}$; noise
level $r$; DDIM parameters $(\eta_k)_{k=K}^{0}$; damping $\lambda$;
relaxation $\rho$; inner steps $P$; CG budget $C$.
\State $\cstate{\rvx} \sim \mathbf{N}(0,\rmI_d)$ \label{line:init}
\For{$k = K-1$ \textbf{to} $1$}
    \State $\hat{\rvx}_0 \leftarrow \xzero{\cstate{\rvx}}{t_{k+1}}$
    \State $\hat{\rvx}_1 \leftarrow
        \big(\cstate{\rvx} - \alpha_{t_{k+1}}\hat{\rvx}_0\big)/\sigma_{t_{k+1}}$
    \State $\rvm \leftarrow \alpha_{t_k}\hat{\rvx}_0
        + \big(\sigma^2_{t_k} - \eta_k^2\big)^{1/2}\hat{\rvx}_1$
    \LineComment{1}{Form \textit{clean-latent prior}}
    \State $\cproxy{\rvx_{t_k}^{\mathrm{pxy}}} \leftarrow \rvm + \eta_k \rvw', \quad \rvw' \sim \mathbf{N}(\vzero,\rmI_d) $
    \State $\cproxy{\hat{\rvx}_1'} \leftarrow \xone{\cproxy{\rvx_{t_k}^{\mathrm{pxy}}}}{t_k}$
         \label{line:proxy}
    \State $\clat{\vmu_z} \leftarrow (\rvm - \sigma_{t_k}\cproxy{\hat{\rvx}_1'})/\alpha_{t_k},
        \quad \eta_z := \eta_k/\alpha_{t_k}$
         \label{line:prior}
    \State $\clat{\rvz_0^{0}} \leftarrow (\cproxy{\rvx_{t_k}^{\mathrm{pxy}}} - \sigma_{t_k}\cproxy{\hat{\rvx}_1'})/\alpha_{t_k}$
    \LineComment{2}{Perturb-and-optimize targets}
    \State $\bxi_y\sim\mathbf{N}(\vzero,\rmI_d)$,
        \quad $\bxi_z\sim\mathbf{N}(\vzero,\rmI_d)$
    \State $\cmeas{\tyz} \leftarrow \cmeas{\yz} + r\,\bxi_y$,
        \quad $\clat{\tilde{\vmu}_z} \leftarrow \clat{\vmu_z} + \eta_z\,\bxi_z$ \phantomsection\label{line:perturb}
    \LineComment{3}{Local measurement update}
    \For{$j = 0$ \textbf{to} $P-1$} \label{line:gn-start}
        \State $\rvh \leftarrow \rmH_{\psi}(\clat{\rvz_0^{j}})$
        \State $\rmJ \leftarrow \partial \rmH_{\psi}/\partial \rvz_0 \big|_{\clat{\rvz_0^{j}}}$
        \State $\rvb \leftarrow \cmeas{\tyz} - \rvh
            - \rmJ(\clat{\tilde{\vmu}_z} - \clat{\rvz_0^{j}})$ \label{line:residual}
        \State \texttt{solve} $\big((r^2+\lambda)\rmI_d
            + \eta_z^2 \rmJ \rmJ^\top\big) \rvv = \rvb$
             \phantomsection\label{line:cg}
        \State $\clat{\rvz_0^{\star}} \leftarrow \clat{\tilde{\vmu}_z} + \eta_z^2 \rmJ^\top \rvv$ \phantomsection\label{line:zstar}
        \State $\clat{\rvz_0^{j+1}} \leftarrow \clat{\rvz_0^{j}} + \rho\,(\clat{\rvz_0^{\star}} - \clat{\rvz_0^{j}})$ \phantomsection\label{line:gn-end}
    \EndFor
    \State $\cstate{\rvx} \leftarrow \alpha_{t_k}\clat{\rvz_0^{P}} + \sigma_{t_k}\cproxy{\hat{\rvx}_1'}$
        \RightStepComment{4}{back to the diffusion state}%
        \label{line:lift}
\EndFor
\State \textbf{Return:} $\gD\big(\xzero{\cstate{\rvx}}{t_1}\big)$
\end{algorithmic}
\end{wrapalgorithm}
\makeatletter
\let\float@makebox\paseo@plainfloatbox
\makeatother

For a linear $\emH$, $\clat{\rvz_0^+}$ is an exact draw from
$\mathbf{N}(\rvm_p,\mathbf{\Sigma}_p)$
(Appendix~\ref{sec:app-perturb-sampling}). This works because the formula is
linear in its two inputs: noise that enters at the prior's and the
measurement's levels leaves with exactly the posterior's spread. The
perturb-and-solve identity itself is known in the sampling
literature~\citep{papandreou2010gaussian,bardsley2014rto}. Our observation is
that, once the proxy fixes the noise prediction, conditioning a latent
diffusion step reduces to exactly this problem; Step~3 extends it to the
learned operator by local linearization. Two consequences follow.

\emph{A sample costs the same as the mean.} The sample uses the same solve as
the mean, only with the noisy inputs:
$\emS\rvv=\cmeas{\tyz}-\emH\clat{\tilde{\vmu}_z}$ gives
$\clat{\rvz_0^+}=\clat{\tilde{\vmu}_z}+\eta_z^2\emH^T\rvv$. The
conjugate gradient (CG) method~\citep{hestenes1952cg} solves it using only
products with $\emH$ and $\emH^T$, so neither $\emS$ nor
$\mathbf{\Sigma}_p$ is ever built (Appendix~\ref{sec:app-cg-solve}).

\emph{The denoiser is never differentiated.} It enters only through the
fixed value $\clat{\vmu_z}$.

For a mask, \eqref{eq:perturb-and-solve} reduces to the closed-form step of
\eqref{eq:b-ding} (Appendix~\ref{sec:app-unification}).

\textbf{\stepnum{3}~Solving with the learned operator.}
So far the operator was a fixed linear map. In practice it is the learned
network $\rmH_\psi$, and the forward degradation model becomes
$\cmeas{\yz}=\rmH_\psi(\clat{\rvz_0})+\rvepsilon$. The identity
\eqref{eq:perturb-and-solve} needs a linear map, so we cannot apply it to
$\rmH_\psi$ directly. Instead, we use the fact that close to the current
estimate $\clat{\rvz_0^{(j)}}$, the network behaves almost like a linear map,
\begin{equation}
\rmH_\psi(\clat{\rvz_0})\approx\rmH_\psi(\clat{\rvz_0^{(j)}})
+\rmJ_j\big(\clat{\rvz_0}-\clat{\rvz_0^{(j)}}\big),
\label{eq:local-linear-model}
\end{equation}
where the Jacobian $\rmJ_j$ tells how the predicted measurement changes when
the latent moves away from $\clat{\rvz_0^{(j)}}$. Within this approximation
we can use \eqref{eq:perturb-and-solve} with $\rmJ_j$ in place of $\emH$.
The update compares the noisy measurement $\cmeas{\tyz}$ with what the
linearized network predicts at $\clat{\tilde{\vmu}_z}$, and uses the mismatch
to correct the estimate, giving an approximate posterior draw
$\clat{\rvz_0^\star}$:
\begin{equation}
\boxed{\;
\clat{\rvz_0^\star}=\clat{\tilde{\vmu}_z}
+\eta_z^2\rmJ_j^T\rvv,\quad
\emS_j\rvv=\cmeas{\tyz}-\rmH_\psi(\clat{\rvz_0^{(j)}})
-\rmJ_j\big(\clat{\tilde{\vmu}_z}-\clat{\rvz_0^{(j)}}\big)\;}
\label{eq:local-perturb-and-solve}
\end{equation}
with $\emS_j:=r^2\rmI_d+\eta_z^2\rmJ_j\rmJ_j^T$. The approximation is only
accurate near the point where it was built, and the correction moves the
estimate away from that point. We therefore linearize again at the new
estimate and repeat the correction, $P$ times in total (the ``refine
$\times P$'' loop in Figure~\ref{fig:method}), always with the same noisy
inputs $\clat{\tilde{\vmu}_z}$ and $\cmeas{\tyz}$. Each solve runs CG, which
only needs the products $\rmJ_j\rvv$ and $\rmJ_j^T\rvu$. Automatic
differentiation computes them through the small network $\rmH_\psi$, without
forming $\rmJ_j$ and without touching the denoiser
(Appendix~\ref{sec:app-jacobian-free-cg}). Because $\rmH_\psi$ is
nonlinear, $\clat{\rvz_0^\star}$ is an exact posterior sample only for the
linearized model (Appendix~\ref{sec:app-local-posterior}).

\textbf{\stepnum{4}~Back to the diffusion state.}
The corrections happen in the clean latent, but the sampler moves through
noisy states, so the result has to become a state at time $t$ before the next
step. We use the relation that built the prior in Step~1, with the same
fixed noise prediction $\cproxy{\hat{\rvx}_1'}$, to get the noisy state
$\cstate{\rvx_t}=\alpha_t\clat{\rvz_0^{(P)}}+\sigma_t\cproxy{\hat{\rvx}_1'}$.
This undoes the change of variables of Step~1, so the corrected clean latent
becomes the next diffusion state, and the next step starts from it. Over the
whole run, the measurement is encoded once and the final clean estimate is
decoded once.

\textbf{Practical choices.}
A few settings keep the corrections of Step~3 cheap and stable. We start from
the proxy's clean latent
$\clat{\rvz_0^{(0)}}:=(\cproxy{\rvx_t^{\mathrm{pxy}}}
-\sigma_t\cproxy{\hat{\rvx}_1'})/\alpha_t$.  We add $\lambda\rmI_d$ to
$\emS_j$, with $\lambda\geq0$, which keeps the solve well conditioned when
$r$ is small. We stop CG after at most $C$ iterations, which bounds the cost
and regularizes the update~\citep{hanke1995conjugate}; it also beats an exact
solve (Section~\ref{sec:main-ablations}). With the denoiser anchor and no
perturbation, the step solves the \textsc{DDS}~\citep{chung2024dds} problem
(Appendix~\ref{sec:unif-dds}). Because the linearization is only accurate nearby, each
correction moves only part of the way,
$\clat{\rvz_0^{(j+1)}}=\clat{\rvz_0^{(j)}}
+\rho\big(\clat{\rvz_0^\star}-\clat{\rvz_0^{(j)}}\big)$ with $0<\rho\leq1$.
For inpainting, we weight the likelihood by $\rmW=\mathrm{diag}(w)$ with
$w_i\propto\|(\rmJ_j)_{i,:}\|$, the sensitivity of measurement $i$ to the latent,
so insensitive measurements get little weight; $\rmW$ multiplies $\rmJ$ and $\rvb$ in
lines~\ref{line:reference-weight}--\ref{line:reference-zstar} of Algorithm~\ref{alg:paseo-reference}
(Appendix~\ref{sec:inpaint-variants}).

One more choice concerns $\clat{\tilde{\vmu}_z}$, the point each correction
starts from, which we call the \emph{anchor}. Early in diffusion sampling the prior \eqref{eq:clean-latent-prior} is
wide, so the anchor can land far from the proxy's clean latent
$\clat{\rvz_0^{(0)}}$, where the first linearization is built. We found that
reusing part of the proxy noise $\rvw'$ in place of the fresh noise
$\bxi_z$ improves quality on most tasks at the same budget:
\begin{equation}
\clat{\tilde{\vmu}_z}\leftarrow\clat{\vmu_z}
+\eta_z\big(c_1\rvw'+c_2\bxi_z\big),\qquad
(c_1,c_2)=\big((1-\beta_k^2)^{1/2},\,\beta_k\big),\qquad
\beta_k=\eta_{z,\min}/\eta_{z,k},
\label{eq:method-practical-anchor}
\end{equation}
where $k$ indexes the current diffusion step, as in
Algorithm~\ref{alg:silo-jding}, and $\eta_{z,\min}$ is the smallest prior
standard deviation in the schedule. We call this the \emph{arc anchor}: $(c_1,c_2)$ lies on the unit
circle, so the anchor keeps the prior's spread $\eta_z$. Since
$\clat{\rvz_0^{(0)}}=\clat{\vmu_z}+\eta_z\rvw'$,
it leans toward the proxy's clean latent while the prior is wide and
recovers $\clat{\tilde{\vmu}_z}$ when $\beta_k=1$. Unlike the fixed anchor
of Step~3, this anchor is recomputed from the current estimate at every
correction (Algorithm~\ref{alg:paseo-reference});
Algorithm~\ref{alg:silo-jding} keeps the fixed original anchor for clarity.
Appendices~\ref{sec:app-anchor-family} and~\ref{sec:app-anchor-ablation} give
the analysis and compare anchors.
\par
\endgroup

\vspace{-5pt}
\section{Experiments}
\label{sec:experiments}
\begingroup
\setlength{\parskip}{.5pc plus 2pt minus 1pt}
\textbf{Models and datasets.}
We evaluate \paseo\ on five 
degradation tasks:
$\times4$ and $\times8$ super-resolution, Gaussian blur ($61\times61$,
$\sigma=3$), $256\times256$ center inpainting, and JPEG
decompression at quality $10\%$, all at $\sigma_{\rvy}=0.01$. We use two latent priors:
diffusion-based RV-v5.1~\citep{sg1612222023realisticvision} and
flow-based Stable Diffusion 3.5 (medium)~\citep{esser2024sd3}. For sampler evaluation, we use $1000$ held-out FFHQ~\citep{karras2019stylegan} images
and $1000$ COCO-val2017~\citep{lin2014coco} images, all at $512\times512$.

\textbf{Evaluation, metrics, and baselines.}
Following \citet{raphaeli2025silo}, we train a separate latent degradation operator $\rmH_{\psi}$ for each task.
For FFHQ, we split the dataset into $68{,}000$ training, $1000$ validation, and $1000$ evaluation images.
For COCO, we train on the $84{,}990$ LSDIR-train images~\citep{li2023lsdir}, validate on $250$ LSDIR validation images, and evaluate on the first $1000$ COCO-val2017 images.
Appendix~\ref{sec:operator-training-details} gives the training details.  We report the following metrics:
PSNR~\citep{hore2010psnr} and LPIPS~\citep{zhang2018lpips}, KID~\citep{binkowski2018kid} for distributional alignment, wall-clock seconds per image and peak device memory in GiB. We emphasize LPIPS, since PSNR favors overly smooth images when exact recovery is impossible~\citep[App.~B.6]{janati2025mixture}. 
Across FFHQ and COCO, we compare \paseo\ against \textsc{SILO},
\textsc{ReSample}~\citep{song2024resample}, \textsc{PSLD} and
\textsc{GML-DPS}~\citep{rout2023psld}, \textsc{LDPS}~\citep{chung2023dps}, \textsc{DING}, \textsc{DAPS},
\textsc{RedDiff}~\citep{mardani2024reddiff}, \textsc{FlowDPS},
\textsc{FlowChef}~\citep{patel2025flowchef}, and \textsc{PnP-Flow}. All our experiments are conducted on an AMD MI210 GPU.
Appendix~\ref{sec:metric-protocol} provides more details on the setup.

\setlength{\columnsep}{8pt}
\setlength{\intextsep}{0pt}
\setlength{\rightskip}{0pt plus 1em}
\newsavebox{\ffhqresultstable}
\begin{lrbox}{\ffhqresultstable}%
\fontsize{8}{9}\selectfont%
\renewcommand{\arraystretch}{1.15}%
\setlength{\tabcolsep}{2.5pt}%
\begin{tabular}{@{}c>{\fontsize{6}{9}\selectfont}lrrrr|crrr@{}}
\toprule
\multirow{2.5}{*}{\rotatebox[origin=c]{90}{Task}}
 & \multicolumn{1}{c}{\multirow{2.5}{*}{\rotatebox[origin=c]{90}{Method}}}
 & \multicolumn{4}{c|}{RV-v5.1}
   & \multicolumn{4}{c}{SD3.5-medium} \\
\cmidrule(lr){3-6}\cmidrule(lr){7-10}
 &
 & LPIPS & KID & Time & Mem
 & LPIPS & KID & Time & Mem \\
\midrule
\rowcolor{ourrow} \cellcolor{white}\multirow{7}{*}{\rotatebox[origin=c]{90}{SR $\times4$}} & \paseo & \underline{0.139} & \textbf{1.81} & \textbf{11.7} & \textbf{3.6}
 & \textbf{0.182} & 10.2 & \textbf{10.6} & \textbf{16.8} \\
 & \cellcolor{ourrow}\paseo$^\dagger$ & \cellcolor{ourrow}\textbf{0.130} & \cellcolor{ourrow}\underline{2.96} & \cellcolor{ourrow}\underline{14.4} & \cellcolor{ourrow}\textbf{3.6}
 & \cellcolor{ourrow}\underline{0.221} & \cellcolor{ourrow}\underline{9.49} & \cellcolor{ourrow}25.9 & \cellcolor{ourrow}\textbf{16.8} \\
 & \textsc{SILO} & 0.182 & 5.10 & 149 & 7.4 & 0.303 & 29.3 & \underline{20} & \underline{21.2} \\
 & \textsc{ReS.} & 0.433 & 25.5 & 1418 & \underline{7.0} & \textbf{0.182} & \textbf{7.70} & 130 & 22.5 \\
 & \textsc{PSLD} & 0.249 & 10.1 & 390 & 7.9 & 0.494 & 47.2 & 47 & 23.4 \\
 & \textsc{GML} & 0.247 & 9.05 & 389 & 7.9 & 0.373 & 46.3 & 48 & 23.4 \\
 & \textsc{LDPS} & 0.281 & 11.6 & 331 & \underline{7.0} & 0.423 & 57.1 & 39 & 22.5 \\
\midrule
\rowcolor{ourrow} \cellcolor{white}\multirow{7}{*}{\rotatebox[origin=c]{90}{SR $\times8$}} & \paseo & \textbf{0.212} & \underline{5.99} & \textbf{11.6} & \textbf{3.6}
 & \textbf{0.330} & \textbf{28.8} & \textbf{8.8} & \textbf{16.8} \\
 & \cellcolor{ourrow}\paseo$^\dagger$ & \cellcolor{ourrow}0.241 & \cellcolor{ourrow}13.13 & \cellcolor{ourrow}\underline{14.4} & \cellcolor{ourrow}\textbf{3.6}
 & \cellcolor{ourrow}0.420 & \cellcolor{ourrow}\underline{30.3} & \cellcolor{ourrow}\underline{10.7} & \cellcolor{ourrow}\textbf{16.8} \\
 & \textsc{SILO} & \underline{0.226} & \textbf{5.23} & 149 & 7.4 & 0.501 & 62.8 & 20 & \underline{21.2} \\
 & \textsc{ReS.} & 0.575 & 118 & 1418 & \underline{7.0} & \underline{0.378} & 31.6 & 130 & 22.5 \\
 & \textsc{PSLD} & 0.320 & 14.9 & 390 & 7.9 & 0.453 & 60.9 & 48 & 23.4 \\
 & \textsc{GML} & 0.327 & 12.9 & 389 & 7.9 & 0.439 & 57.1 & 48 & 23.4 \\
 & \textsc{LDPS} & 0.343 & 12.9 & 331 & \underline{7.0} & 0.456 & 60.2 & 39 & 22.5 \\
\midrule
\rowcolor{ourrow} \cellcolor{white}\multirow{7}{*}{\rotatebox[origin=c]{90}{Gaussian blur}} & \paseo & \textbf{0.162} & \textbf{2.09} & \textbf{17.3} & \textbf{3.6}
 & \textbf{0.200} & \textbf{8.32} & \textbf{8.6} & \textbf{16.8} \\
 & \cellcolor{ourrow}\paseo$^\dagger$ & \cellcolor{ourrow}\underline{0.172} & \cellcolor{ourrow}\underline{6.66} & \cellcolor{ourrow}\underline{121} & \cellcolor{ourrow}\textbf{3.6}
 & \cellcolor{ourrow}0.215 & \cellcolor{ourrow}10.7 & \cellcolor{ourrow}\underline{10.1} & \cellcolor{ourrow}\textbf{16.8} \\
 & \textsc{SILO} & 0.222 & 8.21 & 149 & 7.4 & 0.399 & 47.9 & 20 & \underline{21.2} \\
 & \textsc{ReS.} & 0.253 & 10.7 & 1418 & \underline{7.0} & \underline{0.206} & \underline{9.37} & 138 & 22.5 \\
 & \textsc{PSLD} & 0.288 & 12.2 & 390 & 7.9 & 0.450 & 49.1 & 69 & 25.6 \\
 & \textsc{GML} & 0.309 & 16.5 & 389 & 7.9 & 0.449 & 57.3 & 48 & 23.4 \\
 & \textsc{LDPS} & 0.327 & 19.9& 331 & \underline{7.0} & 0.467 & 61.4 & 40 & 22.5 \\
\midrule
\rowcolor{ourrow} \cellcolor{white}\multirow{8}{*}{\rotatebox[origin=c]{90}{Inpainting}} & \paseo & \underline{0.144} & 14.00 & 38.4 & \textbf{3.6}
 & \textbf{0.118} & \textbf{2.71} & 13.6 & \textbf{16.8} \\
 & \cellcolor{ourrow}\paseo$^\dagger$ & \cellcolor{ourrow}0.147 & \cellcolor{ourrow}13.36 & \cellcolor{ourrow}\underline{31.2} & \cellcolor{ourrow}\textbf{3.6}
 & \cellcolor{ourrow}0.144 & \cellcolor{ourrow}11.4 & \cellcolor{ourrow}\textbf{11.0} & \cellcolor{ourrow}\textbf{16.8} \\
 & \textsc{SILO} & \textbf{0.139} & \textbf{1.80} & 149 & 7.4 & 0.430 & 115 & 20 & \underline{21.2} \\
 & \textsc{DING} & 0.159 & \underline{4.59} & \textbf{8.8} & \textbf{3.6} & \underline{0.134} & \underline{5.89} & \underline{12.6} & \textbf{16.8} \\
 & \textsc{ReS.} & 0.273 & 119 & 1418 & \underline{7.0} & 0.214 & 55.6 & 267 & 22.5 \\
 & \textsc{PSLD} & 0.357 & 17.2 & 390 & 7.9 & 0.459 & 69.9 & 48 & 23.4 \\
 & \textsc{GML} & 0.356 & 16.5 & 389 & 7.9 & 0.509 & 79.2 & 48 & 23.4 \\
 & \textsc{LDPS} & 0.368 & 16.0 & 331 & \underline{7.0} & 0.507 & 78.5 & 39 & 22.5 \\
\midrule
\rowcolor{ourrow} \cellcolor{white}\multirow{6}{*}{\rotatebox[origin=c]{90}{JPEG $q{=}10$}} & \paseo & 0.258 & 26.53 & \textbf{8.2} & \textbf{3.6}
 & \underline{0.350} & 71.2 & \textbf{13.9} & \textbf{16.8} \\
 & \cellcolor{ourrow}\paseo$^\dagger$ & \cellcolor{ourrow}\underline{0.238} & \cellcolor{ourrow}22.73 & \cellcolor{ourrow}\underline{14.0} & \cellcolor{ourrow}\textbf{3.6}
 & \cellcolor{ourrow}0.400 & \cellcolor{ourrow}58.7 & \cellcolor{ourrow}60.5 & \cellcolor{ourrow}\textbf{16.8} \\
 & \textsc{SILO} & \textbf{0.203} & \textbf{4.21} & 138 & 7.4 & 0.503 & 77.4 & \underline{20} & \underline{21.2} \\
 & \textsc{ReS.} & 0.456 & 20.1 & 2438 & \underline{7.0} & \textbf{0.196} & \textbf{19.4} & 140 & 22.5 \\
 & \textsc{GML} & 0.268 & \underline{7.54} & 402 & 7.9 & 0.413 & \underline{50.1} & 49 & 23.4 \\
 & \textsc{LDPS} & 0.373 & 17.7 & 412 & \underline{7.0} & 0.424 & 54.4 & 40 & 22.5 \\
 \bottomrule
\end{tabular}%
\end{lrbox}
\begin{wraptable}[42]{R}{\wd\ffhqresultstable}
\centering
\vspace{-\abovecaptionskip}
\setlength{\belowcaptionskip}{3pt}
\fontsize{8}{9}\selectfont
\caption{FFHQ restoration. Left: \paseo\ with RV-v5.1. Right: SD3.5-medium, baselines at $79$ NFEs and \paseo\ at $53$. Lower is better for all metrics.
Bold/underline mark
best/second-best. \textsc{ReS.} = \textsc{ReSample}; \textsc{GML} = \textsc{GML-DPS} }
\label{tab:ffhq-models}
\usebox{\ffhqresultstable}
\end{wraptable}

\textbf{Hyperparameters.}
Our prompts are \texttt{A high quality photo of a face} for FFHQ and
\texttt{A high quality photo} for COCO. The guidance scale is $1$, except for
SD3.5 on blur, JPEG, and FFHQ $\times4$ super-resolution, where it is $4$.
Each reverse step calls the denoiser twice, at the current state and at the
proxy sample, except the last two, which share one call; $K$ steps thus cost
$2K-3$ NFEs. On FFHQ, both priors use $28$ steps ($53$ NFEs). On COCO, we use
$50$ steps ($97$ NFEs) for $\times4$ super-resolution, blur, and JPEG, and
$100$ steps ($197$ NFEs) for $\times8$ super-resolution and inpainting.
\par
\noindent Each step runs $P$ inner corrections, with $P=3$--$5$ for RV-v5.1
and, for SD3.5, $P=1$--$3$ on FFHQ and $2$--$5$ on COCO. The CG budget is
$C=1$--$3$ iterations, except for FFHQ inpainting ($C=12$ for RV-v5.1, $C=5$
for SD3.5). Appendix~\ref{sec:paseo-hyperparameters} lists these and the
remaining settings (noise level $r$, relaxation $\rho$, damping $\lambda$) for
each task and prior, and Appendix~\ref{sec:ofat} examines their sensitivity.
\par
\vspace{-4pt}
\subsection{Main results}
\vspace{-4pt}

We report two versions of \paseo. Our analysis uses the original anchor of
Algorithm~\ref{alg:silo-jding}, a fresh draw from the prior. Early in sampling,
however, the prior is wide, and this anchor can land far from the proxy's clean
latent. Our default, \paseo, therefore uses the arc anchor of
\eqref{eq:method-practical-anchor}, which reuses part of the proxy noise to
start near that latent. We also report \paseo$^\dagger$, which keeps the
original anchor.

\textbf{FFHQ.}
With both priors, \paseo\ ranks first or second in LPIPS on almost every task,
while running faster and with less memory than every baseline except the
inpainting-only \textsc{DING} (Table~\ref{tab:ffhq-models}).
With RV-v5.1, the baselines are published results that use far more denoiser
evaluations ($999$ NFEs for \textsc{SILO}, against our $53$), yet \paseo\ or
\paseo$^\dagger$ has the lowest LPIPS on both super-resolution tasks and blur;
\textsc{SILO} keeps inpainting and JPEG.
With SD3.5, where every baseline gets more NFEs, \paseo\ has the best LPIPS on
four of the five tasks (Appendix~\ref{sec:baseline-protocol}). Against the strongest baseline, \textsc{ReSample}, it is
$10$--$20\times$ faster and uses about $25\%$ less memory; the price is PSNR,
which is higher for \textsc{ReSample} on four tasks
(Table~\ref{tab:ffhq-models-full}). JPEG, our only nonlinear degradation, is
the hardest case for our sampler with both priors.
\textsc{DING} is faster on inpainting at the same memory, but it handles only
that task, and with SD3.5 \paseo\ beats it on all three quality metrics.

\textbf{COCO.}
At matched NFEs, \paseo\ beats the flow-matching baselines \textsc{FlowDPS},
\textsc{FlowChef}, and \textsc{PnP-Flow} on every task in LPIPS, runtime, and
memory (Table~\ref{tab:coco-models-arc}; reconstructions in
Figure~\ref{fig:coco-qualitative}). Against all baselines
(Table~\ref{tab:coco-models-arc-full}), \paseo\ or \paseo$^\dagger$ still has the
lowest LPIPS on every task except JPEG, which is again the hardest case.
Across the $15$ settings of Tables~\ref{tab:ffhq-models}
and~\ref{tab:coco-models-arc}, \paseo$^\dagger$ beats \paseo\ in $3$ and is
within $0.02$ LPIPS in $5$ more, so the original anchor also works in
practice. We keep the arc anchor as the default because it is better in the
remaining cases and up to $7\times$ faster.

\textbf{Why JPEG is hard.}
Each correction relies on one question: if the latent changes slightly, how
does the measurement change? For blur, super-resolution, and inpainting, the
answer is smooth and reliable, because these degradations are linear. JPEG
instead rounds the image's frequency coefficients to fixed levels: small
changes to the image leave the output unchanged, and the output changes only
in abrupt jumps, when a coefficient crosses to the next level. There is no
reliable local direction to follow, and the learned operator cannot supply
one, so the corrections get little guidance. This makes JPEG our hardest case
with both priors. We test this in Appendix~\ref{sec:operator-validation}, where
the learned operator's local directions for JPEG do not match those of the
true degradation.

\par
\WFclear

\vspace{-8pt}
\subsection{Ablations}
\label{sec:main-ablations}
\vspace{-4pt}
\textbf{Solve depth.}
We stop CG after a few iterations, far from an exact solve. To test whether
this pulls the sampler away from the posterior, we replace $\rmH_\psi$ with an
exact linear latent blur, whose exact update we can compute. We also draw the
ground truth from the model's own prior, so that a calibrated sampler should
see it as one more draw (Appendix~\ref{sec:app-linear-check}).
\par\smallskip
\noindent
\begin{minipage}[t]{0.49\textwidth}
\vspace{0pt}
\begingroup
\centering
\fontsize{8}{9}\selectfont
\renewcommand{\arraystretch}{1.15}
\setlength{\tabcolsep}{2.5pt}
\setlength{\abovecaptionskip}{0pt}
\setlength{\belowcaptionskip}{3pt}
\captionof{table}{COCO-val2017 with SD3.5-medium: \paseo\ versus flow-matching
baselines at matched NFEs.}
\label{tab:coco-models-arc}
\begin{tabular}{@{}c>{\fontsize{7}{9}\selectfont}lrrrr@{}}
\toprule
Task & {\fontsize{8}{9}\selectfont Method} & LPIPS $\downarrow$ & KID $\downarrow$ & Time $\downarrow$ & Mem $\downarrow$ \\
\midrule
\rowcolor{ourrow} \cellcolor{white}\multirow{5}{*}{\rotatebox[origin=c]{90}{\fontsize{6.5}{7.5}\selectfont SR $\times4$}} & \paseo & 0.208 & 1.14 & \textbf{26.4} & \textbf{16.8} \\
 & \cellcolor{ourrow}\paseo$^\dagger$ & \cellcolor{ourrow}\textbf{0.196} & \cellcolor{ourrow}\textbf{0.73} & \cellcolor{ourrow}27.8 & \cellcolor{ourrow}\textbf{16.8} \\
 & \textsc{FlowDPS} & 0.374 & 9.19 & 76.8 & 18.7 \\
 & \textsc{FlowChef} & 0.345 & 8.03 & 228.9 & 18.7 \\
 & \textsc{PnP-Flow} & 0.449 & 19.00 & 35.5 & 18.7 \\
\midrule
\rowcolor{ourrow} \cellcolor{white}\multirow{5}{*}{\rotatebox[origin=c]{90}{\fontsize{6.5}{7.5}\selectfont SR $\times8$}} & \paseo & \textbf{0.344} & \textbf{7.36} & \textbf{41.2} & \textbf{16.8} \\
 & \cellcolor{ourrow}\paseo$^\dagger$ & \cellcolor{ourrow}0.389 & \cellcolor{ourrow}35.92 & \cellcolor{ourrow}92.6 & \cellcolor{ourrow}\textbf{16.8} \\
 & \textsc{FlowDPS} & 0.513 & 24.76 & 156.6 & 18.7 \\
 & \textsc{FlowChef} & 0.550 & 24.13 & 468.3 & 18.7 \\
 & \textsc{PnP-Flow} & 0.494 & 21.24 & 70.8 & 18.7 \\
\midrule
\rowcolor{ourrow} \cellcolor{white}\multirow{5}{*}{\rotatebox[origin=c]{90}{\fontsize{6.5}{7.5}\selectfont Blur}} & \paseo & \textbf{0.229} & \textbf{0.78} & 22.7 & \textbf{16.8} \\
 & \cellcolor{ourrow}\paseo$^\dagger$ & \cellcolor{ourrow}0.280 & \cellcolor{ourrow}2.10 & \cellcolor{ourrow}\textbf{18.3} & \cellcolor{ourrow}\textbf{16.8} \\
 & \textsc{FlowDPS} & 0.456 & 15.11 & 80.2 & 18.7 \\
 & \textsc{FlowChef} & 0.438 & 15.01 & 240.2 & 18.7 \\
 & \textsc{PnP-Flow} & 0.552 & 25.40 & 36.4 & 18.7 \\
\midrule
\rowcolor{ourrow} \cellcolor{white}\multirow{5}{*}{\rotatebox[origin=c]{90}{\fontsize{6.5}{7.5}\selectfont Inpaint}} & \paseo & \textbf{0.174} & \textbf{5.26} & 58.7 & \textbf{16.8} \\
 & \cellcolor{ourrow}\paseo$^\dagger$ & \cellcolor{ourrow}0.186 & \cellcolor{ourrow}5.65 & \cellcolor{ourrow}\textbf{40.3} & \cellcolor{ourrow}\textbf{16.8} \\
 & \textsc{FlowDPS} & 0.391 & 16.02 & 156.4 & 18.7 \\
 & \textsc{FlowChef} & 0.391 & 27.47 & 468.1 & 18.7 \\
 & \textsc{PnP-Flow} & 0.697 & 63.14 & 70.5 & 18.7 \\
\midrule
\rowcolor{ourrow} \cellcolor{white}\multirow{5}{*}{\rotatebox[origin=c]{90}{\fontsize{6.5}{7.5}\selectfont JPEG}} & \paseo & \textbf{0.302} & 18.82 & \textbf{24.8} & \textbf{16.8} \\
 & \cellcolor{ourrow}\paseo$^\dagger$ & \cellcolor{ourrow}0.307 & \cellcolor{ourrow}17.27 & \cellcolor{ourrow}112.6 & \cellcolor{ourrow}\textbf{16.8} \\
 & \textsc{FlowDPS} & 0.378 & 21.56 & 80.7 & 18.7 \\
 & \textsc{FlowChef} & 0.355 & \textbf{9.91} & 243.6 & 18.7 \\
 & \textsc{PnP-Flow} & 0.457 & 26.70 & 36.9 & 18.7 \\
\bottomrule
\end{tabular}
\par\endgroup

\end{minipage}\hfill
\begin{minipage}[t]{0.48\textwidth}
\vspace{0pt}
Solving exactly adds a mesh-like texture (LPIPS $0.75$--$0.80$), while a few
iterations give sharp samples (Figure~\ref{fig:main-cg-curve}a). The arc
anchor is best at $C=3$ and the original anchor at $C=8$ (LPIPS
$0.06$--$0.07$); at these budgets, the draws have the right amount of fine
detail, and their spread matches their distance to the truth within $3\%$.
With the arc anchor, the learned operator is also best at small budgets,
$C=2$--$4$ (Figure~\ref{fig:main-cg-curve}b). Stopping CG early is therefore
not only cheaper: with an exact operator, it keeps the draws closest to the
posterior. Like a step size, $C$ and $\rho$ are tuned per task.
Appendix~\ref{sec:ofat} reports LPIPS at every budget, and
Appendix~\ref{sec:app-anchor-ablation} compares the anchors.
\par\medskip
\begingroup
\centering
\fontsize{8}{9}\selectfont
\setlength{\abovecaptionskip}{2pt}
\definecolor{cgArc}{HTML}{EB6834}
\definecolor{cgBeta}{HTML}{2A78D6}
\begin{tikzpicture}
\begin{groupplot}[
  group style={group size=2 by 1, horizontal sep=0.6cm},
  scale only axis, width=0.39\linewidth, height=2.3cm,
  xmode=log, log basis x=2, axis x line*=bottom, axis y line*=left,
  tick label style={font=\tiny}, label style={font=\tiny},
  title style={font=\scriptsize, yshift=-1.5ex}, tick align=outside,
  major tick length=1.5pt, ymajorgrids, grid style={black!12},
  ylabel style={yshift=-1.2ex},  %
  every axis plot/.append style={line width=0.9pt, mark size=1.2pt},
  clip mode=individual,  %
]
\nextgroupplot[
  title={(a) Exact linear blur}, ylabel={LPIPS}, ymode=log,
  xmin=0.8, xmax=70, ymin=0.05, ymax=0.9,
  xtick={1,2,4,8,16,48}, xticklabels={1,2,4,8,16,exact},
  ytick={0.05,0.1,0.2,0.5}, yticklabels={0.05,0.1,0.2,0.5},
]
\addplot[draw=none, fill=black!6, forget plot] coordinates {(22,0.05) (70,0.05) (70,0.9) (22,0.9)} \closedcycle;
\addplot[cgArc, mark=*] coordinates {(1,0.085) (2,0.074) (3,0.068) (5,0.069) (8,0.110) (16,0.469) (48,0.799)};
\addplot[cgBeta, mark=square*] coordinates {(1,0.099) (2,0.079) (3,0.070) (5,0.063) (8,0.061) (16,0.085) (48,0.751)};
\addplot[cgArc, only marks, mark=o, mark size=2.8pt, line width=0.6pt, forget plot] coordinates {(3,0.068)};
\addplot[cgBeta, only marks, mark=o, mark size=2.8pt, line width=0.6pt, forget plot] coordinates {(8,0.061)};
\node[font=\scriptsize, text=cgArc, anchor=east] at (axis cs:11,0.3) {arc};  %
\node[font=\scriptsize, text=cgBeta, anchor=south west] at (axis cs:0.85,0.11) {original};
\node[font=\scriptsize, fill=white, inner sep=0.4pt] at (axis cs:22,0.05) {$/\!/$};
\nextgroupplot[
  title={(b) Learned operator, arc}, ymode=log,
  xmin=0.8, xmax=20, ymin=0.1, ymax=0.6,
  xtick={1,2,4,8,16}, xticklabels={1,2,4,8,16},
  ytick={0.1,0.2,0.3,0.5}, yticklabels={0.1,0.2,0.3,0.5},
]
\addplot[cgArc, mark=*] coordinates {(1,0.215) (2,0.189) (4,0.320) (8,0.400) (16,0.483)}
  node[pos=0.72, anchor=south east, font=\scriptsize, text=cgArc] {SD3.5};
\addplot[cgArc, mark=*, dashed] coordinates {(1,0.163) (2,0.155) (3,0.150) (4,0.148) (8,0.168) (16,0.217)}
  node[pos=0.2, anchor=north, font=\scriptsize, text=cgArc, yshift=-2pt] {RV-v5.1};
\addplot[cgArc, only marks, mark=o, mark size=2.6pt, line width=0.6pt, forget plot] coordinates {(2,0.189)};
\addplot[cgArc, only marks, mark=o, mark size=2.6pt, line width=0.6pt, forget plot] coordinates {(4,0.148)};
\end{groupplot}
\end{tikzpicture}


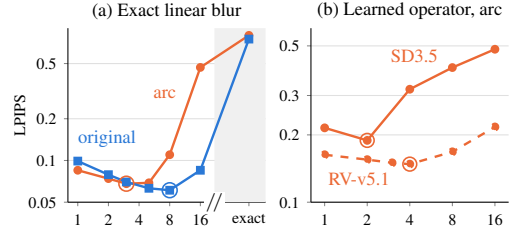
\captionof{figure}{Few CG iterations are best: LPIPS against the number
of CG iterations $C$; circles mark the best budget. (a) Exact linear latent blur, truths from the model's prior ($64$
images, Appendix~\ref{sec:app-linear-check}). (b) The learned operator on
blur (Tables~\ref{tab:ofat-rv} and~\ref{tab:ofat-sd35}).}
\label{fig:main-cg-curve}
\par\endgroup
\end{minipage}
\par
\setlength{\parskip}{.5pc}

\par\medskip
\vspace{-5pt}
\section{Conclusion}
\vspace{-5pt}
\label{sec:conclusion}
Conditioning a latent diffusion model on a learned operator does not require
back-propagating through the denoiser. \paseo\ casts each step as Gaussian
inference over the clean latent, so conditioning becomes a perturb-and-solve
step that is exact for affine operators, forms no Jacobian, and runs the
autoencoder only twice per reconstruction; a few CG iterations beat an exact
solve. Across FFHQ and COCO, with the same or fewer denoiser evaluations than
every baseline, \paseo\ ranks first or second in LPIPS on almost every task,
while running up to $\textcolor{citationblue}{19\times}$ faster and using up to $\textcolor{citationblue}{34\%}$ less peak memory
than the baselines.

\par
\noindent\textbf{Limitations.}
\paseo\ needs a trained latent operator for each degradation--prior pair, and
its quality is bounded by that operator: on JPEG, where the learned operator's
local directions are poorest, \paseo\ is weakest.

\par
\endgroup

\clearpage
\bibliographystyle{iclr2027_conference}
\bibliography{references}

\newpage
\appendix
\addtocontents{toc}{\protect\setcounter{tocdepth}{3}}
\begingroup
\renewcommand{\contentsname}{Supplementary contents}
\newcommand{\supplementarycontentscontinued}{%
  \clearpage\section*{Supplementary contents (continued)}}
\hypersetup{linktoc=all}
\pdfbookmark[0]{Supplementary contents}{supplementary.contents}
\small
\color{black}
\let\supplementarycontentsline\contentsline
\renewcommand{\contentsline}[4]{%
  \supplementarycontentsline{#1}{%
    \begingroup\renewcommand{\color}[2][]{}#2\endgroup}{#3}{#4}}
\tableofcontents
\endgroup
\clearpage

\appendixpart{I}{Theory}
\section{Linear--Gaussian posterior derivations}
\label{sec:posterior-derivation}

This section derives the posterior forms, conjugate-gradient properties, and
perturbation sampler used in Section~\ref{sec:method}. Each subsection first
provides the necessary context and states its central result before giving the
derivation.

\subsection{Precision-form posterior}
\label{sec:app-precision-form}

At a fixed target time $t$, condition on the current diffusion state. In the
notation of Section~\ref{sec:background}, the DDIM next-state bridge is
$\cstate{\rvx_t}\sim\mathbf{N}(\rvm_t,\eta_t^2\rmI_d)$, with $\rvm_t$ the
deterministic part of the transition \eqref{eq:b-ddim}. Draw the auxiliary proxy
$\cproxy{\rvx_t^{\mathrm{pxy}}}$ independently from this bridge, condition on its
realization, and hold $\cproxy{\hat{\rvx}_1'}=\xone{\cproxy{\rvx_t^{\mathrm{pxy}}}}{t}$ fixed.
For a candidate bridge state $\cstate{\rvx_t}$, define its clean-latent coordinate
$\clat{\rvz_0}$ by $\cstate{\rvx_t}=\alpha_t\clat{\rvz_0}+\sigma_t\cproxy{\hat{\rvx}_1'}$. The resulting affine
transformation of the bridge gives
\begin{equation*}
    \clat{\rvz_0}\sim\mathbf{N}(\clat{\vmu_z},\eta_z^2\rmI_d),
    \qquad
    \clat{\vmu_z}=\frac{\rvm_t-\sigma_t\cproxy{\hat{\rvx}_1'}}{\alpha_t},
    \qquad
    \eta_z=\frac{\eta_t}{\alpha_t}.
\end{equation*}
We suppress the fixed conditioning variables and the step index below. Under
the temporary linear-operator assumption of Section~\ref{sec:method}, the
likelihood is
$\cmeas{\rvy}\mid\clat{\rvz_0}\sim
\mathbf{N}(\emH\clat{\rvz_0},\sigma_{\rvy}^2\rmI_{d_{\rvy}})$.

\paragraph{Result.}
The posterior is
\begin{equation}
\boxed{
    p(\rvz_0 \mid \rvy)
    = \mathbf{N}\!\left(\emLambda^{-1}\rvb,\emLambda^{-1}\right)}.
\label{eq:app-precision-result}
\end{equation}
Here
$\emLambda:=\rmI_{d}/\eta_z^2
+\emH^T\emH/\sigma_{\rvy}^2$ is the posterior precision and
$\rvb:=\vmu_z/\eta_z^2+\emH^T\rvy/\sigma_{\rvy}^2$ is its information vector.
Equivalently, $\rvm_p=\emLambda^{-1}\rvb$ and
$\mathbf{\Sigma}_p=\emLambda^{-1}$.

\paragraph{Derivation.}
Bayes' rule gives
$p(\rvz_0\mid\rvy)\propto
p(\rvy\mid\rvz_0)p(\rvz_0)$. Taking $-2\log$ turns this product into
the sum
\begin{equation*}
    -2\log p(\rvz_0\mid\rvy)
    = \frac{\|\rvz_0-\vmu_z\|^2}{\eta_z^2}
      + \frac{\|\rvy-\emH\rvz_0\|^2}{\sigma_{\rvy}^2}
      + \text{const}.
\end{equation*}
Expanding the two squared norms gives
\begin{align*}
    \|\rvz_0-\vmu_z\|^2
        &= \rvz_0^T\rvz_0-2\vmu_z^T\rvz_0+\vmu_z^T\vmu_z, \\
    \|\rvy-\emH\rvz_0\|^2
        &= \rvz_0^T\emH^T\emH\rvz_0
           -2\rvy^T\emH\rvz_0+\rvy^T\rvy.
\end{align*}
The final term in each line is independent of $\rvz_0$ and can be absorbed into
the constant. Grouping the remaining quadratic and linear terms yields
\begin{align*}
    -2\log p(\rvz_0\mid\rvy)
    &= \rvz_0^T\!\left(
        \frac{\rmI_{d}}{\eta_z^2}
        +\frac{\emH^T\emH}{\sigma_{\rvy}^2}\right)\!\rvz_0
       -2\left(
        \frac{\vmu_z}{\eta_z^2}
        +\frac{\emH^T\rvy}{\sigma_{\rvy}^2}\right)^{\!T}\rvz_0
       +\text{const} \\
    &= \rvz_0^T\emLambda\rvz_0-2\rvb^T\rvz_0+\text{const}.
\end{align*}
Thus, $\emLambda$ and $\rvb$ are precisely the quadratic and linear
coefficients of the posterior log-density. Both $\rmI_{d}$ and
$\emH^T\emH$ are symmetric. Moreover, for any nonzero $\rvu$, provided
$\eta_z>0$ and $\sigma_{\rvy}>0$,
\begin{equation*}
    \rvu^T\emLambda\rvu
    = \frac{\|\rvu\|^2}{\eta_z^2}
      +\frac{\|\emH\rvu\|^2}{\sigma_{\rvy}^2}>0,
\end{equation*}
so $\emLambda$ is symmetric positive definite and therefore invertible.

To complete the square, set $\rvc=\emLambda^{-1}\rvb$. Symmetry of
$\emLambda$ gives
\begin{align*}
    (\rvz_0-\rvc)^T\emLambda(\rvz_0-\rvc)
    &= \rvz_0^T\emLambda\rvz_0-2\rvb^T\rvz_0
       +\rvb^T\emLambda^{-1}\rvb,
\end{align*}
and hence
\begin{equation*}
    p(\rvz_0\mid\rvy)
    \propto \exp\!\left[-\frac12
        (\rvz_0-\emLambda^{-1}\rvb)^T
        \emLambda(\rvz_0-\emLambda^{-1}\rvb)\right].
\end{equation*}
Comparing this expression with the density of a Gaussian shows that its mean is
$\emLambda^{-1}\rvb$ and its covariance is $\emLambda^{-1}$, which proves
\eqref{eq:app-precision-result}.

\subsection{Measurement-space gain form}
\label{sec:app-gain-form}

The precision form in Appendix~\ref{sec:app-precision-form} is exact, but it
expresses both posterior moments through the inverse of the
$d\times d$ matrix $\emLambda$. Its identity-plus-sandwich
structure allows the inverse to be moved to measurement space.

\paragraph{Result.}
Define the measurement-space matrix and gain by
$\emS:=\sigma_{\rvy}^2\rmI_{d_{\rvy}}+\eta_z^2\emH\emH^T$ and
$\emK:=\eta_z^2\emH^T\emS^{-1}$. The posterior mean is
\begin{equation}
\boxed{\rvm_p=\vmu_z+\emK(\rvy-\emH\vmu_z)}.
\label{eq:app-gain-result}
\end{equation}
The same rearrangement gives the posterior covariance
\begin{equation*}
    \mathbf{\Sigma}_p
    = \eta_z^2\rmI_{d}
      -\eta_z^4\emH^T\emS^{-1}\emH.
\end{equation*}
Thus, the only inverse left in either moment is the
$d_{\rvy}\times d_{\rvy}$ matrix $\emS$.

\paragraph{Derivation.}
The Sherman--Morrison--Woodbury identity states that
\begin{equation*}
    (\emA+\emU\emC\emV)^{-1}
    =\emA^{-1}-\emA^{-1}\emU
      (\emC^{-1}+\emV\emA^{-1}\emU)^{-1}
      \emV\emA^{-1}.
\end{equation*}
The precision matrix $\emLambda$ matches its left-hand side with
\begin{equation*}
    \emA = \eta_z^{-2}\rmI_{d},
    \qquad
    \emU = \emH^T,
    \qquad
    \emC = \sigma_{\rvy}^{-2}\rmI_{d_{\rvy}},
    \qquad
    \emV = \emH.
\end{equation*}
Indeed, $\emU\emC\emV=\sigma_{\rvy}^{-2}\emH^T\emH$. The pieces on the
right-hand side are
\begin{align*}
    \emA^{-1} &= \eta_z^2\rmI_{d},
    & \emA^{-1}\emU &= \eta_z^2\emH^T, \\
    \emV\emA^{-1} &= \eta_z^2\emH,
    & \emC^{-1}+\emV\emA^{-1}\emU
        &= \sigma_{\rvy}^2\rmI_{d_{\rvy}}
           +\eta_z^2\emH\emH^T=\emS.
\end{align*}
Substituting them into the identity gives
\begin{equation*}
    \mathbf{\Sigma}_p
    = \emLambda^{-1}
    = \eta_z^2\rmI_{d}
      - \eta_z^4\emH^T\emS^{-1}\emH.
\end{equation*}
The two factors of $\eta_z^2$ in the correction come from
$\emA^{-1}\emU$ and $\emV\emA^{-1}$, respectively.

For the mean, substitute
$\rvb=\vmu_z/\eta_z^2+\emH^T\rvy/\sigma_{\rvy}^2$ and the covariance above
into $\rvm_p=\emLambda^{-1}\rvb$:
\begin{align*}
    \rvm_p
    &= \vmu_z
       - \eta_z^2\emH^T\emS^{-1}\emH\vmu_z \\
    &\quad
       + \frac{\eta_z^2}{\sigma_{\rvy}^2}\emH^T
         \left(\rmI_{d_{\rvy}}
         - \eta_z^2\emS^{-1}\emH\emH^T\right)\rvy.
\end{align*}
Multiplying the definition of $\emS$ by $\emS^{-1}$ on the left gives
\begin{equation*}
    \rmI_{d_{\rvy}}
    =\sigma_{\rvy}^2\emS^{-1}
     +\eta_z^2\emS^{-1}\emH\emH^T,
\end{equation*}
or, equivalently,
\begin{equation*}
    \rmI_{d_{\rvy}}
    - \eta_z^2\emS^{-1}\emH\emH^T
    = \sigma_{\rvy}^2\emS^{-1}.
\end{equation*}
Substituting this identity into the expanded mean and collecting the two
measurement-space terms yields
\begin{align*}
    \rvm_p
    &=\vmu_z+\eta_z^2\emH^T\emS^{-1}\rvy
      -\eta_z^2\emH^T\emS^{-1}\emH\vmu_z \\
    &=\vmu_z+\eta_z^2\emH^T\emS^{-1}
      (\rvy-\emH\vmu_z)
     =\vmu_z+\emK(\rvy-\emH\vmu_z),
\end{align*}
which proves \eqref{eq:app-gain-result}.

\subsection{Matrix-free conjugate-gradient solve}
\label{sec:app-cg-solve}

The gain form moves the inverse to measurement space, but it still contains
$\emS^{-1}$. The posterior mean never needs this inverse as a matrix: it needs
its action only on the single residual $\rvy-\emH\vmu_z$. This is exactly the
setting for conjugate gradients.

\paragraph{Result.}
Let $\rvd:=\rvy-\emH\vmu_z$ and
$\emS:=\sigma_{\rvy}^2\rmI_{d_{\rvy}}+\eta_z^2\emH\emH^T$. The gain-form mean
is obtained from the single solve and update
\begin{equation}
\boxed{\emS\rvv=\rvd,
       \qquad \rvm_p=\vmu_z+\eta_z^2\emH^T\rvv.}
\label{eq:app-cg-result}
\end{equation}
CG can evaluate this solve without forming $\emS$ or $\emH\emH^T$: one
iteration needs one application of $\emH^T$ followed by one of $\emH$.
With a zero initial guess and $k$ executed iterations, forming $\rvd$, running
CG, and lifting $\rvv$ into $\rvm_p$ uses $k+1$ applications of each operator,
or $2k+2$ operator applications in total.

\paragraph{Derivation.}
From Appendix~\ref{sec:app-gain-form},
$\rvm_p=\vmu_z+\eta_z^2\emH^T\emS^{-1}\rvd$. Naming
$\rvv:=\emS^{-1}\rvd$ converts this one inverse--vector product into
\eqref{eq:app-cg-result}.

CG accesses $\emS$ only through products with its search vectors. For any
$\rvu\in\R^{d_{\rvy}}$,
\begin{equation*}
    \emS\rvu
    = \sigma_{\rvy}^2\rvu
      + \eta_z^2\emH\bigl(\emH^T\rvu\bigr).
\end{equation*}
This product is evaluated from right to left and therefore uses one application
of $\emH^T$ followed by one application of $\emH$, without forming $\emS$ or
$\emH\emH^T$.

CG requires $\emS$ to be symmetric positive definite. Symmetry follows from
$\emS=\sigma_{\rvy}^2\rmI_{d_{\rvy}}+\eta_z^2\emH\emH^T$. For every nonzero
$\rvu$ and $\sigma_{\rvy}>0$,
\begin{equation*}
    \rvu^T\emS\rvu
    = \sigma_{\rvy}^2\|\rvu\|^2
      + \eta_z^2\|\emH^T\rvu\|^2 > 0,
\end{equation*}
because the first term is strictly positive. Thus $\emS$ remains positive
definite even when $\emH$ is rank deficient.

The extreme eigenvalues satisfy
\begin{equation*}
    \lambda_{\min}(\emS) \geq \sigma_{\rvy}^2,
    \qquad
    \lambda_{\max}(\emS)
    = \sigma_{\rvy}^2+\eta_z^2\|\emH\|_2^2.
\end{equation*}
Consequently,
\begin{equation*}
    \kappa_2(\emS)
    = \frac{\lambda_{\max}(\emS)}{\lambda_{\min}(\emS)}
    \leq 1+\frac{\eta_z^2\|\emH\|_2^2}{\sigma_{\rvy}^2}.
\end{equation*}
The bound approaches one as $\eta_z$ decreases, so the later systems are
increasingly well conditioned.

\subsection{Posterior sampling by input perturbation}
\label{sec:app-perturb-sampling}

The solve above produces the posterior mean, whereas a posterior draw would
seem to require a square root of $\mathbf{\Sigma}_p$. Instead, randomness can
be introduced through the Gaussian prior and likelihood before applying the
same measurement-space solve.

\paragraph{Result.}
Draw independent $\bxi_z\sim\mathbf{N}(0,\rmI_{d})$ and
$\bxi_y\sim\mathbf{N}(0,\rmI_{d_{\rvy}})$, and set
$\tilde{\vmu}_z:=\vmu_z+\eta_z\bxi_z$ and
$\tilde{\rvy}:=\rvy+\sigma_{\rvy}\bxi_y$. With
$\emS:=\sigma_{\rvy}^2\rmI_{d_{\rvy}}+\eta_z^2\emH\emH^T$, solve and update
\begin{equation*}
    \emS\rvv=\tilde{\rvy}-\emH\tilde{\vmu}_z,
    \qquad
    \rvz_0^+=\tilde{\vmu}_z+\eta_z^2\emH^T\rvv.
\end{equation*}
For a fixed linear model and an exact solve, the central distributional claim is
\begin{equation}
\boxed{\rvz_0^+\sim\mathbf{N}(\rvm_p,\mathbf{\Sigma}_p).}
\label{eq:app-perturb-result}
\end{equation}
Here
\begin{equation*}
    \rvm_p=\vmu_z+\eta_z^2\emH^T\emS^{-1}(\rvy-\emH\vmu_z),
    \qquad
    \mathbf{\Sigma}_p
    =\eta_z^2\rmI_{d}-\eta_z^4\emH^T\emS^{-1}\emH.
\end{equation*}
The randomized solve replaces the deterministic mean solve; it is not an
additional solve.

\paragraph{Derivation.}
Let $\emK=\eta_z^2\emH^T\emS^{-1}$. Substituting the perturbed inputs into the
gain-form update gives
\begin{align*}
    \rvz_0^+
    &= \tilde{\vmu}_z
       +\emK(\tilde{\rvy}-\emH\tilde{\vmu}_z) \\
    &= \underbrace{\vmu_z+\emK(\rvy-\emH\vmu_z)}_{=\,\rvm_p}
       +\eta_z(\rmI_{d}-\emK\emH)\bxi_z
       +\sigma_{\rvy}\emK\bxi_y.
\end{align*}
This is an affine function of two independent Gaussian vectors and is therefore
Gaussian. Since both perturbations have zero mean,
$\mathbb{E}[\rvz_0^+]=\rvm_p$.

Independence also makes the cross-covariance terms vanish, giving
\begin{equation*}
    \operatorname{Cov}(\rvz_0^+)
    = \eta_z^2(\rmI_{d}-\emK\emH)
      (\rmI_{d}-\emK\emH)^T
      +\sigma_{\rvy}^2\emK\emK^T.
\end{equation*}
Because $\emS$ is symmetric, $\emK\emH$ is symmetric. Expanding the covariance
and using the definition of $\emS$ yields
\begin{align*}
    \operatorname{Cov}(\rvz_0^+)
    &= \eta_z^2\rmI_{d}
       -2\eta_z^2\emK\emH
       +\emK\bigl(\eta_z^2\emH\emH^T
                    +\sigma_{\rvy}^2\rmI_{d_{\rvy}}\bigr)\emK^T \\
    &= \eta_z^2\rmI_{d}
       -2\eta_z^2\emK\emH+\emK\emS\emK^T.
\end{align*}
The identity $\emK\emS=\eta_z^2\emH^T$ implies
$\emK\emS\emK^T=\eta_z^2\emK\emH$. Hence
\begin{equation*}
    \operatorname{Cov}(\rvz_0^+)
    = \eta_z^2\rmI_{d}-\eta_z^2\emK\emH
    = \eta_z^2\rmI_{d}
      -\eta_z^4\emH^T\emS^{-1}\emH
    = \mathbf{\Sigma}_p.
\end{equation*}
Together with the mean calculation, this proves the distributional statement
in \eqref{eq:app-perturb-result}. Equivalently, the two perturbations provide
the rectangular covariance factor
\begin{equation*}
    \left[\;\eta_z(\rmI_{d}-\emK\emH)
    \;\middle|\;\sigma_{\rvy}\emK\;\right],
\end{equation*}
whose product with its transpose is $\mathbf{\Sigma}_p$. A symmetric square
root is therefore unnecessary.

This exactness result assumes the fixed linear model above and an exact linear
solve. Truncated CG, damping, or relinearization yields an approximate draw,
as in the practical nonlinear update of Algorithm~\ref{alg:silo-jding}.
Appendix~\ref{sec:app-tractable} checks this result numerically.

\clearpage

\section{Nonlinear learned-operator extension}
\label{sec:app-nonlinear-operator}

The preceding derivations assume a fixed linear measurement map. We now derive
the local linear model used when that map is the nonlinear learned operator
$\rmH_{\psi}$.

\subsection{Local linearization of the learned operator}
\label{sec:app-local-linearization}

At one inner update, the DDIM bridge supplies the clean-latent belief
$\rvz_0\sim\mathbf{N}(\vmu_z,\eta_z^2\rmI_d)$, while the encoded observation obeys
$\yz=\rmH_{\psi}(\rvz_0)+\rvepsilon$ with
$\rvepsilon\sim\mathbf{N}(0,r^2\rmI_d)$. The only part that prevents the
linear--Gaussian analysis above is the nonlinearity of $\rmH_{\psi}$.
At a fixed expansion point $\rvz_0^{(j)}$, let
$\rvh_j:=\rmH_{\psi}(\rvz_0^{(j)})$ and
$\rmJ_j:=\left.\partial\rmH_{\psi}/\partial\rvz_0\right|_{\rvz_0^{(j)}}$.

\paragraph{Result.}
The central approximation is the first-order linearization
\begin{equation}
\boxed{\rmH_{\psi}(\rvz_0)
       \approx \rvh_j+\rmJ_j(\rvz_0-\rvz_0^{(j)}).}
\label{eq:app-local-linearization-result}
\end{equation}
Moving the known affine offset to the measurement side defines
$\rvy_{\mathrm{eff}}^{(j)}:=\yz-\rvh_j+\rmJ_j\rvz_0^{(j)}$, for which
\begin{align*}
    \rvy_{\mathrm{eff}}^{(j)}
        &\approx \rmJ_j\rvz_0+\rvepsilon, \\
    \rvy_{\mathrm{eff}}^{(j)}-\rmJ_j\vmu_z
        &=\yz-\rvh_j-\rmJ_j(\vmu_z-\rvz_0^{(j)}).
\end{align*}
Thus the nonlinear likelihood is locally the same linear--Gaussian model as
before, with $\emH$ replaced by $\rmJ_j$ and $\rvy$ by
$\rvy_{\mathrm{eff}}^{(j)}$.

\paragraph{Derivation.}
Taylor's theorem writes the operator as
\begin{equation*}
    \rmH_{\psi}(\rvz_0)
    = \rvh_j+\rmJ_j(\rvz_0-\rvz_0^{(j)})+\mathbf{R}_j(\rvz_0),
\end{equation*}
where $\mathbf{R}_j(\rvz_0)$ is the remainder. If the Jacobian is
$L$-Lipschitz on the line segment joining $\rvz_0^{(j)}$ and $\rvz_0$, then
\begin{equation*}
    \|\mathbf{R}_j(\rvz_0)\|
    \leq \frac{L}{2}\|\rvz_0-\rvz_0^{(j)}\|^2.
\end{equation*}
Dropping this second-order term gives
\eqref{eq:app-local-linearization-result}. Substituting the expansion into the
observation model and moving the known affine offset gives
\begin{align*}
    \yz
        &= \rvh_j+\rmJ_j(\rvz_0-\rvz_0^{(j)})
           +\mathbf{R}_j(\rvz_0)+\rvepsilon \\
    \yz-\rvh_j+\rmJ_j\rvz_0^{(j)}
        &= \rmJ_j\rvz_0+\mathbf{R}_j(\rvz_0)+\rvepsilon.
\end{align*}
After the remainder is discarded, this is the claimed linear--Gaussian model.
Its residual at the Gaussian-prior center is
\begin{align*}
    \rvy_{\mathrm{eff}}^{(j)}-\rmJ_j\vmu_z
    &= \yz-\rvh_j+\rmJ_j\rvz_0^{(j)}-\rmJ_j\vmu_z \\
    &= \yz-\rvh_j-\rmJ_j(\vmu_z-\rvz_0^{(j)}),
\end{align*}
as stated above. If $\rmH_{\psi}$ is affine, the remainder vanishes and the
reduction is exact. Otherwise its neglected error is second order in the
distance from the expansion point.

\subsection{Perturb-and-solve sampling for a fixed local model}
\label{sec:app-local-posterior}

Fix $\rvz_0^{(j)}$ independently of the perturbations used below, and write
$\rvh_j:=\rmH_{\psi}(\rvz_0^{(j)})$ and
$\rmJ_j:=\left.\partial\rmH_{\psi}/\partial\rvz_0\right|_{\rvz_0^{(j)}}$.
With the expansion point fixed, the surrogate from
Appendix~\ref{sec:app-local-linearization} is an ordinary linear--Gaussian
model, so the preceding posterior and perturbation results apply directly.

Define its residual and measurement-space matrix by
\begin{equation*}
    \rvd_j:=\yz-\rvh_j-\rmJ_j(\vmu_z-\rvz_0^{(j)}),
    \qquad
    \emS_j:=r^2\rmI_d+\eta_z^2\rmJ_j\rmJ_j^T,
\end{equation*}
and its posterior moments by
\begin{equation*}
    \rvm_j:=\vmu_z+\eta_z^2\rmJ_j^T\emS_j^{-1}\rvd_j,
    \qquad
    \mathbf{\Sigma}_j
       :=\eta_z^2\rmI_d-\eta_z^4\rmJ_j^T\emS_j^{-1}\rmJ_j.
\end{equation*}
Draw independent $\bxi_z\sim\mathbf{N}(0,\rmI_d)$ and
$\bxi_y\sim\mathbf{N}(0,\rmI_d)$, and set
$\tilde{\vmu}_z:=\vmu_z+\eta_z\bxi_z$ and $\tyz:=\yz+r\bxi_y$.

\paragraph{Result.}
When the following local system is solved exactly, the update is an exact draw
from the fixed affine-surrogate posterior:
\begin{equation}
\boxed{
\begin{aligned}
    \emS_j\rvv
        &= \tyz-\rvh_j-\rmJ_j(\tilde{\vmu}_z-\rvz_0^{(j)}), \\
    \rvz_0^{\star}
        &= \tilde{\vmu}_z+\eta_z^2\rmJ_j^T\rvv,
    & \rvz_0^{\star} &\sim\mathbf{N}(\rvm_j,\mathbf{\Sigma}_j).
\end{aligned}}
\label{eq:app-local-sampling-result}
\end{equation}

\paragraph{Derivation.}
The fixed affine surrogate is
\begin{equation*}
    \rvz_0\sim\mathbf{N}(\vmu_z,\eta_z^2\rmI_d),
    \qquad
    \rvy_{\mathrm{eff}}^{(j)}\mid\rvz_0
    \sim\mathbf{N}(\rmJ_j\rvz_0,r^2\rmI_d).
\end{equation*}
Applying Appendices~\ref{sec:app-gain-form} and
\ref{sec:app-perturb-sampling} to this fixed affine model requires only the
substitution
\begin{equation*}
    (\rvy,\sigma_{\rvy},\emH)
    \longmapsto
    (\rvy_{\mathrm{eff}}^{(j)},r,\rmJ_j).
\end{equation*}
Under this substitution, those results give the moments above. In particular,
the unperturbed residual is
\begin{align*}
    \rvy_{\mathrm{eff}}^{(j)}-\rmJ_j\vmu_z
    &= \yz-\rvh_j+\rmJ_j\rvz_0^{(j)}-\rmJ_j\vmu_z \\
    &= \yz-\rvh_j-\rmJ_j(\vmu_z-\rvz_0^{(j)})
     = \rvd_j.
\end{align*}

For sampling, perturb the prior center and the original measurement before
forming the effective measurement:
\begin{equation*}
    \tilde{\rvy}_{\mathrm{eff}}^{(j)}
    := \tyz-\rvh_j+\rmJ_j\rvz_0^{(j)}.
\end{equation*}
We now reuse the solve from Step 2 with this adjusted measurement.
The solve subtracts the adjusted model's prediction at the perturbed prior mean,
$\rmJ_j\tilde{\vmu}_z$, giving
\begin{align*}
    \tilde{\rvy}_{\mathrm{eff}}^{(j)}-\rmJ_j\tilde{\vmu}_z
    &= \tyz-\rvh_j+\rmJ_j\rvz_0^{(j)}-\rmJ_j\tilde{\vmu}_z \\
    &= \tyz-\rvh_j-\rmJ_j(\tilde{\vmu}_z-\rvz_0^{(j)}),
\end{align*}
which is the right-hand side of the boxed solve. The Jacobian is still
evaluated at $\rvz_0^{(j)}$. To verify the distribution of the output, let
$\emK_j:=\eta_z^2\rmJ_j^T\emS_j^{-1}$.
Substituting the two perturbed inputs into the local update gives
\begin{align*}
    \rvz_0^{\star}
    &= \tilde{\vmu}_z
       +\emK_j\bigl(\tyz-\rvh_j
       -\rmJ_j(\tilde{\vmu}_z-\rvz_0^{(j)})\bigr) \\
    &= \underbrace{\vmu_z+\emK_j\rvd_j}_{=\,\rvm_j}
       +\eta_z(\rmI_d-\emK_j\rmJ_j)\bxi_z
       +r\emK_j\bxi_y.
\end{align*}
Thus $\rvz_0^{\star}$ is Gaussian with mean $\rvm_j$. Since $\emS_j$ is
symmetric, $\emK_j\rmJ_j=\eta_z^2\rmJ_j^T\emS_j^{-1}\rmJ_j$ is symmetric.
Independence of the two perturbations then gives
\begin{align*}
    \operatorname{Cov}(\rvz_0^{\star})
    &= \eta_z^2(\rmI_d-\emK_j\rmJ_j)
       (\rmI_d-\emK_j\rmJ_j)^T+r^2\emK_j\emK_j^T \\
    &= \eta_z^2\rmI_d-2\eta_z^2\emK_j\rmJ_j
       +\emK_j\bigl(\eta_z^2\rmJ_j\rmJ_j^T+r^2\rmI_d\bigr)\emK_j^T \\
    &= \eta_z^2\rmI_d-2\eta_z^2\emK_j\rmJ_j
       +\emK_j\emS_j\emK_j^T.
\end{align*}
Because $\emK_j\emS_j=\eta_z^2\rmJ_j^T$, the final term is
$\emK_j\emS_j\emK_j^T=\eta_z^2\emK_j\rmJ_j$. Therefore
\begin{equation*}
    \operatorname{Cov}(\rvz_0^{\star})
    = \eta_z^2\rmI_d-\eta_z^2\emK_j\rmJ_j
    = \eta_z^2\rmI_d-\eta_z^4\rmJ_j^T\emS_j^{-1}\rmJ_j
    = \mathbf{\Sigma}_j,
\end{equation*}
which proves \eqref{eq:app-local-sampling-result}.

The claim is exact for the fixed affine surrogate when the two perturbations
are independent and the linear system is solved exactly. For the original
nonlinear measurement model it is a local approximation. When
$\rmH_{\psi}$ is affine, the surrogate equals the original model and the claim
is exact.

\subsection{Jacobian-free conjugate gradients}
\label{sec:app-jacobian-free-cg}

The fixed-local-model update above is written with a Jacobian matrix, but a learned
operator may have far too many columns for that matrix to be assembled. At a
fixed expansion point, automatic differentiation supplies exactly the two
actions required by CG.

Let $\rvh_j:=\rmH_{\psi}(\rvz_0^{(j)})$,
$\rmJ_j:=\left.\partial\rmH_{\psi}/\partial\rvz_0\right|_{\rvz_0^{(j)}}$, and
\begin{equation*}
    \emS_j:=r^2\rmI_d+\eta_z^2\rmJ_j\rmJ_j^T,
    \qquad
    \rvb_j:=\tyz-\rvh_j-\rmJ_j(\tilde{\vmu}_z-\rvz_0^{(j)}).
\end{equation*}
The local draw solves $\emS_j\rvv=\rvb_j$ and returns
$\rvz_0^{\star}=\tilde{\vmu}_z+\eta_z^2\rmJ_j^T\rvv$. Define the automatic-%
differentiation closures
\begin{align*}
    \operatorname{JVP}_j(\rvw)
        &:=\left.\frac{\mathrm{d}}{\mathrm{d}\epsilon}
          \rmH_{\psi}(\rvz_0^{(j)}+\epsilon\rvw)
          \right|_{\epsilon=0}=\rmJ_j\rvw, \\
    \operatorname{VJP}_j(\rvu)
        &:=\left.\nabla_{\rvz_0}
          \langle\rvu,\rmH_{\psi}(\rvz_0)\rangle
          \right|_{\rvz_0=\rvz_0^{(j)}}=\rmJ_j^T\rvu.
\end{align*}

\paragraph{Result.}
For every search vector $\rvu\in\R^d$, the matrix-free CG product is
\begin{equation}
\boxed{\emS_j\rvu
       =r^2\rvu+\eta_z^2
        \operatorname{JVP}_j\!\bigl(\operatorname{VJP}_j(\rvu)\bigr).}
\label{eq:app-jacobian-free-result}
\end{equation}
For every nonzero $\rvu$ and $r>0$, the same operator satisfies
\begin{equation*}
    \rvu^T\emS_j\rvu
      =r^2\|\rvu\|^2+\eta_z^2\|\rmJ_j^T\rvu\|^2>0,
    \qquad
    \kappa_2(\emS_j)
      \leq1+\frac{\eta_z^2\|\rmJ_j\|_2^2}{r^2}.
\end{equation*}
Thus one CG iteration uses one VJP followed by one JVP and never forms
$\rmJ_j$, $\rmJ_j\rmJ_j^T$, or $\emS_j$. Beyond evaluating $\rvh_j$, forming
$\rvb_j$ requires one JVP and the final lift requires one VJP.

\paragraph{Derivation.}
The derivative of $\rmH_{\psi}$ at $\rvz_0^{(j)}$ is, by definition, the linear
map that sends a direction $\rvw$ to $\rmJ_j\rvw$; this is the JVP above. For
the transpose action, let
$g(\rvz_0):=\langle\rvu,\rmH_{\psi}(\rvz_0)\rangle$. For every direction $\rvw$,
the chain rule gives
\begin{equation*}
    Dg(\rvz_0^{(j)})[\rvw]
      =\rvu^T\rmJ_j\rvw
      =\rvw^T\rmJ_j^T\rvu.
\end{equation*}
Therefore $\nabla g(\rvz_0^{(j)})=\rmJ_j^T\rvu$, which is the VJP computed by
reverse-mode automatic differentiation.

CG accesses $\emS_j=r^2\rmI_d+\eta_z^2\rmJ_j\rmJ_j^T$ only through products
with its search vectors. Associativity gives
\begin{equation*}
    \emS_j\rvu
    = r^2\rvu+\eta_z^2\rmJ_j(\rmJ_j^T\rvu),
\end{equation*}
which is exactly \eqref{eq:app-jacobian-free-result} when evaluated from right
to left. Forming $\rvb_j$ needs the additional JVP
$\rmJ_j(\tilde{\vmu}_z-\rvz_0^{(j)})$, and the final correction needs the
additional VJP $\rmJ_j^T\rvv$.

Symmetry follows from the definition of $\emS_j$. For every nonzero
$\rvu\in\R^d$,
\begin{equation*}
    \rvu^T\emS_j\rvu
    = r^2\|\rvu\|^2+\eta_z^2\|\rmJ_j^T\rvu\|^2>0,
\end{equation*}
whenever $r>0$, irrespective of the rank of $\rmJ_j$. Thus $\emS_j$ is
symmetric positive definite. Finally,
\begin{equation*}
    \lambda_{\min}(\emS_j)\geq r^2,
    \qquad
    \lambda_{\max}(\emS_j)=r^2+\eta_z^2\|\rmJ_j\|_2^2,
\end{equation*}
and therefore
\begin{equation*}
    \kappa_2(\emS_j)
    \leq 1+\frac{\eta_z^2\|\rmJ_j\|_2^2}{r^2}.
\end{equation*}
This is the same positive-definiteness and conditioning guarantee as in the
linear case, now obtained entirely through automatic-differentiation products.

\subsection{Relinearization as Gauss--Newton}
\label{sec:app-relinearized-gn}

One affine surrogate is reliable only near its expansion point.
For the fixed perturbed-prior anchor, we
rebuild it at each new expansion point, while keeping one realization of the
perturbed targets $\tilde{\vmu}_z$ and $\tyz$ fixed throughout the inner loop.
Those fixed targets define a single randomized nonlinear objective in the
randomize-then-optimize framework~\citep{bardsley2014rto}:
\begin{equation*}
    \widetilde{\Phi}(\rvz_0)
    := \frac{1}{2\eta_z^2}\|\rvz_0-\tilde{\vmu}_z\|^2
       +\frac{1}{2r^2}\|\tyz-\rmH_{\psi}(\rvz_0)\|^2.
\end{equation*}
At $\rvz_0^{(j)}$, let $\rvh_j:=\rmH_{\psi}(\rvz_0^{(j)})$,
$\rve_j:=\tyz-\rvh_j$, and
$\rmJ_j:=\left.\partial\rmH_{\psi}/\partial\rvz_0\right|_{\rvz_0^{(j)}}$.
Linearizing only the learned operator gives the local quadratic
\begin{equation*}
    Q_j(\rvz_0)
    :=\frac{1}{2\eta_z^2}\|\rvz_0-\tilde{\vmu}_z\|^2
      +\frac{1}{2r^2}
       \|\rve_j-\rmJ_j(\rvz_0-\rvz_0^{(j)})\|^2.
\end{equation*}
Define
\begin{align*}
    \emA_j&:=\frac{1}{\eta_z^2}\rmI_d+\frac{1}{r^2}\rmJ_j^T\rmJ_j,
    &
    \rvb_j&:=\rve_j-\rmJ_j(\tilde{\vmu}_z-\rvz_0^{(j)}), \\
    \emS_j&:=r^2\rmI_d+\eta_z^2\rmJ_j\rmJ_j^T,
    &
    \emS_j\rvv_j&=\rvb_j.
\end{align*}

\paragraph{Result.}
With an exact local solve, the measurement-space proposal, the minimizer of the
local quadratic, and the full Gauss--Newton step are the same point:
\begin{equation}
\boxed{
\begin{aligned}
    \rvz_{0,j}^{\star}
       &:=\tilde{\vmu}_z+\eta_z^2\rmJ_j^T\rvv_j \\
       &=\argmin_{\rvz_0}Q_j(\rvz_0)
        =\rvz_0^{(j)}-\emA_j^{-1}
          \nabla\widetilde{\Phi}(\rvz_0^{(j)}).
\end{aligned}}
\label{eq:app-relinearized-gn-result}
\end{equation}
For the ideal full step, $\rvz_0^{(j+1)}:=\rvz_{0,j}^{\star}$. Recomputing $\rvh_j$
and $\rmJ_j$ at each such iterate is therefore exactly full-step
Gauss--Newton on $\widetilde{\Phi}$.

\paragraph{Derivation.}
Replacing $\rmH_{\psi}$ in $\widetilde{\Phi}$ by its first-order model at
$\rvz_0^{(j)}$ gives $Q_j$. At the expansion point, the local and nonlinear
objectives have the same gradient:
\begin{align*}
    \left.\nabla_{\rvz_0}Q_j(\rvz_0)\right|_{\rvz_0=\rvz_0^{(j)}}
    &= \frac{\rvz_0^{(j)}-\tilde{\vmu}_z}{\eta_z^2}
       -\frac{\rmJ_j^T\rve_j}{r^2}
     = \nabla_{\rvz_0}\widetilde{\Phi}(\rvz_0^{(j)}),
\end{align*}
while the Hessian of $Q_j$ is the constant positive-definite matrix
\begin{equation*}
    \nabla_{\rvz_0}^2 Q_j(\rvz_0)
    = \frac{1}{\eta_z^2}\rmI_d+\frac{1}{r^2}\rmJ_j^T\rmJ_j
    = \emA_j \succ 0.
\end{equation*}
The full Hessian of the nonlinear objective is
\begin{equation*}
    \nabla_{\rvz_0}^2\widetilde{\Phi}(\rvz_0^{(j)})
    = \emA_j-\frac{1}{r^2}\sum_{i=1}^{d}
      (\rve_j)_i\,\nabla_{\rvz_0}^2\rmH_{\psi,i}(\rvz_0^{(j)}).
\end{equation*}
Dropping the residual-weighted second-derivative term leaves $\emA_j$; this is
the Gauss--Newton curvature approximation. Since $Q_j$ has this curvature and
matches the nonlinear gradient at $\rvz_0^{(j)}$, its unique minimizer is
\begin{equation*}
    \argmin_{\rvz_0}Q_j(\rvz_0)
    = \rvz_0^{(j)}-\emA_j^{-1}
      \nabla\widetilde{\Phi}(\rvz_0^{(j)}).
\end{equation*}

It remains to show that the measurement-space solve computes this minimizer.
Let $\bm{\delta}:=\rvz_0-\tilde{\vmu}_z$. The residual in $Q_j$ is then
$\rvb_j-\rmJ_j\bm{\delta}$, so its first-order condition is
\begin{equation*}
    \emA_j\bm{\delta}=\frac{1}{r^2}\rmJ_j^T\rvb_j.
\end{equation*}
If $\emS_j\rvv_j=\rvb_j$ and
$\bm{\delta}=\eta_z^2\rmJ_j^T\rvv_j$, then
\begin{align*}
    \emA_j\bm{\delta}
    &= \left(\frac{1}{\eta_z^2}\rmI_d
       +\frac{1}{r^2}\rmJ_j^T\rmJ_j\right)
       \eta_z^2\rmJ_j^T\rvv_j \\
    &= \frac{1}{r^2}\rmJ_j^T
       \left(r^2\rmI_d+\eta_z^2\rmJ_j\rmJ_j^T\right)\rvv_j
     = \frac{1}{r^2}\rmJ_j^T\rvb_j.
\end{align*}
Thus $\bm{\delta}=\eta_z^2\rmJ_j^T\rvv_j$ satisfies the normal equations, whose solution is unique,
and $\tilde{\vmu}_z+\bm{\delta}=\rvz_{0,j}^{\star}$. This proves
\eqref{eq:app-relinearized-gn-result}. Recomputing $\rvh_j$ and $\rmJ_j$ at
the accepted point builds the next quadratic model of the same objective.

Keeping $\tilde{\vmu}_z$ and $\tyz$ fixed is essential: redrawing either changes
$\widetilde{\Phi}$, so the sequence no longer optimizes one objective. For
nonlinear $\rmH_{\psi}$, the final iterate is therefore an approximate
update; if $\rmH_{\psi}$ is affine, the first surrogate is exact and one
exact solve recovers the linear--Gaussian draw. Because we omit the determinant
weighting and accept--reject correction, the nonlinear iteration has an
optimization interpretation rather than an exact posterior-sampling guarantee.
This interpretation assumes a fixed anchor. The practical variant in
Section~\ref{sec:method} updates the anchor between inner corrections;
Appendix~\ref{sec:app-anchor-family} explains that construction and its implications.

\section{Anchor construction and analysis}
\label{sec:app-anchor-family}

Section~\ref{sec:method} introduced the interpolated anchor in
\eqref{eq:method-practical-anchor}: it replaces the perturbed-prior mean
$\tilde{\vmu}_z$ in Algorithm~\ref{alg:silo-jding}'s measurement correction
by interpolating toward the denoiser's clean estimate. This appendix explains
that construction and derives its statistical effect on the first correction.

\subsection{Why interpolate the anchor?}
\label{sec:app-anchor-mechanism}

Early in sampling, the clean latent is inferred from a highly noisy state,
and its Gaussian belief is broad: $\eta_z=\eta_t/\alpha_t$ is large.
A fresh draw $\tilde{\vmu}_z=\vmu_z+\eta_z\bxi_z$ can therefore lie far
from the denoiser's clean estimate, where we linearize the learned operator.
The measurement correction must then account for a large displacement before
it can refine the reconstruction. This is demanding for the short CG solves
used in our sampler.

The interpolation in \eqref{eq:method-practical-anchor} reduces this displacement by
favoring the denoiser's estimate while the belief is broad. As the belief
narrows, it returns to the original perturbed-prior anchor. This trades some
of the ideal update's conditional variability for a correction that is easier
to compute with a small solver budget; the anchor ablation in
Appendix~\ref{sec:app-anchor-ablation} measures the reconstruction consequences.

\textbf{Why the distance can be large.}
The denoiser's estimate can be written as
$\rvz_0^{(0)}=\vmu_z+\eta_z\rvw'$; we derive this identity in the next
subsection. Here $\rvw'$ is the bridge noise already used to construct the
clean-latent belief. The original anchor
$\tilde{\vmu}_z=\vmu_z+\eta_z\bxi_z$ instead uses independent fresh
noise $\bxi_z$. Subtracting gives
\[
\tilde{\vmu}_z-\rvz_0^{(0)}=\eta_z(\bxi_z-\rvw').
\]
Each coordinate of these independent standard Gaussian noises has variance
$1$, so their difference has variance $1+1=2$. Summing over the $d$ latent
coordinates and averaging over both noise draws gives
\[
\mathbb{E}\!\left[\|\tilde{\vmu}_z-\rvz_0^{(0)}\|^2\right]
=\eta_z^2\sum_{i=1}^{d}\mathbb{E}\!\left[(\xi_{z,i}-w'_i)^2\right]
=2d\,\eta_z^2.
\]
The root-mean-square distance is therefore $\sqrt{2d}\,\eta_z$: the broader
the belief, the larger the gap. For illustration, the SD3.5 SR$\times8$
configuration uses a reverse-diffusion schedule with $100$ timesteps.
Across its anchor updates, $\eta_z$ falls from about $250$ to $0.03$.
These timesteps describe progress from noise toward the reconstructed image;
they are separate from the CG iterations used to solve each local correction.

\textbf{How the anchor enters the correction.}
Let $\rvz_a$ denote the replacement anchor: the value substituted for
$\tilde{\vmu}_z$ in Algorithm~\ref{alg:silo-jding}'s residual and proposal.
We write $\tilde{\vmu}_z^{\mathrm{new}}:=\rvz_a$, while reserving
$\tilde{\vmu}_z=\vmu_z+\eta_z\bxi_z$ for the original perturbed-prior
anchor. The algorithm keeps its general form; the anchor choice changes
these two occurrences of its input.

The learned operator $\rmH_\psi$ predicts a measurement from a clean latent.
To compute a correction, we approximate this operator near the denoiser's
estimate $\rvz_0^{(0)}$ by a linear model. Its Jacobian $\rmJ_0$ describes
how the predicted measurement changes when the latent moves. At the
replacement anchor, this model predicts
\[
\rmH_\psi(\rvz_a)\approx\rmH_\psi(\rvz_0^{(0)})
+\rmJ_0(\rvz_a-\rvz_0^{(0)}).
\]
Thus moving from the denoiser's estimate to the anchor also changes the
predicted measurement. Subtracting this prediction from the perturbed
measurement target $\tyz$ used in Algorithm~\ref{alg:silo-jding} gives
\[
\underbrace{\tyz-\rmH_\psi(\rvz_0^{(0)})}_{\text{mismatch at the denoiser's estimate}}
\; - \;
\underbrace{\rmJ_0(\rvz_a-\rvz_0^{(0)})}_{\text{predicted change from moving to the anchor}}.
\]
This is the right-hand side of the linear system in
\eqref{eq:local-perturb-and-solve}. The second term is the
\emph{extrapolation term}: it accounts for using an anchor away from the
point where the local model was built.

\textbf{Why the limited solve matters.}
Conjugate gradients (CG) approximately solves that linear system. We allow
at most three iterations per solve ($C\le3$), so the computed correction
need not satisfy the equation exactly. The relative CG residual measures
the remaining error in that equation, divided by the norm of its right-hand
side. Its mean value on SR$\times8$ is $0.69$, indicating that the solves
are far from converged. The resulting correction error can leave noise for
the next denoiser call to remove.

\textbf{Why interpolation helps.}
Moving the anchor toward the denoiser's estimate reduces their separation
and the associated change in the linear model's prediction. At the denoiser
endpoint, $\rvz_a=\rvz_0^{(0)}$, the extrapolation term above is zero.
The correction then starts with the measurement mismatch at the denoiser's
estimate alone. This is the computational motivation for favoring that
endpoint when the belief is broad.

\subsection{Constructing the anchor}

\textbf{Recovering the denoiser endpoint.}
We first derive the identity used above. Fix one diffusion transition and
omit its index. The bridge has mean $\rvm$ and noise scale $\eta_t$;
drawing standard Gaussian noise $\rvw'$ gives the proxy state
\[
\rvx^{\mathrm{pxy}}=\rvm+\eta_t\rvw',\qquad
\rvw'\sim\mathbf{N}(0,\rmI_d).
\]
Evaluating the denoiser at this proxy gives a clean estimate
$\rvz_0^{(0)}=\xzero{\rvx^{\mathrm{pxy}}}{t}$ and a noise estimate
$\hat{\rvx}_1'=\xone{\rvx^{\mathrm{pxy}}}{t}$. These are two ways of
expressing the same prediction: the clean and noise components recombine
into the input,
\[
\rvx^{\mathrm{pxy}}=\alpha_t\rvz_0^{(0)}+\sigma_t\hat{\rvx}_1'.
\]
Here $\alpha_t$ and $\sigma_t$ are the signal and noise coefficients of the
diffusion schedule. For the flow-matching parameterization,
$\alpha_t+\sigma_t=1$ and both estimates are affine in one velocity evaluation.
The clean-latent belief in \eqref{eq:clean-latent-prior} is defined by
\[
\vmu_z=\frac{\rvm-\sigma_t\hat{\rvx}_1'}{\alpha_t},\qquad
\eta_z=\frac{\eta_t}{\alpha_t}.
\]
Substituting the bridge sample into the recombination identity and dividing
by $\alpha_t$ therefore gives
\begin{equation}
    \boxed{\;\rvz_0^{(0)}=\vmu_z+\eta_z\rvw'.\;}
    \label{eq:app-anchor-identity}
\end{equation}
This explains the two endpoints: the denoiser's estimate uses the bridge
noise already involved in constructing the belief, while the original
perturbed-prior anchor uses a fresh, independent draw.

\begin{figure}[ht]
\centering
\begin{tikzpicture}[scale=2.5,>=stealth]
  \draw[->,black!45] (0,0) -- (1.32,0);
  \draw[->,black!45] (0,0) -- (0,1.32);
  \draw[very thick,blue!60!black] (1,0) arc[start angle=0,end angle=90,radius=1];
  \draw[very thick,orange!85!black,dashed] (1,0) -- (0,1);
  \draw[thin,orange!85!black,dotted] (0,0) -- (0.5,0.5);
  \draw[thin,blue!60!black,dotted]   (0,0) -- (0.866,0.5);
  \draw[thin,black!55,dotted] (0.5,0.5) -- (0.866,0.5);
  \fill (1,0) circle (1.1pt); \fill (0,1) circle (1.1pt);
  \fill[black!70] (1,1) circle (1.1pt);
  \fill[orange!85!black] (0.5,0.5) circle (1.1pt);
  \fill[blue!60!black]   (0.866,0.5) circle (1.1pt);
  \node[font=\scriptsize,below right] at (1,0) {$(1,0)$ denoiser est.};
  \node[font=\scriptsize,above right] at (0,1) {$(0,1)$ exact draw};
  \node[font=\scriptsize,right] at (1,1) {$(1,1)$ variance only};
  \draw[thin,black!55] (0.866,0.5) -- (1.12,0.60);
  \node[font=\scriptsize,right] at (1.12,0.60) {$\beta=\tfrac12$};
  \node[font=\scriptsize] at (0.26,0.49) {$0.71$};
  \node[font=\scriptsize,below] at (0.58,0.20) {$1.00$};
  \node[font=\scriptsize,black!60,below] at (0.66,-0.07) {$c_1$ (recycled, $\rvw'$)};
  \node[font=\scriptsize,black!60,rotate=90,above] at (-0.08,0.55)
        {$c_2$ (fresh, $\bxi_z$)};
\end{tikzpicture}
\caption{Anchor choices in coefficient space, corresponding to
Table~\ref{tab:anchor-ablation}. The horizontal axis weights recycled bridge
noise; the vertical axis weights fresh noise. The blue arc keeps
$c_1^2+c_2^2=1$, preserving the variance of the lifted prior-only proposal.
The orange line joins the same endpoints but reduces its standard deviation
to $0.71$ of the bridge value at $\beta=\tfrac12$. At a fixed $\beta$, both
choices use the same fresh-noise coefficient and differ in recycled noise.
The point $(1,1)$ is the added-noise (``variance only'') variant, with
standard deviation $\sqrt{2}$ times the bridge value. These are statements
about the prior-only proposal, not posterior exactness.}
\label{fig:anchor-family}
\end{figure}
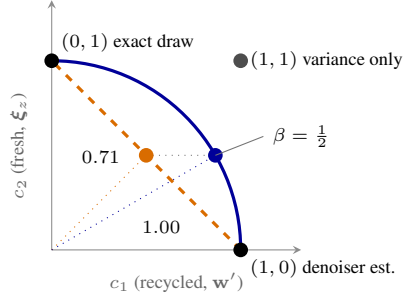

\textbf{Mixing the two noise contributions.}
The coefficient $c_1$ controls how much bridge noise we reuse, and $c_2$
controls how much fresh noise we add. The mixing parameter $\beta\in[0,1]$
sets the balance between the denoiser's estimate ($\beta=0$) and the original
perturbed-prior anchor ($\beta=1$), as in \eqref{eq:method-practical-anchor}.
At the first inner step, the anchor is
\[
\rvz_a=\vmu_z+\eta_z(c_1\rvw'+c_2\bxi_z),\qquad
(c_1,c_2)=\big(\sqrt{1-\beta^2},\beta\big).
\]
At $\beta=0$, only the bridge noise remains and we recover the denoiser's
estimate. At $\beta=1$, only fresh noise remains and we recover the original
perturbed-prior anchor. Between these endpoints, $c_1^2+c_2^2=1$: the
variance of the combined noise stays constant when averaging over both draws.
Figure~\ref{fig:anchor-family} contrasts this arc with the linear blend
$(c_1,c_2)=(1-\beta,\beta)$, whose combined noise has smaller variance
between the endpoints. The statistical analysis below explains what this
variance statement does and does not imply for the posterior update.

\textbf{Choosing the mixture along the trajectory.}
The existing diffusion noise schedule determines the belief widths
$\eta_{z,k}$. We use $\eta_{z,\min}:=\min_k\eta_{z,k}$ simply as notation
for their smallest value across the scheduled anchor updates; it is not a
new tunable parameter. At diffusion step $k$, we set
$\beta_k=\eta_{z,\min}/\eta_{z,k}$, as in \eqref{eq:method-practical-anchor}.
Thus $\beta_k$ and both coefficients
$c_{1,k}=\sqrt{1-\beta_k^2}$ and $c_{2,k}=\beta_k$ follow from that same
schedule, with no additional hyperparameter.
When the current width $\eta_{z,k}$ is large, $\beta_k$ is small and the
anchor stays near the denoiser's estimate. At the narrowest belief,
$\beta_k=1$ and it becomes the original perturbed-prior anchor.
This choice requires no extra denoiser evaluation. The Jacobian--vector
product for the extrapolation term is already part of the correction.

\subsection{Statistical effect of the first correction}

\textbf{Isolating the effect of the anchor.}
The preceding subsection explained why interpolation can help a short solve.
We now examine its statistical cost: how does changing the anchor affect
the mean and spread of the proposed reconstruction? To make this comparison,
we hold the local measurement model fixed and solve its linear system exactly.
We use the original operator-noise level ($\lambda=0$) and give all
measurement coordinates equal weight. These choices isolate the anchor's
effect from the solver approximations described in Section~\ref{sec:method}.
The calculation concerns the first inner correction ($j=0$) and its full
proposal, before relaxation takes a partial step toward that proposal.

\textbf{What is fixed, and what remains random?}
Imagine repeating this first proposal from the same incoming diffusion state
and the same bridge sample. The denoiser's estimates, the belief
$(\vmu_z,\eta_z)$, and the first local Jacobian $\rmJ_0$ then stay fixed.
Only the fresh prior and measurement perturbations, $\bxi_z$ and $\bxi_y$,
are redrawn as independent standard Gaussians, independent of the fixed
state and bridge noise.

In the formulas below, the incoming diffusion state is held fixed throughout.
We write $\mid\rvw'$ to indicate that the already defined bridge noise
$\rvw'$ is also held fixed. Thus these distributions, means, and covariances
average only over the two fresh perturbations. Where we instead average
over bridge samples, we state that change explicitly.

\textbf{The local correction and its reference distribution.}
For this first correction, write $\emH=\rmJ_0$. The local approximation to
the learned operator is $\emH\rvz_0+\rvb$, where the offset
$\rvb=\rmH_\psi(\rvz_0^{(0)})-\emH\rvz_0^{(0)}$ makes the approximation
agree with the operator at the denoiser's estimate. The measurement model is
\[
\yz=\emH\rvz_0+\rvb+\rvepsilon,\qquad
\rvepsilon\sim\mathbf{N}(0,r^2\rmI_d).
\]
Here $d$ is the shared latent dimension, and $r$ is the uncertainty
assigned to this operator model. As in Algorithm~\ref{alg:silo-jding}, we
perturb the measurement to $\tyz=\yz+r\bxi_y$ before computing a proposal.

Two matrices describe the correction. The matrix $\emS$ is the system
solved in measurement space; $\emK$ maps a measurement discrepancy back to
a correction in the clean latent:
\[
\emS=r^2\rmI_d+\eta_z^2\emH\emH^T,\qquad
\emK=\eta_z^2\emH^T\emS^{-1}.
\]
Starting from the anchor and correcting its predicted measurement gives
\begin{align*}
\rvz_0^\star&=\rvz_a+\emK(\tyz-\emH\rvz_a-\rvb)\\
&=(\rmI_d-\emK\emH)\rvz_a+\emK(\tyz-\rvb).
\end{align*}
We compare the distribution of this proposal with the exact posterior for
the original prior $\mathbf{N}(\vmu_z,\eta_z^2\rmI_d)$. Its mean and
covariance, derived in Appendix~\ref{sec:posterior-derivation}, are
\[
\rvm_p=(\rmI_d-\emK\emH)\vmu_z+\emK(\yz-\rvb),\qquad
\mathbf{\Sigma}_p=\eta_z^2(\rmI_d-\emK\emH).
\]
Matching these moments gives the correct Gaussian posterior in this ideal
local model; the next paragraphs show how interpolation changes them.

\textbf{Conditional distribution of the anchor.}
In the repeated-proposal experiment, the bridge contribution
$c_1\eta_z\rvw'$ is fixed and shifts the anchor's center. Only
$c_2\eta_z\bxi_z$ varies across draws. Calling the resulting center
$\mch$ and standard deviation $\ech$, we obtain
\begin{equation}
    \rvz_a\mid\rvw'\;\sim\;\mathbf{N}\big(\mch,\ech^2\rmI_d\big),
    \qquad \mch=\vmu_z+c_1\eta_z\rvw',\qquad \ech=c_2\eta_z.
    \label{eq:app-anchor-condlaw}
\end{equation}
Interpolation therefore shifts the conditional center and reduces the fresh
prior variance. At $\beta=1$ the original conditional prior is recovered.
For $\beta<1$, the unit-circle constraint alone does not provide the
conditional anchor law required by Appendix~\ref{sec:app-perturb-sampling}.

\textbf{Preserving bridge variance does not imply posterior exactness.}
The preceding distribution holds the bridge sample fixed. We now consider
a different check: with the measurement removed ($\emK=0$), do the anchors
reproduce the original diffusion bridge when we average over bridge samples?
The proposal is then the anchor itself. Converting it back to diffusion
coordinates, using the same noise estimate as in the construction, gives
\[
\alpha_t\rvz_a+\sigma_t\hat{\rvx}_1'
=\rvm+\eta_t(c_1\rvw'+c_2\bxi_z).
\]
Here $\alpha_t\vmu_z+\sigma_t\hat{\rvx}_1'=\rvm$ and
$\alpha_t\eta_z=\eta_t$. Holding the incoming diffusion state fixed but
averaging over both noises gives
$\mathbf{N}(\rvm,(c_1^2+c_2^2)\eta_t^2\rmI_d)$, exactly the bridge on the
unit circle. The linear blend instead reduces its standard deviation by
$\sqrt{(1-\beta)^2+\beta^2}$, which is $0.707$ at $\beta=\tfrac12$.
This statement averages over the bridge noise; the posterior analysis above
holds that noise and the resulting local model fixed. Preserving the first
distribution therefore does not guarantee the second. The bridge statement
also applies only before relaxation or repeated corrections.

\textbf{Conditional mean and covariance.}
Substitute the anchor distribution into the local correction. The fixed
bridge contribution shifts the proposal's mean. The fresh prior noise is
transformed by $(\rmI_d-\emK\emH)$, while the measurement noise is
transformed by $\emK$; because these noises are independent, their
covariances add. This gives
\begin{align}
    \mathbb{E}[\rvz_0^\star\mid\rvw']
        &=\rvm_p+c_1\eta_z(\rmI_d-\emK\emH)\rvw',
        \label{eq:app-anchor-bias}\\
    \operatorname{Cov}[\rvz_0^\star\mid\rvw']
        &=c_2^2\eta_z^2(\rmI_d-\emK\emH)^2+r^2\emK\emK^T
         =\mathbf{\Sigma}_p\big[c_2^2(\rmI_d-\emK\emH)+\emK\emH\big].
        \label{eq:app-anchor-cov}
\end{align}
The first covariance expression displays the two noise contributions.
The second rewrites their sum in terms of the reference posterior covariance
$\mathbf{\Sigma}_p$, making the variance comparison below possible.
To obtain it, use $r^2\emK=\eta_z^2(\rmI_d-\emK\emH)\emH^T$, which
follows from $(\rmI_d-\emK\emH)\emH^T=\emH^T\emS^{-1}(\emS-\eta_z^2\emH\emH^T)
=r^2\emH^T\emS^{-1}$, together with the symmetry of $\emK\emH$.

\textbf{Interpretation by measurement direction.}
Some changes in the latent strongly affect the measurement; others barely
change it. A singular value decomposition $\emH=\emU\bm{\Gamma}\emV^T$
separates these directions: the columns of $\emV$ are latent directions,
and each singular value $\gamma_i$ measures the operator's sensitivity
along one of them. Define
\[
g_i=\frac{\eta_z^2\gamma_i^2}{r^2+\eta_z^2\gamma_i^2}.
\]
This number measures the influence of the measurement in that direction:
$g_i=0$ for an invisible direction, while $g_i$ approaches $1$ when the
measurement strongly constrains it relative to the prior. In this basis,
$\emK\emH=\emV\,\mathrm{diag}(g_i)\emV^T$ and
$\mathbf{\Sigma}_p=\eta_z^2\emV\,\mathrm{diag}(1-g_i)\emV^T$.
Substituting into \eqref{eq:app-anchor-cov} and dividing by the exact
posterior variance in the same direction gives the retained fraction
\begin{equation}
    \boxed{\;g_i+c_2^2(1-g_i)\;}
    \label{eq:app-anchor-fraction}
\end{equation}
of the exact posterior variance. At $\beta=1$ the mean shift and variance
deficit vanish, recovering \eqref{eq:app-perturb-result}. At $\beta=0$,
the proposal becomes
\[
\rvz_0^{(0)}+\emK(\tyz-\emH\rvz_0^{(0)}-\rvb),
\]
a damped Gauss--Newton correction on the data term, related to the updates
used by $\Pi$GDM~\citep{song2023pigdm} and DDS. The fresh prior perturbation
is absent. Its retained variance fraction is $g_i$, which is zero in
directions that the measurement operator does not see.

\textbf{Comparing the arc and the linear blend.}
Both use $c_2=\beta$, so they retain the same conditional variance. They
differ in $c_1$, which controls the mean shift and the displacement from
the initial denoiser estimate. Subtracting that estimate from the anchor gives
\[
\rvz_a-\rvz_0^{(0)}=\eta_z\big[(c_1-1)\rvw'+c_2\bxi_z\big].
\]
Before fixing the bridge draw, the two independent noises contribute
variances $(c_1-1)^2$ and $c_2^2$ per coordinate. The displacement therefore
has standard deviation $\eta_z\sqrt{(1-c_1)^2+c_2^2}$ per coordinate.
The two choices can consequently give different extrapolation terms even
though they add the same amount of fresh noise.

\subsection{Repeated inner corrections}

Within one diffusion transition, the sampler can apply several local
corrections. Let $j$ count these inner corrections and $\rvz_0^{(j)}$
be the current estimate. After each correction, the next local model is
built at the updated estimate, and the anchor is recomputed using that
estimate in place of the initial denoiser output. The same original
perturbed-prior draw is reused throughout. Thus, for the arc rule, the
replacement $\tilde{\vmu}_z^{\mathrm{new},(j)}:=\rvz_a^{(j)}$ is
\begin{equation}
    \rvz_a^{(j)}=\vmu_z+c_1(\rvz_0^{(j)}-\vmu_z)
        +c_2(\tilde{\vmu}_z-\vmu_z),
    \qquad (c_1,c_2)=(\sqrt{1-\beta^2},\beta).
    \label{eq:app-anchor-inner}
\end{equation}
The quantities $\vmu_z$, $\tilde{\vmu}_z$, $\tyz$, $\eta_z$ and $\beta$
stay fixed within the transition, while $\rvz_0^{(j)}$ and its local
linearization change. At $j=0$, \eqref{eq:app-anchor-identity} reduces
\eqref{eq:app-anchor-inner} to \eqref{eq:method-practical-anchor}.

\textbf{Why the first-step law cannot be reused.}
The first-step calculation treated the denoiser's estimate as fixed and
independent of the two fresh perturbations. After a correction, the current
estimate already contains effects of those same perturbations,
$\bxi_z$ and $\bxi_y$. Reusing them in the next anchor therefore creates
dependence between its two contributions. Its displacement from $\vmu_z$
is no longer the original bridge noise $\eta_z\rvw'$. Consequently,
\eqref{eq:app-anchor-condlaw}--\eqref{eq:app-anchor-fraction} cannot be
reapplied as conditional laws of each later correction, and do not establish
the distribution of the final $P$-step output.

\textbf{Endpoint behavior.}
At $\beta=0$ the anchor is the current iterate; at $\beta=1$ it is the
fixed perturbed prior mean $\tilde{\vmu}_z$ on every inner step.
For a fixed affine operator $\rmH_\psi(\rvz)=\emH\rvz+\rvb$, the latter
choice gives the same residual $\tyz-\emH\tilde{\vmu}_z-\rvb$ at every
linearization. With $\lambda=0$, $\rho=1$, the unweighted likelihood, and
exact solves, the first correction is the affine posterior draw and every
later correction returns that same draw: $P=1$ suffices but is not necessary.
For nonlinear operators this fixed-anchor case has the Gauss--Newton
interpretation of Appendix~\ref{sec:app-relinearized-gn}. For $0\le\beta<1$,
the changing anchors do not retain that single fixed-target objective, and
we make no exact posterior-sampling claim for the repeated practical updates.

\noindent\begin{minipage}{\linewidth}
\subsection{Technical remark: a different prior interpretation}

The interpolated anchor has conditional center $\mch$ and standard
deviation $\ech=c_2\eta_z$. One alternative would treat these as a
new prior and sample its posterior exactly. To do this in the ideal local
model, the linear solve must use the same prior variance as the random
anchor.

Specifically, if the declared prior were $\mathbf{N}(\mch,\ech^2\rmI_d)$,
we would use
\[
\widehat{\emS}=r^2\rmI_d+\ech^2\emH\emH^T,\qquad
\widehat{\emK}=\ech^2\emH^T\widehat{\emS}^{-1}
\]
along with the anchor drawn from that prior. The noise in the anchor and
the correction matrices then share the same scale, satisfying
$\widehat{\emK}\widehat{\emS}=\ech^2\emH^T$.
Appendix~\ref{sec:app-perturb-sampling} gives exact sampling for this
different prior under the same ideal assumptions. At $\ech=0$, all prior
mass is at one point and the posterior stays at that point.

The implemented variant instead uses the original width $\eta_z$ in its
solve. Its first-step distribution is therefore the one derived in
\eqref{eq:app-anchor-bias}--\eqref{eq:app-anchor-cov}, without the above
guarantee for the alternative prior. With our schedule,
$\ech=\beta_k\eta_{z,k}=\eta_{z,\min}$, so this alternative would use
one constant prior width throughout the trajectory. We have not evaluated it.

\end{minipage}

\clearpage
\section{Connections to existing methods}
\label{sec:method-connections}

\subsection{Exact special cases of the measurement step}
\label{sec:app-unification}

This section relates the measurement step of Section~\ref{sec:method}
(Step~3; the ideal measurement update underlying
Algorithm~\ref{alg:silo-jding}) to four published
samplers. The step is a linear--Gaussian posterior update taken from an
anchor in the clean-latent space, and it is determined by three choices: the
anchor from which the update is taken, the ratio $\eta_z^2/r^2$ that weights
the measurement against the prior, and the solver that applies the resulting
gain. Fixing these three choices recovers the measurement steps of
DDS~\citep{chung2024dds}, of DiffPIR~\citep{zhu2023diffpir} for a linear
operator, of DDNM~\citep{wang2023ddnm} with the Moore--Penrose inverse, and of
DING~\citep{moufad2026ding} in distribution. Each subsection states the
published update and reduces it to ours in four steps. Exactness is
understood in exact arithmetic under the stated idealization: the reductions
identify the point the step converges to, not the iterate a truncated solver
returns. The reductions concern the measurement step only: recovering a
method's step does not recover its outer sampling loop. Two further methods
that match the step only to first order are treated in
Appendix~\ref{sec:app-unification-approx}.

\paragraph{The \paseo\ measurement step.}
Consider one sampling step. The clean-latent prior is
$\mathbf{N}(\vmu_z,\eta_z^2\rmI_d)$, the measurement model is
$\yz=\emH\rvz_0+\rvepsilon$ with $\rvepsilon\sim\mathbf{N}(0,r^2\rmI_d)$, and
$\emS=r^2\rmI_d+\eta_z^2\emH\emH^T$ and $\emK=\eta_z^2\emH^T\emS^{-1}$ are
as in Section~\ref{sec:method}. Starting from an anchor
$\rvz_a$, the measurement step is
\begin{equation}
    \rvz_0^\star=\rvz_a+\emK\big(\tyz-\emH\rvz_a\big),
    \qquad \tyz=\yz+r\bxi_y ,
    \label{eq:unif-update}
\end{equation}
where $\tyz$ is the measurement perturbed by its own noise. The anchor is one
of two points. It is either the denoised estimate $\rvz_0^{(0)}$, or the
prior mean perturbed by its own noise, $\vmu_z+\eta_z\bxi_z$. With the second
anchor, \eqref{eq:unif-update} is an exact draw from the posterior
when the solve is exact
(Appendix~\ref{sec:app-perturb-sampling}); with the first, it is a
correction of the denoised estimate, deterministic when the
measurement is not perturbed.

DDS, DiffPIR and DDNM are compared in the simplest setting: a linear operator
$\emH$, a single iteration, no mask or damping, the anchor at the denoised
estimate, and no measurement perturbation. DING keeps the second anchor and
the perturbation, and is compared with \eqref{eq:unif-update} directly. Under
the simplest setting \eqref{eq:unif-update} becomes
\begin{equation}
    \boxed{\;
    \rvz_0^\star=\rvz_0^{(0)}
    +\eta_z^2\emH^T\big(r^2\rmI_d+\eta_z^2\emH\emH^T\big)^{-1}
    \big(\yz-\emH\rvz_0^{(0)}\big),
    \;}
    \label{eq:unif-canonical}
\end{equation}
which is the minimiser of
\begin{equation}
    \frac{1}{2r^2}\big\|\yz-\emH\rvz\big\|^2
    +\frac{1}{2\eta_z^2}\big\|\rvz-\rvz_0^{(0)}\big\|^2
    \label{eq:unif-objective}
\end{equation}
(Appendix~\ref{sec:app-gain-form}). Scaling \eqref{eq:unif-objective} by
$\eta_z^2$ does not change its minimiser and gives
\begin{equation}
    \frac{\tau}{2}\big\|\yz-\emH\rvz\big\|^2
    +\frac{1}{2}\big\|\rvz-\rvz_0^{(0)}\big\|^2,
    \qquad \tau=\frac{\eta_z^2}{r^2},
    \label{eq:unif-objective-kappa}
\end{equation}
so only the ratio $\tau$ matters: it sets how far the measurement is
allowed to move the denoised estimate. The four samplers below differ in this
ratio, in the anchor, and in how the linear system in
\eqref{eq:unif-canonical} is solved.

\subsubsection{DDS: the same objective, solved in the other space}
\label{sec:unif-dds}

DDS~\citep{chung2024dds} conditions each sampling step by minimizing their
Eq.~(26), written here with $\emH$ for their operator $\bm{A}$ and $\emH^T$
for its adjoint $\bm{A}^{*}$,
\begin{equation}
    \ell(\rvx)=\frac{\gamma}{2}\big\|\rvy-\emH\rvx\big\|_2^2
    +\frac12\big\|\rvx-\hat{\rvx}_t\big\|_2^2,
    \label{eq:unif-dds}
\end{equation}
with $M$ conjugate-gradient iterations on the normal equations
$(\gamma\emH^T\emH+\rmI_d)\,\rvx=\hat{\rvx}_t+\gamma\emH^T\rvy$, started
at the Tweedie estimate $\hat{\rvx}_t$.

\paragraph{Reduction to the \paseo\ step.}
The objective \eqref{eq:unif-dds} is strictly convex, so its minimiser
$\rvx^\star$ is the unique zero of its gradient. From that condition,
\eqref{eq:unif-canonical} follows in four steps:
\begin{align}
    0&=\nabla\ell(\rvx^\star)
    =-\gamma\emH^T\big(\rvy-\emH\rvx^\star\big)+\big(\rvx^\star-\hat{\rvx}_t\big),
    \label{eq:unif-dds-normal}\\
    \rvx^\star&\overset{\mathrm{(i)}}{=}\hat{\rvx}_t
    +\big(\rmI_d+\gamma\emH^T\emH\big)^{-1}\gamma\emH^T\big(\rvy-\emH\hat{\rvx}_t\big)
    \label{eq:unif-dds-solution}\\
    &\overset{\mathrm{(ii)}}{=}\hat{\rvx}_t
    +\gamma\emH^T\big(\rmI_d+\gamma\emH\emH^T\big)^{-1}\big(\rvy-\emH\hat{\rvx}_t\big)
    \label{eq:unif-dds-gain}\\
    &\overset{\mathrm{(iii)}}{=}\hat{\rvx}_t
    +\eta_z^2\emH^T\big(r^2\rmI_d+\eta_z^2\emH\emH^T\big)^{-1}\big(\rvy-\emH\hat{\rvx}_t\big)
    \label{eq:unif-dds-kappa}\\
    &\overset{\mathrm{(iv)}}{=}\rvz_0^{(0)}
    +\eta_z^2\emH^T\big(r^2\rmI_d+\eta_z^2\emH\emH^T\big)^{-1}\big(\yz-\emH\rvz_0^{(0)}\big).
    \label{eq:unif-dds-final}
\end{align}
The last line is \eqref{eq:unif-canonical}. The steps are the following.
\begin{itemize}
\item[(i)] Rearranging \eqref{eq:unif-dds-normal} gives
    $(\rmI_d+\gamma\emH^T\emH)\rvx^\star=\hat{\rvx}_t+\gamma\emH^T\rvy$, the
    system DDS solves by CG. Subtracting $(\rmI_d+\gamma\emH^T\emH)\hat{\rvx}_t$
    from both sides writes it for the correction,
    \begin{equation*}
        \big(\rmI_d+\gamma\emH^T\emH\big)\big(\rvx^\star-\hat{\rvx}_t\big)
        =\gamma\emH^T\big(\rvy-\emH\hat{\rvx}_t\big).
    \end{equation*}
\item[(ii)] The inverse moves from latent space to measurement space by the
    push-through identity
    \begin{equation*}
        \big(\rmI_d+\gamma\emH^T\emH\big)^{-1}\gamma\emH^T
        =\gamma\emH^T\big(\rmI_d+\gamma\emH\emH^T\big)^{-1},
    \end{equation*}
    obtained from $\emH^T(\rmI_d+\gamma\emH\emH^T)=(\rmI_d+\gamma\emH^T\emH)\emH^T$
    by multiplying on the left by $(\rmI_d+\gamma\emH^T\emH)^{-1}$ and on the
    right by $(\rmI_d+\gamma\emH\emH^T)^{-1}$.
\item[(iii)] Substituting $\gamma=\eta_z^2/r^2$ and multiplying the bracket
    by $r^2/r^2$,
    \begin{equation*}
        \gamma\emH^T\big(\rmI_d+\gamma\emH\emH^T\big)^{-1}
        =\frac{\eta_z^2}{r^2}\emH^T\Big(\frac{r^2\rmI_d+\eta_z^2\emH\emH^T}{r^2}\Big)^{-1}
        =\eta_z^2\emH^T\big(r^2\rmI_d+\eta_z^2\emH\emH^T\big)^{-1}.
    \end{equation*}
\item[(iv)] $\hat{\rvx}_t$ is read as $\rvz_0^{(0)}$ and $\rvy$ as $\yz$,
    both being the denoiser's clean estimate and the measurement in latent
    coordinates.
\end{itemize}
Their noiseless variant (Eqs.~22 and~24, CG on $\emH^T\emH$ with
right-hand side $\emH^T\rvy$) is the limit $\gamma\to\infty$, that is
$r\to0$. The one difference is where the solve runs: DDS applies CG to the
latent-space system in (i); we apply it to the measurement-space system
$\emS\rvv=\yz-\emH\rvz_0^{(0)}$ of Appendix~\ref{sec:app-cg-solve}, and by
(ii) the two agree at convergence.

\subsubsection{DiffPIR: the same objective, solved in closed form}
\label{sec:unif-diffpir}

DiffPIR~\citep{zhu2023diffpir} conditions each sampling step through the
data subproblem of its half-quadratic splitting, their Eq.~(12b), written
here with $\emH$ for their operator $\mathcal{H}$,
\begin{equation}
    \hat{\rvx}_0^{(t)}=\arg\min_{\rvx}\;
    \big\|\rvy-\emH\rvx\big\|^2+\rho_t\big\|\rvx-\rvx_0^{(t)}\big\|^2,
    \qquad
    \rho_t=\lambda\Big(\frac{\sigma_n}{\bar\sigma_t}\Big)^{2},
    \label{eq:unif-diffpir}
\end{equation}
anchored at the denoised estimate $\rvx_0^{(t)}$ and solved in closed form,
by an FFT for deblurring and elementwise for inpainting.

\paragraph{Reduction to the \paseo\ step.}
The objective \eqref{eq:unif-diffpir} is strictly convex, so its minimiser
is the unique zero of its gradient. From that condition,
\eqref{eq:unif-canonical} follows in four steps:
\begin{align}
    0&=-2\emH^T\big(\rvy-\emH\hat{\rvx}_0^{(t)}\big)+2\rho_t\big(\hat{\rvx}_0^{(t)}-\rvx_0^{(t)}\big),
    \label{eq:unif-diffpir-normal}\\
    \hat{\rvx}_0^{(t)}&\overset{\mathrm{(i)}}{=}\rvx_0^{(t)}
    +\big(\emH^T\emH+\rho_t\rmI_d\big)^{-1}\emH^T\big(\rvy-\emH\rvx_0^{(t)}\big)
    \label{eq:unif-diffpir-solution}\\
    &\overset{\mathrm{(ii)}}{=}\rvx_0^{(t)}
    +\emH^T\big(\emH\emH^T+\rho_t\rmI_d\big)^{-1}\big(\rvy-\emH\rvx_0^{(t)}\big)
    \label{eq:unif-diffpir-gain}\\
    &\overset{\mathrm{(iii)}}{=}\rvx_0^{(t)}
    +\eta_z^2\emH^T\big(r^2\rmI_d+\eta_z^2\emH\emH^T\big)^{-1}\big(\rvy-\emH\rvx_0^{(t)}\big)
    \label{eq:unif-diffpir-rho}\\
    &\overset{\mathrm{(iv)}}{=}\rvz_0^{(0)}
    +\eta_z^2\emH^T\big(r^2\rmI_d+\eta_z^2\emH\emH^T\big)^{-1}\big(\yz-\emH\rvz_0^{(0)}\big).
    \label{eq:unif-diffpir-final}
\end{align}
The last line is \eqref{eq:unif-canonical}. The steps are the following.
\begin{itemize}
\item[(i)] Dividing \eqref{eq:unif-diffpir-normal} by two and rearranging
    gives $(\emH^T\emH+\rho_t\rmI_d)\hat{\rvx}_0^{(t)}=\emH^T\rvy+\rho_t\rvx_0^{(t)}$,
    the system DiffPIR solves in closed form. Subtracting
    $(\emH^T\emH+\rho_t\rmI_d)\rvx_0^{(t)}$ from both sides writes it for the
    correction,
    \begin{equation*}
        \big(\emH^T\emH+\rho_t\rmI_d\big)\big(\hat{\rvx}_0^{(t)}-\rvx_0^{(t)}\big)
        =\emH^T\big(\rvy-\emH\rvx_0^{(t)}\big).
    \end{equation*}
\item[(ii)] The inverse moves from latent space to measurement space by the
    push-through identity
    \begin{equation*}
        \big(\emH^T\emH+\rho_t\rmI_d\big)^{-1}\emH^T
        =\emH^T\big(\emH\emH^T+\rho_t\rmI_d\big)^{-1},
    \end{equation*}
    obtained from $\emH^T(\emH\emH^T+\rho_t\rmI_d)=(\emH^T\emH+\rho_t\rmI_d)\emH^T$
    by multiplying on the left by $(\emH^T\emH+\rho_t\rmI_d)^{-1}$ and on the
    right by $(\emH\emH^T+\rho_t\rmI_d)^{-1}$.
\item[(iii)] Substituting $\rho_t=r^2/\eta_z^2$ and multiplying the bracket
    by $\eta_z^2/\eta_z^2$,
    \begin{equation*}
        \emH^T\big(\emH\emH^T+\rho_t\rmI_d\big)^{-1}
        =\emH^T\Big(\frac{\eta_z^2\emH\emH^T+r^2\rmI_d}{\eta_z^2}\Big)^{-1}
        =\eta_z^2\emH^T\big(r^2\rmI_d+\eta_z^2\emH\emH^T\big)^{-1}.
    \end{equation*}
\item[(iv)] $\rvx_0^{(t)}$ is read as $\rvz_0^{(0)}$ and $\rvy$ as $\yz$,
    both being the denoiser's clean estimate and the measurement in latent
    coordinates.
\end{itemize}
Equivalently, \eqref{eq:unif-diffpir} is \eqref{eq:unif-objective} scaled by
$2r^2$, so $\rho_t$ is $1/\tau$. The idealization is a linear operator whose
structure admits the closed-form solve, which is where the two differ: DiffPIR
inverts $\emH^T\emH+\rho_t\rmI_d$ exactly, by an FFT when $\emH$ is a
circular convolution and elementwise when it is a mask, whereas we apply CG to
$\emS\rvv=\yz-\emH\rvz_0^{(0)}$ for a general $\emH$; by (ii) the two agree
at convergence. Where no closed form is available, as for their bicubic
super-resolution, DiffPIR replaces the solve by iterative back-projection,
their Eq.~(30), a gradient iteration on \eqref{eq:unif-diffpir} with a tuned
step size, which is not this reduction. The per-step
objectives match, but the schedules do not: DiffPIR sets $\rho_t$ from
$\bar\sigma_t$ and the sensor noise $\sigma_n$, whereas $r^2/\eta_z^2$ follows
the bridge width and the operator's trust level.

\subsubsection{DDNM: the noiseless limit}
\label{sec:unif-ddnm}

DDNM~\citep{wang2023ddnm} conditions each sampling step by a
range--null-space decomposition, their Eq.~(13), written here with $\emH$
for their operator $\mathbf{A}$ and $\emH^{+}$ for its pseudo-inverse
$\mathbf{A}^{\dagger}$,
\begin{equation}
    \hat{\rvx}_{0|t}=\emH^{+}\rvy+\big(\rmI_d-\emH^{+}\emH\big)\rvx_{0|t},
    \label{eq:unif-ddnm}
\end{equation}
anchored at the denoised estimate $\rvx_{0|t}$: the component of the signal
the operator observes is taken from the measurement, the rest from the
denoiser.

\paragraph{Reduction to the \paseo\ step.}
Write $\emH=\rmU\mathbf{\Sigma}\rmV^T$ for the singular value decomposition,
with singular values $s_i\ge0$. The update \eqref{eq:unif-ddnm} is
\eqref{eq:unif-canonical} in the limit $r\to0$, in four steps:
\begin{align}
    \hat{\rvx}_{0|t}
    &\overset{\mathrm{(i)}}{=}\rvx_{0|t}+\emH^{+}\big(\rvy-\emH\rvx_{0|t}\big)
    \label{eq:unif-ddnm-correction}\\
    &\overset{\mathrm{(ii)}}{=}\rvx_{0|t}
    +\rmV\operatorname{diag}\big(s_i^{+}\big)\rmU^T\big(\rvy-\emH\rvx_{0|t}\big)
    \label{eq:unif-ddnm-svd}\\
    &\overset{\mathrm{(iii)}}{=}\rvx_{0|t}
    +\lim_{r\to0}\eta_z^2\emH^T\big(r^2\rmI_d+\eta_z^2\emH\emH^T\big)^{-1}\big(\rvy-\emH\rvx_{0|t}\big)
    \label{eq:unif-ddnm-limit}\\
    &\overset{\mathrm{(iv)}}{=}\lim_{r\to0}\Big[\rvz_0^{(0)}
    +\eta_z^2\emH^T\big(r^2\rmI_d+\eta_z^2\emH\emH^T\big)^{-1}\big(\yz-\emH\rvz_0^{(0)}\big)\Big].
    \label{eq:unif-ddnm-final}
\end{align}
The bracket in the last line is \eqref{eq:unif-canonical}. The steps are the
following.
\begin{itemize}
\item[(i)] Expanding $\rvx_{0|t}+\emH^{+}(\rvy-\emH\rvx_{0|t})
    =\emH^{+}\rvy+(\rmI_d-\emH^{+}\emH)\rvx_{0|t}$ recovers
    \eqref{eq:unif-ddnm}: the update is the denoised estimate plus the
    pseudo-inverse applied to its residual.
\item[(ii)] The Moore--Penrose inverse is
    $\emH^{+}=\rmV\operatorname{diag}(s_i^{+})\rmU^T$ with $s_i^{+}=1/s_i$
    for $s_i>0$ and $s_i^{+}=0$ for $s_i=0$.
\item[(iii)] In the same basis, $\emH\emH^T=\rmU\operatorname{diag}(s_i^2)\rmU^T$,
    so the gain of \eqref{eq:unif-canonical} is diagonal,
    \begin{equation*}
        \eta_z^2\emH^T\big(r^2\rmI_d+\eta_z^2\emH\emH^T\big)^{-1}
        =\rmV\operatorname{diag}\Big(\frac{\eta_z^2s_i}{r^2+\eta_z^2s_i^2}\Big)\rmU^T .
    \end{equation*}
    As $r\to0$, each entry with $s_i>0$ tends to $\eta_z^2s_i/(\eta_z^2s_i^2)=1/s_i$,
    and each entry with $s_i=0$ is zero for every $r$. The limit is therefore
    $\rmV\operatorname{diag}(s_i^{+})\rmU^T=\emH^{+}$.
\item[(iv)] $\rvx_{0|t}$ is read as $\rvz_0^{(0)}$ and $\rvy$ as $\yz$,
    both being the denoiser's clean estimate and the measurement in latent
    coordinates.
\end{itemize}
The limit $r\to0$ is $\tau\to\infty$: the measurement is believed
absolutely, and the gain times $\emH$ becomes the projection onto the range of $\emH^T$.
The idealization is that DDNM's $\mathbf{A}^{\dagger}$ is the Moore--Penrose
inverse. The paper defines the pseudo-inverse only through
$\mathbf{A}\mathbf{A}^{\dagger}\mathbf{A}=\mathbf{A}$ and composes it by hand
for chained degradations, $\mathbf{A}^{\dagger}=\mathbf{A}_n^{\dagger}\cdots\mathbf{A}_1^{\dagger}$;
the reduction holds when their choice coincides with the Moore--Penrose
inverse, as it does for their mask, colorization and average-pooling
operators. The other difference is
the solver: DDNM applies $\emH^{+}$ in closed form, which requires the
singular vectors of $\emH$; we apply CG to $\emS\rvv=\yz-\emH\rvz_0^{(0)}$,
which requires only products with $\emH$ and $\emH^T$, at a finite $r$.

\subsubsection{DING: the same draw, for a coordinate mask}
\label{sec:unif-ding}

DING~\citep{moufad2026ding} conditions each sampling step by lines 10--12 of
its Algorithm~1, written in state coordinates. With $\bar{\mathbf{m}}$ the
observed and $\mathbf{m}$ the missing coordinates, $\rvm$ the bridge mean
(their $\bm{\mu}$), $\eta_t$ the bridge width, $\hat{\rvx}_1'$ the proxy
noise estimate (their $\hat{\rvx}_1^{\mathrm{pxy}}$), $\sigma_{\rvy}$ the
sensor noise and $\rvw'\sim\mathbf{N}(0,\rmI_d)$ fresh,
\begin{equation}
\begin{aligned}
    \gamma&=\frac{\eta_t^2}{\eta_t^2+\alpha_t^2\sigma_{\rvy}^2},\\
    \rvx[\mathbf{m}]&=\rvm[\mathbf{m}]+\eta_t\rvw'[\mathbf{m}],\\
    \rvx[\bar{\mathbf{m}}]&=(1-\gamma)\rvm[\bar{\mathbf{m}}]
    +\gamma\big(\alpha_t\rvy+\sigma_t\hat{\rvx}_1'[\bar{\mathbf{m}}]\big)
    +\alpha_t\sigma_{\rvy}\sqrt{\gamma}\,\rvw'[\bar{\mathbf{m}}].
\end{aligned}
\label{eq:unif-ding}
\end{equation}
Missing coordinates take the bridge sample; observed ones take a weighted
average of the bridge mean and the measurement, plus noise. This is the one
case anchored at the perturbed prior mean, so the target is
\eqref{eq:unif-update} itself rather than \eqref{eq:unif-canonical}: the
anchor is $\tilde{\vmu}_z=\vmu_z+\eta_z\bxi_z$, the measurement is
$\tyz=\yz+r\bxi_y$, and $r=\sigma_{\rvy}$. For this coordinate-selection
case, let $q$ be the number of observed coordinates and take the measurement
vectors in $\mathbb{R}^q$. The operator is the row selection
$\rmM\in\mathbb{R}^{q\times d}$ of the observed coordinates, so that
$\rmM\rmM^T=\rmI_q$.

\paragraph{Reduction to the \paseo\ step.}
The clean-latent coordinate is $\rvx=\alpha_t\rvz+\sigma_t\hat{\rvx}_1'$, with
$\vmu_z=(\rvm-\sigma_t\hat{\rvx}_1')/\alpha_t$ and $\eta_z=\eta_t/\alpha_t$
as in \eqref{eq:clean-latent-prior}; in that coordinate
$\gamma=\eta_z^2/(\eta_z^2+\sigma_{\rvy}^2)$. The update \eqref{eq:unif-ding}
is \eqref{eq:unif-update} in distribution, in four steps, of which (ii) is
the distributional one:
\begin{align}
    \rvz_0^\star[\mathbf{m}]
    &\overset{\mathrm{(i)}}{=}\vmu_z[\mathbf{m}]+\eta_z\rvw'[\mathbf{m}],
    \label{eq:unif-ding-clean-m}\\
    \rvz_0^\star[\bar{\mathbf{m}}]
    &\overset{\mathrm{(i)}}{=}(1-\gamma)\vmu_z[\bar{\mathbf{m}}]
    +\gamma\rvy+\sigma_{\rvy}\sqrt{\gamma}\,\rvw'[\bar{\mathbf{m}}],
    \label{eq:unif-ding-clean}\\
    \rvz_0^\star[\mathbf{m}]
    &\overset{\mathrm{(ii)}}{=}\vmu_z[\mathbf{m}]+\eta_z\bxi_z[\mathbf{m}],
    \label{eq:unif-ding-twonoise-m}\\
    \rvz_0^\star[\bar{\mathbf{m}}]
    &\overset{\mathrm{(ii)}}{=}(1-\gamma)\big(\vmu_z[\bar{\mathbf{m}}]+\eta_z\bxi_z[\bar{\mathbf{m}}]\big)
    +\gamma\big(\rvy+\sigma_{\rvy}\bxi_y\big),
    \label{eq:unif-ding-twonoise}\\
    \rvz_0^\star
    &\overset{\mathrm{(iii)}}{=}\tilde{\vmu}_z+\gamma\rmM^T\big(\tyz-\rmM\tilde{\vmu}_z\big)
    \label{eq:unif-ding-assembled}\\
    &\overset{\mathrm{(iv)}}{=}\tilde{\vmu}_z
    +\eta_z^2\rmM^T\big(\sigma_{\rvy}^2\rmI_q+\eta_z^2\rmM\rmM^T\big)^{-1}\big(\tyz-\rmM\tilde{\vmu}_z\big).
    \label{eq:unif-ding-final}
\end{align}
The last line is \eqref{eq:unif-update} with $\emH=\rmM$ and
$r=\sigma_{\rvy}$. The steps are the following.
\begin{itemize}
\item[(i)] Subtracting $\sigma_t\hat{\rvx}_1'$ from each line of
    \eqref{eq:unif-ding} and dividing by $\alpha_t$ moves the update to the
    clean coordinate: $(1-\gamma)\rvm+\gamma\sigma_t\hat{\rvx}_1'-\sigma_t\hat{\rvx}_1'
    =(1-\gamma)(\rvm-\sigma_t\hat{\rvx}_1')$ on the observed block, and
    $\gamma$ is unchanged by dividing numerator and denominator by
    $\alpha_t^2$.
\item[(ii)] Both sides are Gaussian with the same mean and independent
    blocks, so equality in distribution needs only the variances to agree.
    On the missing block both are $\eta_z^2$. On the observed block,
    with $1-\gamma=\sigma_{\rvy}^2/(\eta_z^2+\sigma_{\rvy}^2)$,
    \begin{equation*}
        (1-\gamma)^2\eta_z^2+\gamma^2\sigma_{\rvy}^2
        =\frac{\sigma_{\rvy}^4\eta_z^2+\eta_z^4\sigma_{\rvy}^2}{(\eta_z^2+\sigma_{\rvy}^2)^2}
        =\frac{\sigma_{\rvy}^2\eta_z^2}{\eta_z^2+\sigma_{\rvy}^2}
        =\gamma\sigma_{\rvy}^2,
    \end{equation*}
    which is the variance of $\sigma_{\rvy}\sqrt{\gamma}\,\rvw'$. DING draws
    one noise vector; \eqref{eq:unif-update} draws two with the same total
    variance.
\item[(iii)] With $\tilde{\vmu}_z=\vmu_z+\eta_z\bxi_z$ and
    $\tyz=\rvy+\sigma_{\rvy}\bxi_y$, the observed block of
    $\tilde{\vmu}_z+\gamma\rmM^T(\tyz-\rmM\tilde{\vmu}_z)$ is
    $(1-\gamma)\tilde{\vmu}_z[\bar{\mathbf{m}}]+\gamma\tyz$, and the missing
    block is $\tilde{\vmu}_z[\mathbf{m}]$ since $\rmM^T$ places zeros there.
    These are \eqref{eq:unif-ding-twonoise} and
    \eqref{eq:unif-ding-twonoise-m}.
\item[(iv)] $\rmM\rmM^T=\rmI_q$, so
    $\sigma_{\rvy}^2\rmI_q+\eta_z^2\rmM\rmM^T=(\sigma_{\rvy}^2+\eta_z^2)\rmI_q$
    and $\eta_z^2(\sigma_{\rvy}^2+\eta_z^2)^{-1}=\gamma$.
\end{itemize}
The reduction is exact in distribution, not pathwise, because of the noise
bookkeeping in (ii). The trust is the sensor noise itself, $r=\sigma_{\rvy}$,
because the operator is exact and needs no surrogate error absorbed into
$r$. The one difference is again the solver: the system in (iv) is a multiple
of the identity, so CG terminates in one iteration, and this is the only
reduction of the four in which the truncation caveat of the opening does not
arise. DING is stated in pixel space; carrying its coordinate mask into the
latent is the subject of Appendix~\ref{sec:competitors-ding}.

\subsubsection{Summary of the reductions}
\label{sec:unif-summary}

Table~\ref{tab:unif-map} collects the substitutions above. Three of the four
anchor at the denoised estimate; DING anchors at the perturbed prior mean and
is the only one that also perturbs the measurement, which is what makes its
step a posterior draw rather than a correction. Our sampler exposes the anchor
as a setting (Appendix~\ref{sec:app-anchor-ablation}); each of the four fixes
it. The rows follow the order of the subsections.

\begin{table}[H]
\centering
\footnotesize
\setlength{\tabcolsep}{4pt}
\renewcommand{\arraystretch}{1.2}
\caption{The four measurement steps as instances of \eqref{eq:unif-update}.
Trust is $\tau=\eta_z^2/r^2$; each row is recovered by one substitution
under the stated idealization.}
\label{tab:unif-map}
\begin{tabular}{@{}lllll@{}}
\toprule
method & anchor & trust $\tau$ & how the system is solved & idealization \\
\midrule
DDS      & denoised estimate & finite & $M$-step CG in $\mathbb{R}^d$ & converged solve \\
DiffPIR  & denoised estimate & finite & closed form (FFT) & linear $\emH$ \\
DDNM     & denoised estimate & $\tau\to\infty$ & projection by $\emH^{+}$ & Moore--Penrose inverse \\
DING     & perturbed prior mean & $\eta_z^2/\sigma_{\rvy}^2$ & closed form, trivial system & in distribution \\
\midrule
\paseo\ & either & tuned $r$ & $M$-step CG in $\mathbb{R}^d$ & \\
\bottomrule
\end{tabular}
\end{table}

\subsection{First-order special cases of the measurement step}
\label{sec:app-unification-approx}

Two published samplers, FlowChef~\citep{patel2025flowchef} and
PnP-Flow~\citep{martin2025pnpflow}, apply the measurement as a single
gradient step on the data misfit rather than as a solve. Neither is an exact
special case of \eqref{eq:unif-canonical}; both are its first-order
truncation, and the term they discard is exactly the one our solver computes. This
section states that truncation once and then reduces each method to it. The
notation and the simplest setting of Appendix~\ref{sec:app-unification}
apply throughout.

\paragraph{The first-order step.}
In the $\tau$ form the gain of \eqref{eq:unif-canonical} is
$\tau\emH^T(\rmI_d+\tau\emH\emH^T)^{-1}$. Whenever $\tau\|\emH\emH^T\|<1$
the inverse has a convergent Neumann series,
$(\rmI_d+\tau\emH\emH^T)^{-1}=\rmI_d-\tau\emH\emH^T+\tau^2(\emH\emH^T)^2-\cdots$,
and collecting powers of $\tau$,
\begin{equation}
    \rvz_0^\star
    =\rvz_0^{(0)}
    +\tau\emH^T\big(\yz-\emH\rvz_0^{(0)}\big)
    -\tau^2\emH^T\emH\emH^T\big(\yz-\emH\rvz_0^{(0)}\big)
    +\mathcal{O}(\tau^3).
    \label{eq:unif-firstorder}
\end{equation}
The first term is a gradient step on the data misfit; everything after it is
the effect of the preconditioner $(\rmI_d+\tau\emH\emH^T)^{-1}$ beyond the
identity, which a solve computes and a gradient step does not. Keeping the
first term only,
\begin{equation}
    \boxed{\;
    \rvz_0^{\star}\approx\rvz_0^{(0)}+\tau\emH^T\big(\yz-\emH\rvz_0^{(0)}\big),
    \;}
    \label{eq:unif-gradstep}
\end{equation}
whose error is $\mathcal{O}(\tau^2)$ wherever the series converges. Lifted
to the diffusion state through
$\rvx_t=\alpha_t\rvz_0+\sigma_t\hat{\rvx}_1'$, the same step reads
\begin{equation}
    \rvx_t\approx
    \underbrace{\alpha_t\rvz_0^{(0)}+\sigma_t\hat{\rvx}_1'}_{\text{unconditional move}}
    +\underbrace{\alpha_t\tau\,\emH^T\big(\yz-\emH\rvz_0^{(0)}\big)}_{\text{measurement correction}},
    \label{eq:unif-gradstep-state}
\end{equation}
so in state coordinates the correction carries a factor $\alpha_t$.

\subsubsection{FlowChef: the gradient step, taken in the state}
\label{sec:unif-flowchef}

FlowChef~\citep{patel2025flowchef} steers a rectified flow by its Eq.~(9),
with the inverse-problem cost of their Eq.~(4) written here with $\emH$ for
their degradation operator $\mathcal{F}$ and $\rvy$ for their degraded
reference $\rvx_0^{\mathrm{ref}}$,
\begin{equation}
    \rvx_{t-\Delta t}
    =\underbrace{\rvx_t+\Delta t\,u_\theta(\rvx_t,t)}_{\text{(a) unconditional move}}
    \;\underbrace{-\;s'\,\nabla_{\hat{\rvx}_0}L}_{\text{(b) measurement correction}},
    \qquad
    L=\big\|\rvy-\emH\hat{\rvx}_0\big\|^2,
    \label{eq:unif-flowchef}
\end{equation}
with $\hat{\rvx}_0$ the clean estimate implied by the current velocity and
$s'$ a scalar guidance scale. Only (b) is comparable to us; (a) is their
Euler step on the flow ODE and plays the role of our bridge transition. The
gradient is taken with respect to the clean estimate on a graph that contains
no network: their Algorithm~1 evaluates the velocity $\rvv$ before enabling
gradients, so $\hat{\rvx}_0=\rvx_t+t\rvv$ is affine in $\rvx_t$ with
$\partial\hat{\rvx}_0/\partial\rvx_t=\rmI_d$. This is what ``gradient-free''
means in their abstract: free of gradients through the network.

\paragraph{Reduction to the first-order step.}
The correction (b) is the measurement correction of
\eqref{eq:unif-gradstep-state}, in four steps:
\begin{align}
    -s'\,\nabla_{\hat{\rvx}_0}L
    &\overset{\mathrm{(i)}}{=}2s'\,\emH^T\big(\rvy-\emH\hat{\rvx}_0\big)
    \label{eq:unif-flowchef-grad}\\
    &\overset{\mathrm{(ii)}}{=}\alpha_t\,\frac{2s'}{\alpha_t}\,\emH^T\big(\rvy-\emH\hat{\rvx}_0\big)
    \label{eq:unif-flowchef-alpha}\\
    &\overset{\mathrm{(iii)}}{=}\alpha_t\tau\,\emH^T\big(\yz-\emH\rvz_0^{(0)}\big),
    \qquad \tau=\frac{2s'}{\alpha_t},
    \label{eq:unif-flowchef-tau}\\
    &\overset{\mathrm{(iv)}}{=}\alpha_t\Big[\big(\rvz_0^{(0)}+\tau\emH^T(\yz-\emH\rvz_0^{(0)})\big)-\rvz_0^{(0)}\Big].
    \label{eq:unif-flowchef-final}
\end{align}
The bracket in the last line is the correction of \eqref{eq:unif-gradstep}.
The steps are the following.
\begin{itemize}
\item[(i)] Differentiating $L=(\rvy-\emH\hat{\rvx}_0)^T(\rvy-\emH\hat{\rvx}_0)$
    gives $\nabla_{\hat{\rvx}_0}L=-2\emH^T(\rvy-\emH\hat{\rvx}_0)$, the
    $-\emH^T$ from the chain rule and the $2$ from the square.
\item[(ii)] Multiplying and dividing by $\alpha_t$ exposes the factor the
    lift \eqref{eq:unif-gradstep-state} attaches to a correction applied in
    the state.
\item[(iii)] $\hat{\rvx}_0$ is read as $\rvz_0^{(0)}$ and $\rvy$ as $\yz$,
    both being the denoiser's clean estimate and the measurement in latent
    coordinates, and $\tau$ is defined by matching the scalars.
\item[(iv)] Rewriting the correction as the difference between the
    first-order step and its anchor, which is what
    \eqref{eq:unif-gradstep-state} lifts.
\end{itemize}
FlowChef's correction is therefore ours to first order in $\tau$, with
$\tau=2s'/\alpha_t$. Their released settings hold $s'$ fixed per task, so the
$1/\alpha_t$ is a timestep dependence a constant $s'$ ignores: a scale tuned
at one timestep is mistuned at another. One choice is folded into (ii): the
corrections are equated in the state, which is what the next network call
sees. Equating them in the clean estimate instead gives $\tau=2s'$; the two
differ because the two methods linearize the clean--state relation
differently, ours holding $\hat{\rvx}_1'$ fixed so that
$\partial\rvx_t/\partial\rvz_0=\alpha_t\rmI_d$, theirs holding the velocity
fixed so that $\partial\hat{\rvx}_0/\partial\rvx_t=\rmI_d$. Two further
differences. Their Algorithm~1 repeats the correction $N$ times per sampling
step with the velocity frozen, which refines a step against a fixed model,
whereas our Gauss--Newton loop re-linearizes the operator; at their reported
setting $N=1$ the distinction is immaterial, and for $N>1$ the composition
is a polynomial in $\emH^T\emH$ applied to the residual, not a larger
$\tau$. And the reduction is to their stated rule, Theorem~4.3, in pixel
space; their latent-space settings pass the loss through the decoder and
optimize with Adam, so there even the direction is not $\emH^T(\cdot)$.

\subsubsection{PnP-Flow: the gradient step, taken in the clean estimate}
\label{sec:unif-pnpflow}

PnP-Flow~\citep{martin2025pnpflow} runs a forward--backward splitting whose
data step, the first line of their Algorithm~3, is
\begin{equation}
    \rvz_n=\rvx_n-\gamma_n\nabla F(\rvx_n),
    \qquad
    F(\rvx)=\frac{1}{2\sigma^2}\big\|\emH\rvx-\rvy\big\|^2,
    \label{eq:unif-pnpflow}
\end{equation}
followed by a re-noising interpolation and a denoiser call. Here $\rvx_n$ is
the output of the flow-matching denoiser at the previous step, an estimate of
the clean image, and $\gamma_n$ a step size with schedule
$\gamma_n=(1-t_n)^{a}$, $a\in(0,1]$ (their exponent $\alpha$, renamed here
to keep $\alpha_t$ for the DDIM coefficient).

\paragraph{Reduction to the first-order step.}
The data step \eqref{eq:unif-pnpflow} is \eqref{eq:unif-gradstep}, in three
steps:
\begin{align}
    \rvz_n
    &\overset{\mathrm{(i)}}{=}\rvx_n+\frac{\gamma_n}{\sigma^2}\emH^T\big(\rvy-\emH\rvx_n\big)
    \label{eq:unif-pnpflow-grad}\\
    &\overset{\mathrm{(ii)}}{=}\rvz_0^{(0)}+\tau\emH^T\big(\yz-\emH\rvz_0^{(0)}\big),
    \qquad \tau=\frac{\gamma_n}{\sigma^2},
    \label{eq:unif-pnpflow-tau}\\
    &\overset{\mathrm{(iii)}}{=}\rvz_0^{(0)}
    +\tau\emH^T\big(\rmI_d+\tau\emH\emH^T\big)^{-1}\big(\yz-\emH\rvz_0^{(0)}\big)
    +\mathcal{O}(\tau^2).
    \label{eq:unif-pnpflow-final}
\end{align}
The last line is \eqref{eq:unif-canonical} up to the discarded tail of
\eqref{eq:unif-firstorder}. The steps are the following.
\begin{itemize}
\item[(i)] $\nabla F(\rvx)=\sigma^{-2}\emH^T(\emH\rvx-\rvy)$, and the sign
    is absorbed into the residual.
\item[(ii)] $\rvx_n$ is read as $\rvz_0^{(0)}$ and $\rvy$ as $\yz$; the
    correction is applied in clean coordinates on both sides, so no
    $\alpha_t$ appears, and $\tau$ is defined by matching the scalars.
\item[(iii)] The Neumann expansion \eqref{eq:unif-firstorder} in reverse.
\end{itemize}
PnP-Flow's data step is therefore ours to first order in $\tau$, with
$\tau=\gamma_n/\sigma^2$. Its schedule $\gamma_n=(1-t_n)^{a}$ decays as the
sampler approaches the data, as does our $\eta_z^2/r^2$ through the bridge
width; the two are the same quantity, how far the data may move the denoised
estimate, set by two different arguments. Their Algorithm~3 has no inner
loop, so the remark on $N$ above does not apply.

\paragraph{What both discard.}
The gap between these two methods and \eqref{eq:unif-canonical} is not a
modeling difference but a solver one. Both take the steepest-descent
direction $\emH^T(\yz-\emH\rvz_0^{(0)})$; \eqref{eq:unif-canonical} takes the
preconditioned direction $\emH^T(\rmI_d+\tau\emH\emH^T)^{-1}(\yz-\emH\rvz_0^{(0)})$,
and the preconditioner is what the $M$ conjugate-gradient iterations of
Appendix~\ref{sec:app-cg-solve} apply. Table~\ref{tab:unif-approx} records the
two substitutions beside the exact four.

\begin{table}[H]
\centering
\footnotesize
\setlength{\tabcolsep}{4pt}
\renewcommand{\arraystretch}{1.2}
\caption{The two first-order cases. Both anchor at the denoised estimate,
keep only the gradient term of \eqref{eq:unif-firstorder}, and discard the
preconditioner.}
\label{tab:unif-approx}
\begin{tabular}{@{}llll@{}}
\toprule
method & trust $\tau$ & where the correction is applied & inner loop \\
\midrule
FlowChef & $2s'/\alpha_t$ & the state & $N$ steps, velocity frozen \\
PnP-Flow & $\gamma_n/\sigma^2$ & the clean estimate & none \\
\bottomrule
\end{tabular}
\end{table}

\clearpage
\appendixpart{II}{Experiments}
\section{Experimental setup and evaluation}
\label{sec:additional}

\subsection{Latent-operator training details}
\label{sec:operator-training-details}

We follow the recipe of \citet{raphaeli2025silo}, training one operator per
degradation and per latent space, offline. For a clean image $\rvx$, we form a
measurement as in \eqref{eq:i-forward} with
$\rvn\sim\mathbf{N}(0,s^2\rmI)$ and $s\sim\mathcal{U}(0,0.04)$, resample it to
the encoder's resolution, and minimize the loss
\begingroup
\begin{equation}
    \mathcal{L}(\psi)
    =\mathbb{E}_{\rvx,s,\rvn}
     \big\|\,\rmH_{\psi}\big(\gE(\rvx),s\big)
            -\gE\big(\gA(\rvx)+\rvn\big)\big\|_1 .
\label{eq:operator-training}
\end{equation}
\endgroup
Conditioning on $s$ lets one operator serve a range of noise levels; at
sampling time we pass the known $\sigma_{\rvy}$. $\rmH_{\psi}$ is a
$1.2$M-parameter CNN used in \citet{raphaeli2025silo} acting at latent resolution, so a forward pass and a
Jacobian--vector product cost far less than one denoiser evaluation. We use
FFHQ-$512$ with the evaluation images held out, Adam with a learning rate of
$10^{-4}$, and a batch size of $16$; the prior and autoencoder stay frozen, and the operator never sees a
diffusion state. Its nonlinearity is load-bearing:
$\gE\circ\gA\circ\gD$ is nonlinear even for linear $\gA$, and a matched affine
operator fits the latent measurement $4.8$--$8.1$\,dB worse on
super-resolution.
Appendix~\ref{sec:operator-validation} compares the trained
operator's values and Jacobian with those of the true chain.

\FloatBarrier
\subsection{Validation of the learned operator and its Jacobian}
\label{sec:operator-validation}

A useful learned operator must predict both the encoded measurement and
how that measurement changes when the latent moves. Table~\ref{tab:jacobian}
checks these separately against the true decode--degrade--encode chain.
The saved reconstructions use the arc anchor; inpainting uses the hard mask
rather than the learned weighting of the main results
(Appendix~\ref{sec:inpaint-variants}).

\textbf{Prediction and direction.} Panel A compares clean-image latents
with re-encoded saved reconstructions. Value error often exceeds the noise
scale $r$ assumed by the solver: for RV-v5.1 blur, the RMS error is $0.31$
against $r=0.005$. This comparison does not validate $r$ as a calibrated
error model. On the four non-JPEG tasks, direction agreement on clean latents
is higher for SD3.5 (cosine $0.87$--$0.96$) than RV-v5.1 ($0.37$--$0.63$).
Errors on re-encoded reconstructions are not uniformly smaller; SD3.5
super-resolution is a counterexample. JPEG needs particular care because
the reference derivative itself is sensitive to the finite-difference step.

\textbf{Numerical consistency.} Panel B asks whether the forward and
transpose Jacobian products agree and estimates a condition-number bound.
The small adjoint discrepancies support the consistency of those two
implemented products. They do not establish that the learned Jacobian is
accurate, or that the CG solve converges. Likewise, a large estimated bound
suggests potential numerical difficulty but is not a measurement of the
actual condition number.

\begin{table}[H]
\centering
\footnotesize
\setlength{\tabcolsep}{4pt}
\renewcommand{\arraystretch}{0.9}
\caption{Does the learned operator predict both measurements and their local changes?
Eight test images and four random directions per image and configuration.
Panel A: value error is RMS distance from $\gE\circ\gA\circ\gD$; compare it with
$r$, the solver's assumed operator-noise scale. Direction agreement is the
cosine between learned and reference Jacobian--vector products ($1$ is best).
``Clean'' means $\gE(\rvx)$; ``re-encoded'' means the VAE mean encoding of a
saved arc reconstruction, not an internal sampler state.
Panel B: magnitude error compares directional-change lengths ($0.1$ means
$10\%$ relative error). Adjoint error compares $\langle\rvu,\rmJ\rvv\rangle$
with $\langle\rmJ^T\rvu,\rvv\rangle$. Both errors should approach $0$.
The norm is estimated by power iteration; the condition bound estimate is
$1+250^2\|\rmJ\|_2^2/(r^2+\lambda)$ (Appendix~\ref{sec:app-cg-solve}), not a measured
condition number. Reference derivatives use the fp32 chain with finite-difference
checks; the JPEG check is unreliable across step sizes.}
\label{tab:jacobian}
\begin{tabular}{@{}clrrrrr@{}}
\toprule
\multicolumn{7}{l}{\textbf{A. Agreement with the reference operator}} \\
\addlinespace
Prior & Task & $r$ & \multicolumn{2}{c}{Value RMS error $\downarrow$} & \multicolumn{2}{c}{Direction cosine $\uparrow$} \\
 &  &  & Clean & Re-encoded & Clean & Re-encoded \\
\midrule
\multirow{5}{*}{\rotatebox[origin=c]{90}{RV-v5.1}} & SR $\times4$ & 0.05 & 0.19 & 0.18 & 0.37 & 0.44 \\
 & SR $\times8$ & 0.02 & 0.16 & 0.14 & 0.59 & 0.64 \\
 & Blur & 0.005 & 0.31 & 0.31 & 0.63 & 0.62 \\
 & Inpaint & 0.05 & 0.22 & 0.20 & 0.37 & 0.55 \\
 & JPEG & 0.08 & 0.38 & 0.36 & 0.01 & 0.02 \\
\midrule
\multirow{5}{*}{\rotatebox[origin=c]{90}{SD3.5}} & SR $\times4$ & 0.01 & 0.08 & 0.12 & 0.90 & 0.89 \\
 & SR $\times8$ & 0.025 & 0.05 & 0.06 & 0.96 & 0.95 \\
 & Blur & 0.02 & 0.26 & 0.26 & 0.91 & 0.92 \\
 & Inpaint & 0.01 & 0.08 & 0.06 & 0.87 & 0.87 \\
 & JPEG & 0.02 & 0.28 & 0.25 & 0.02 & 0.03 \\
\bottomrule
\end{tabular}

\par\smallskip

\begin{tabular}{@{}clrrrr@{}}
\toprule
\multicolumn{6}{l}{\textbf{B. Numerical checks on clean latents}} \\
\addlinespace
Prior & Task & \shortstack{Magnitude\\error $\downarrow$} & \shortstack{Adjoint\\error $\downarrow$} & Estimated $\|\rmJ\|_2$ & \shortstack{Condition\\bound estimate} \\
\midrule
\multirow{5}{*}{\rotatebox[origin=c]{90}{RV-v5.1}} & SR $\times4$ & 0.67 & 3e-07 & 6.7 & 4e+08 \\
 & SR $\times8$ & 0.46 & 7e-07 & 12.2 & 2e+09 \\
 & Blur & 0.42 & 2e-06 & 8.3 & 9e+08 \\
 & Inpaint & 0.63 & 5e-07 & 5.2 & 2e+08 \\
 & JPEG & 0.98 & 9e-07 & 10.5 & 1e+08 \\
\midrule
\multirow{5}{*}{\rotatebox[origin=c]{90}{SD3.5}} & SR $\times4$ & 0.14 & 3e-06 & 3.5 & 8e+09 \\
 & SR $\times8$ & 0.04 & 2e-06 & 2.3 & 5e+08 \\
 & Blur & 0.22 & 4e-06 & 1.6 & 4e+08 \\
 & Inpaint & 0.11 & 4e-06 & 6.1 & 2e+10 \\
 & JPEG & 0.97 & 2e-06 & 27.1 & 1e+11 \\
\bottomrule
\end{tabular}
\end{table}

\FloatBarrier

\subsection{Metric protocol}
\label{sec:metric-protocol}

KID is reported relative to the task, not on an absolute scale: the
observations handed to the sampler score $65.68$, $176.31$, $46.39$, $117.64$,
and $42.90$ for SR $\times4$, SR $\times8$, Gaussian blur, center inpainting,
and JPEG, respectively, so a larger reconstruction KID on SR $\times8$ than on
SR $\times4$ reflects how far the measurement starts from the data
distribution, not a worse reconstruction. KID is therefore comparable within a
task and across methods, and not across tasks.

Memory is the peak allocated memory reported
by PyTorch, which is measured per process and is unaffected by co-tenancy.

\begingroup
\subsection{PASEO sampling hyperparameters}
\label{sec:paseo-hyperparameters}

Tables~\ref{tab:hyperparameters-ffhq} and~\ref{tab:hyperparameters-coco} list
the settings used for the main results. We chose these settings by LPIPS on
the validation images of Section~\ref{sec:experiments}, which are disjoint
from the evaluation images. Here $P$ is the number of inner
corrections per reverse step, $C$ is the maximum number of CG iterations per
correction, $\rho$ is the relaxation factor, and $\lambda$ is the damping
added to $r^2$ in the linear solve. We report the effective latent likelihood
standard deviation $r$, including any scaling applied in the configuration;
CFG denotes the classifier-free guidance scale. All runs use the arc anchor
in \eqref{eq:method-practical-anchor}, a relative CG tolerance of $10^{-5}$,
and both prior and measurement perturbations.

\begin{table}[htbp]
\centering
\small
\setlength{\tabcolsep}{10pt}
\renewcommand{\arraystretch}{1.08}
\caption{\paseo\ hyperparameters on FFHQ. Both priors use $28$ reverse
steps and $53$ NFEs for every task.}
\label{tab:hyperparameters-ffhq}
\begin{tabular}{@{}lrrrrrr@{}}
\toprule
Task & $P$ & $C$ & $\rho$ & $\lambda$ & $r$ & CFG \\
\midrule
\multicolumn{7}{@{}l}{\textbf{RV-v5.1}} \\
\addlinespace[2pt]
SR $\times4$ & 3 & 3  & 0.15 & 0.005 & 0.05  & 1 \\
SR $\times8$ & 4 & 2  & 0.15 & 0.005 & 0.02  & 1 \\
Gaussian blur & 5 & 3 & 0.15 & 0.005 & 0.005 & 1 \\
Inpainting & 3 & 12 & 0.6 & 0.005 & 0.05 & 1 \\
JPEG & 4 & 1 & 0.3 & 0.05 & 0.08 & 1 \\
\midrule
\multicolumn{7}{@{}l}{\textbf{SD3.5-medium}} \\
\addlinespace[2pt]
SR $\times4$ & 2 & 2 & 0.6 & 0 & 0.01 & 4 \\
SR $\times8$ & 1 & 2 & 1 & 0 & 0.025 & 1 \\
Gaussian blur & 1 & 2 & 0.6 & 0 & 0.02 & 4 \\
Inpainting & 1 & 5 & 1 & 0 & 0.01 & 1 \\
JPEG & 3 & 3 & 0.3 & 0 & 0.02 & 4 \\
\bottomrule
\end{tabular}
\end{table}

\begin{table}[htbp]
\centering
\small
\setlength{\tabcolsep}{5.5pt}
\renewcommand{\arraystretch}{1.08}
\caption{\paseo\ hyperparameters on COCO with SD3.5-medium and
LSDIR-trained operators. Steps denotes the number of reverse steps, and
NFEs the number of denoiser evaluations.}
\label{tab:hyperparameters-coco}
\begin{tabular}{@{}lrrrrrrrr@{}}
\toprule
Task & Steps & NFEs & $P$ & $C$ & $\rho$ & $\lambda$ & $r$ & CFG \\
\midrule
SR $\times4$ & 50 & 97 & 3 & 3 & 0.3 & 0 & 0.01 & 1 \\
SR $\times8$ & 100 & 197 & 3 & 1 & 0.45 & 0.005 & 0.025 & 1 \\
Gaussian blur & 50 & 97 & 4 & 1 & 0.4 & 0.005 & 0.02 & 4 \\
Inpainting & 100 & 197 & 2 & 2 & 1 & 0 & 0.005 & 1 \\
JPEG & 50 & 97 & 5 & 1 & 0.3 & 0.02 & 0.08 & 4 \\
\bottomrule
\end{tabular}
\end{table}

For consecutive sampling times $t_{k+1}>t_k$, define the DDPM bridge
standard deviation as
\[
\eta_k^{\mathrm{DDPM}}
=\sigma_{t_k}\sqrt{1-
\left(\frac{\alpha_{t_{k+1}}\sigma_{t_k}}
{\alpha_{t_k}\sigma_{t_{k+1}}}\right)^2}.
\]
RV-v5.1 uses this schedule for all tasks. SD3.5 uses
$\eta_k=\sigma_{t_k}(1-\alpha_{t_k})$ throughout FFHQ and for COCO
inpainting. On COCO, SR $\times4$ uses
$\eta_k=\sigma_{t_k}\sqrt{1-\alpha_{t_k}}$, while SR $\times8$, blur, and
JPEG use $\eta_k^{\mathrm{DDPM}}$.

We initialize the inner corrections at the proxy's clean-latent estimate,
except for COCO SR $\times4$, which starts at the local prior mean $\vmu_z$.
Inpainting uses soft likelihood weights derived from the learned operator's
sensitivity, estimated with four random probes at each correction
(Appendix~\ref{sec:inpaint-variants}). The likelihood receives no supplied
mask. We use the prompt ``A high quality photo of a face'' for FFHQ and
``A high quality photo'' for COCO, except for COCO SR $\times8$, which uses
the image caption.

\FloatBarrier
\endgroup

\clearpage
\subsection{PASEO algorithm}
\label{sec:paseo-algorithm-reference}

\begingroup
\algrenewcommand\algorithmicrequire{\textbf{Input:}}
\definecolor{algcomment}{gray}{0.64}
\algrenewcommand{\algorithmiccomment}[1]{%
  \hfill\textcolor{algcomment}{\(\triangleright\) #1}}
\newcommand{\ReferenceLineComment}[2]{\Statex
  \hspace{\algorithmicindent}\textcolor{algcomment}{%
    \texttt{/* \phantom{\stepnum{#1}~}#2 */}}%
  \hfill\phantom{\(\triangleright\) }\stepnum{#1}}

\begin{algorithm}[ht]
\caption{\paseo: Matrix-free local learned-operator conditioning}
\label{alg:paseo-reference}
\small
\begin{algorithmic}[1]
\Require decreasing timesteps $(t_k)_{k=K}^{0}$ with $t_K = 1$, $t_0 = 0$; latent
measurement $\cmeas{\yz} = \gE(\cmeas{\rvy}) \in \R^d$; learned latent operator $\rmH_{\psi}$; noise
level $r$; DDIM parameters $(\eta_k)_{k=K}^{0}$; damping $\lambda\geq0$;
relaxation $\rho\in(0,1]$; inner steps $P$; CG budget $C$.
\State $\cstate{\rvx} \sim \mathbf{N}(0,\rmI_d)$ \label{line:reference-init}
\For{$k = K-1$ \textbf{to} $1$}
    \State $\hat{\rvx}_0 \leftarrow \xzero{\cstate{\rvx}}{t_{k+1}}$
    \State $\hat{\rvx}_1 \leftarrow
        \big(\cstate{\rvx} - \alpha_{t_{k+1}}\hat{\rvx}_0\big)/\sigma_{t_{k+1}}$
    \State $\rvm \leftarrow \alpha_{t_k}\hat{\rvx}_0
        + \big(\sigma^2_{t_k} - \eta_k^2\big)^{1/2}\hat{\rvx}_1$
        \Comment{bridge $\rvx_{t_k} \mid \rvx_{t_{k+1}} \approx \mathbf{N}(\rvm,\eta_k^2\rmI_d)$}
    \ReferenceLineComment{1}{Bridge sample and clean-latent prior}
    \State $\rvw' \sim \mathbf{N}(\vzero,\rmI_d)$
    \State $\cproxy{\rvx_{t_k}^{\mathrm{pxy}}} \leftarrow \rvm + \eta_k \rvw'$
    \State $\cproxy{\hat{\rvx}_1'} \leftarrow \xone{\cproxy{\rvx_{t_k}^{\mathrm{pxy}}}}{t_k}$
        \Comment{noise latent re-estimated at $t_k$} \label{line:reference-proxy}
    \State $\clat{\vmu_z} \leftarrow (\rvm - \sigma_{t_k}\cproxy{\hat{\rvx}_1'})/\alpha_{t_k}$,
        \quad $\eta_z := \eta_k/\alpha_{t_k}$
        \Comment{clean prior given the current state and proxy} \label{line:reference-prior}
    \State $\clat{\rvz_0^{(0)}} \leftarrow (\cproxy{\rvx_{t_k}^{\mathrm{pxy}}} - \sigma_{t_k}\cproxy{\hat{\rvx}_1'})/\alpha_{t_k}$
        \Comment{Gauss--Newton expansion point}
    \ReferenceLineComment{2}{Perturb-and-optimize targets}
    \State $\bxi_y\sim\mathbf{N}(\vzero,\rmI_d)$,
        \quad $\bxi_z\sim\mathbf{N}(\vzero,\rmI_d)$
    \State $\cmeas{\tyz} \leftarrow \cmeas{\yz} + r\,\bxi_y$,
        \quad $\clat{\tilde{\vmu}_z} \leftarrow \clat{\vmu_z} + \eta_z\,\bxi_z$ \phantomsection\label{line:reference-perturb}
    \State $\beta_k \leftarrow \eta_{z,\min}/\eta_z$, \quad $c_1 \leftarrow (1-\beta_k^2)^{1/2}$
        \Comment{$\eta_{z,\min}$: smallest $\eta_z$ in the schedule}
    \ReferenceLineComment{3}{Local measurement update}
    \For{$j = 0$ \textbf{to} $P-1$} \label{line:reference-gn-start}
        \State $\clat{\rvz_a} \leftarrow \clat{\vmu_z} + c_1(\clat{\rvz_0^{(j)}} - \clat{\vmu_z}) + \beta_k(\clat{\tilde{\vmu}_z} - \clat{\vmu_z})$
            \Comment{arc anchor, Appendix~\ref{sec:app-anchor-family}} \label{line:reference-arc}
        \State $\rvh \leftarrow \rmH_{\psi}(\clat{\rvz_0^{(j)}})$
        \State $\rmJ \leftarrow \partial \rmH_{\psi}/\partial \rvz_0 \big|_{\clat{\rvz_0^{(j)}}}$
            \Comment{used only through JVPs and VJPs}
        \State $\rmW \leftarrow \mathrm{diag}(w)$, $w_i \propto \|\rmJ_{i,:}\|$
            \Comment{inpainting; otherwise $\rmW = \rmI_d$} \phantomsection\label{line:reference-weight}
        \State $\rvb \leftarrow \rmW\big(\cmeas{\tyz} - \rvh
            - \rmJ(\clat{\rvz_a} - \clat{\rvz_0^{(j)}})\big)$ \label{line:reference-residual}
        \State \texttt{solve} $\big((r^2+\lambda)\rmI_d
            + \eta_z^2 \rmW\rmJ \rmJ^\top\rmW\big) \rvv = \rvb$
            \Comment{at most $C$ CG iterations} \phantomsection\label{line:reference-cg}
        \State $\clat{\rvz_0^{\star}} \leftarrow \clat{\rvz_a} + \eta_z^2 \rmJ^\top\rmW \rvv$ \phantomsection\label{line:reference-zstar}
        \State $\clat{\rvz_0^{(j+1)}} \leftarrow \clat{\rvz_0^{(j)}} + \rho\,(\clat{\rvz_0^{\star}} - \clat{\rvz_0^{(j)}})$ \phantomsection\label{line:reference-gn-end}
    \EndFor
    \State $\cstate{\rvx} \leftarrow \alpha_{t_k}\clat{\rvz_0^{(P)}} + \sigma_{t_k}\cproxy{\hat{\rvx}_1'}$
        \Comment{back to the diffusion state~\stepnum{4}}%
        \label{line:reference-lift}
\EndFor
\State \textbf{Return:} $\gD\big(\xzero{\cstate{\rvx}}{t_1}\big)$
\end{algorithmic}
\end{algorithm}
\endgroup

\clearpage
\addtocontents{toc}{\protect\supplementarycontentscontinued}
\section{Ablations and diagnostics}
\label{sec:ablations-diagnostics}

We examine the practical sampler through four questions: whether it
matches the posterior when that posterior is known, how the anchor
and solver settings affect reconstruction, how computation is allocated,
and what changes specifically in
inpainting. The learned operator is validated separately alongside its
training details in Appendix~\ref{sec:operator-validation}.

\subsection{Checking the sampler against known posteriors}
\label{sec:app-posterior-checks}

We check the sampler on two problems whose posterior is known. The first
is a small linear--Gaussian problem that tests the ideal update of
Appendix~\ref{sec:app-perturb-sampling}. The second runs the full SD3.5
sampler with exact linear operators in place of the learned one.

\subsubsection{A linear--Gaussian problem}
\label{sec:app-tractable}

Table~\ref{tab:tractable} checks whether the update in
\eqref{eq:app-perturb-result} produces the right distribution, including its
uncertainty. We use a linear--Gaussian problem whose posterior mean and
covariance are known exactly, and ask three questions.

\textbf{1. Does the ideal update work?}
An exact solve reproduces the posterior mean, covariance and interval
coverage up to sampling error. This provides a numerical check of the derivation.

\textbf{2. What changes when we approximate the update?}
Stopping conjugate gradients (CG) after one or two iterations shifts the
sample mean and distorts the covariance. By $C=8$ iterations, this
$32$-dimensional linear solve matches the exact-solve reference.
Taking only a fraction $\rho<1$ of the correction also shifts the mean and
makes the samples too spread out.

\textbf{3. What does changing the anchor do?}
Here the distinction is whether we redraw the bridge noise $\rvw'$ for
each sample or keep it fixed. With fresh bridge noise, every tested
$\beta$ recovers the posterior in this affine experiment.
With fixed bridge noise, as in repeated updates from one bridge state,
$\beta<1$ shifts the center and narrows the distribution: $\beta=0.5$
retains about $57\%$ of the posterior spread, and $\beta=0$ about $30\%$.
Appendix~\ref{sec:app-anchor-family} derives this shift in
\eqref{eq:app-anchor-bias} and the variance change.
Averaging coverage over different fixed bridge noises can still give
about $90\%$, so coverage alone does not reveal this mismatch.
These checks concern the affine update; they do not establish exact
posterior sampling for the full nonlinear algorithm.

\begin{table}[ht]
\centering
\footnotesize
\setlength{\tabcolsep}{4pt}
\caption{Checking samples against a known Gaussian posterior.
\textbf{Reading guide:} mean and covariance errors should approach $0$,
coverage $0.90$, and spread $1$; spread below $1$ means too little variation.
The problem has state dimension $d=64$, measurement dimension $q=32$,
dense random $\emH$, $\eta=1$, and $r=0.3$.
Each rule uses $20{,}000$ draws, except the fixed-noise rows: their statistics
are computed from $2{,}000$ draws per fixed $\rvw'$ and averaged over $20$
choices of $\rvw'$. Mean error is
$\|\bar{\rvz}-\rvm_p\|/\|\rvm_p-\vmu_z\|$; covariance error is the relative
Frobenius error to $\mathbf{\Sigma}_p$. Coverage is the fraction of sample
coordinates inside the true $90\%$ marginal intervals; spread is
$\sqrt{\operatorname{tr}\operatorname{Cov}/\operatorname{tr}\mathbf{\Sigma}_p}$.}
\label{tab:tractable}
\begin{tabular}{lrrrr}
\toprule
Update rule & mean err.\ & cov.\ err.\ & $90\%$ coverage & spread \\
\midrule
\multicolumn{5}{l}{\textbf{1. Ideal update: reference}} \\
exact solve & 0.008 & 0.045 & 0.900 & 1.000 \\
\midrule
\multicolumn{5}{l}{\textbf{2. Approximate update: fewer CG steps or partial correction}} \\
CG $C{=}$1 & 0.438 & 0.272 & 0.842 & 1.093 \\
CG $C{=}$2 & 0.197 & 0.107 & 0.885 & 1.027 \\
CG $C{=}$4 & 0.051 & 0.047 & 0.899 & 1.003 \\
CG $C{=}$8 & 0.008 & 0.045 & 0.900 & 1.000 \\
CG $C{=}$16 & 0.007 & 0.045 & 0.900 & 1.000 \\
$\rho{=}$0.6 & 0.400 & 0.147 & 0.857 & 1.060 \\
$\rho{=}$0.3 & 0.702 & 0.428 & 0.781 & 1.174 \\
\midrule
\multicolumn{5}{l}{\textbf{3. Anchor interpolation}} \\
\multicolumn{5}{l}{\emph{Fresh bridge noise for each sample}} \\
$\beta{=}$1.0 & 0.007 & 0.046 & 0.900 & 1.000 \\
$\beta{=}$0.5 & 0.008 & 0.046 & 0.900 & 1.000 \\
$\beta{=}$0.25 & 0.007 & 0.045 & 0.900 & 1.000 \\
$\rvz_0^{(0)}$ ($\beta{=}0$) & 0.008 & 0.044 & 0.900 & 1.001 \\
\addlinespace
\multicolumn{5}{l}{\emph{Fixed bridge noise within each experiment}} \\
$\beta{=}$1.0 & 0.026 & 0.142 & 0.900 & 0.999 \\
$\beta{=}$0.5 & 0.968 & 0.743 & 0.894 & 0.565 \\
$\beta{=}$0.25 & 1.057 & 0.927 & 0.896 & 0.386 \\
$\beta{=}$0.0 & 1.117 & 0.989 & 0.892 & 0.304 \\
\bottomrule
\end{tabular}
\end{table}

\subsubsection{The full sampler with exact linear operators}
\label{sec:app-linear-check}

The image experiments run a different algorithm: a
learned operator, a few CG iterations, and the arc anchor.
Table~\ref{tab:lincheck} checks this full SD3.5 sampler on problems whose
posterior is known. We keep
the sampler and replace the learned operator $\rmH_\psi$ with an exactly
linear operator in latent space. Every linearization is then exact, so any
departure from the posterior comes from the solver, the anchor, or the
diffusion prior, not from operator error. We ask three questions.

\begin{figure}[ht]
\centering
\begin{tikzpicture}
\node[anchor=south west, inner sep=0] (ipgrid) at (0,0) {\includegraphics[width=0.95\linewidth]{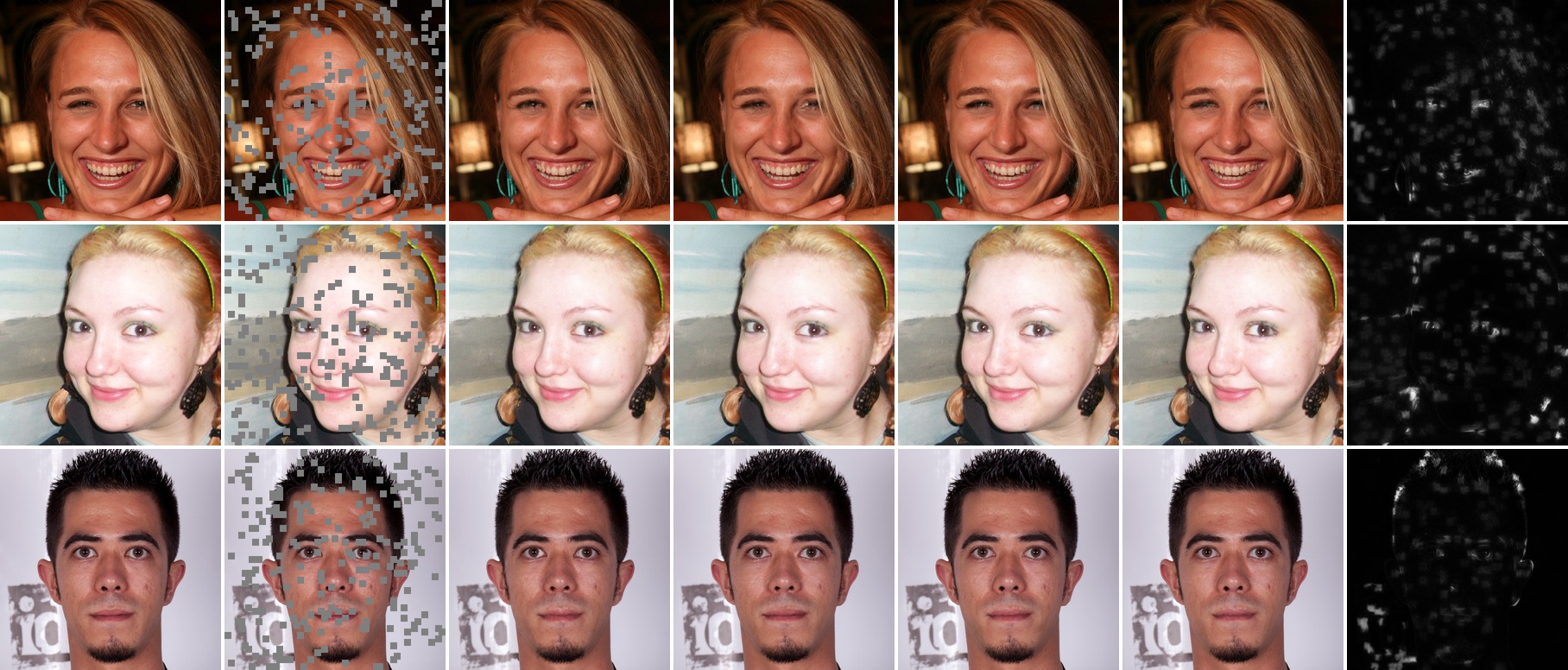}};
\foreach \f/\t in {0.070485/Ground truth, 0.213656/Measurement, 0.356828/\paseo\ draw 1, 0.500000/\paseo\ draw 2, 0.643172/\paseo\ draw 3, 0.786344/\textsc{DING} draw, 0.929515/Std.\ over draws}
  \node[anchor=south, font=\scriptsize] at ($(ipgrid.north west)!\f!(ipgrid.north east)+(0,2pt)$) {\t};
\end{tikzpicture}
\caption{Random-hole inpainting with the exact mask operator on three FFHQ
faces. \paseo\ uses the arc anchor. The last column is the per-pixel standard
deviation over $20$ draws: the variation sits in the holes.}
\label{fig:lincheck-inpaint}
\end{figure}

\textbf{Setup.}
Both operators act on the $16\times64\times64$ SD3.5 latent of a
$512\times512$ image. The \emph{random-hole mask}
$\rmH\rvz=\rmM\odot\rvz$ hides $20\%$ of the latent cells in
$16\times16$-pixel holes aligned to the latent grid
(Figure~\ref{fig:lincheck-inpaint}). The \emph{circular latent blur} blurs
each channel with a Gaussian of one latent cell ($8$ pixels); it is diagonal
in the Fourier basis, so the local system of Appendix~\ref{sec:app-cg-solve}
can be solved exactly at every step. Each image has one measurement
$\rvy=\rmH\rvz^\star+r\bxi$ with $r=0.01$. A calibration test needs a truth
that is itself a posterior draw, so we draw $\rvz^\star$ from the model's own
prior by running the sampler with nothing observed; we also report real FFHQ
faces, passed through the same final denoiser step as the draws. The sampler
is SD3.5-medium with $28$ steps, $P=1$, $\rho=1$, and $\lambda=0$. We compare
three anchors (Appendix~\ref{sec:app-anchor-family}): the perturbed prior
($\beta=1$), the arc, and the denoiser anchor ($\beta=0$). Each draws $20$
samples per image, with shared measurements and seeds, so the differences
are paired.

\textbf{Checks.}
If the sampler is correct, the truth behaves like one more draw. The
\emph{spread ratio} $S$ divides the RMS distance between draws by the RMS
distance from the draws to the truth; $S=1$ when calibrated, and below $1$
means the draws vary too little or are off-center. Unlike the spread in Table~\ref{tab:tractable},
it compares the draws with the truth. The \emph{rank} of the truth among the
$20$ draws is uniform when calibrated; TV is the distance of its histogram
from uniform. Both look at one coordinate at a time
and can miss structured errors, so we also divide the energy of the draws in
each Fourier band by that of the truth; $1$ is the right amount of detail.
Intervals are $95\%$ bootstrap intervals over images. The mask and the main
blur comparison use $100$ images; the CG sweep uses $64$.

\begin{table}[ht]
\centering
\footnotesize
\setlength{\tabcolsep}{4pt}
\caption{Checking the full sampler against a known posterior, with exact linear latent operators. \textbf{Reading guide:} spread and finest energy should be $1$, and TV $0$; spread below $1$ means the draws vary too little or are off-center, and finest energy above $1$ means detail that the truth does not have. Spread is the RMS distance between draws divided by the RMS distance from the draws to the truth (in the holes, for the mask). TV is the total-variation distance of the rank histogram of the truth from uniform. Finest energy is the energy of the draws in the finest Fourier band divided by that of the truth (Table~\ref{tab:lincheck-bands} gives all bands). Blocks 1 and 2 use $100$ images and block 3 uses $64$, each with $20$ draws. Bold (block 3): lowest LPIPS for each anchor; ---: not measured.}
\label{tab:lincheck}

\begin{tabular}{lcccccc}
\toprule
& \multicolumn{4}{c}{Truth drawn from the model's prior} & \multicolumn{2}{c}{Real FFHQ face} \\
\cmidrule(lr){2-5}\cmidrule(lr){6-7}
Sampler & spread & TV & finest energy & LPIPS & spread & LPIPS \\
\midrule
\multicolumn{7}{l}{\textbf{1. Exact update: random-hole mask}} \\
\textsc{DING} & 1.029 & 0.019 & --- & 0.038 & 1.085 & 0.045 \\
\paseo, $\beta{=}1$ & 1.029 & 0.020 & --- & 0.038 & 1.085 & 0.045 \\
\paseo, arc & 1.024 & 0.016 & --- & 0.037 & 1.077 & 0.044 \\
\paseo, denoiser anchor & 1.023 & 0.015 & --- & 0.037 & 1.077 & 0.044 \\
\midrule
\multicolumn{7}{l}{\textbf{2. Exact solve or a few CG steps: latent blur}} \\
$\beta{=}1$, exact solve & 1.010 & 0.007 & 14.54 & 0.735 & 0.928 & 0.277 \\
$\beta{=}1$, $C{=}5$ & 0.999 & 0.001 & 0.92 & 0.063 & 0.946 & 0.087 \\
arc, exact solve & 1.039 & 0.033 & 18.73 & 0.781 & 0.974 & 0.327 \\
arc, $C{=}5$ & 1.063 & 0.061 & 1.19 & 0.070 & 0.987 & 0.085 \\
denoiser, exact solve & 1.039 & 0.033 & 18.69 & 0.781 & 0.974 & 0.326 \\
denoiser, $C{=}5$ & 1.063 & 0.060 & 1.18 & 0.069 & 0.986 & 0.085 \\
\midrule
\multicolumn{7}{l}{\textbf{3. CG budget: latent blur}} \\
\multicolumn{7}{l}{\emph{Arc anchor}} \\
$C{=}1$ & 0.984 & --- & 0.88 & 0.085 & 0.897 & 0.156 \\
$C{=}2$ & 1.010 & --- & 0.95 & 0.074 & 0.939 & 0.117 \\
$C{=}3$ & 1.030 & --- & 1.01 & \textbf{0.068} & 0.961 & 0.099 \\
$C{=}5$ & 1.063 & --- & 1.19 & 0.069 & 0.988 & 0.083 \\
$C{=}8$ & 1.090 & --- & 1.61 & 0.110 & 1.008 & 0.079 \\
$C{=}16$ & 1.110 & --- & 4.81 & 0.469 & 1.018 & 0.131 \\
exact solve & 1.039 & --- & 18.93 & 0.799 & 0.975 & 0.336 \\
\addlinespace
\multicolumn{7}{l}{\emph{Perturbed prior, $\beta{=}1$}} \\
$C{=}1$ & 0.904 & --- & 0.70 & 0.099 & 0.848 & 0.169 \\
$C{=}2$ & 0.956 & --- & 0.82 & 0.079 & 0.906 & 0.122 \\
$C{=}3$ & 0.973 & --- & 0.86 & 0.070 & 0.925 & 0.103 \\
$C{=}5$ & 0.998 & --- & 0.92 & 0.063 & 0.947 & 0.085 \\
$C{=}8$ & 1.024 & --- & 1.01 & \textbf{0.061} & 0.964 & 0.074 \\
$C{=}16$ & 1.052 & --- & 1.35 & 0.085 & 0.981 & 0.072 \\
exact solve & 1.011 & --- & 14.76 & 0.751 & 0.930 & 0.286 \\
\bottomrule
\end{tabular}
\end{table}

\begin{figure}[ht]
\centering
\includegraphics[width=\linewidth]{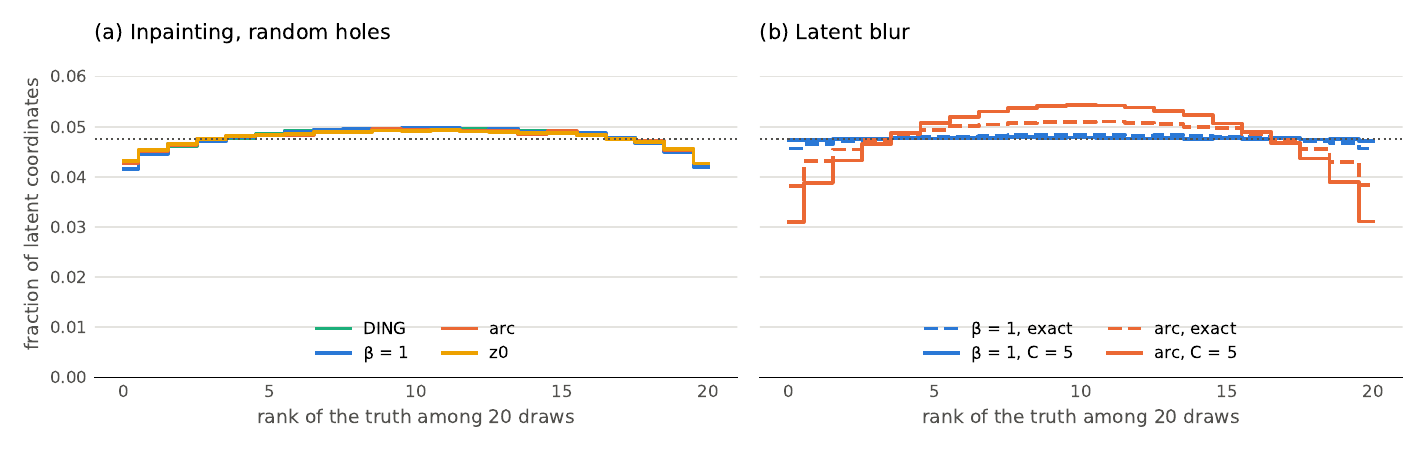}
\caption{Rank of the truth among $20$ draws, pooled over latent
coordinates and $100$ images, with truths from the model's prior. A flat
histogram at the dotted line is calibrated. (a) Inpainting: all samplers
overlap. (b) Blur: the exact solve looks as calibrated as $C=5$ here;
Table~\ref{tab:lincheck-bands} shows what these histograms miss.}
\label{fig:lincheck-ranks}
\end{figure}

\textbf{1. Does the sampler match the posterior when the update is exact?}
With a coordinate mask, one CG iteration solves the linear system exactly, so
\paseo\ with $\beta=1$ takes the same closed-form step as \textsc{DING}
(Appendix~\ref{sec:unif-ding}). Block~1 of Table~\ref{tab:lincheck}
confirms that the two agree, to within $\pm0.001$ in spread. Every sampler is within
$3\%$ of calibrated (spread $1.02$--$1.03$), and the rank histograms are
nearly flat (Figure~\ref{fig:lincheck-ranks}a). The small excess is shared
with \textsc{DING}, so it comes from the $28$-step discretization and the final denoiser
step, not from the measurement update. The anchors barely differ, because the correction is
zero inside the holes; the mask tests the outer loop, and the blur tests the
anchor and the solver. With real faces, every sampler, \textsc{DING}
included, reads about $1.08$, because the SD3.5 prior is broader than FFHQ.

\begin{figure}[ht]
\centering
\begin{tikzpicture}
\node[anchor=south west, inner sep=0] (blgrid) at (0,0) {\includegraphics[width=0.9\linewidth]{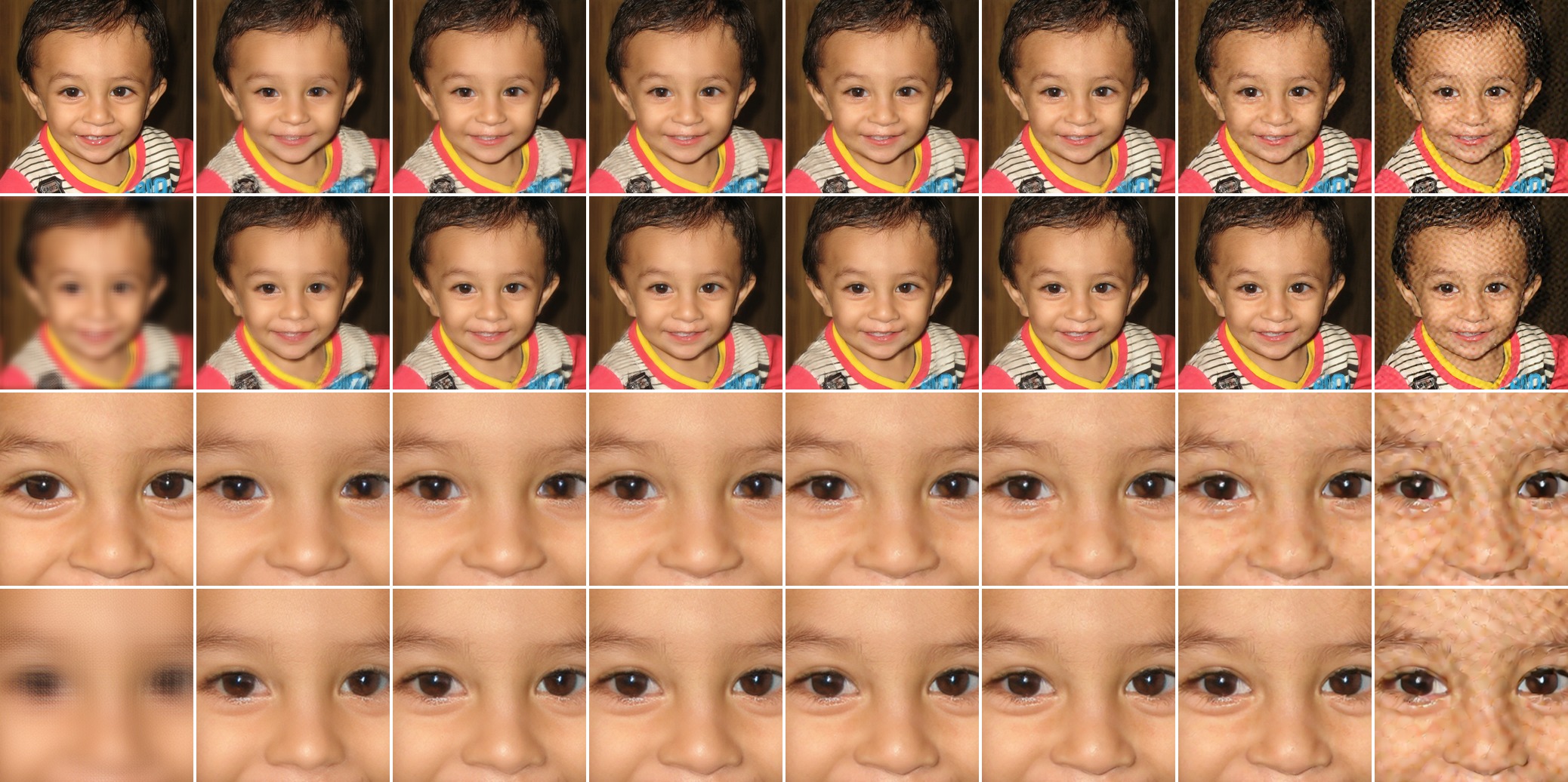}};
\foreach \f/\t in {0.061657/{Truth / meas.}, 0.186898/$C=1$, 0.312139/$C=2$, 0.437380/$C=3$, 0.562620/$C=5$, 0.687861/$C=8$, 0.813102/$C=16$, 0.938343/exact}
  \node[anchor=south, font=\scriptsize] at ($(blgrid.north west)!\f!(blgrid.north east)+(0,2pt)$) {\t};
\foreach \f/\t in {0.123552/arc, 0.374517/$\beta=1$, 0.625483/arc (zoom), 0.876448/$\beta=1$ (zoom)}
  \node[anchor=south, rotate=90, font=\scriptsize] at ($(blgrid.north west)!\f!(blgrid.south west)+(-2pt,0)$) {\t};
\end{tikzpicture}
\caption{Circular latent blur on one FFHQ face, one draw per CG budget. The
first column shows the ground truth and the blurred measurement (top two rows)
with the same zoom below. Few iterations give a smooth face; a few more give a
sharp one; many iterations and the exact solve add a mesh-like texture that
the truth does not have.}
\label{fig:lincheck-blur}
\end{figure}
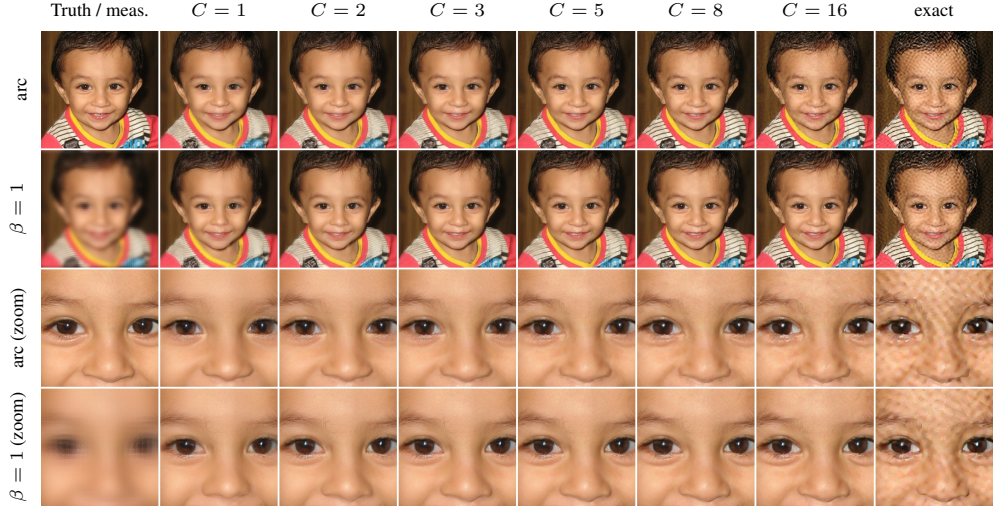

\textbf{2. What changes when we solve exactly instead of stopping CG early?}
The blur couples neighboring cells, so the solver now matters
(Table~\ref{tab:lincheck}, block~2). The exact solve adds a mesh-like texture
(Figure~\ref{fig:lincheck-blur}): its LPIPS is $0.73$--$0.78$, against
$0.06$--$0.07$ with $C=5$ CG iterations, and real faces show the same.
The spread ratio misses this, since the texture changes from draw to draw,
but the band energies show it (Table~\ref{tab:lincheck-bands}): in the finest
band, where the blur keeps $1\%$ of the signal, the exact solve has
$15$--$19$ times the energy of the truth, and $C=5$ has $0.9$--$1.2$. The
cause is the per-step prior. Early in sampling $\eta_z$ is large, so the
exact update fits even the directions that the blur almost erases and
amplifies the noise in them; the isotropic per-step Gaussian does not know
that natural latents carry little energy there. CG started from zero fits the
strongly observed directions first, so a few iterations leave the rest to the
diffusion prior. This does not contradict
Appendix~\ref{sec:app-tractable}, where the exact solve is the reference:
there the per-step Gaussian is the true prior, so fitting every direction is
correct.

\begin{table}[ht]
\centering
\footnotesize
\setlength{\tabcolsep}{4pt}
\caption{Energy of the draws per frequency band, divided by that of the truth, on the latent blur with truths from the model's prior ($100$ images). Blur keeps is the mean gain of the operator in the band. The exact solve adds content where the blur keeps almost nothing.}
\label{tab:lincheck-bands}
\resizebox{\linewidth}{!}{%
\begin{tabular}{lcccccc}
\toprule
& & \multicolumn{2}{c}{$\beta=1$} & \multicolumn{2}{c}{arc} \\
\cmidrule(lr){3-4}\cmidrule(lr){5-6}
Band (cycles per latent cell) & Blur keeps & exact & $C=5$ & exact & $C=5$ \\
\midrule
coarse ($0$--$0.1$) & 91\% & 1.00 {\scriptsize[1.00, 1.00]} & 1.00 {\scriptsize[1.00, 1.00]} & 1.00 {\scriptsize[1.00, 1.00]} & 1.00 {\scriptsize[1.00, 1.00]} \\
medium ($0.1$--$0.3$) & 41\% & 1.15 {\scriptsize[1.14, 1.16]} & 1.09 {\scriptsize[1.09, 1.10]} & 1.16 {\scriptsize[1.15, 1.17]} & 1.14 {\scriptsize[1.13, 1.15]} \\
fine ($0.3$--$0.4$) & 9\% & 2.56 {\scriptsize[2.49, 2.64]} & 0.79 {\scriptsize[0.79, 0.80]} & 2.84 {\scriptsize[2.74, 2.93]} & 1.19 {\scriptsize[1.18, 1.21]} \\
finest ($0.4$--$0.75$) & 1\% & 14.54 {\scriptsize[13.79, 15.30]} & 0.92 {\scriptsize[0.91, 0.93]} & 18.73 {\scriptsize[17.76, 19.69]} & 1.19 {\scriptsize[1.17, 1.20]} \\
\bottomrule
\end{tabular}}
\end{table}

\begin{figure}[ht]
\centering
\includegraphics[width=\linewidth]{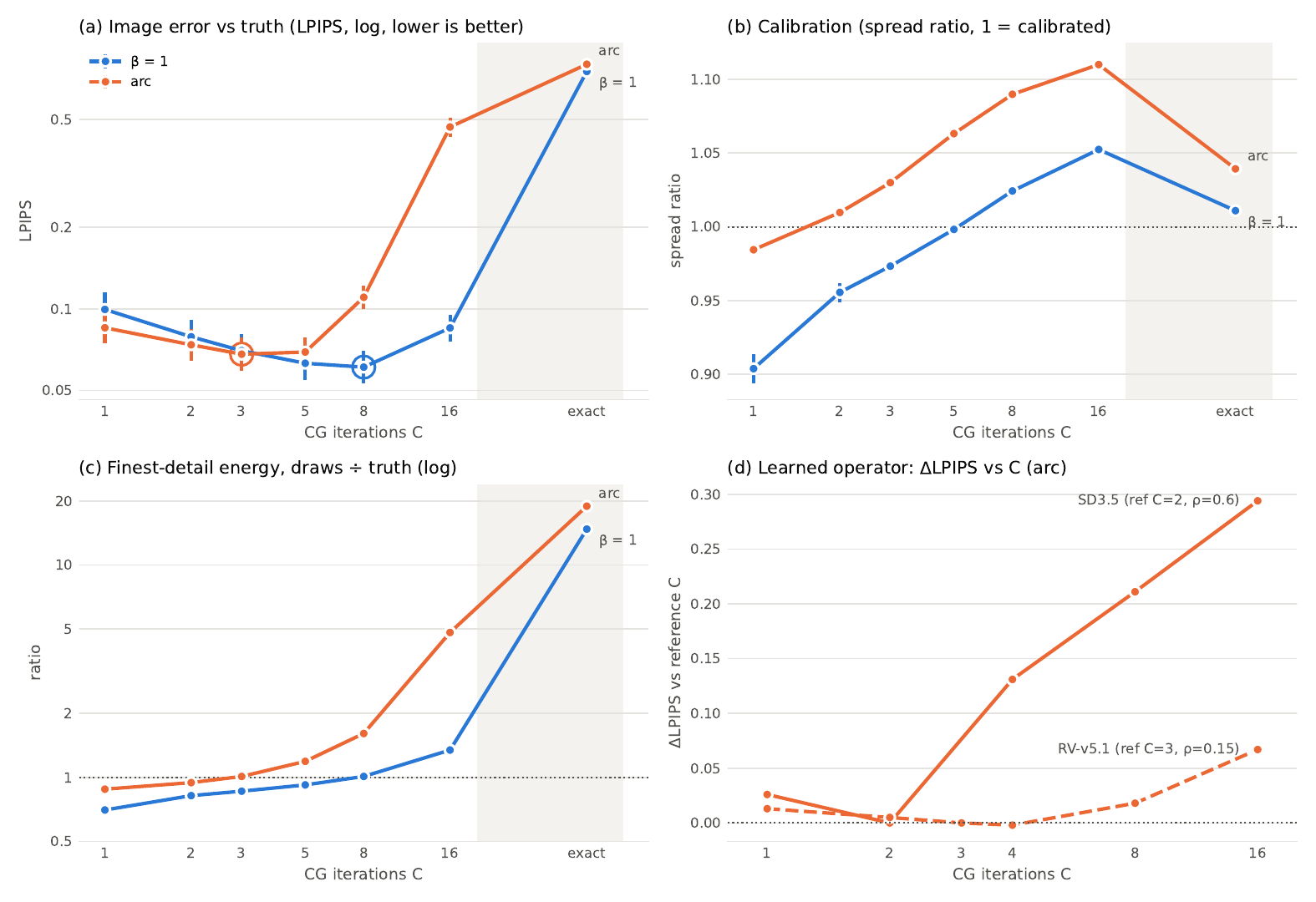}
\caption{CG budget on the exact latent blur (a--c, truths from the model's
prior, $64$ images, $95\%$ intervals) and on the learned-operator blur of the
main experiments (d, Tables~\ref{tab:ofat-sd35} and~\ref{tab:ofat-rv}). The
budget with the lowest LPIPS (circles) is where the fine-detail energy
crosses $1$, with the spread closest to $1$ one budget earlier; the arc is
best near the $C=2$--$3$ used with the learned operator.}
\label{fig:lincheck-sweep}
\end{figure}

\textbf{3. Which CG budget is best, and is it also closest to the posterior?}
Figure~\ref{fig:lincheck-sweep} and block~3 of Table~\ref{tab:lincheck} vary $C$
from $1$ to the exact solve. LPIPS is U-shaped: too few iterations give draws
that are too smooth and too similar, and too many add detail that the truth
does not have. The arc is best at $C=3$ and $\beta=1$ at $C=8$. At these
budgets the fine-detail energy is closest to $1$, and the spread is closest
one budget earlier. The budget that gives the best images is therefore also
close to the one that best matches the posterior. The learned operator on blur is best
at the same small budgets, $C=2$ with SD3.5 and $C=4$ with RV-v5.1
(Tables~\ref{tab:ofat-sd35} and~\ref{tab:ofat-rv}), although its curve rises
sooner; the preference for few iterations is therefore not caused by
operator error. $\beta=1$ needs more iterations because CG must also cancel
its fresh noise, so it trails the arc at small budgets, more so with the
learned operator (Table~\ref{tab:anchor-ablation}). With real faces the best
budget moves up ($C=8$ for the arc), but the exact solve is still the worst.

These checks use latent operators chosen so that the answer is
computable; they are not the pixel degradations of the main experiments and
say nothing direct about the learned operator. Truths from the prior test
calibration against the SD3.5 prior, not natural images, and we fix $P=1$,
$\rho=1$, and $\lambda=0$, whereas the learned-operator blur experiments use
relaxation ($\rho<1$).

\FloatBarrier
\subsection{Anchor choice}
\label{sec:app-anchor-ablation}

\subsubsection{Anchors at the same solver budget}

The anchor is the latent from which the measurement correction is taken
(Section~\ref{sec:method}, Appendix~\ref{sec:app-anchor-family}). Early in
sampling, the original perturbed-prior anchor can lie far from the denoiser's
estimate. Table~\ref{tab:anchor-ablation} asks how the choice between these
anchors affects reconstruction when the solver budget is held fixed.

\textbf{Defining the five anchors.}
At the first inner correction, the denoiser estimate is
$\rvz_0^{(0)}=\vmu_z+\eta_z\rvw'$, while the original perturbed-prior
anchor is $\tilde\vmu_z=\vmu_z+\eta_z\bxi_z$.
Here $\vmu_z$ and $\eta_z$ are the center and standard deviation of the
clean-latent belief, $\rvw'$ is the bridge noise already used by the
denoiser, and $\bxi_z$ is an independent standard Gaussian draw.
All five choices can therefore be written as
\[
\rvz_a=\vmu_z+\eta_z(c_1\rvw'+c_2\bxi_z).
\]
The coefficients $c_1$ and $c_2$ specify how much bridge noise we reuse
and how much fresh noise we add. In the same order as the table columns,
\[
\begin{aligned}
\text{Arc interpolation:}\quad &(c_1,c_2)=(\sqrt{1-\beta_k^2},\beta_k),\\
\text{Denoiser estimate:}\quad &(c_1,c_2)=(1,0),\\
\text{Linear interpolation:}\quad &(c_1,c_2)=(1-\beta_k,\beta_k),\\
\text{Denoiser + noise:}\quad &(c_1,c_2)=(1,1),\\
\text{Perturbed prior mean:}\quad &(c_1,c_2)=(0,1).
\end{aligned}
\]
For both interpolation rules, $\beta_k=\eta_{z,\min}/\eta_{z,k}$ is
fixed by the diffusion noise schedule, where $\eta_{z,\min}$ is the
smallest scheduled belief width; it adds no new hyperparameter.
Arc and linear interpolation connect the denoiser estimate to the
perturbed prior mean along the two paths in
Appendix~\ref{sec:app-anchor-family}. ``Denoiser + noise'' instead adds
$\eta_z\bxi_z$ to the denoiser estimate.
For later inner corrections, the current iterate replaces the initial
denoiser estimate, as described in \eqref{eq:app-anchor-inner}.

The perturbed prior mean is also called the ``exact anchor'' because
it is the anchor used in the ideal derivation. Exact posterior sampling
additionally requires the model and solve conditions stated in
Section~\ref{sec:method}.

At the tested settings, the denoiser anchor has lower LPIPS than the arc
on all five RV-v5.1 tasks. On SD3.5, the arc, denoiser and linear blend are
close on most tasks, with a larger denoiser advantage on JPEG. The original
perturbed-prior anchor generally performs worse than the arc, although it
slightly improves RV-v5.1 inpainting. Thus reducing the fresh perturbation
can help reconstruction at this budget, but the arc is not uniformly the
best choice. Inpainting uses the supplied hard mask for every column.

{
\begin{table}[!htbp]
\centering
\footnotesize

\setlength{\tabcolsep}{4pt}
\caption{How does the anchor affect reconstruction quality?
The table compares LPIPS ($\downarrow$) on the first $200$ FFHQ images at the
reference solver settings for each task and model, shared by all five
anchors within each row. Inpainting uses the hard mask.
Bold marks the lowest LPIPS and underline the second-lowest distinct value
within each row; ties at the displayed precision share the same marking.}
\label{tab:anchor-ablation}
\vspace{6pt}
\begin{tabular}{@{}clrrrrr@{}}
\toprule
Prior & Task & \shortstack{Arc\\interpolation} & \shortstack{Denoiser\\estimate} & \shortstack{Linear\\interpolation} & \shortstack{Denoiser\\+ noise} & \shortstack{Perturbed\\prior mean} \\
\midrule
\multirow{5}{*}{\rotatebox[origin=c]{90}{RV-v5.1}} & SR $\times4$ & \underline{0.135} & \textbf{0.134} & 0.145 & 0.161 & 0.214 \\
 & SR $\times8$ & \underline{0.205} & \textbf{0.198} & 0.281 & 0.667 & 0.374 \\
 & Blur & \underline{0.157} & \textbf{0.151} & 0.214 & 0.534 & 0.292 \\
 & Inpaint & 0.140 & \textbf{0.132} & 0.147 & 0.430 & \underline{0.139} \\
 & JPEG & \underline{0.255} & \textbf{0.223} & 0.362 & 0.689 & 0.679 \\
\midrule
\multirow{5}{*}{\rotatebox[origin=c]{90}{SD3.5}} & SR $\times4$ & \underline{0.180} & \textbf{0.178} & \textbf{0.178} & 0.534 & 0.659 \\
 & SR $\times8$ & \textbf{0.328} & \textbf{0.328} & \underline{0.330} & 0.628 & 0.843 \\
 & Blur & \textbf{0.197} & \underline{0.198} & 0.199 & 0.383 & 0.233 \\
 & Inpaint & \textbf{0.117} & \textbf{0.117} & \underline{0.118} & 0.322 & 0.141 \\
 & JPEG & \underline{0.347} & \textbf{0.340} & \underline{0.347} & 0.596 & 0.685 \\
\bottomrule
\end{tabular}
\end{table}}

\FloatBarrier
\subsection{Sensitivity to solver settings}
\label{sec:ofat}

The practical sampler requires choices about how many corrections to
make, how accurately to solve each linear system, and how far to move at
each correction. Reporting results at selected settings does not show
how much reconstruction quality depends on these choices. This section
identifies which settings need careful tuning and which have little effect
over the tested range, helping readers understand the sensitivity of the
reported performance.

Tables~\ref{tab:ofat-rv} and~\ref{tab:ofat-sd35} change one setting at a time,
keeping the others at the reference configuration for that task and prior.
Each configuration is evaluated on the first $100$ FFHQ images. All runs
use the arc anchor. Inpainting uses the supplied hard mask, so those
rows describe the masked variant rather than the learned-weight variant
in the main results (Appendix~\ref{sec:inpaint-variants}).

The row groups correspond to the operations in Algorithm~\ref{alg:silo-jding}.
$P$ counts how many times the operator is linearized and a correction is
proposed within one reverse step. $C$ limits the CG iterations used to solve
each resulting linear system. The fraction $\rho$ controls how far we move
toward each proposal. The noise scale $r$ controls trust in the learned
operator, while $\lambda$ adds noise inflation to the solve for stability.
The final group varies the fresh noise: $\bxi_y$ is the measurement
perturbation and $\bxi_z$ is the prior perturbation. ``$\bxi_y$ only''
retains measurement noise, ``$\bxi_z$ only'' retains prior noise, and
``neither'' removes both fresh perturbations. All three choices retain
the bridge randomness.

\textbf{Reading the tables.} Each column starts with its reference LPIPS.
Entries below report $\Delta\mathrm{LPIPS}=\mathrm{LPIPS}_{\mathrm{tested}}
-\mathrm{LPIPS}_{\mathrm{reference}}$: negative is better, positive is worse,
and zero means no change at the reported precision. For example, SD3.5 SR $\times8$ starts at $0.314$; the entry $-0.053$
for one CG iteration corresponds to LPIPS $0.261$. Bold zero identifies a
reference setting included in the tested grid. If a parameter group has no
bold zero for a task, that task's reference setting was not among the
levels tested in that group.

\textbf{Main observations.} Quality is sensitive to the correction
fraction and the number of inner updates; changing the fraction can increase
LPIPS by more than $0.2$. More CG iterations can also worsen reconstruction.
The effect of fresh noise depends on the task: retaining only the measurement
perturbation improves RV-v5.1 JPEG by $0.027$ LPIPS. These are individual
changes around the reference settings, not a joint optimization of all
parameters. Appendix~\ref{sec:cg-diagnostics} separately examines how accurately
the linear systems are solved.

{
\begin{table}[!htbp]
\centering
\footnotesize

\setlength{\tabcolsep}{4pt}
\caption{Solver sensitivity on RV-v5.1: one setting changes at a time
on the first $100$ FFHQ images, using the arc anchor and a hard mask for
inpainting. The first row gives reference LPIPS; subsequent rows give
$\Delta\mathrm{LPIPS}$ (negative: better; positive: worse).
Bold zero marks the reference setting, not the best result.
These are individual changes, not joint retuning.}
\label{tab:ofat-rv}
\vspace{6pt}
\begin{tabular}{@{}llrrrrr@{}}
\toprule
Setting & Level & SR $\times4$ & SR $\times8$ & Blur & Inpaint & JPEG \\
\midrule
\multicolumn{2}{l}{Reference LPIPS} & 0.131 & 0.199 & 0.150 & 0.138 & 0.248 \\
\midrule
\multicolumn{7}{l}{\textbf{Relinearization count $P$}} \\
 & 1 & +0.127 & +0.182 & +0.176 & +0.002 & +0.244 \\
 & 2 & +0.021 & +0.042 & +0.050 & -0.007 & +0.066 \\
 & 3 & \textbf{0} & +0.012 & +0.021 & \textbf{0} & +0.015 \\
 & 5 & +0.007 & -0.006 & \textbf{0} & +0.096 & +0.011 \\
\midrule
\multicolumn{7}{l}{\textbf{CG iteration budget $C$}} \\
 & 1 & +0.007 & +0.006 & +0.013 & -0.002 & \textbf{0} \\
 & 2 & +0.002 & \textbf{0} & +0.005 & -0.002 & -0.023 \\
 & 4 & +0.002 & +0.000 & -0.002 & -0.000 & +0.053 \\
 & 8 & +0.043 & +0.055 & +0.018 & -0.000 & +0.186 \\
 & 16 & +0.102 & +0.143 & +0.067 & +0.000 & +0.358 \\
\midrule
\multicolumn{7}{l}{\textbf{Correction fraction $\rho$}} \\
 & 0.15 & \textbf{0} & \textbf{0} & \textbf{0} & +0.025 & +0.051 \\
 & 0.3 & +0.027 & +0.012 & +0.014 & +0.000 & \textbf{0} \\
 & 0.6 & +0.145 & +0.268 & +0.179 & \textbf{0} & +0.220 \\
 & 1.0 & +0.249 & +0.448 & +0.308 & +0.102 & +0.422 \\
\midrule
\multicolumn{7}{l}{\textbf{Assumed operator-noise scale $r$}} \\
 & 0.005 & -0.000 & -0.000 & \textbf{0} & -0.000 & +0.002 \\
 & 0.01 & -0.000 & -0.000 & +0.000 & -0.001 & +0.002 \\
 & 0.02 & -0.001 & \textbf{0} & +0.000 & -0.000 & +0.002 \\
 & 0.05 & \textbf{0} & +0.000 & +0.000 & \textbf{0} & +0.002 \\
 & 0.08 & +0.001 & -0.000 & +0.001 & +0.001 & \textbf{0} \\
\midrule
\multicolumn{7}{l}{\textbf{Additional damping $\lambda$}} \\
 & 0.0 & +0.002 & -0.001 & +0.003 & +0.001 & +0.056 \\
 & 0.005 & \textbf{0} & \textbf{0} & \textbf{0} & \textbf{0} & +0.046 \\
 & 0.02 & +0.007 & +0.006 & +0.007 & +0.001 & +0.022 \\
 & 0.05 & +0.026 & +0.016 & +0.024 & +0.003 & \textbf{0} \\
\midrule
\multicolumn{7}{l}{\textbf{Fresh perturbations retained}} \\
 & $\bxi_y$ only & +0.003 & +0.012 & +0.006 & +0.003 & -0.027 \\
 & $\bxi_z$ only & +0.005 & +0.002 & +0.002 & -0.001 & -0.000 \\
 & neither & +0.002 & +0.025 & +0.011 & +0.003 & -0.019 \\
\bottomrule
\end{tabular}
\end{table}}

{
\begin{table}[!htbp]
\centering
\footnotesize

\setlength{\tabcolsep}{4pt}
\caption{Solver sensitivity on SD3.5: one setting changes at a time
on the first $100$ FFHQ images, using the arc anchor and a hard mask for
inpainting. The first row gives reference LPIPS; subsequent rows give
$\Delta\mathrm{LPIPS}$ (negative: better; positive: worse).
Bold zero marks the reference setting, not the best result.
These are individual changes, not joint retuning.
For SR $\times8$, the effective $r$ is half the listed value.}
\label{tab:ofat-sd35}
\vspace{6pt}
\begin{tabular}{@{}llrrrrr@{}}
\toprule
Setting & Level & SR $\times4$ & SR $\times8$ & Blur & Inpaint & JPEG \\
\midrule
\multicolumn{2}{l}{Reference LPIPS} & 0.173 & 0.314 & 0.189 & 0.117 & 0.335 \\
\midrule
\multicolumn{7}{l}{\textbf{Relinearization count $P$}} \\
 & 1 & +0.039 & \textbf{0} & \textbf{0} & \textbf{0} & +0.141 \\
 & 2 & \textbf{0} & +0.128 & +0.013 & +0.001 & +0.040 \\
 & 3 & +0.025 & +0.135 & +0.038 & +0.001 & \textbf{0} \\
 & 5 & +0.050 & +0.140 & +0.057 & +0.001 & -0.011 \\
\midrule
\multicolumn{7}{l}{\textbf{CG iteration budget $C$}} \\
 & 1 & -0.000 & -0.053 & +0.026 & +0.004 & +0.086 \\
 & 2 & \textbf{0} & \textbf{0} & \textbf{0} & +0.001 & +0.020 \\
 & 4 & +0.021 & +0.365 & +0.131 & -0.000 & -0.001 \\
 & 8 & +0.097 & +0.304 & +0.211 & +0.000 & +0.025 \\
 & 16 & +0.133 & +0.310 & +0.294 & +0.000 & +0.074 \\
\midrule
\multicolumn{7}{l}{\textbf{Correction fraction $\rho$}} \\
 & 0.15 & +0.165 & +0.214 & +0.268 & +0.240 & +0.074 \\
 & 0.3 & +0.074 & +0.078 & +0.090 & +0.081 & \textbf{0} \\
 & 0.6 & \textbf{0} & +0.003 & \textbf{0} & +0.017 & +0.045 \\
 & 1.0 & +0.076 & \textbf{0} & +0.051 & \textbf{0} & +0.096 \\
\midrule
\multicolumn{7}{l}{\textbf{Assumed operator-noise scale $r$}} \\
 & 0.005 & +0.002 & +0.008 & +0.004 & -0.000 & +0.013 \\
 & 0.01 & \textbf{0} & +0.002 & +0.005 & \textbf{0} & +0.010 \\
 & 0.02 & +0.001 & -0.004 & \textbf{0} & +0.001 & \textbf{0} \\
 & 0.05 & +0.003 & \textbf{0} & +0.004 & +0.004 & +0.001 \\
 & 0.08 & +0.006 & +0.008 & +0.006 & +0.007 & -0.004 \\
\midrule
\multicolumn{7}{l}{\textbf{Additional damping $\lambda$}} \\
 & 0.0 & \textbf{0} & \textbf{0} & \textbf{0} & \textbf{0} & \textbf{0} \\
 & 0.005 & +0.002 & +0.001 & +0.004 & +0.006 & -0.002 \\
 & 0.02 & +0.004 & -0.003 & +0.014 & +0.012 & -0.000 \\
 & 0.05 & +0.009 & -0.007 & +0.028 & +0.019 & +0.002 \\
\midrule
\multicolumn{7}{l}{\textbf{Fresh perturbations retained}} \\
 & $\bxi_y$ only & +0.007 & +0.016 & +0.006 & +0.004 & -0.003 \\
 & $\bxi_z$ only & +0.007 & +0.011 & +0.003 & +0.004 & -0.005 \\
 & neither & +0.002 & -0.013 & -0.002 & +0.010 & -0.001 \\
\bottomrule
\end{tabular}
\end{table}}

\FloatBarrier
\textbf{One-CG quality--cost comparison.}
Table~\ref{tab:cg-cost-quality} expands the SD3.5 $C=1$ row into absolute
Alex-LPIPS, paired uncertainty, sampler time, and decoded peak
allocated memory. The quality comparison uses the first $100$ FFHQ images,
whereas the time columns come from a separate $20$-image run of the two-call sampler, both arms in the same job, and memory from the earlier synchronized benchmark; neither is remeasured on the larger set. At the paired 95\%
interval, one CG iteration is worse for blur, inpainting, and JPEG,
indistinguishable from the reference for SR $\times4$, and better for
SR $\times8$. It reduces sampler time by $2.9$--$25.1\%$, while decoded peak
allocated memory is unchanged at the reported precision. This is a controlled
ablation, not a recommendation of one common budget for every task.

{
\begin{table}[!htbp]
\centering
\scriptsize
\setlength{\tabcolsep}{2.4pt}
\caption{Full-derivative ($C=1$) quality--cost ablation for SD3.5 at
seed $0$, using the arc anchor and the supplied hard mask for inpainting.
With zero-initialized CG, $C=1$ is the full-derivative case: its single
iterate gives a correction in the complete Jacobian-transpose measurement
derivative direction $\rmJ^T\rvb$, scaled by the CG line-search coefficient.
It is not a converged linear-system solve. Alex-LPIPS
($\downarrow$) uses the first $100$ FFHQ images; the interval is a paired
bootstrap over images. Sampler time is the two-call sampler's mean over the first $20$ FFHQ images (one run per arm, both arms in the same job); peak memory comes unchanged from the earlier synchronized benchmark. The reference uses the task-specific CG budget shown;
the full-derivative case keeps every other setting fixed, including the native
correction fraction $\rho$. The LPIPS difference is full derivative minus
reference.}
\label{tab:cg-cost-quality}
\vspace{6pt}
\resizebox{\linewidth}{!}{%
\begin{tabular}{@{}lrrcccrrrrrr@{}}
\toprule
 & & & \multicolumn{3}{c}{Alex-LPIPS ($n=100$)}
 & \multicolumn{4}{c}{Sampler cost ($20$ images, two-call sampler)}
 & \multicolumn{2}{c}{Decoded peak allocated (GiB)} \\
\cmidrule(lr){4-6}\cmidrule(lr){7-10}\cmidrule(lr){11-12}
Task & Native $\rho$ & Ref. $C$ & Ref. & \shortstack{Full deriv.\\($C=1$)}
 & $\Delta$ [95\% CI] & Ref. (s) & \shortstack{Full deriv.\\($C=1$) (s)}
 & Reduction & Speedup & Ref. & \shortstack{Full deriv.\\($C=1$)} \\
\midrule
SR $\times4$ & 0.6 & 2 & 0.1727 & 0.1723
 & $-0.0004$ [$-0.0030$, $+0.0024$]
 & 11.461 & 9.428 & 17.7\% & $1.216\times$ & 16.764 & 16.764 \\
SR $\times8$ & 1.0 & 2 & 0.3136 & 0.2606
 & $-0.0530$ [$-0.0591$, $-0.0470$]
 & 8.997 & 7.827 & 13.0\% & $1.149\times$ & 16.751 & 16.751 \\
Blur & 0.6 & 2 & 0.1889 & 0.2151
 & $+0.0262$ [$+0.0196$, $+0.0330$]
 & 8.467 & 8.222 & 2.9\% & $1.030\times$ & 16.751 & 16.751 \\
Inpaint & 1.0 & 5 & 0.1172 & 0.1212
 & $+0.0040$ [$+0.0028$, $+0.0054$]
 & 10.716 & 8.029 & 25.1\% & $1.335\times$ & 16.794 & 16.794 \\
JPEG & 0.3 & 3 & 0.3350 & 0.4212
 & $+0.0861$ [$+0.0760$, $+0.0963$]
 & 14.577 & 10.980 & 24.7\% & $1.328\times$ & 16.784 & 16.784 \\
\bottomrule
\end{tabular}%
}
\end{table}
}

\FloatBarrier
\subsection{Computational budget and reconstruction quality}
\label{sec:computational-budget}

To assess the sampler's efficiency, we need to understand what extra
computation buys in reconstruction quality. We can take more reverse
sampling steps, apply more measurement corrections within each step, or
allow more iterations to solve each correction. These choices increase
the work in different ways, so we examine where additional computation
helps and how much the limited solver leaves unresolved.

We denote the number of reverse steps by $K$, the number of measurement
corrections per step by $P$, and the maximum CG iterations per correction
by $C$. We first measure how changing $K$ affects quality and runtime.
We then check how closely the solutions returned by the limited CG solver
satisfy their linear equations.

Both diagnostics use the arc anchor.
Each experiment starts from the reference settings for its task and prior.
All inpainting runs here use the supplied hard mask, whereas the main
results use learned weights (Appendix~\ref{sec:inpaint-variants}).

\subsubsection{Reverse-step count and runtime}
\label{sec:steps}

Table~\ref{tab:pareto} asks whether spending more reverse steps improves
reconstruction. It compares $K=14,28,56,112$, while keeping all other solver
settings at those selected for $K=28$. Each step-count group shows LPIPS, PSNR,
and seconds per image side by side; its header also gives the number of
denoiser evaluations.

More steps consistently take more time, but their quality benefit depends
on the task and metric. The $28$-step schedule is competitive on most tasks.
For SD3.5, $56$ steps improves LPIPS on inpainting and JPEG, while blur ties
at the displayed precision. PSNR can improve even when LPIPS worsens.
These results support the short schedule at the tested settings; they do not
establish the best quality attainable after retuning each schedule. The reported timings are the two-call sampler's, one job per row on the first $20$ images, and are separate from the production timings in Table~\ref{tab:ffhq-models}.

{
\begin{table}[!htbp]
\centering
\scriptsize

\setlength{\tabcolsep}{3pt}
\caption{Reconstruction quality and runtime as the reverse-step count changes,
using the same first $200$ FFHQ images per configuration. Other solver settings
are fixed at those selected for $K=28$. All runs use the arc anchor;
inpainting uses the hard mask. Timings are separate from Table~\ref{tab:ffhq-models}. PSNR is in dB and time in seconds per image.
For each task and each quality metric, bold marks the best and underline
the second-best distinct value across step counts; ties at the reported
precision share the same marking. Denoiser-evaluation counts and the Time column are those of the two-call sampler ($2(K-2)+1$ evaluations for $K$ steps); time is the mean over the first $20$ images, one job per row.}
\label{tab:pareto}
\vspace{6pt}
\begin{tabular}{@{}cl*{4}{rrr}@{}}
\toprule
 & & \multicolumn{3}{c}{\shortstack{14 steps\\25 denoiser evaluations}} & \multicolumn{3}{c}{\shortstack{28 steps\\53 denoiser evaluations}} & \multicolumn{3}{c}{\shortstack{56 steps\\109 denoiser evaluations}} & \multicolumn{3}{c}{\shortstack{112 steps\\221 denoiser evaluations}} \\
\cmidrule(lr){3-5}\cmidrule(lr){6-8}\cmidrule(lr){9-11}\cmidrule(lr){12-14}
Prior & Task & LPIPS $\downarrow$ & PSNR $\uparrow$ & Time $\downarrow$ & LPIPS $\downarrow$ & PSNR $\uparrow$ & Time $\downarrow$ & LPIPS $\downarrow$ & PSNR $\uparrow$ & Time $\downarrow$ & LPIPS $\downarrow$ & PSNR $\uparrow$ & Time $\downarrow$ \\
\midrule
\multirow{5}{*}{\rotatebox[origin=c]{90}{RV-v5.1}} & SR $\times4$ & 0.213 & 26.6 & 6 & \textbf{0.135} & \underline{28.1} & 11 & \underline{0.163} & \textbf{28.4} & 23 & 0.238 & \underline{28.1} & 46 \\
 & SR $\times8$ & 0.272 & 25.5 & 5 & \textbf{0.205} & 26.8 & 12 & \underline{0.224} & \textbf{27.1} & 24 & 0.303 & \underline{26.9} & 47 \\
 & Blur & 0.216 & 26.6 & 8 & \textbf{0.157} & 27.8 & 17 & \underline{0.165} & \textbf{28.4} & 34 & 0.205 & \underline{28.3} & 70 \\
 & Inpaint & \underline{0.146} & 21.5 & 15 & \textbf{0.140} & 21.5 & 30 & 0.152 & \underline{21.8} & 65 & 0.158 & \textbf{22.0} & 127 \\
 & JPEG & 0.438 & 20.5 & 4 & \textbf{0.255} & 24.6 & 8 & \underline{0.295} & \textbf{25.9} & 17 & 0.410 & \underline{25.5} & 34 \\
\midrule
\multirow{5}{*}{\rotatebox[origin=c]{90}{SD3.5}} & SR $\times4$ & 0.235 & 29.3 & 5 & \textbf{0.180} & \underline{30.0} & 11 & \underline{0.203} & \textbf{30.1} & 22 & 0.220 & \textbf{30.1} & 46 \\
 & SR $\times8$ & 0.414 & \underline{26.3} & 4 & \textbf{0.328} & \textbf{27.3} & 8 & \underline{0.388} & \textbf{27.3} & 17 & 0.428 & \textbf{27.3} & 38 \\
 & Blur & 0.292 & 28.2 & 4 & \textbf{0.197} & \underline{29.6} & 9 & \textbf{0.197} & \textbf{30.2} & 18 & \underline{0.230} & \textbf{30.2} & 36 \\
 & Inpaint & 0.145 & 21.5 & 6 & \underline{0.117} & \underline{22.6} & 11 & \textbf{0.111} & \textbf{23.0} & 23 & \underline{0.117} & \textbf{23.0} & 45 \\
 & JPEG & 0.478 & 22.5 & 7 & 0.347 & 25.1 & 14 & \textbf{0.338} & \underline{25.8} & 32 & \underline{0.343} & \textbf{26.2} & 60 \\
\bottomrule
\end{tabular}
\end{table}}

\FloatBarrier

\subsubsection{Accuracy of the truncated CG solves}
\label{sec:cg-diagnostics}

\begin{figure}[ht]
\centering
\includegraphics[width=0.9\linewidth]{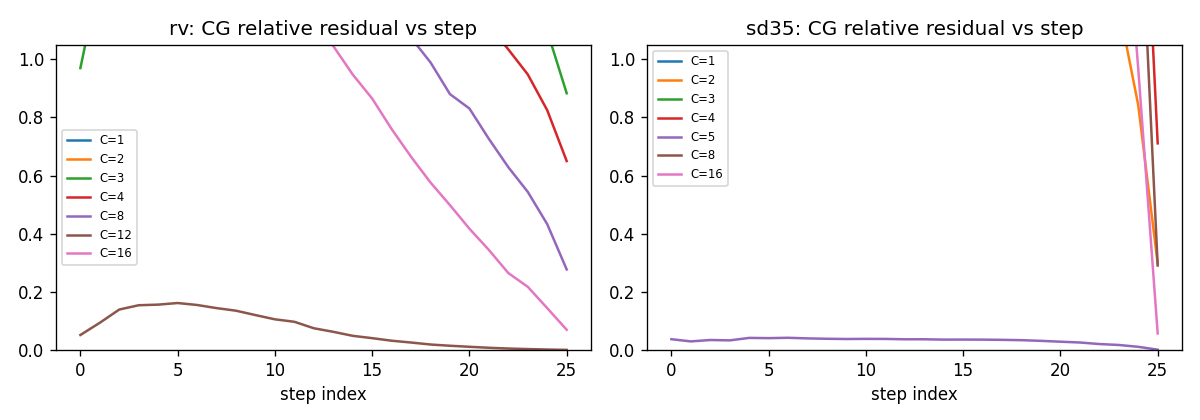}
\caption{Mean relative CG residual by step index, pooled by prior and
CG budget. The plot shows only the range up to $1.05$; larger values are
clipped, and Table~\ref{tab:cg} reports the pooled summaries.
Because task and parameter mixtures differ between curves, this is not a
controlled comparison of CG budgets on one fixed problem.}
\label{fig:cg}
\end{figure}

Reconstruction scores alone do not tell us how accurately the sampler
solves its internal linear systems. This matters because the ideal update
assumes an exact solve, while the practical sampler limits the number of
CG iterations. We therefore measure the error remaining in the linear
equations when CG stops. Table~\ref{tab:cg} summarizes these errors, and
Figure~\ref{fig:cg} shows how they vary over the sampling steps.

\textbf{Reading the residuals.} For a system with matrix $\emS_j$ and
right-hand side $\rvb$, let $\rvv$ be the solution returned by CG. Its
\emph{relative residual} is
\[
\frac{\|\emS_j\rvv-\rvb\|}{\|\rvb\|}.
\]
This measures the remaining error in the equation relative to the size of
its right-hand side. Zero means an exact solve; $1$ is the residual at the
zero initial solution, and $0.5$ means half that initial residual.
A value above $1$ means that the residual is larger than at initialization.
The table reports the mean, median, and fraction of solves with relative
residual above $0.5$. The logged-solve counts use $k$ for thousands.

These summaries combine solves from the parameter sweeps, grouping them
by prior and CG budget. Most groups have mean ratios above $1$. The
RV-v5.1 $C=12$ and SD3.5 $C=5$ groups have much smaller means, $0.07$ and
$0.03$; these budgets are used for inpainting. However, the recorded groups
lack task identifiers and contain different mixtures of tasks and settings.
Comparing their rows or the curves in Figure~\ref{fig:cg} therefore cannot
isolate what increasing $C$ does to one fixed problem. A small mean also
does not imply that every solve converged.

The practical solves often fall short of the exact-solve assumption in
\eqref{eq:local-perturb-and-solve}. The conditioning estimates in
Table~\ref{tab:jacobian} suggest potential numerical difficulty, without
establishing its cause. These measurements concern the linear equations;
the separate sensitivity experiment in Appendix~\ref{sec:ofat} measures the
quality of the final images.

{
\begin{table}[!htbp]
\centering
\footnotesize

\setlength{\tabcolsep}{4pt}
\caption{Accuracy of the truncated CG solves, grouped by prior and CG
iteration budget $C$. Smaller relative residuals indicate more accurate
solutions; $k$ denotes thousands of logged solves. Rows pool different
tasks and parameter settings, so their differences do not isolate the
effect of changing $C$ alone.}
\label{tab:cg}
\vspace{6pt}
\begin{tabular}{@{}ccrrrr@{}}
\toprule
Prior & \shortstack{CG budget\\$C$} & \shortstack{Logged\\solves} & \shortstack{Mean relative\\residual $\downarrow$} & \shortstack{Median relative\\residual $\downarrow$} & \shortstack{Fraction with\\residual $>0.5$ $\downarrow$} \\
\midrule
\multirow{7}{*}{\rotatebox[origin=c]{90}{RV-v5.1}} & 1 & 202k & 2.44 & 2.65 & 0.97 \\
 & 2 & 202k & 1.75 & 1.32 & 0.95 \\
 & 3 & 306k & 1.42 & 1.27 & 0.98 \\
 & 4 & 49k & 1.41 & 1.31 & 0.84 \\
 & 8 & 49k & 1.26 & 1.14 & 0.74 \\
 & 12 & 122k & 0.07 & 0.03 & 0.00 \\
 & 16 & 49k & 1.01 & 0.78 & 0.60 \\
\midrule
\multirow{7}{*}{\rotatebox[origin=c]{90}{SD3.5}} & 1 & 20k & 3.25 & 0.96 & 0.79 \\
 & 2 & 221k & 1.19 & 0.97 & 0.92 \\
 & 3 & 122k & 5.56 & 5.69 & 0.99 \\
 & 4 & 20k & 2.71 & 1.64 & 0.85 \\
 & 5 & 59k & 0.03 & 0.02 & 0.00 \\
 & 8 & 20k & 2.77 & 1.87 & 0.84 \\
 & 16 & 20k & 2.66 & 1.70 & 0.81 \\
\bottomrule
\end{tabular}
\end{table}}

\FloatBarrier
\subsection{Repeatability and sample variation}
\label{sec:diagnostics}

The reconstruction pipeline uses random draws, so repeating an evaluation
can change both its average scores and individual reconstructions.
We examine variation across eight seeded runs. In the newly rerun
inpainting configurations, measurement construction is held fixed and only
the sampling seed changes; the retained non-inpainting results use the legacy
joint-seed protocol described below.

This experiment uses the arc anchor and the reference settings for each
task and prior. Its two inpainting rows use the mask-free soft likelihood
weights from the main inpainting results; these weights are examined in
Appendix~\ref{sec:inpaint-variants}.

\FloatBarrier
\subsubsection{Variation across repeated seeded reconstructions}
\label{sec:diversity}

\begin{figure}[ht]
\centering
\begin{tikzpicture}
\node[anchor=south west, inner sep=0] (drawgrid) at (0,0) {\includegraphics[width=0.81\linewidth]{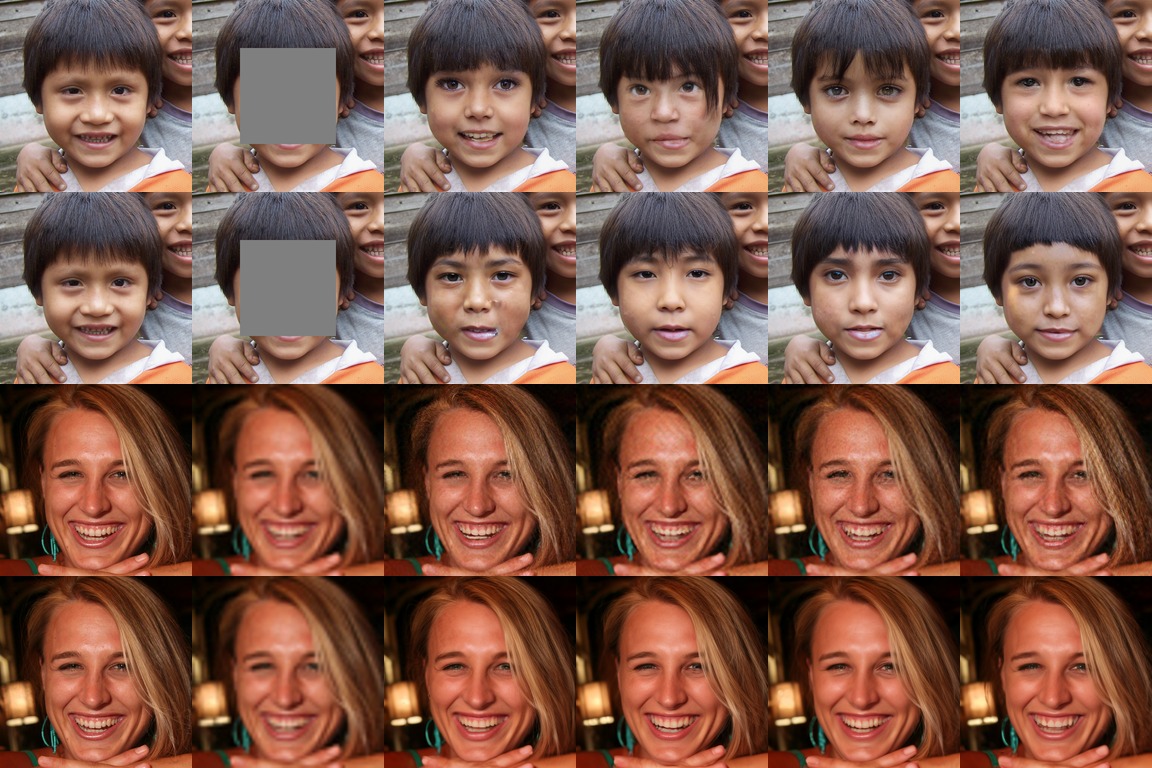}};
\node[anchor=south, font=\scriptsize] at ($(drawgrid.north west)!0.083333!(drawgrid.north east)+(0,2pt)$) {Ground truth};
\node[anchor=south, font=\scriptsize] at ($(drawgrid.north west)!0.250000!(drawgrid.north east)+(0,2pt)$) {Measurement};
\node[anchor=south, align=center, font=\scriptsize] at ($(drawgrid.north west)!0.416667!(drawgrid.north east)+(0,2pt)$) {Reconstruction\\run 1};
\node[anchor=south, align=center, font=\scriptsize] at ($(drawgrid.north west)!0.583333!(drawgrid.north east)+(0,2pt)$) {Reconstruction\\run 2};
\node[anchor=south, align=center, font=\scriptsize] at ($(drawgrid.north west)!0.750000!(drawgrid.north east)+(0,2pt)$) {Reconstruction\\run 3};
\node[anchor=south, align=center, font=\scriptsize] at ($(drawgrid.north west)!0.916667!(drawgrid.north east)+(0,2pt)$) {Reconstruction\\run 4};
\node[anchor=east, align=right, font=\scriptsize] at ($(drawgrid.south west)!0.875!(drawgrid.north west)+(-3pt,0)$) {Centre inpainting\\SD3.5};
\node[anchor=east, align=right, font=\scriptsize] at ($(drawgrid.south west)!0.625!(drawgrid.north west)+(-3pt,0)$) {Centre inpainting\\RV-v5.1};
\node[anchor=east, align=right, font=\scriptsize] at ($(drawgrid.south west)!0.375!(drawgrid.north west)+(-3pt,0)$) {SR $\times8$\\SD3.5};
\node[anchor=east, align=right, font=\scriptsize] at ($(drawgrid.south west)!0.125!(drawgrid.north west)+(-3pt,0)$) {SR $\times8$\\RV-v5.1};
\end{tikzpicture}
\caption{Repeated \paseo\ reconstructions. In the inpainting rows, the
ground-truth image and measurement are held fixed while the sampling seed
varies; the four runs are seeds $12$, $13$, $14$ and $16$. The inpainting
image and these four seeds were selected by visual screening of all $50$
measurements under both priors and all eight seeds, prioritizing clean
boundaries and coherent facial structure rather than LPIPS.
These rows use the current mask-free soft likelihood weights. The retained SR
$\times8$ rows show legacy end-to-end repetitions with seeds $10$--$13$: the
displayed measurement is from run seed $10$, and measurement noise is
regenerated with the run seed for the other reconstructions.
Table~\ref{tab:draws} summarizes all $50$ images.}
\label{fig:draws}
\end{figure}
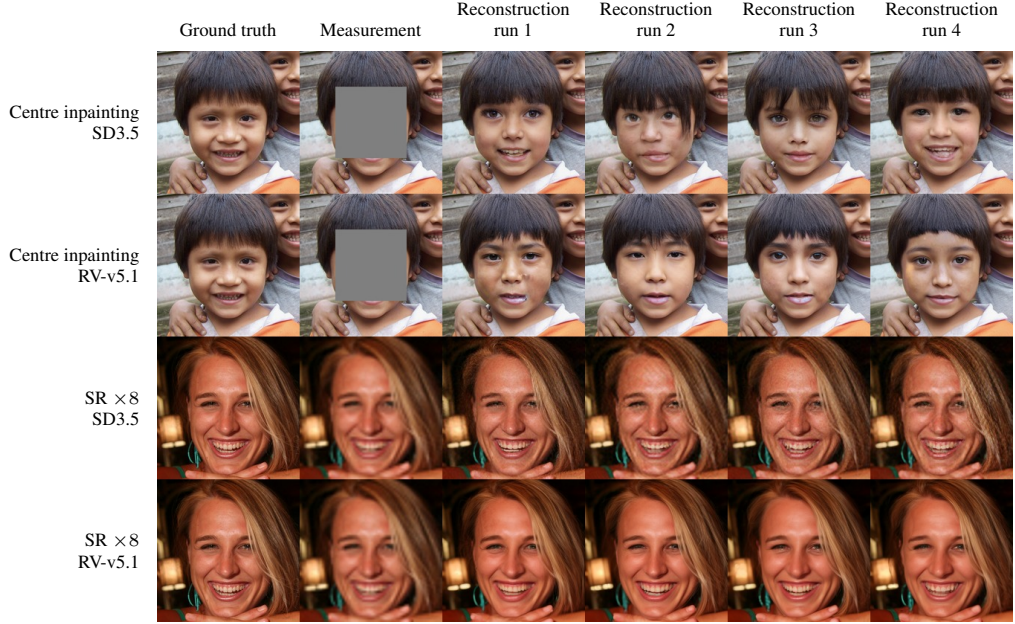

Similar average scores can hide differences between reconstructions of
the same input or between repeated runs of the full stochastic pipeline.
Here we examine how much the outputs vary, where that variation occurs, and
how closely they agree with the reference measurement after degradation.
For each of the first $50$ test images, the two inpainting configurations
construct the measurement with seed $0$ and reconstruct it with eight sampling
seeds ($10$--$17$). These rows therefore isolate reconstruction variation and
use the current mask-free soft likelihood weighting configuration. The
retained SR, blur, and JPEG rows use the legacy joint-seed protocol, in which
the run seed affects both measurement construction and sampling; their
statistics describe end-to-end repeatability and mix the two sources of
randomness. Table~\ref{tab:draws} summarizes the results, and
Figure~\ref{fig:draws} shows four runs for selected inputs.

\textbf{Where the images vary.} Panel A summarizes three aspects of
variation. Pixel standard deviation (std.) measures how much each pixel
changes across the eight seeded runs, then averages over pixels and test images.
Pixels are measured on the $[-1,1]$ scale. For inpainting, the parentheses
report the hole and visible region separately; variation inside the hole
is about $9$--$20$ times larger. LPIPS mean and standard deviation describe
quality and its variation across runs for each image, then average these
quantities over images.

The RMS distance ratio compares the mean pairwise distance between
reconstructions with their mean reconstruction error to the reference
image, using root mean square (RMS) distance for both. The denominator
includes reconstruction error and bias, so this ratio does not measure
posterior calibration or a fraction of missing uncertainty.

\textbf{Agreement after degradation.} Panel B applies the true degradation
to each reconstruction and compares it with the degraded reference image.
Errors are measured as RMS distances. The table reports both the average
error and a worst-draw summary: for each image,
take the largest error among its eight draws, then average over images.
The comparison is to the noiseless reference $\gA(\rvx)$, rather than the
noisy input $\rvy$. Together, the panels describe variation across outputs
and their agreement with the reference measurement. Calibration is tested
separately in the controlled affine experiment of
Appendix~\ref{sec:app-tractable}.

{
\begin{table}[!htbp]
\centering
\footnotesize

\setlength{\tabcolsep}{4pt}
\caption{Variation across eight seeded reconstructions ($10$--$17$) on the
first $50$ images per configuration, using the arc anchor. The inpainting rows
use the corrected fixed-measurement protocol (measurement seed $0$), together
with the mask-free soft likelihood weights from the main results. The retained
SR, blur, and JPEG rows use the legacy joint-seed protocol, in which the run
seed affects measurement construction and sampling. Panel A describes
variation across runs; panel B measures RMS error against the noiseless
degraded reference $\gA(\rvx)$. These are descriptive statistics, not a
posterior-calibration test.}
\label{tab:draws}
\vspace{6pt}
\begin{tabular}{@{}clrll@{}}
\toprule
\multicolumn{5}{l}{\textbf{A. Variation across draws}} \\
\addlinespace
Prior & Task & \shortstack{RMS distance\\ratio} & \shortstack{Pixel std.\\whole (hole / visible)} & \shortstack{LPIPS\\mean $\pm$ std.} \\
\midrule
\multirow{5}{*}{\rotatebox[origin=c]{90}{RV-v5.1}} & SR $\times4$ & 0.89 & 0.039 & 0.137 $\pm$ 0.008 \\
 & SR $\times8$ & 0.79 & 0.039 & 0.210 $\pm$ 0.012 \\
 & Blur & 0.83 & 0.037 & 0.159 $\pm$ 0.009 \\
 & Inpaint & 0.87 & 0.070 (0.210 / 0.024) & 0.158 $\pm$ 0.014 \\
 & JPEG & 0.90 & 0.065 & 0.262 $\pm$ 0.015 \\
\midrule
\multirow{5}{*}{\rotatebox[origin=c]{90}{SD3.5}} & SR $\times4$ & 0.77 & 0.030 & 0.176 $\pm$ 0.009 \\
 & SR $\times8$ & 0.78 & 0.042 & 0.323 $\pm$ 0.043 \\
 & Blur & 0.78 & 0.030 & 0.196 $\pm$ 0.026 \\
 & Inpaint & 0.86 & 0.055 (0.192 / 0.010) & 0.131 $\pm$ 0.014 \\
 & JPEG & 1.02 & 0.071 & 0.354 $\pm$ 0.036 \\
\bottomrule
\end{tabular}

\par\medskip

\begin{tabular}{@{}clrr@{}}
\toprule
\multicolumn{4}{l}{\textbf{B. Error after applying the true degradation}} \\
\addlinespace
Prior & Task & Mean RMS error & \shortstack{Mean worst-draw\\RMS error} \\
\midrule
\multirow{5}{*}{\rotatebox[origin=c]{90}{RV-v5.1}} & SR $\times4$ & 0.047 & 0.050 \\
 & SR $\times8$ & 0.036 & 0.039 \\
 & Blur & 0.033 & 0.036 \\
 & Inpaint & 0.057 & 0.058 \\
 & JPEG & 0.120 & 0.128 \\
\midrule
\multirow{5}{*}{\rotatebox[origin=c]{90}{SD3.5}} & SR $\times4$ & 0.025 & 0.026 \\
 & SR $\times8$ & 0.022 & 0.024 \\
 & Blur & 0.023 & 0.025 \\
 & Inpaint & 0.035 & 0.036 \\
 & JPEG & 0.093 & 0.098 \\
\bottomrule
\end{tabular}
\end{table}}

\FloatBarrier
\subsection{Inpainting case study}
\label{sec:inpainting-analysis}

Inpainting is the task for which the main results and the preceding
ablations use different measurement weights: the main results use learned
weights, while the ablations use a supplied mask. We examine how this choice
affects reconstruction quality, whether its effect depends on the anchor,
and which image regions account for the differences between anchors.

\subsubsection{Supplied mask versus learned likelihood weighting}
\label{sec:inpaint-variants}

The missing region contains no observed image content and should therefore
be excluded from the measurement constraint. A supplied \emph{hard mask}
identifies this region explicitly. The learned weighting rule instead uses
the operator's sensitivity to determine which measurement coordinates
receive weight, without receiving the mask. To measure the effect of this
choice, we first compare the two rules with the arc anchor fixed in
Table~\ref{tab:inpaint-variants}. This comparison explains the quality and
runtime consequences of the mask-free configuration used in the main results.

\textbf{How the learned weights are obtained.} At each inner correction, we
estimate how strongly each predicted measurement coordinate responds to a
change in the latent. Coordinates with greater sensitivity receive larger
weights in the likelihood. With $J$ denoting the local operator Jacobian,
this gives $w_i\propto\sqrt{(JJ^{\top})_{ii}}$, the norm of its $i$th row.
We estimate these quantities using four random probes, known as Hutchinson
probes, without forming the Jacobian. The table calls this rule ``learned
weight''; we also refer to it as the soft weight below.

\textbf{What changes when the mask is withheld.} On FFHQ with RV-v5.1, the
soft weight improves PSNR, hole LPIPS and KID relative to the hard mask,
while whole-image LPIPS worsens by $0.003$ and runtime increases by about
$25\%$. With SD3.5, PSNR and both LPIPS scores are similar, while KID
increases by about $0.7$. On COCO, the soft weight loses $0.12$\,dB PSNR,
adds $0.008$ LPIPS and increases KID by about $4.3$. Thus the cost of avoiding a supplied
mask depends on the dataset and prior. The main tables report the
mask-free rule consistently, including where the supplied mask performs better.

\subsubsection{Interaction between anchor and weighting}
\label{sec:inpaint-anchor-weighting}

The learned weights are estimated during each correction, so their effect
may depend on the anchor used for that correction. We therefore extend the
comparison in Table~\ref{tab:inpaint-variants} to the original perturbed-prior
anchor, testing whether the weighting rule used with the arc also works
well with this anchor. We also examine a rule that estimates the weights
once and keeps them fixed throughout sampling.

The perturbed-prior (``exact'') anchor corresponds to $\beta=1$ in
\eqref{eq:method-practical-anchor} and adds the full fresh prior perturbation,
whereas the arc reduces that perturbation early in sampling. On RV-v5.1,
combining the perturbed-prior anchor with the per-correction soft weight gives
LPIPS $0.247$, compared with $0.139$ when that anchor receives the hard mask.
One possible explanation is that the stronger perturbation changes the
operator's sensitivity inside the hole, making the estimated weights less
useful for excluding missing content. The table measures the resulting
quality difference but does not isolate this mechanism.

To test another weighting rule, we estimate the weights once at the encoded
measurement and keep them fixed. We remove weights below $0.2$ times the
largest weight and shrink the retained region by one latent cell along its
boundary. The table calls this thresholded and eroded rule ``fixed weight.''
For the RV-v5.1 perturbed-prior anchor, this
rule improves LPIPS from $0.247$ to $0.147$; for the arc, it worsens LPIPS
relative to the per-correction soft weight under both priors.

Among the reported RV-v5.1 combinations, the denoiser anchor with the hard
mask has the best quality scores and shortest runtime. The perturbed-prior
anchor with the mask has the second-best quality scores. The arc variants
have KID around $10$ points above the denoiser-plus-mask result across the
three tested weighting rules. The next experiment examines where this
image-quality difference occurs.

\textbf{Reading the quality metrics.} Table~\ref{tab:inpaint-variants}
reports LPIPS for both the whole image and the $256\times256$ missing region
(the hole). KID compares the reconstructions with the paired reference
images and is reported multiplied by $10^3$. Higher PSNR and lower LPIPS
and KID indicate better scores; runtime is measured in seconds per image.

\begin{table}[!ht]
\centering
\footnotesize
\setlength{\tabcolsep}{3pt}
\renewcommand{\arraystretch}{1.35}
\caption{Time re-measured with the two-call sampler on the first $20$ images; the rows printed in Tables~\ref{tab:ffhq-models} and~\ref{tab:coco-models-arc} carry those tables' values. Anchor and weighting choices for centre inpainting on $1000$ FFHQ
images (first two groups) and $1000$ COCO images (last group).
Within each group and metric, bold marks the best value and underline the
second-best distinct value. Shaded rows marked $\dagger$ are the configurations
reported in Tables~\ref{tab:ffhq-models} and~\ref{tab:coco-models-arc}.}
\label{tab:inpaint-variants}
\vspace{6pt}
\begin{tabular}{@{}cllrrrrr@{}}
\toprule
Prior & Anchor & Weighting & \shortstack{PSNR $\uparrow$\\(dB)} & \shortstack{LPIPS $\downarrow$\\Whole image} & \shortstack{LPIPS $\downarrow$\\Hole} & \shortstack{KID $\downarrow$\\($\times10^3$)} & \shortstack{Time $\downarrow$\\(s/image)} \\
\midrule
 & Denoiser & Hard mask & \textbf{21.97} & \textbf{0.133} & \textbf{0.330} & \textbf{3.50} & \textbf{30.4} \\
 & Arc & Hard mask & 21.64 & 0.141 & 0.354 & 16.26 & \underline{30.9} \\
\rowcolor{black!6} \cellcolor{white} & Arc & Learned weight$^\dagger$ & 21.76 & 0.144 & 0.351 & 14.00 & 38.4 \\
 & Arc & Fixed weight & 21.13 & 0.150 & 0.362 & 14.88 & 33.0 \\
 & Perturbed prior & Hard mask & \underline{21.77} & \underline{0.139} & \underline{0.342} & \underline{8.89} & 31.8 \\
 & Perturbed prior & Learned weight & 19.85 & 0.247 & 0.676 & 60.78 & 36.0 \\
\multirow{-7}{*}{\rotatebox[origin=c]{90}{RV-v5.1}} & Perturbed prior & Fixed weight & 21.67 & 0.147 & 0.344 & 13.36 & 31.2 \\
\midrule
 & Arc & Hard mask & \textbf{22.62} & \textbf{0.117} & \textbf{0.346} & \textbf{2.04} & \textbf{10.7} \\
\rowcolor{black!6} \cellcolor{white} & Arc & Learned weight$^\dagger$ & \underline{22.61} & \underline{0.118} & \underline{0.349} & 2.71 & 13.6 \\
\multirow{-3}{*}{\rotatebox[origin=c]{90}{\shortstack{SD3.5\\medium}}} & Arc & Fixed weight & 22.52 & 0.119 & 0.350 & \underline{2.07} & \textbf{10.7} \\
\midrule
 & Arc & Hard mask & \textbf{19.43} & \textbf{0.166} & \textbf{0.496} & \textbf{1.00} & \textbf{39.9} \\
\rowcolor{black!6} \cellcolor{white} \multirow{-2}{*}{\rotatebox[origin=c]{90}{\shortstack{SD3.5\\COCO}}} & Arc & Learned weight$^\dagger$ & \underline{19.31} & \underline{0.174} & \underline{0.521} & \underline{5.26} & \underline{58.7} \\
\bottomrule
\end{tabular}
\end{table}

\FloatBarrier
\newpage
\subsubsection{Where the reconstruction differences occur}
\label{sec:kid-vs-lpips}

The denoiser and arc anchors produce different whole-image scores, but
these scores do not show whether the difference lies in the filled hole or
the visible background. We separate these regions to locate the quality
gap. Table~\ref{tab:kid-spatial} compares both anchors on the same $1000$ test
images under RV-v5.1, with the hard mask in both runs. In addition to scoring
complete images and hole crops, we replace one region at a time with
reference pixels so that only errors in the other region remain.

\textbf{Isolating the regions.} Replacing the reconstructed hole with the
reference hole leaves only background errors; both KID estimates then become
close to zero. Conversely, inserting the reconstructed hole into the
reference background leaves only errors inside the hole. This largely
reproduces the whole-image gap: KID is $2.89$ for the denoiser anchor and
$17.31$ for the arc, compared with $3.50$ and $16.26$ for the complete
reconstructions. The difference therefore comes mainly from the filled
region. Compare anchors within each row: cropping changes the input to the
feature extractor, so crop scores and complete-image scores are not directly
comparable. FID provides a complementary comparison.

\begin{figure}[ht]
\centering
\includegraphics[width=\linewidth]{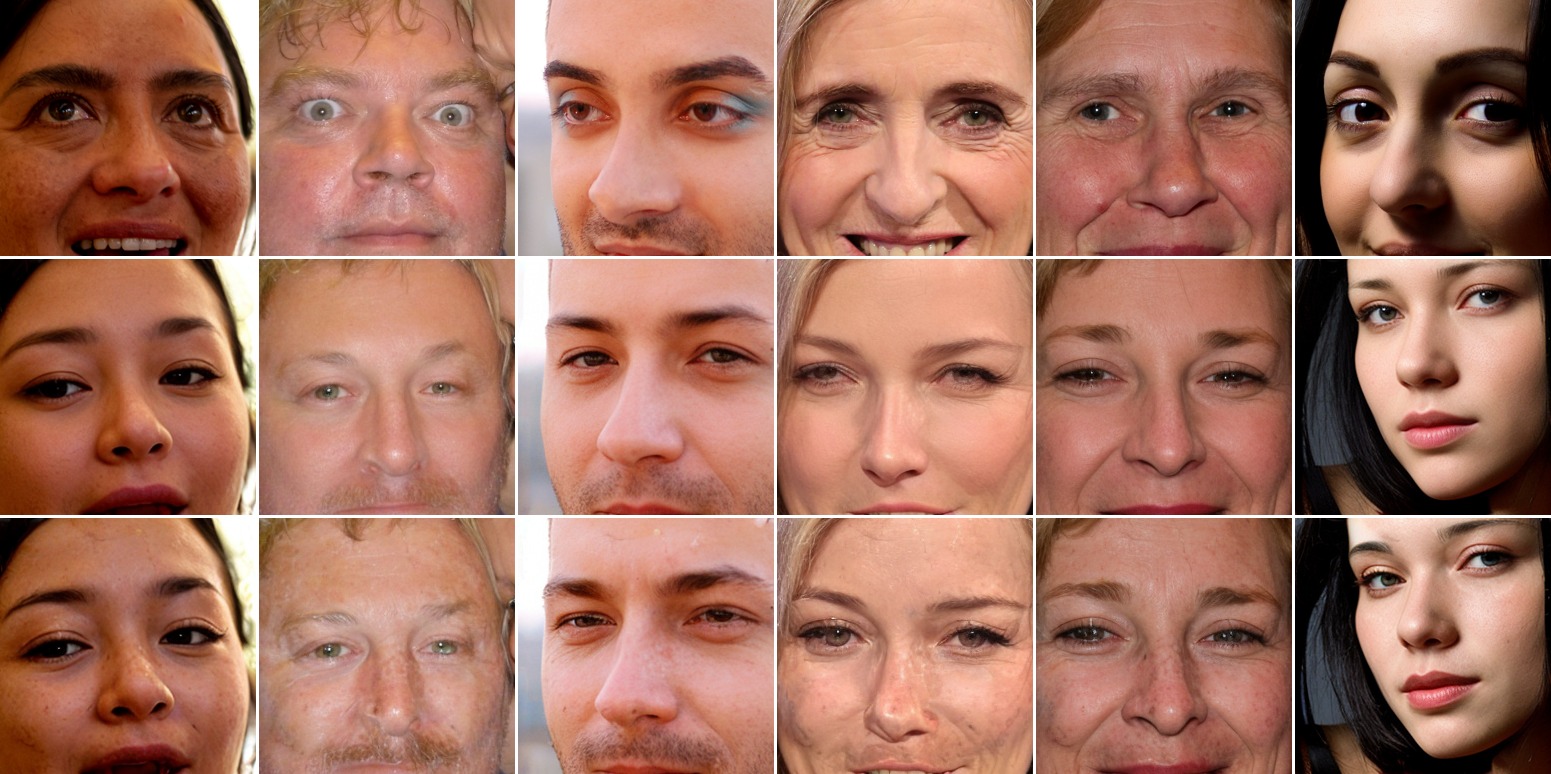}
\caption{The $256\times256$ hole, RV-v5.1 center inpainting, six test
images. Top: the true image. Middle: the $\rvz_0^{(0)}$ anchor. Bottom: the
arc. The arc examples show recurring small dark blemishes on skin. Across the
full dataset, the whole-image KID is about $13$ points higher than with the
denoiser anchor (Table~\ref{tab:kid-spatial}). The regional comparison
locates the gap in the filled hole, without establishing what causes the
artifacts.}
\label{fig:inpaint-rows}
\end{figure}

\textbf{Interpreting the metrics.} LPIPS compares each reconstruction
with its own reference, whereas KID compares distributions of image features
across the dataset. Repeated texture artifacts, visible in
Figure~\ref{fig:inpaint-rows}, can therefore produce a distributional gap
while having a smaller effect on the average paired-image distance. This
interpretation is consistent with the measurements; the regional experiment
locates the gap without identifying the mechanism that causes the artifacts.

\textbf{Evaluation details.} The regional comparison uses paired reference
images, a $256\times256$ hole, and torchmetrics $1.5.2$ with a KID subset
size of $999$. KID is reported multiplied by $10^3$. Small negative KID
estimates can arise from the unbiased finite-sample estimator and do not
establish that the two distributions are identical. The table highlights
numerical differences, without testing their statistical significance.

{
\begin{table}[!htbp]
\centering
\footnotesize

\setlength{\tabcolsep}{4pt}
\caption{Locating the inpainting quality gap on the same $1000$ test images
with RV-v5.1 and the hard mask. For each metric within a row, bold marks
the lower value and underline the higher value. Compare anchors within
rows; scores from hole crops and complete images are not directly comparable.}
\label{tab:kid-spatial}
\vspace{6pt}
\begin{tabular}{@{}lrrrr@{}}
\toprule
Image scored & \multicolumn{2}{c}{Denoiser anchor} & \multicolumn{2}{c}{Arc anchor} \\
\cmidrule(lr){2-3}\cmidrule(lr){4-5}
 & KID ($\times10^3$) $\downarrow$ & FID $\downarrow$ & KID ($\times10^3$) $\downarrow$ & FID $\downarrow$ \\
\midrule
Whole reconstruction & \textbf{3.50} & \textbf{21.2} & \underline{16.26} & \underline{34.4} \\
Reconstructed hole crop & \textbf{12.06} & \textbf{35.3} & \underline{27.30} & \underline{50.8} \\
Reconstructed background + reference hole & \textbf{-0.23} & \textbf{6.1} & \underline{-0.05} & \underline{6.6} \\
Reconstructed hole + reference background & \textbf{2.89} & \textbf{17.9} & \underline{17.31} & \underline{32.6} \\
\bottomrule
\end{tabular}
\end{table}}

\clearpage
\appendixpart{III}{Competitors}
\section{Competitor implementation and evaluation}
\label{sec:competitor-evaluation}
\label{sec:baseline-protocol}

The left block of Table~\ref{tab:ffhq-models} takes its quality metrics from
\citet[Tables~13--15]{raphaeli2025silo}: $1{,}000$ FFHQ images, CFG $1$. We
re-measured runtime and memory on our MI210. \textsc{SILO} uses RV-v5.1 there;
the other methods use their published priors, and \textsc{ReSample} outputs $256\times256$
images. Their single non-JPEG runtime is measured on SR $\times8$. This block is not
equal-budget: its baselines run at about $1{,}000$ denoiser steps ($500$ for
\textsc{ReSample}), against our $28$. ``n/a'' marks a cell the method cannot
express, not one we declined to run.

No published baseline reports peak memory, so every entry of that column is our
own measurement of peak allocated memory on the MI210, one process per cell.
We measured peak memory at $512\times512$ and CFG $1$, RV-v5.1
being architecturally identical to SD1.5 and therefore allocating identically.
Peak allocation is fixed by the shape of the autograd graph that a single step
builds and frees, so it does not depend on the step count: we measured over
$50$ steps rather than the longer published schedules behind the quality
columns of the same block. It is likewise flat across images ($\leq0.01$\,GiB) and,
for every cell except \textsc{PSLD} on Gaussian blur
(Appendix~\ref{sec:comp-psld}), across tasks ($\leq0.03$\,GiB).

Every baseline in the SD3.5 block uses $79$ denoiser evaluations, the budget our three-call sampler spent in $28$ steps; the two-call sampler reported in Table~\ref{tab:ffhq-models} spends $53$ at the same step count.

The COCO baselines use the implementations released with
\textsc{DING}~\citep{moufad2026ding}, the SD3.5-medium prior, the true forward
operator and the same prompts as our runs. For each, we tune one setting, its
data-consistency step size, on $20$ COCO-val2017 images ($1000$--$1049$,
disjoint from the test set) at $50$ steps by LPIPS; every grid contains the
published default, and inpainting keeps the authors' settings. The step count
is set so that each method spends the NFEs listed in
Table~\ref{tab:coco-models-arc-full}.

\subsection{\textsc{SILO}}
\label{sec:comp-silo}

\textsc{SILO}'s $7.41$\,GiB is measured from its released solver.
Within the SD3.5 block, our extension of \textsc{SILO} was tuned per task over $50$ grid
cells, with scale and flow-shift pairs
$(8,0.125),(8,0.125),(32,1),(2,1),(32,0.25)$ in task order.

\paragraph{\textsc{SILO} at $999$ steps.} Table~\ref{tab:ffhq-models} reports
\textsc{SILO} at $79$ NFEs, the budget given to every baseline in that block (\paseo\ spends $53$).
Its full $999$-step configuration is recorded here for reference:

\begin{center}
\footnotesize
\begin{tabular}{@{}lrrrrr@{}}
\toprule
Task & PSNR $\uparrow$ & LPIPS $\downarrow$ & KID $\downarrow$ & Time $\downarrow$ & Mem $\downarrow$ \\
\midrule
SR $\times4$ & 28.44 & 0.293 & 21.62 & 237 & 21.2 \\
SR $\times8$ & 25.55 & 0.543 & 63.37 & 237 & 21.2 \\
Gaussian blur & 27.19 & 0.369 & 39.42 & 237 & 21.2 \\
Center inpainting & 16.31 & 0.242 & 130.36 & 237 & 21.2 \\
JPEG $q{=}10$ & 24.37 & 0.441 & 51.96 & 237 & 21.2 \\
\bottomrule
\end{tabular}
\end{center}

Relative to the 79-NFE run, the full configuration changes PSNR by at most
$0.84$\,dB and LPIPS by at most $0.062$ on SR $\times4$, SR $\times8$,
Gaussian blur, and JPEG, while increasing runtime from $20$ to $237$\,s per
image. Center inpainting is the exception and is
discussed separately --- in \textsc{SILO}'s released solver under a prior it was
not designed for, the masked region is not reconstructed at any step count.

\subsection{\textsc{ReSample}}
\label{sec:comp-resample}

\textsc{ReSample}'s quality row in the left block is $256\times256$ on an LDM-VQ4 prior, so its
$6.97$-GiB peak-memory entry is an upper bound for that configuration.
Because \textsc{ReSample}'s step count is a
tuning choice rather than a fixed property, we ran it both at its shipped
setting and at the step count corresponding to $79$ NFEs. The budget arm had
lower LPIPS on all five tasks and is therefore reported.
\textsc{ReSample}'s
latent-optimization rate is transferred per task from our COCO sweep, except
for inpainting, which takes $5\times10^{-3}$ from the sampler's own task block.
On Gaussian blur, none of its six settings is mistuned: the largest
single-parameter perturbation costs $0.079$ LPIPS and the shipped configuration
ranks second among seventeen, a check we performed on Gaussian blur.
Its SD3.5 rows are at $512\times512$ and so are not comparable to the $256\times256$
\textsc{ReSample} results in the left block.

\paragraph{\textsc{ReSample} at $999$ NFEs.}
\label{sec:nfe-scaling}
We reran \textsc{ReSample} at
$999$ NFEs on $30$ images per task, paired against the same images from the
$79$-NFE production run, so the two arms differ only in step count. The
runtime columns are not measured alike: the 79-NFE runtime is the production
per-image measurement under eight-way GPU contention, whereas the 999-NFE runtime is instrumented
wall-clock time including roughly $41$\,s of model loading. The stated ratios
therefore overstate the true cost by about $3\%$.

\begin{table}[!ht]
\centering
\footnotesize
\caption{\textsc{ReSample} under SD3.5 at $999$ NFEs, compared with the $79$-NFE budget of
Table~\ref{tab:ffhq-models}, paired on the same $30$ images per task.
``Wins'' counts images on which the 999-NFE run attains the lower LPIPS, and a
negative $\Delta$LPIPS favors $999$. The two time columns are measured under
different conventions; see the text.}
\label{tab:nfe-scaling}
\begin{tabular}{@{}lrrrrrrrrr@{}}
\toprule
 & & \multicolumn{3}{c}{PSNR $\uparrow$} & \multicolumn{3}{c}{LPIPS $\downarrow$}
 & \multicolumn{2}{c}{Time (s)} \\
\cmidrule(lr){3-5}\cmidrule(lr){6-8}\cmidrule(lr){9-10}
Task & Wins & $79$ & $999$ & $\Delta$ & $79$ & $999$ & $\Delta$ & $79$ & $999$ \\
\midrule
SR $\times4$      & $21/30$ & 30.34 & \textbf{31.08} & $+0.74$ & 0.1779 & \textbf{0.1730} & $-0.0049$ & 130 & 1565 \\
SR $\times8$      & $12/30$ & 27.53 & 27.90 & $+0.37$ & \textbf{0.3652} & 0.3682 & $+0.0030$ & 130 & 1565 \\
Gaussian blur     & $19/30$ & 30.30 & \textbf{31.13} & $+0.83$ & 0.2028 & \textbf{0.1931} & $-0.0097$ & 138 & 1657 \\
Center inpainting & $12/30$ & \textbf{21.19} & 20.47 & $-0.71$ & \textbf{0.2207} & 0.2226 & $+0.0020$ & 267 & 3625 \\
JPEG $q{=}10$     & $\phantom{0}4/30$ & 29.59 & \textbf{31.17} & $+1.58$ & \textbf{0.1940} & 0.2229 & $+0.0289$ & 141 & 1668 \\
\bottomrule
\end{tabular}
\end{table}

Three regimes appear. On $\times4$ super-resolution and Gaussian blur, the
budget does cost \textsc{ReSample}: both metrics improve at $999$ NFEs, with win rates
near two-thirds and gains of $+0.74$ and $+0.83$\,dB, so the $79$-NFE rows
of Table~\ref{tab:ffhq-models} understate \textsc{ReSample}'s performance on
those tasks. On $\times8$ super-resolution and inpainting, the extra compute
buys nothing: win rates remain near chance, the $\Delta$LPIPS values are small,
and inpainting
is $0.71$\,dB \emph{worse} for $13.6\times$ the compute. Since inpainting is
also where \textsc{ReSample} is weakest in absolute terms, the hypothesis that
it is compute-starved is not supported. On JPEG, the metrics disagree: the
999-NFE run gains $1.58$\,dB but increases LPIPS by $0.029$ and wins only $4$
of $30$ images. The distortion--perception trade-off therefore appears within
one method as a function of compute alone. The efficiency claim is unaffected:
on the tasks where both metrics improve, the gain remains below $0.9$\,dB for
roughly $12\times$ the wall-clock time, against our own $8.6$--$13.9$\,s per image.

\subsection{\textsc{PSLD}}
\label{sec:comp-psld}

We followed the setup in \textsc{SILO}'s Appendix~D: \textsc{PSLD} uses its
official repository with Stable Diffusion at $512\times512$. We checked our
\textsc{PSLD} implementation numerically against the released code; it agreed
with it to within the repeat-to-repeat floor ($\leq2\times10^{-7}$ absolute,
cosine $1.0$).

Of the baselines in
Table~\ref{tab:ffhq-models}, \textsc{PSLD} is the only one whose update needs the
operator adjoint: its gluing term evaluates $A^{\top}y$ and $A^{\top}AD(z)$
explicitly, so an adjoint has to exist as a linear map.
\textsc{GML-DPS}, \textsc{LDPS} and \textsc{ReSample} only apply $H$ and
differentiate through it, which a nonlinear operator supports, so they run on
every task. JPEG compression quantises transform coefficients, and
quantisation is not linear, so there is no adjoint to supply: the obstruction
is in the method--operator pair, not in our implementation of either.
Substituting a plausible surrogate such as the identity would let
\textsc{PSLD} run to completion while enforcing a different constraint, so we
report the cell as unsupported instead.
\citet{raphaeli2025silo} report no \textsc{PSLD} entry for this task either,
so the left block's cell is ``n/a'' as well.

\textsc{PSLD}'s memory on Gaussian blur is higher because it is the only
sampler that needs the operator adjoint: the exact adjoint of the
$61\times61$ blur kernel is an autograd vector--Jacobian product whose
double-backward graph costs a further $2.1$\,GiB under both priors, so that
cell is reported on its own. What separates the two groups is the gluing term,
which puts the encoder in the graph: \textsc{PSLD} sits at $7.87$\,GiB on its
three remaining tasks and \textsc{GML-DPS} at $7.89$\,GiB, while
\textsc{ReSample} and \textsc{LDPS} backpropagate through the decoder alone at
$6.97$\,GiB -- a cost of $0.90$ and $0.92$\,GiB respectively.

\paragraph{At $1{,}000$ steps.}
\label{app:sd35-1000step}
Table~\ref{tab:ffhq-models} runs \textsc{PSLD}, \textsc{GML-DPS}, and
\textsc{LDPS} at $79$ NFEs, the budget of our former three-call sampler ($53$ for the two-call sampler of Table~\ref{tab:ffhq-models}). Their official implementations
default to $1{,}000$ DDIM steps, so
we report all three at $1{,}000$ steps on their supported tasks using the same $1000$ images, with $\omega$ and
$\gamma$ tuned for each solver--task pair at this step count. Peak memory is
unchanged from the 79-NFE arm because the same autograd graph is built at every
step, regardless of the total step count.

\begin{table}[!ht]
\centering
\footnotesize
\caption{\textsc{GML-DPS}, \textsc{LDPS}, and \textsc{PSLD} on FFHQ with SD3.5-medium at $1{,}000$
steps, the authors' default, with $1000$ samples per cell. Each entry is
PSNR / LPIPS / KID$\times10^{3}$ / seconds per image; peak memory is $23.4$,
$22.5$, and $23.4$ GiB, respectively.
For \textsc{PSLD}, JPEG is unsupported (n/a) because its update requires a
linear operator adjoint (Appendix~\ref{sec:comp-psld}).}
\label{tab:sd35-1000step}
\begin{tabular}{@{}lrrr@{}}
\toprule
Task & \textsc{GML-DPS} & \textsc{LDPS} & \textsc{PSLD} \\
\midrule
SR $\times4$ & 28.65 / 0.253 / 20.51 / 606 & 26.42 / 0.343 / 29.71 / 497 & 22.44 / 0.490 / 40.36 / 606 \\
SR $\times8$ & 25.52 / 0.471 / 31.08 / 606 & 20.61 / 0.519 / 90.35 / 497 & 22.56 / 0.553 / 40.06 / 607 \\
Gaussian blur & 22.15 / 0.458 / 44.58 / 615 & 22.14 / 0.514 / 68.82 / 506 & 21.28 / 0.529 / 64.56 / 615 \\
Center inpainting & 16.90 / 0.313 / 80.73 / 606 & 18.07 / 0.363 / 64.87 / 497 & 17.84 / 0.339 / 63.28 / 605 \\
JPEG $q{=}10$ & 29.47 / 0.295 / 13.69 / 617 & 29.50 / 0.296 / 22.63 / 512 & n/a \\
\bottomrule
\end{tabular}
\end{table}

This arm is not uniformly better than the $79$-NFE one; the likeliest cause is
the guidance weight $\omega$, discussed below. Spending $12.7\times$ the compute improves LPIPS on SR $\times4$ and
inpainting for all three solvers and on JPEG for \textsc{GML-DPS} and
\textsc{LDPS} (by about $0.12$), but costs over $5$\,dB of
PSNR on Gaussian blur for every solver and degrades SR $\times8$ sharply for
\textsc{LDPS}. A loss of the same size across three solvers argues against a
solver-specific cause; a sign that flips with the task argues against a single
mis-set weight. The likeliest explanation is the guidance weight $\omega$,
whose optimum moves with step count ($10$ at $50$ steps, $3$ at $150$ and
above) and which was selected on a held-out tuning image per solver--task pair. Until that is
settled, we quote the $79$-NFE arm in Table~\ref{tab:ffhq-models} and report
this one for completeness. KID tells a different story from PSNR: the extra
compute helps it substantially on SR $\times4$ for all three solvers and on
JPEG for \textsc{GML-DPS} and \textsc{LDPS}, while inpainting stays poor
throughout.

\subsection{\textsc{GML-DPS}}
\label{sec:comp-gmldps}

Following \textsc{SILO}'s Appendix~D, \textsc{GML-DPS} replaces
\textsc{PSLD}'s gluing step with the goodness step. We checked it against its
equations; it agreed with them to within the same floor ($\leq2\times10^{-7}$
absolute, cosine $1.0$). Its peak memory ($7.89$\,GiB) and its $1{,}000$-step
results (Table~\ref{tab:sd35-1000step}) are discussed with \textsc{PSLD}
(Appendix~\ref{sec:comp-psld}).

\subsection{\textsc{LDPS}}
\label{sec:comp-ldps}

Following \textsc{SILO}'s Appendix~D, \textsc{LDPS} removes \textsc{PSLD}'s
gluing step. We checked our implementation numerically against the released
\textsc{PSLD} code evaluated at $\gamma=0$, which yields \textsc{LDPS} by
definition; it agreed to within the same floor. Its peak memory
($6.97$\,GiB) and its $1{,}000$-step results (Table~\ref{tab:sd35-1000step})
are discussed with \textsc{PSLD} (Appendix~\ref{sec:comp-psld}).

\subsection{\textsc{DING}}
\label{sec:competitors}
\label{sec:ding-detail}

\textsc{DING}~\citep{moufad2026ding} is the method Section~\ref{sec:method}
builds on, and the only row of Table~\ref{tab:ffhq-models} needing neither a
trained operator nor a denoiser Jacobian. For a coordinate mask, its
measurement step is an exact special case of ours
(Appendix~\ref{sec:unif-ding}). Two choices behind that row affect
the reported results, and neither is forced by the algorithm.

\addtocontents{toc}{\protect\setcounter{tocdepth}{2}}
\subsubsection{\textsc{DING}: carrying a pixel mask into the latent}
\label{sec:competitors-ding}

\textsc{DING}'s update is closed-form because its measurement enters as a
coordinate mask on the clean latent: the resulting conditional is Gaussian with
a diagonal covariance. Our benchmark instead measures in pixels,
$\rvy=\rmM_{\mathrm{pix}}\odot\rvx+\sigma_{\rvy}\rvn$, so the mask has to be
carried into the latent, and there are two ways to do it.

\textbf{Encode the reference, then mask: $\rmM\odot\gE(\rvx)$.} This is what a
latent inpainting task yields by default and what the released code constructs.
It is inadmissible on a pixel-measurement benchmark. $\gE$ has a receptive
field wider than one latent cell, so a cell just outside the hole is computed
partly from pixels inside it; because $\gE$ is applied to the true image, those
border cells carry the hidden content. A sampler pinned to them as observed
reads part of the fill it is meant to infer. Restricted to the visible cells
the two observations differ by $6.4\%$ in relative $\ell_1$ under SD3.5 and
$14.2\%$ under RV-v5.1, with maximum per-cell deviations of $1.24$ and $2.16$.

\textbf{Encode the measurement, then mask: $\rmM\odot\gE(\rvy)$.} This is what
we use. It cannot leak, because $\rvy$ contains nothing inside the hole, and it
is the only construction under which \textsc{DING} and our sampler receive the
same information. We verify that the pixel observation \textsc{DING} is handed
is identical, entry for entry, to our own sampler's on every image.

\subsubsection{What that choice costs \textsc{DING}}
\label{sec:competitors-fairness}

The adaptation is not neutral, and we would rather state its cost than claim it
is.

\textsc{DING}'s conjugacy assumes the observation \emph{is} the clean latent
restricted to the visible cells. Under $\rmM\odot\gE(\rvy)$ the cells adjacent
to the hole are not: they encode a blank region, so they differ from the
corresponding entries of $\gE(\rvx)$ by the same receptive-field effect that
makes the other construction inadmissible. \textsc{DING} is therefore pinned to
slightly incorrect values in a one-cell ring around the hole. This is a model
misspecification we impose; on \textsc{DING}'s own benchmark the measurement is
latent by definition, $\rmM\odot\gE(\rvx)$ is not leakage there, and no such
mismatch arises. The reported row is consequently a measurement of
\textsc{DING} on \emph{our} problem and is not comparable to the authors'
published results.

We do not have a measurement isolating the size of this effect. The check
reported below --- that the visible region is
reconstructed at the autoencoder's ceiling, $29.57$\,dB with the four-channel
latent and $34.29$\,dB with the sixteen-channel one --- establishes that the
coordinate mask is honoured rather than approximated, but it is an average over
the whole visible region and a one-cell ring at the boundary would be swamped
by the interior. It should not be read as evidence that the boundary is
unaffected.

A third construction would remove both problems: erode the latent mask by one
cell, so that the ambiguous border cells enter neither the observation nor the
leakage path. It costs a small number of observed cells and would leave
\textsc{DING} with an exactly specified, non-leaking measurement. We did not
run it, and we flag it as the fairer comparison rather than presenting
$\rmM\odot\gE(\rvy)$ as the last word.

We report $\rmM\odot\gE(\rvy)$ because between the two constructions we did
evaluate, it is the only one that gives every method the same information; the
alternative would hand \textsc{DING} an advantage no other method receives.

\subsubsection{Evaluation budget and regional metrics}

\textsc{DING} spends $2(n-2)+1$ denoiser evaluations at
$n$ steps: two for each of its $n-2$ iterations, plus a final clean-latent
prediction. A batched classifier-free pair counts once. At $n=41$ this is $79$,
the baselines' budget, a count confirmed by instrumenting the network rather
than inferred from the step count. We report a single arm: the two-arm protocol of
Appendix~\ref{sec:comp-resample} exists because a data-consistency step
size chosen at one step count is detuned at another, and \textsc{DING} has no
such parameter --- its only knobs are the step count and the noise schedule
$\eta$. The argument is structural: \textsc{DING} exposes no data-consistency
step size that could be detuned by a change of budget.

\textbf{What the row does not show.} Whole-image metrics understate the task,
since three quarters of the image is observed. Restricted to the
$256\times256$ fill, \textsc{DING} attains $16.13$\,dB and $0.377$ LPIPS under
RV-v5.1 and $15.77$\,dB and $0.378$ under SD3.5. The visible region is
reconstructed at the autoencoder's ceiling in both cases, $29.57$\,dB with the
four-channel latent and $34.29$\,dB with the sixteen-channel one. This confirms
that the coordinate mask is honored rather than approximated. All values are
computed over the same $1000$ images as the rest of
Table~\ref{tab:ffhq-models}; all runs completed without failure under eight-way
GPU contention.
\addtocontents{toc}{\protect\setcounter{tocdepth}{3}}

\subsection{\textsc{FlowDPS}}
\label{sec:comp-flowdps}
\label{sec:flowdps-27nfe-comparison}

In the COCO tables, \textsc{FlowDPS}~\citep{kim2025flowdps} takes three
data-consistency iterations per step at one NFE per step. Tuned step size:
$10$ (SR and blur) and $60$ (JPEG); the published default is $5$.

\paragraph{Matched 27-NFE comparison.}
We additionally ran a paired equal-budget comparison against
\textsc{FlowDPS} on the canonical COCO-val2017 extension slice 1000--1199
($n=200$, disjoint from the 0--999 evaluation set).  Both methods used
SD3.5-medium at $512\times512$, observation noise $0.01$, seed $0$, the same
images and observations, and the same task-specific prompt policy.
\textsc{FlowDPS} used its transferred author setting with 28 solver points;
\paseo\ used 15 solver points and shared the deterministic proxy velocity
between its predictor and correction.  Instrumented per-image counters
verified exactly 27 denoiser evaluations for both methods in every cell.

\begin{table}[!ht]
\centering
\scriptsize
\setlength{\tabcolsep}{3.5pt}
\caption{Exact-budget comparison on the same 200 COCO-val2017 images at 27
denoiser evaluations.  PSNR, LPIPS and sampler time are mean $\pm$ sample
standard deviation.  ``Wins'' is the number of paired images on which \paseo\
has lower LPIPS; \paseo\ is faster on all 800 pairs.}
\label{tab:flowdps-27nfe-coco200}
\begin{tabular}{@{}lrrrrrrrr@{}}
\toprule
 & \multicolumn{2}{c}{PSNR $\uparrow$}
 & \multicolumn{2}{c}{LPIPS $\downarrow$}
 & \multicolumn{2}{c}{Time (s/image) $\downarrow$}
 & \multicolumn{2}{c}{\paseo} \\
\cmidrule(lr){2-3}\cmidrule(lr){4-5}\cmidrule(lr){6-7}\cmidrule(lr){8-9}
Task & \textsc{FlowDPS} & \paseo & \textsc{FlowDPS} & \paseo & \textsc{FlowDPS} & \paseo & Wins & Speedup \\
\midrule
SR $\times4$      & $13.44\pm2.20$ & $\mathbf{25.43\pm3.08}$ & $.866\pm.083$ & $\mathbf{.241\pm.102}$ & $20.88\pm.01$ & $\mathbf{7.55\pm.32}$ & $200/200$ & $2.77\times$ \\
SR $\times8$      & $10.01\pm1.73$ & $\mathbf{22.27\pm3.00}$ & $.808\pm.081$ & $\mathbf{.537\pm.126}$ & $20.87\pm.02$ & $\mathbf{5.59\pm.17}$ & $197/200$ & $3.73\times$ \\
Gaussian blur     & $23.44\pm3.02$ & $\mathbf{23.81\pm2.80}$ & $.510\pm.146$ & $\mathbf{.397\pm.088}$ & $43.35\pm.05$ & $\mathbf{6.24\pm.21}$ & $168/200$ & $6.95\times$ \\
Center inpainting & $17.04\pm2.34$ & $\mathbf{18.65\pm2.42}$ & $.396\pm.090$ & $\mathbf{.193\pm.030}$ & $20.80\pm.01$ & $\mathbf{7.75\pm.32}$ & $199/200$ & $2.68\times$ \\
\bottomrule
\end{tabular}
\end{table}

The runtime is sampler-recorded time on one AMD MI210 GPU.  The
comparison matches denoiser calls, inputs, randomness, and prompt policy, but
each method retains its configured observation preprocessing; equal NFE
therefore does not imply identical preprocessing.  \textsc{FlowDPS} runs the
authors' published hyperparameters rather than a locally tuned setting, and
its low super-resolution PSNR reflects that transfer; the tuned
\textsc{FlowDPS} rows appear in the main COCO tables.

\subsection{\textsc{FlowChef}}
\label{sec:comp-flowchef}

\textsc{FlowChef}~\citep{patel2025flowchef} takes ten data-consistency
iterations per step at one NFE per step. Tuned step size: $7.2$ (SR $\times4$
and blur), $1.8$ (SR $\times8$) and $57.6$ (JPEG); the published default is
$0.45$.

\subsection{\textsc{PnP-Flow}}
\label{sec:comp-pnpflow}

\textsc{PnP-Flow}~\citep{martin2025pnpflow} spends one NFE per step and
keeps its published constant learning-rate schedule, $\alpha=1$ and five
samples per step. Tuned learning rate: $1.5$ (SR $\times4$), $9.88$
(SR $\times8$), $0.1$ (blur, the default) and $0.4$ (JPEG).

\subsection{\textsc{DAPS}}
\label{sec:comp-daps}

\textsc{DAPS}~\citep{zhang2025daps} spends three NFEs per step and runs
$50$ steps ($100$ for SR $\times8$ and inpainting). Its annealing and Langevin
settings stay at the published values ($2$ diffusion steps, $\tau=10^{-2}$,
$20$ MCMC steps). Tuned Langevin step size: $5\times10^{-6}$ (SR $\times4$),
$1.1\times10^{-3}$ (SR $\times8$) and $3\times10^{-5}$ (blur and JPEG); the
published default is $1.1\times10^{-6}$.

\subsection{\textsc{RedDiff}}
\label{sec:comp-reddiff}

\textsc{RedDiff}~\citep{mardani2024reddiff} spends one NFE per step and
keeps its published loss weights. Tuned learning rate: $1.6$ (SR $\times4$ and
SR $\times8$), $3.2$ (blur) and $0.8$ (JPEG); the published default is $0.1$.

\clearpage
\appendixpart{IV}{Additional results}
\section{Additional quantitative results}
\label{sec:additional-quantitative}

\subsection{Full FFHQ comparison}
\label{sec:ffhq-full}

Table~\ref{tab:ffhq-models-full} gives the full FFHQ results, including the
PSNR columns omitted from Table~\ref{tab:ffhq-models}.

\paseo$^\dagger$ keeps the original anchor of
Algorithm~\ref{alg:silo-jding} and changes only the settings in
Table~\ref{tab:paseo-dagger-settings}, chosen the same way as \paseo's
(Appendix~\ref{sec:paseo-hyperparameters}); its NFEs match \paseo. Mem is taken from
\paseo\ without re-measuring, as \paseo's peak memory does not change across
tasks with different settings.

\begin{table}[!ht]
\centering
\fontsize{8}{9}\selectfont
\setlength{\tabcolsep}{2.7pt}
\caption{FFHQ restoration. Left: \paseo\ with RV-v5.1 and published
baselines; right: SD3.5-medium, baselines at $79$ NFEs and \paseo\ at $53$ NFEs. Bold/underline mark
best/second-best within each block. KID is scaled by $10^3$.
\paseo$^\dagger$ uses the original anchor of Algorithm~\ref{alg:silo-jding}.}
\label{tab:ffhq-models-full}
\begin{tabular}{@{}clrrrrr@{\hspace{10.5pt}}|@{\hspace{2pt}}rrrrr@{}}
\toprule
 & & \multicolumn{5}{c@{\hspace{-2pt}\vrule\hspace{2pt}}}{RV-v5.1}
   & \multicolumn{5}{c}{SD3.5-medium} \\
\cmidrule(lr){3-7}\cmidrule(lr){8-12}
Task & Method
 & PSNR $\uparrow$ & LPIPS $\downarrow$ & KID $\downarrow$ & Time $\downarrow$ & Mem $\downarrow$\hspace{-8.5pt}
 & PSNR $\uparrow$ & LPIPS $\downarrow$ & KID $\downarrow$ & Time $\downarrow$ & Mem $\downarrow$ \\
\midrule
\rowcolor{ourrow} \cellcolor{white}\multirow{7}{*}{\rotatebox[origin=c]{90}{SR $\times4$}} & \paseo & 27.98 & \underline{0.139} & \textbf{1.81} & \textbf{11.7} & \textbf{3.6}
 & \underline{29.92} & \textbf{0.182} & 10.27 & \textbf{10.6} & \textbf{16.8} \\
 & \cellcolor{ourrow}\paseo$^\dagger$ & \cellcolor{ourrow}28.15 & \cellcolor{ourrow}\textbf{0.130} & \cellcolor{ourrow}\underline{2.96} & \cellcolor{ourrow}\underline{14.4} & \cellcolor{ourrow}\textbf{3.6}
 & \cellcolor{ourrow}28.76 & \cellcolor{ourrow}\underline{0.221} & \cellcolor{ourrow}\underline{9.49} & \cellcolor{ourrow}25.9 & \cellcolor{ourrow}\textbf{16.8} \\
 & \textsc{SILO} & 27.03 & 0.182 & 5.10 & 149 & 7.4 & 29.28 & 0.303 & 29.36 & \underline{20} & \underline{21.2} \\
 & \textsc{ReSample} & 24.62 & 0.433 & 25.50 & 1418 & \underline{7.0} & \textbf{30.35} & \textbf{0.182} & \textbf{7.70} & 130.4 & 22.5 \\
 & \textsc{PSLD} & 28.23 & 0.249 & 10.11 & 390 & 7.9 & 22.88 & 0.494 & 47.23 & 47 & 23.4 \\
 & \textsc{GML-DPS} & \textbf{29.34} & 0.247 & 9.05 & 389 & 7.9 & 28.50 & 0.373 & 46.34 & 48 & 23.4 \\
 & \textsc{LDPS} & \underline{29.06} & 0.281 & 11.69 & 331 & \underline{7.0} & 28.38 & 0.423 & 57.12 & 39 & 22.5 \\
\midrule
\rowcolor{ourrow} \cellcolor{white}\multirow{7}{*}{\rotatebox[origin=c]{90}{SR $\times8$}} & \paseo & 26.74 & \textbf{0.212} & \underline{5.99} & \textbf{11.6} & \textbf{3.6}
 & 27.25 & \textbf{0.330} & \textbf{28.80} & \textbf{8.8} & \textbf{16.8} \\
 & \cellcolor{ourrow}\paseo$^\dagger$ & \cellcolor{ourrow}26.37 & \cellcolor{ourrow}0.241 & \cellcolor{ourrow}13.13 & \cellcolor{ourrow}\underline{14.4} & \cellcolor{ourrow}\textbf{3.6}
 & \cellcolor{ourrow}27.23 & \cellcolor{ourrow}0.420 & \cellcolor{ourrow}\underline{30.33} & \cellcolor{ourrow}\underline{10.7} & \cellcolor{ourrow}\textbf{16.8} \\
 & \textsc{SILO} & 26.28 & \underline{0.226} & \textbf{5.23} & 149 & 7.4 & 25.79 & 0.501 & 62.88 & 20 & \underline{21.2} \\
 & \textsc{ReSample} & 22.80 & 0.575 & 118.5 & 1418 & \underline{7.0} & \underline{27.55} & \underline{0.378} & 31.61 & 130.4 & 22.5 \\
 & \textsc{PSLD} & 25.08 & 0.320 & 14.90 & 390 & 7.9 & 27.48 & 0.453 & 60.94 & 48 & 23.4 \\
 & \textsc{GML-DPS} & \textbf{27.01} & 0.327 & 12.99 & 389 & 7.9 & \textbf{27.56} & 0.439 & 57.12 & 48 & 23.4 \\
 & \textsc{LDPS} & \underline{26.89} & 0.343 & 12.94 & 331 & \underline{7.0} & 27.46 & 0.456 & 60.17 & 39 & 22.5 \\
\midrule
\rowcolor{ourrow} \cellcolor{white}\multirow{7}{*}{\rotatebox[origin=c]{90}{Gaussian blur}} & \paseo & 27.68 & \textbf{0.162} & \textbf{2.09} & \textbf{17.3} & \textbf{3.6}
 & 29.54 & \textbf{0.200} & \textbf{8.32} & \textbf{8.6} & \textbf{16.8} \\
 & \cellcolor{ourrow}\paseo$^\dagger$ & \cellcolor{ourrow}27.90 & \cellcolor{ourrow}\underline{0.172} & \cellcolor{ourrow}\underline{6.66} & \cellcolor{ourrow}\underline{121} & \cellcolor{ourrow}\textbf{3.6}
 & \cellcolor{ourrow}\underline{29.61} & \cellcolor{ourrow}0.215 & \cellcolor{ourrow}10.69 & \cellcolor{ourrow}\underline{10.1} & \cellcolor{ourrow}\textbf{16.8} \\
 & \textsc{SILO} & 26.70 & 0.222 & 8.21 & 149 & 7.4 & 26.61 & 0.399 & 47.98 & 20 & \underline{21.2} \\
 & \textsc{ReSample} & 27.92 & 0.253 & 10.74 & 1418 & \underline{7.0} & \textbf{30.25} & \underline{0.206} & \underline{9.37} & 138.5 & 22.5 \\
 & \textsc{PSLD} & \underline{28.63} & 0.288 & 12.23 & 390 & 7.9 & 26.63 & 0.450 & 49.18 & 69 & 25.6 \\
 & \textsc{GML-DPS} & \textbf{28.74} & 0.309 & 16.58 & 389 & 7.9 & 27.57 & 0.449 & 57.38 & 48 & 23.4 \\
 & \textsc{LDPS} & 28.00 & 0.327 & 19.95 & 331 & \underline{7.0} & 27.49 & 0.467 & 61.41 & 40 & 22.5 \\
\midrule
\rowcolor{ourrow} \cellcolor{white}\multirow{8}{*}{\rotatebox[origin=c]{90}{Inpainting}} & \paseo & \underline{21.76} & \underline{0.144} & 14.00 & 38.4 & \textbf{3.6}
 & \textbf{22.61} & \textbf{0.118} & \textbf{2.71} & 13.6 & \textbf{16.8} \\
 & \cellcolor{ourrow}\paseo$^\dagger$ & \cellcolor{ourrow}21.67 & \cellcolor{ourrow}0.147 & \cellcolor{ourrow}13.36 & \cellcolor{ourrow}\underline{31.2} & \cellcolor{ourrow}\textbf{3.6}
 & \cellcolor{ourrow}20.94 & \cellcolor{ourrow}0.144 & \cellcolor{ourrow}11.39 & \cellcolor{ourrow}\textbf{11.0} & \cellcolor{ourrow}\textbf{16.8} \\
 & \textsc{SILO} & \textbf{22.51} & \textbf{0.139} & \textbf{1.80} & 149 & 7.4 & 14.64 & 0.430 & 115.1 & 20 & \underline{21.2} \\
 & \textsc{DING} & 21.49 & 0.159 & \underline{4.59} & \textbf{8.8} & \textbf{3.6} & \underline{21.57} & \underline{0.134} & \underline{5.89} & \underline{12.6} & \textbf{16.8} \\
 & \textsc{ReSample} & 16.91 & 0.273 & 119.3 & 1418 & \underline{7.0} & 21.30 & 0.214 & 55.67 & 267.2 & 22.5 \\
 & \textsc{PSLD} & 20.58 & 0.357 & 17.23 & 390 & 7.9 & 17.36 & 0.459 & 69.94 & 48 & 23.4 \\
 & \textsc{GML-DPS} & 20.64 & 0.356 & 16.54 & 389 & 7.9 & 18.04 & 0.509 & 79.27 & 48 & 23.4 \\
 & \textsc{LDPS} & 20.58 & 0.368 & 16.02 & 331 & \underline{7.0} & 18.15 & 0.507 & 78.53 & 39 & 22.5 \\
\midrule
\rowcolor{ourrow} \cellcolor{white}\multirow{7}{*}{\rotatebox[origin=c]{90}{JPEG $q{=}10$}} & \paseo & 24.60 & 0.258 & 26.53 & \textbf{8.2} & \textbf{3.6}
 & 25.04 & \underline{0.350} & 71.24 & \textbf{13.9} & \textbf{16.8} \\
 & \cellcolor{ourrow}\paseo$^\dagger$ & \cellcolor{ourrow}25.25 & \cellcolor{ourrow}\underline{0.238} & \cellcolor{ourrow}22.73 & \cellcolor{ourrow}\underline{14.0} & \cellcolor{ourrow}\textbf{3.6}
 & \cellcolor{ourrow}26.40 & \cellcolor{ourrow}0.400 & \cellcolor{ourrow}58.74 & \cellcolor{ourrow}60.5 & \cellcolor{ourrow}\textbf{16.8} \\
 & \textsc{SILO} & 25.52 & \textbf{0.203} & \textbf{4.21} & 138 & 7.4 & 24.64 & 0.503 & 77.48 & \underline{20} & \underline{21.2} \\
 & \textsc{ReSample} & \underline{25.69} & 0.456 & 20.17 & 2438 & \underline{7.0} & \textbf{29.81} & \textbf{0.196} & \textbf{19.46} & 140.6 & 22.5 \\
 & \textsc{PSLD} & \multicolumn{5}{c@{\hspace{-2pt}\vrule\hspace{2pt}}}{n/a} & \multicolumn{5}{c}{n/a} \\
 & \textsc{GML-DPS} & \textbf{27.60} & 0.268 & \underline{7.54} & 402 & 7.9 & 27.94 & 0.413 & \underline{50.16} & 49 & 23.4 \\
 & \textsc{LDPS} & 24.53 & 0.373 & 17.71 & 412 & \underline{7.0} & \underline{27.98} & 0.424 & 54.48 & 40 & 22.5 \\
 \bottomrule
\end{tabular}
\end{table}
\vspace{-5pt}

\begin{table}[!ht]
\centering
\footnotesize
\caption{Settings of \paseo$^\dagger$ that differ from \paseo\
(Tables~\ref{tab:hyperparameters-ffhq} and~\ref{tab:hyperparameters-coco}); all
other settings, including the reverse steps, are unchanged. Fixed weight is the
inpainting rule of Appendix~\ref{sec:inpaint-variants}.}
\label{tab:paseo-dagger-settings}

\begin{tabular}{@{}lll@{}}
\toprule
Prior / data & Task & Changes from \paseo \\
\midrule
RV-v5.1 / FFHQ & SR $\times4$ & $P$: $3\to4$, $\rho$: $0.15\to0.3$ \\
 & SR $\times8$ & $C$: $2\to3$, $\rho$: $0.15\to0.6$ \\
 & Gaussian blur & $C$: $3\to32$ \\
 & Inpainting & fixed weight \\
 & JPEG & $P$: $4\to1$, $C$: $1\to16$, $\rho$: $0.3\to0.6$ \\
\midrule
SD3.5 / FFHQ & SR $\times4$ & $P$: $2\to3$, $C$: $2\to8$, $r$: $0.01\to0.005$ \\
 & SR $\times8$ & $P$: $1\to2$, $\rho$: $1\to0.25$ \\
 & Gaussian blur & $C$: $2\to4$ \\
 & Inpainting & fixed weight \\
 & JPEG & $P$: $3\to4$, $C$: $3\to20$, $r$: $0.02\to0.08$ \\
\midrule
SD3.5 / COCO & SR $\times4$ & none \\
 & SR $\times8$ & $P$: $3\to2$, $C$: $1\to12$, $\rho$: $0.45\to1$ \\
 & Gaussian blur & $P$: $4\to1$, $C$: $1\to4$, $\rho$: $0.4\to0.6$, $\lambda$: $0.005\to0$ \\
 & Inpainting & fixed weight \\
 & JPEG & $C$: $1\to16$ \\
\bottomrule
\end{tabular}
\end{table}
\FloatBarrier

\subsection{Full COCO-val2017 comparison}
\label{sec:coco-full}

Table~\ref{tab:coco-models-arc-full} reports all eight methods on the same
$1000$ COCO-val2017 images with SD3.5-medium, including PSNR.
It differs from Table~\ref{tab:coco-models-arc} in the main text in two
ways. First, it adds the diffusion-based baselines. Second, every baseline
here runs at its original budget of $145$ or $295$ NFEs, while \paseo\ uses
$97$ or $197$. The main-text table instead reruns the three flow-based
baselines at \paseo's own budget, where their quality is within noise of
the larger budget.

\paseo\ uses the arc anchor and LSDIR-trained latent operators, while
the baselines receive the true forward operator. LPIPS is measured over
the whole image, including inpainting. For inpainting, \paseo\ receives no
mask and weights its likelihood using the learned operator's sensitivity
(Appendix~\ref{sec:inpaint-variants}). \textsc{PSLD} is omitted for JPEG
because its update requires a linear operator adjoint
(Appendix~\ref{sec:comp-psld}).

\begin{table}[!ht]
\centering
\footnotesize
\setlength{\tabcolsep}{2.5pt}
\caption{Full COCO-val2017 comparison supporting Table~\ref{tab:coco-models-arc}, $n{=}1000$ (canonical $0$--$999$), SD3.5-medium
with LSDIR-trained latent operators, reported from the \emph{arc} anchor
(\texttt{interp\_arc\_std}) run on our shipped configurations. Every method runs at the NFE
listed in the table. KID is $\times10^3$, Time is
s/image, Mem is GiB. LPIPS is computed over the whole image on every
task, inpainting included, matching Table~\ref{tab:ffhq-models}; as there, our
inpainting row is never given the mask and restricts the likelihood with the
learned operator's sensitivity weight (Appendix~\ref{sec:inpaint-variants}).
Baselines are given the \emph{true} forward operator while ours has only a
learned latent surrogate. \textsc{PSLD} is omitted for JPEG because its update
requires a linear operator adjoint (Appendix~\ref{sec:comp-psld}).
\paseo$^\dagger$ uses the original anchor of
Algorithm~\ref{alg:silo-jding}, at the same NFEs as \paseo.}
\label{tab:coco-models-arc-full}
\begin{tabular}{@{}llrrrrrr@{}}
\toprule
Task & Method & NFE $\downarrow$ & PSNR $\uparrow$ & LPIPS $\downarrow$ & KID $\downarrow$ & Time $\downarrow$ & Mem $\downarrow$ \\
\midrule
\rowcolor{ourrow} SR $\times4$ & \paseo (arc) & \textbf{97} & 26.69 & \underline{0.208} & \underline{1.14} & \textbf{26.4} & \textbf{16.8} \\
\rowcolor{ourrow} & \paseo$^\dagger$ & \textbf{97} & 26.76 & \textbf{0.196} & \textbf{0.73} & \underline{27.8} & \textbf{16.8} \\
 & \textsc{ReSample} & \underline{145} & \underline{27.23} & 0.219 & 1.77 & 208.9 & 22.5 \\
 & \textsc{DAPS} & 145 & 23.01 & 0.681 & 29.65 & 233.4 & \underline{18.7} \\
 & \textsc{RedDiff} & 145 & 23.91 & 0.387 & 9.23 & 51.2 & 18.7 \\
 & \textsc{FlowDPS} & 145 & 26.31 & 0.380 & 12.38 & 115.7 & 18.7 \\
 & \textsc{FlowChef} & 145 & \textbf{27.63} & 0.339 & 7.65 & 343.4 & 18.7 \\
 & \textsc{PnP-Flow} & 145 & 26.34 & 0.449 & 20.93 & 50.4 & 18.7 \\
 & \textsc{PSLD} & 145 & 16.30 & 0.717 & 55.27 & 88.2 & 23.4 \\
\midrule
\rowcolor{ourrow} SR $\times8$ & \paseo (arc) & \textbf{197} & 23.64 & \textbf{0.344} & \underline{7.36} & \textbf{41.2} & \textbf{16.8} \\
\rowcolor{ourrow} & \paseo$^\dagger$ & \textbf{197} & 22.62 & 0.389 & 35.92 & \underline{92.6} & \textbf{16.8} \\
 & \textsc{ReSample} & \underline{295} & 24.32 & 0.462 & 15.93 & 465.5 & 22.5 \\
 & \textsc{DAPS} & 295 & \underline{24.56} & \underline{0.368} & \textbf{6.87} & 476.3 & 18.7 \\
 & \textsc{RedDiff} & 295 & 21.01 & 0.521 & 25.85 & 103.4 & \underline{18.7} \\
 & \textsc{FlowDPS} & 295 & 24.45 & 0.510 & 24.95 & 235.5 & 18.7 \\
 & \textsc{FlowChef} & 295 & 24.36 & 0.556 & 23.45 & 700.3 & 18.7 \\
 & \textsc{PnP-Flow} & 295 & \textbf{24.83} & 0.493 & 21.24 & 102.2 & 18.7 \\
 & \textsc{PSLD} & 295 & 17.22 & 0.636 & 11.41 & 179.4 & 23.4 \\
\midrule
\rowcolor{ourrow} Gaussian blur & \paseo (arc) & \textbf{97} & \underline{26.98} & \textbf{0.229} & \textbf{0.78} & \underline{22.7} & \textbf{16.8} \\
\rowcolor{ourrow} & \paseo$^\dagger$ & \textbf{97} & 26.88 & 0.280 & 2.10 & \textbf{18.3} & \textbf{16.8} \\
 & \textsc{ReSample} & \underline{145} & \textbf{27.14} & \underline{0.256} & \underline{1.09} & 221.2 & 22.5 \\
 & \textsc{DAPS} & 145 & 26.89 & 0.410 & 16.74 & 242.4 & 18.7 \\
 & \textsc{RedDiff} & 145 & 23.91 & 0.419 & 16.85 & 52.6 & \underline{18.7} \\
 & \textsc{FlowDPS} & 145 & 25.06 & 0.466 & 19.82 & 119.6 & 18.7 \\
 & \textsc{FlowChef} & 145 & 26.73 & 0.430 & 14.19 & 356.2 & 18.7 \\
 & \textsc{PnP-Flow} & 145 & 24.47 & 0.555 & 27.00 & 51.7 & 18.7 \\
 & \textsc{PSLD} & 145 & 18.51 & 0.631 & 30.98 & 89.5 & 23.4 \\
\midrule
\rowcolor{ourrow} Center inpainting & \paseo (arc) & \textbf{197} & \underline{19.31} & \textbf{0.174} & 5.26 & \underline{58.7} & \textbf{16.8} \\
\rowcolor{ourrow} & \paseo$^\dagger$ & \textbf{197} & 18.33 & \underline{0.186} & 5.65 & \textbf{40.3} & \textbf{16.8} \\
 & \textsc{ReSample} & \underline{295} & 18.31 & 0.278 & 38.61 & 1056.2 & 22.5 \\
 & \textsc{DAPS} & 295 & \textbf{19.37} & 0.253 & \underline{4.98} & 476.2 & \underline{18.7} \\
 & \textsc{RedDiff} & 295 & 17.13 & 0.445 & 32.54 & 104.1 & 18.7 \\
 & \textsc{FlowDPS} & 295 & 18.53 & 0.387 & 16.27 & 234.4 & 18.7 \\
 & \textsc{FlowChef} & 295 & 18.54 & 0.369 & 26.00 & 697.3 & 18.7 \\
 & \textsc{PnP-Flow} & 295 & 14.59 & 0.676 & 57.80 & 102.5 & 18.7 \\
 & \textsc{PSLD} & 295 & 17.47 & 0.492 & \textbf{3.16} & 179.6 & 23.4 \\
\midrule
\rowcolor{ourrow} JPEG $q{=}10$ & \paseo (arc) & \textbf{97} & 24.62 & 0.302 & 18.82 & \textbf{24.8} & \textbf{16.8} \\
\rowcolor{ourrow} & \paseo$^\dagger$ & \textbf{97} & 26.21 & 0.307 & 17.27 & 112.6 & \textbf{16.8} \\
 & \textsc{ReSample} & \underline{145} & \textbf{28.49} & \textbf{0.193} & \textbf{3.19} & 224.3 & 22.5 \\
 & \textsc{DAPS} & 145 & \underline{27.97} & \underline{0.263} & \underline{5.80} & 244.3 & \underline{18.7} \\
 & \textsc{RedDiff} & 145 & 25.06 & 0.364 & 21.15 & 53.0 & 18.7 \\
 & \textsc{FlowDPS} & 145 & 26.63 & 0.379 & 25.19 & 121.4 & 18.7 \\
 & \textsc{FlowChef} & 145 & 26.96 & 0.346 & 8.61 & 360.0 & 18.7 \\
 & \textsc{PnP-Flow} & 145 & 26.00 & 0.452 & 27.10 & \underline{52.3} & 18.7 \\
\bottomrule
\end{tabular}
\end{table}

\clearpage
\section{Additional qualitative results}
\label{sec:appendix-qualitative}

{
Following the appendix of \citet{raphaeli2025silo}, each figure shows one task
under one prior: the ground truth $\rvx$, the measurement $\rvy$, \paseo, and
every baseline that the corresponding table reports under that prior, all at
$\sigma_{\rvy}=0.01$. FFHQ has one figure per prior
(Figures~\ref{fig:appendix-qual-ffhq-rv-sr4}--\ref{fig:appendix-qual-ffhq-sd35-jpeg});
COCO-val2017 is reported with SD3.5-medium only
(Figures~\ref{fig:appendix-qual-coco-sd35-sr4}--\ref{fig:appendix-qual-coco-sd35-jpeg}).

\paragraph{Rows.} Each figure shows the five images with the lowest \paseo\
LPIPS under that figure's prior among the first $100$ test images, and every
method on those same images. These are \paseo's best cases under a stated
rule, chosen without looking at any baseline; they are not a random sample, and
Tables~\ref{tab:ffhq-models-full} and~\ref{tab:coco-models-arc-full} remain the
measure of typical performance. Below each row, every column's crop of the
$256\times128$ region boxed on $\rvx$ is shown at native resolution. The box is
placed without looking at any reconstruction: a fixed box on the eyes for FFHQ,
whose faces are aligned; the center of the mask for COCO inpainting; and the
most detailed window of $\rvx$ (largest mean gradient magnitude) for the other
COCO tasks.

\paragraph{Sources.} Every \paseo\ panel, and every baseline panel on FFHQ with
SD3.5-medium and on COCO, is taken from the run behind its row in
Tables~\ref{tab:ffhq-models-full} and~\ref{tab:coco-models-arc-full}; $\rvy$ is
the measurement \paseo's run received, and each baseline draws its own
observation noise. The RV-v5.1 block of Table~\ref{tab:ffhq-models} instead
quotes the numbers published by \citet{raphaeli2025silo}, for which no images
exist. In those figures \textsc{SILO} is the authors' released code at its
default settings, and \textsc{ReSample}, \textsc{PSLD}, \textsc{GML} and
\textsc{LDPS} are our implementations with the true forward operator, run with
the settings \citet{raphaeli2025silo} report using, those of the official
repositories: $1000$ DDIM steps for the \textsc{PSLD} family and $500$ for
\textsc{ReSample}, with hard data consistency every $10$ steps.
As a check, the same settings were run on three further images per task that
are not among the rows. \textsc{PSLD} and \textsc{GML} land within $0.06$ LPIPS
of the published values, and \textsc{LDPS} below them on every task.
\textsc{ReSample} is well below its published values on SR and JPEG; on blur
the official values gave $0.375$ against the published $0.253$, so its blur
panels use a larger latent step size ($0.08$) with consistency every $5$ steps,
the best of a small grid on those three images ($0.285$). On inpainting no
setting we tried makes \textsc{ReSample} fill the box under RV-v5.1 ($0.392$
against $0.273$), so its column in
Figure~\ref{fig:appendix-qual-ffhq-rv-inpaint} is worse than its published row.
\textsc{PSLD} has no JPEG panel, as its update needs the operator's adjoint, and
\textsc{DING} is shown for inpainting only.
}

\ifdefined\aqw\else\newlength{\aqw}\fi
\definecolor{aqzoom}{rgb}{1.00,0.98,0.01}
\providecommand*{\aqbox}[5]{\begin{tikzpicture}[baseline=0pt]%
  \node[inner sep=0pt, anchor=south west] at (0,0) {\includegraphics[width=\aqw]{#1}};%
  \draw[aqzoom, line width=0.6pt] ({#2/512*\aqw}, {\aqw-#3/512*\aqw})
    rectangle ({#4/512*\aqw}, {\aqw-#5/512*\aqw});%
\end{tikzpicture}}
\begin{figure}[p]
\centering
\setlength{\aqw}{\dimexpr(\linewidth-7pt)/8\relax}
\setlength{\tabcolsep}{0pt}
\renewcommand{\arraystretch}{0}
\begin{tabular}{@{}c@{\hspace{1pt}}c@{\hspace{1pt}}c@{\hspace{1pt}}c@{\hspace{1pt}}c@{\hspace{1pt}}c@{\hspace{1pt}}c@{\hspace{1pt}}c@{}}
\footnotesize $\rvx$ & \footnotesize $\rvy$ & \footnotesize \paseo & \footnotesize \textsc{SILO} & \footnotesize \textsc{ReSample} & \footnotesize \textsc{PSLD} & \footnotesize \textsc{GML} & \footnotesize \textsc{LDPS} \\[3pt]
\aqbox{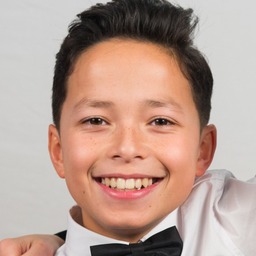}{128}{176}{384}{304} &
\includegraphics[width=\aqw]{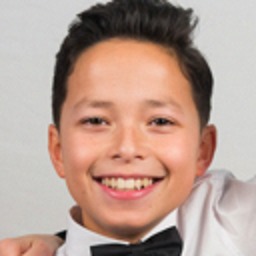} &
\includegraphics[width=\aqw]{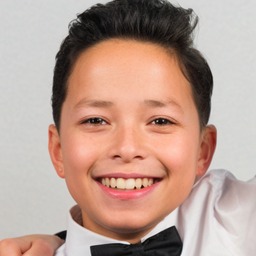} &
\includegraphics[width=\aqw]{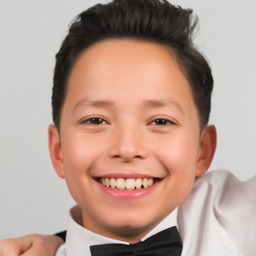} &
\includegraphics[width=\aqw]{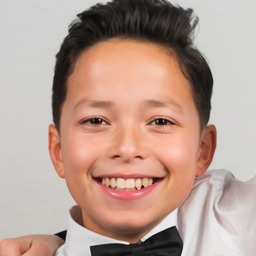} &
\includegraphics[width=\aqw]{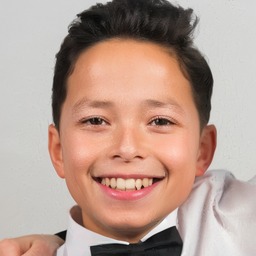} &
\includegraphics[width=\aqw]{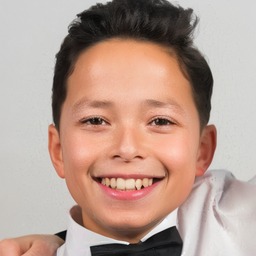} &
\includegraphics[width=\aqw]{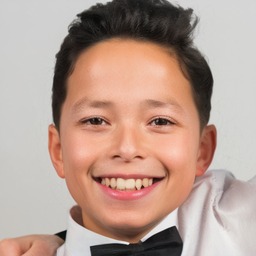} \\[1pt]
\includegraphics[width=\aqw]{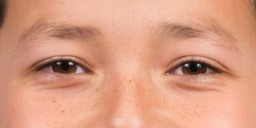} &
\includegraphics[width=\aqw]{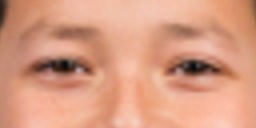} &
\includegraphics[width=\aqw]{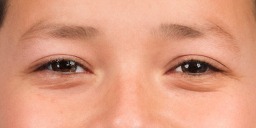} &
\includegraphics[width=\aqw]{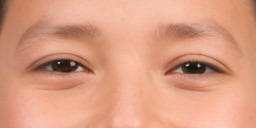} &
\includegraphics[width=\aqw]{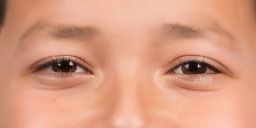} &
\includegraphics[width=\aqw]{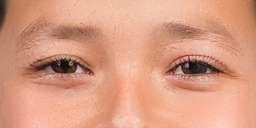} &
\includegraphics[width=\aqw]{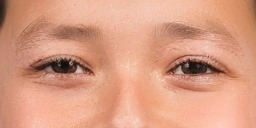} &
\includegraphics[width=\aqw]{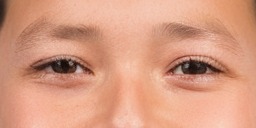} \\
\end{tabular}
\caption{SR $\times4$ with $\sigma_{\rvy}=0.01$, FFHQ, RV-v5.1 prior. Below each row, the region boxed on $\rvx$ in every column. \paseo\ is from the runs of Table~\ref{tab:ffhq-models-full}. The other baselines' rows there quote SILO's published numbers, so their panels come from our own runs (Appendix~\ref{sec:appendix-qualitative}).}
\label{fig:appendix-qual-ffhq-rv-sr4}
\end{figure}

\begin{figure}[p]
\centering
\setlength{\aqw}{\dimexpr(\linewidth-7pt)/8\relax}
\setlength{\tabcolsep}{0pt}
\renewcommand{\arraystretch}{0}
\begin{tabular}{@{}c@{\hspace{1pt}}c@{\hspace{1pt}}c@{\hspace{1pt}}c@{\hspace{1pt}}c@{\hspace{1pt}}c@{\hspace{1pt}}c@{\hspace{1pt}}c@{}}
\footnotesize $\rvx$ & \footnotesize $\rvy$ & \footnotesize \paseo & \footnotesize \textsc{SILO} & \footnotesize \textsc{ReSample} & \footnotesize \textsc{PSLD} & \footnotesize \textsc{GML} & \footnotesize \textsc{LDPS} \\[3pt]
\aqbox{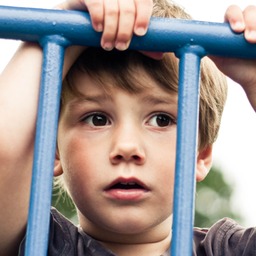}{128}{176}{384}{304} &
\includegraphics[width=\aqw]{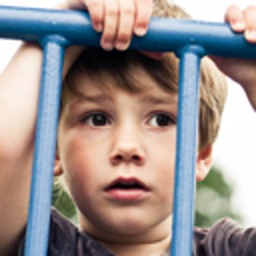} &
\includegraphics[width=\aqw]{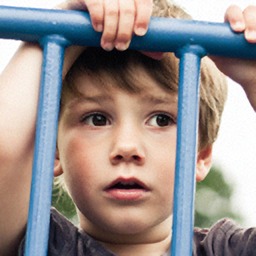} &
\includegraphics[width=\aqw]{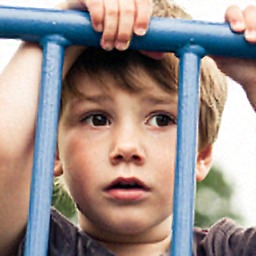} &
\includegraphics[width=\aqw]{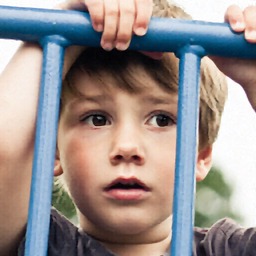} &
\includegraphics[width=\aqw]{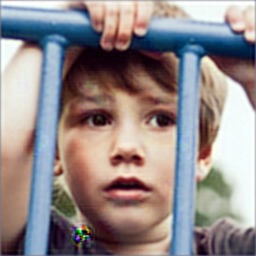} &
\includegraphics[width=\aqw]{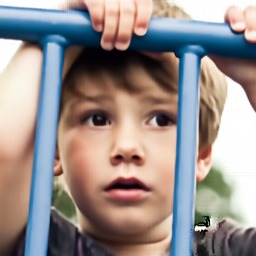} &
\includegraphics[width=\aqw]{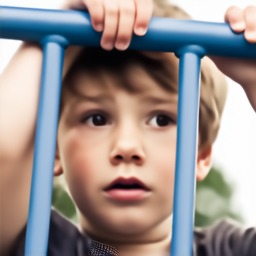} \\[1pt]
\includegraphics[width=\aqw]{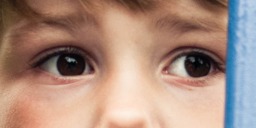} &
\includegraphics[width=\aqw]{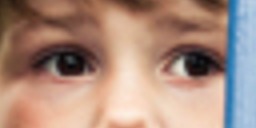} &
\includegraphics[width=\aqw]{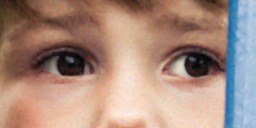} &
\includegraphics[width=\aqw]{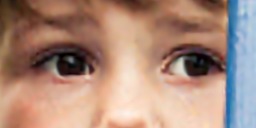} &
\includegraphics[width=\aqw]{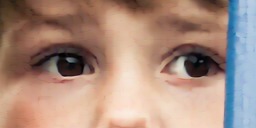} &
\includegraphics[width=\aqw]{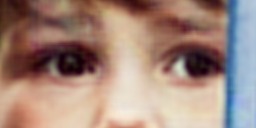} &
\includegraphics[width=\aqw]{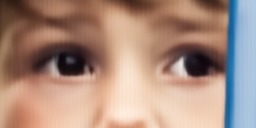} &
\includegraphics[width=\aqw]{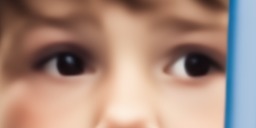} \\
\end{tabular}
\caption{SR $\times4$ with $\sigma_{\rvy}=0.01$, FFHQ, SD3.5-medium prior. Below each row, the region boxed on $\rvx$ in every column. Panels are from the runs of Table~\ref{tab:ffhq-models-full}.}
\label{fig:appendix-qual-ffhq-sd35-sr4}
\end{figure}

\begin{figure}[p]
\centering
\setlength{\aqw}{\dimexpr(\linewidth-7pt)/8\relax}
\setlength{\tabcolsep}{0pt}
\renewcommand{\arraystretch}{0}
\begin{tabular}{@{}c@{\hspace{1pt}}c@{\hspace{1pt}}c@{\hspace{1pt}}c@{\hspace{1pt}}c@{\hspace{1pt}}c@{\hspace{1pt}}c@{\hspace{1pt}}c@{}}
\footnotesize $\rvx$ & \footnotesize $\rvy$ & \footnotesize \paseo & \footnotesize \textsc{SILO} & \footnotesize \textsc{ReSample} & \footnotesize \textsc{PSLD} & \footnotesize \textsc{GML} & \footnotesize \textsc{LDPS} \\[3pt]
\aqbox{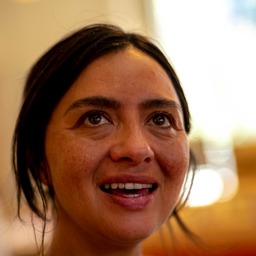}{128}{176}{384}{304} &
\includegraphics[width=\aqw]{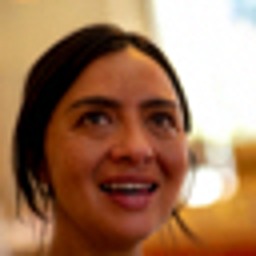} &
\includegraphics[width=\aqw]{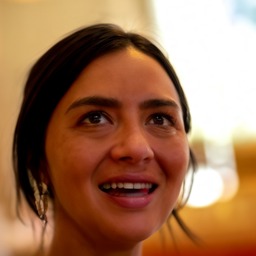} &
\includegraphics[width=\aqw]{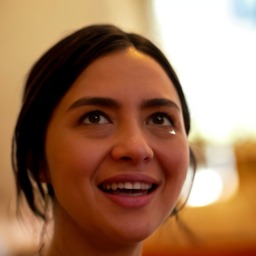} &
\includegraphics[width=\aqw]{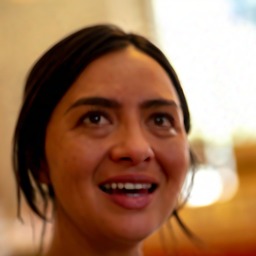} &
\includegraphics[width=\aqw]{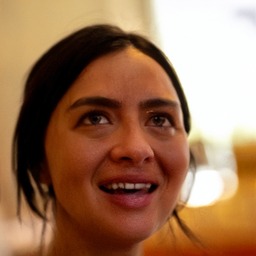} &
\includegraphics[width=\aqw]{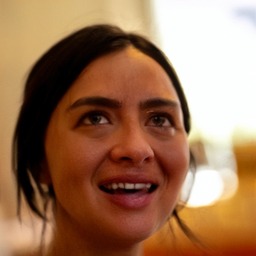} &
\includegraphics[width=\aqw]{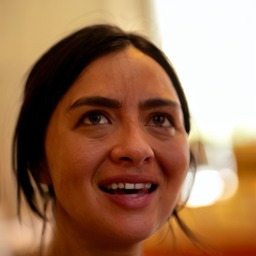} \\[1pt]
\includegraphics[width=\aqw]{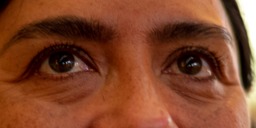} &
\includegraphics[width=\aqw]{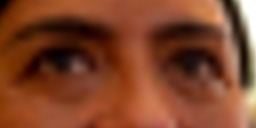} &
\includegraphics[width=\aqw]{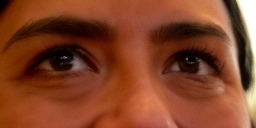} &
\includegraphics[width=\aqw]{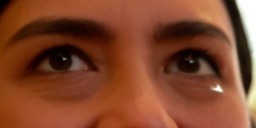} &
\includegraphics[width=\aqw]{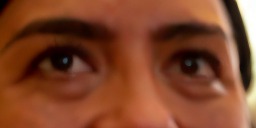} &
\includegraphics[width=\aqw]{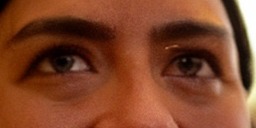} &
\includegraphics[width=\aqw]{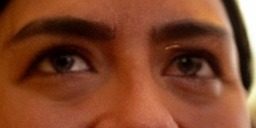} &
\includegraphics[width=\aqw]{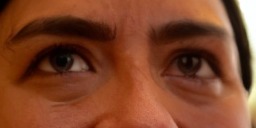} \\
\end{tabular}
\caption{SR $\times8$ with $\sigma_{\rvy}=0.01$, FFHQ, RV-v5.1 prior. Below each row, the region boxed on $\rvx$ in every column. \paseo\ is from the runs of Table~\ref{tab:ffhq-models-full}. The other baselines' rows there quote SILO's published numbers, so their panels come from our own runs (Appendix~\ref{sec:appendix-qualitative}).}
\label{fig:appendix-qual-ffhq-rv-sr8}
\end{figure}

\begin{figure}[p]
\centering
\setlength{\aqw}{\dimexpr(\linewidth-7pt)/8\relax}
\setlength{\tabcolsep}{0pt}
\renewcommand{\arraystretch}{0}
\begin{tabular}{@{}c@{\hspace{1pt}}c@{\hspace{1pt}}c@{\hspace{1pt}}c@{\hspace{1pt}}c@{\hspace{1pt}}c@{\hspace{1pt}}c@{\hspace{1pt}}c@{}}
\footnotesize $\rvx$ & \footnotesize $\rvy$ & \footnotesize \paseo & \footnotesize \textsc{SILO} & \footnotesize \textsc{ReSample} & \footnotesize \textsc{PSLD} & \footnotesize \textsc{GML} & \footnotesize \textsc{LDPS} \\[3pt]
\aqbox{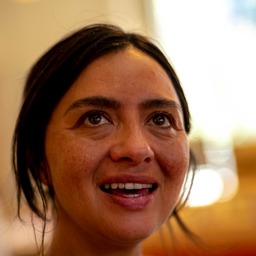}{128}{176}{384}{304} &
\includegraphics[width=\aqw]{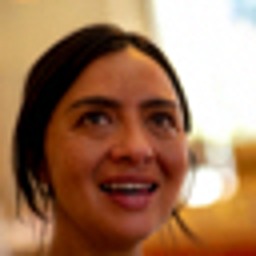} &
\includegraphics[width=\aqw]{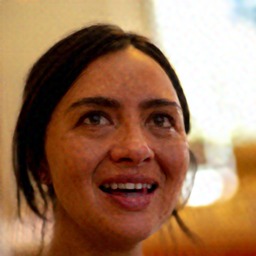} &
\includegraphics[width=\aqw]{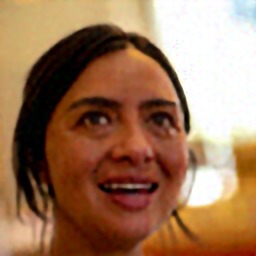} &
\includegraphics[width=\aqw]{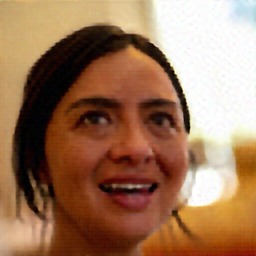} &
\includegraphics[width=\aqw]{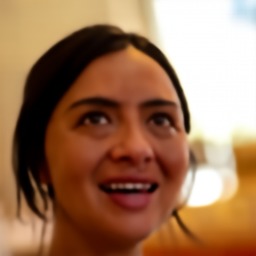} &
\includegraphics[width=\aqw]{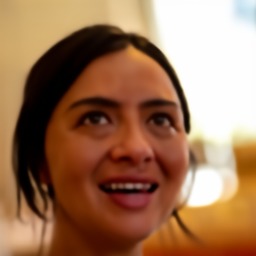} &
\includegraphics[width=\aqw]{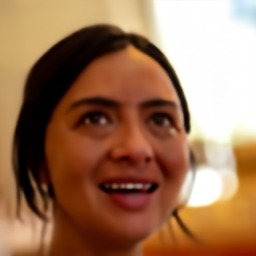} \\[1pt]
\includegraphics[width=\aqw]{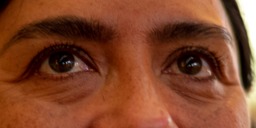} &
\includegraphics[width=\aqw]{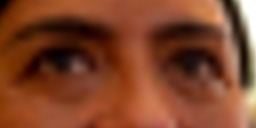} &
\includegraphics[width=\aqw]{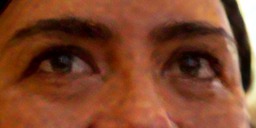} &
\includegraphics[width=\aqw]{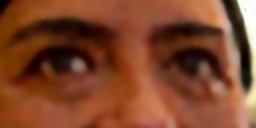} &
\includegraphics[width=\aqw]{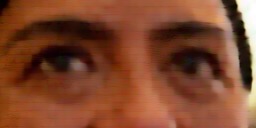} &
\includegraphics[width=\aqw]{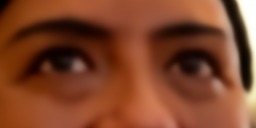} &
\includegraphics[width=\aqw]{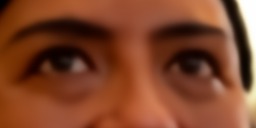} &
\includegraphics[width=\aqw]{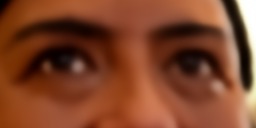} \\
\end{tabular}
\caption{SR $\times8$ with $\sigma_{\rvy}=0.01$, FFHQ, SD3.5-medium prior. Below each row, the region boxed on $\rvx$ in every column. Panels are from the runs of Table~\ref{tab:ffhq-models-full}.}
\label{fig:appendix-qual-ffhq-sd35-sr8}
\end{figure}

\begin{figure}[p]
\centering
\setlength{\aqw}{\dimexpr(\linewidth-7pt)/8\relax}
\setlength{\tabcolsep}{0pt}
\renewcommand{\arraystretch}{0}
\begin{tabular}{@{}c@{\hspace{1pt}}c@{\hspace{1pt}}c@{\hspace{1pt}}c@{\hspace{1pt}}c@{\hspace{1pt}}c@{\hspace{1pt}}c@{\hspace{1pt}}c@{}}
\footnotesize $\rvx$ & \footnotesize $\rvy$ & \footnotesize \paseo & \footnotesize \textsc{SILO} & \footnotesize \textsc{ReSample} & \footnotesize \textsc{PSLD} & \footnotesize \textsc{GML} & \footnotesize \textsc{LDPS} \\[3pt]
\aqbox{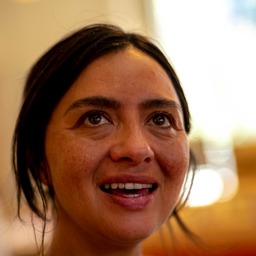}{128}{176}{384}{304} &
\includegraphics[width=\aqw]{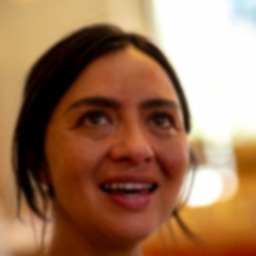} &
\includegraphics[width=\aqw]{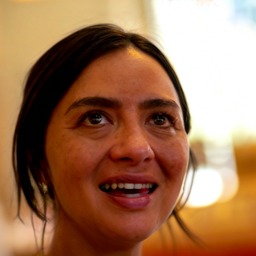} &
\includegraphics[width=\aqw]{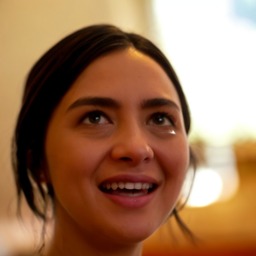} &
\includegraphics[width=\aqw]{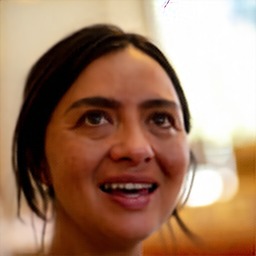} &
\includegraphics[width=\aqw]{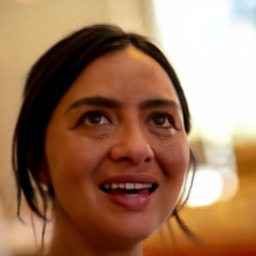} &
\includegraphics[width=\aqw]{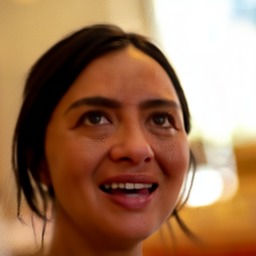} &
\includegraphics[width=\aqw]{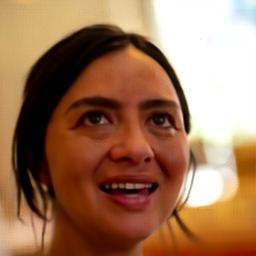} \\[1pt]
\includegraphics[width=\aqw]{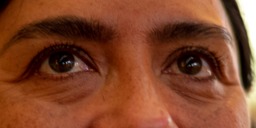} &
\includegraphics[width=\aqw]{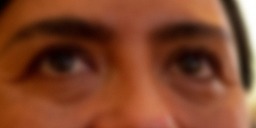} &
\includegraphics[width=\aqw]{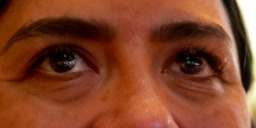} &
\includegraphics[width=\aqw]{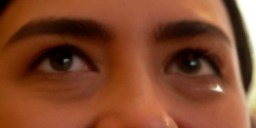} &
\includegraphics[width=\aqw]{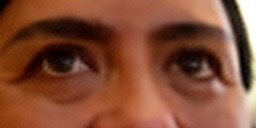} &
\includegraphics[width=\aqw]{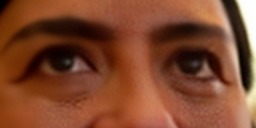} &
\includegraphics[width=\aqw]{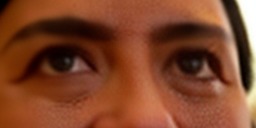} &
\includegraphics[width=\aqw]{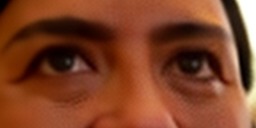} \\
\end{tabular}
\caption{Gaussian blur with $\sigma_{\rvy}=0.01$, FFHQ, RV-v5.1 prior. Below each row, the region boxed on $\rvx$ in every column. \paseo\ is from the runs of Table~\ref{tab:ffhq-models-full}. The other baselines' rows there quote SILO's published numbers, so their panels come from our own runs (Appendix~\ref{sec:appendix-qualitative}).}
\label{fig:appendix-qual-ffhq-rv-blur}
\end{figure}

\begin{figure}[p]
\centering
\setlength{\aqw}{\dimexpr(\linewidth-7pt)/8\relax}
\setlength{\tabcolsep}{0pt}
\renewcommand{\arraystretch}{0}
\begin{tabular}{@{}c@{\hspace{1pt}}c@{\hspace{1pt}}c@{\hspace{1pt}}c@{\hspace{1pt}}c@{\hspace{1pt}}c@{\hspace{1pt}}c@{\hspace{1pt}}c@{}}
\footnotesize $\rvx$ & \footnotesize $\rvy$ & \footnotesize \paseo & \footnotesize \textsc{SILO} & \footnotesize \textsc{ReSample} & \footnotesize \textsc{PSLD} & \footnotesize \textsc{GML} & \footnotesize \textsc{LDPS} \\[3pt]
\aqbox{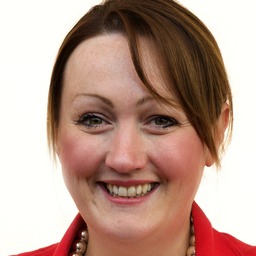}{128}{176}{384}{304} &
\includegraphics[width=\aqw]{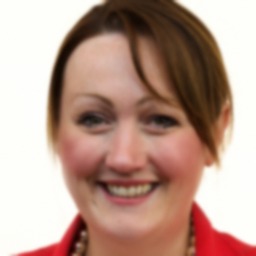} &
\includegraphics[width=\aqw]{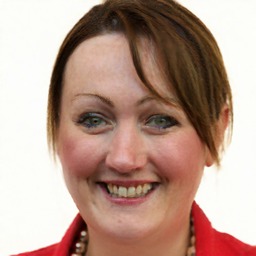} &
\includegraphics[width=\aqw]{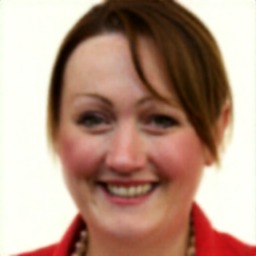} &
\includegraphics[width=\aqw]{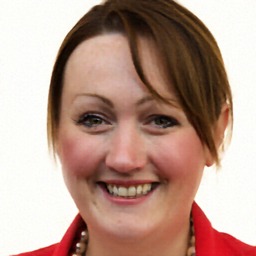} &
\includegraphics[width=\aqw]{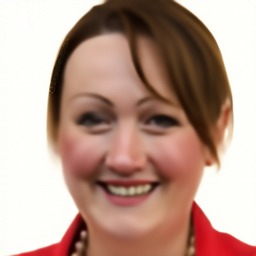} &
\includegraphics[width=\aqw]{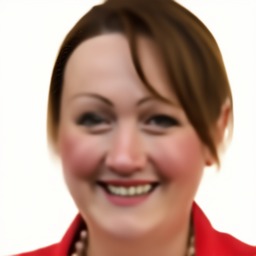} &
\includegraphics[width=\aqw]{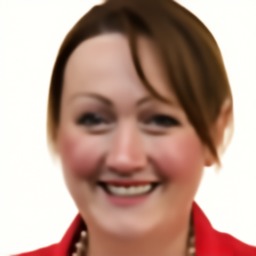} \\[1pt]
\includegraphics[width=\aqw]{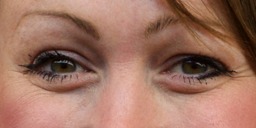} &
\includegraphics[width=\aqw]{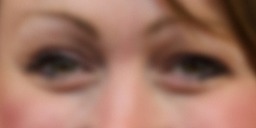} &
\includegraphics[width=\aqw]{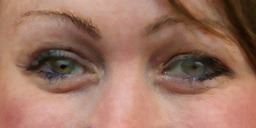} &
\includegraphics[width=\aqw]{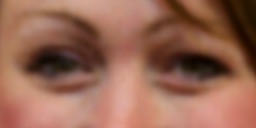} &
\includegraphics[width=\aqw]{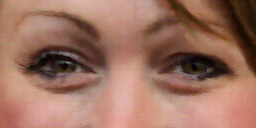} &
\includegraphics[width=\aqw]{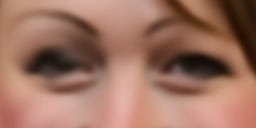} &
\includegraphics[width=\aqw]{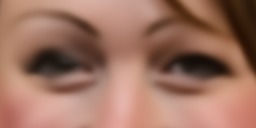} &
\includegraphics[width=\aqw]{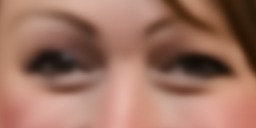} \\
\end{tabular}
\caption{Gaussian blur with $\sigma_{\rvy}=0.01$, FFHQ, SD3.5-medium prior. Below each row, the region boxed on $\rvx$ in every column. Panels are from the runs of Table~\ref{tab:ffhq-models-full}.}
\label{fig:appendix-qual-ffhq-sd35-blur}
\end{figure}

\begin{figure}[p]
\centering
\setlength{\aqw}{\dimexpr(\linewidth-8pt)/9\relax}
\setlength{\tabcolsep}{0pt}
\renewcommand{\arraystretch}{0}
\begin{tabular}{@{}c@{\hspace{1pt}}c@{\hspace{1pt}}c@{\hspace{1pt}}c@{\hspace{1pt}}c@{\hspace{1pt}}c@{\hspace{1pt}}c@{\hspace{1pt}}c@{\hspace{1pt}}c@{}}
\scriptsize $\rvx$ & \scriptsize $\rvy$ & \scriptsize \paseo & \scriptsize \textsc{SILO} & \scriptsize \textsc{ReSample} & \scriptsize \textsc{PSLD} & \scriptsize \textsc{GML} & \scriptsize \textsc{LDPS} & \scriptsize \textsc{DING} \\[3pt]
\aqbox{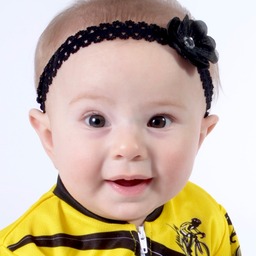}{128}{176}{384}{304} &
\includegraphics[width=\aqw]{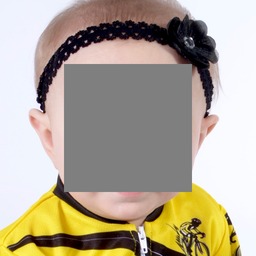} &
\includegraphics[width=\aqw]{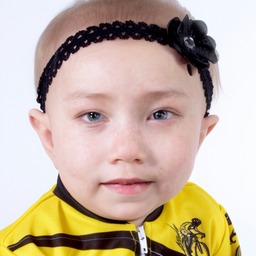} &
\includegraphics[width=\aqw]{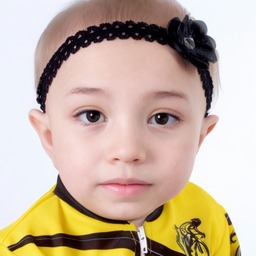} &
\includegraphics[width=\aqw]{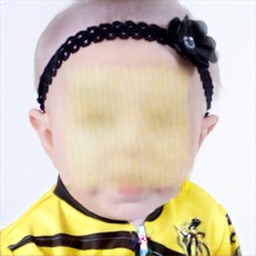} &
\includegraphics[width=\aqw]{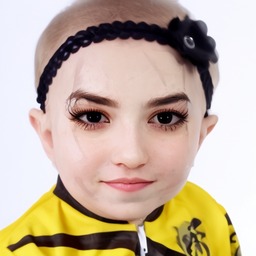} &
\includegraphics[width=\aqw]{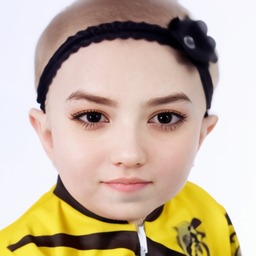} &
\includegraphics[width=\aqw]{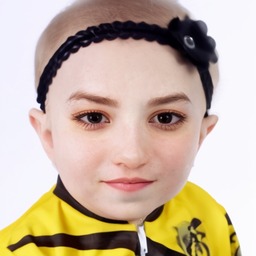} &
\includegraphics[width=\aqw]{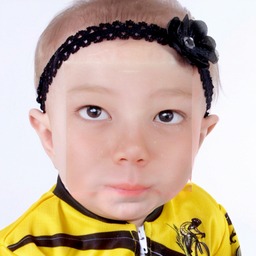} \\[1pt]
\includegraphics[width=\aqw]{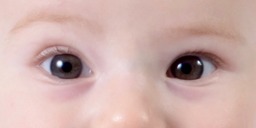} &
\includegraphics[width=\aqw]{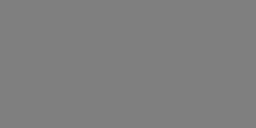} &
\includegraphics[width=\aqw]{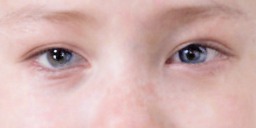} &
\includegraphics[width=\aqw]{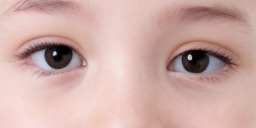} &
\includegraphics[width=\aqw]{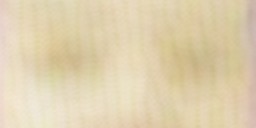} &
\includegraphics[width=\aqw]{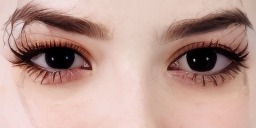} &
\includegraphics[width=\aqw]{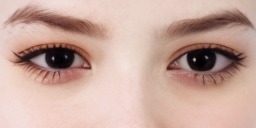} &
\includegraphics[width=\aqw]{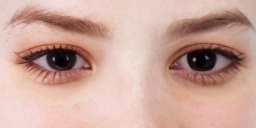} &
\includegraphics[width=\aqw]{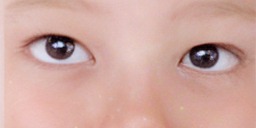} \\
\end{tabular}
\caption{Box inpainting with $\sigma_{\rvy}=0.01$, FFHQ, RV-v5.1 prior. Below each row, the region boxed on $\rvx$ in every column. \paseo\ and \textsc{DING} are from the runs of Table~\ref{tab:ffhq-models-full}. The other baselines' rows there quote SILO's published numbers, so their panels come from our own runs (Appendix~\ref{sec:appendix-qualitative}). In those runs \textsc{ReSample} does not fill the box, so its column is worse than its published row.}
\label{fig:appendix-qual-ffhq-rv-inpaint}
\end{figure}

\begin{figure}[p]
\centering
\setlength{\aqw}{\dimexpr(\linewidth-8pt)/9\relax}
\setlength{\tabcolsep}{0pt}
\renewcommand{\arraystretch}{0}
\begin{tabular}{@{}c@{\hspace{1pt}}c@{\hspace{1pt}}c@{\hspace{1pt}}c@{\hspace{1pt}}c@{\hspace{1pt}}c@{\hspace{1pt}}c@{\hspace{1pt}}c@{\hspace{1pt}}c@{}}
\scriptsize $\rvx$ & \scriptsize $\rvy$ & \scriptsize \paseo & \scriptsize \textsc{SILO} & \scriptsize \textsc{ReSample} & \scriptsize \textsc{PSLD} & \scriptsize \textsc{GML} & \scriptsize \textsc{LDPS} & \scriptsize \textsc{DING} \\[3pt]
\aqbox{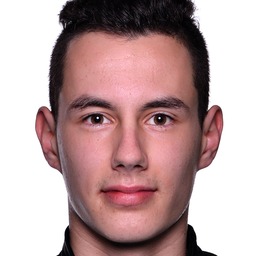}{128}{176}{384}{304} &
\includegraphics[width=\aqw]{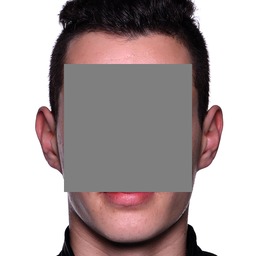} &
\includegraphics[width=\aqw]{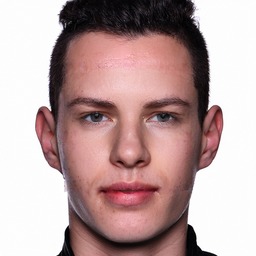} &
\includegraphics[width=\aqw]{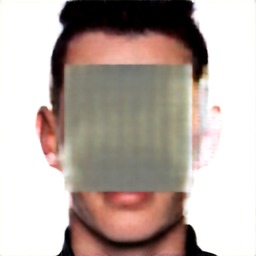} &
\includegraphics[width=\aqw]{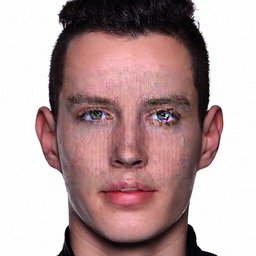} &
\includegraphics[width=\aqw]{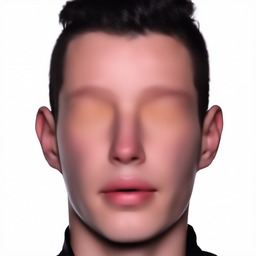} &
\includegraphics[width=\aqw]{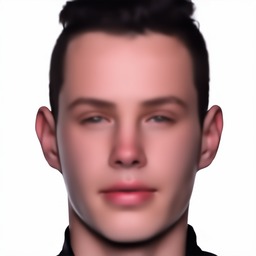} &
\includegraphics[width=\aqw]{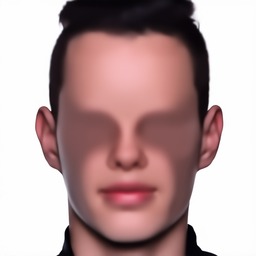} &
\includegraphics[width=\aqw]{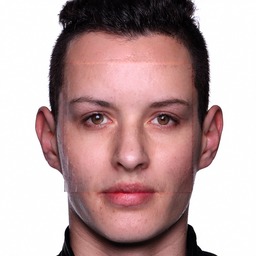} \\[1pt]
\includegraphics[width=\aqw]{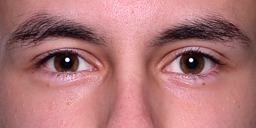} &
\includegraphics[width=\aqw]{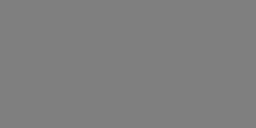} &
\includegraphics[width=\aqw]{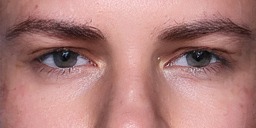} &
\includegraphics[width=\aqw]{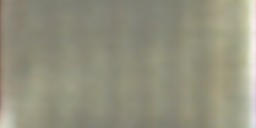} &
\includegraphics[width=\aqw]{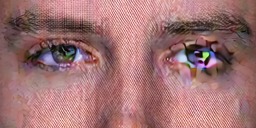} &
\includegraphics[width=\aqw]{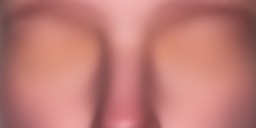} &
\includegraphics[width=\aqw]{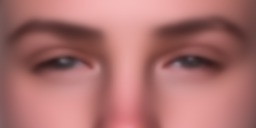} &
\includegraphics[width=\aqw]{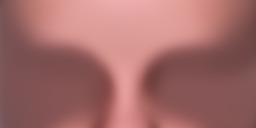} &
\includegraphics[width=\aqw]{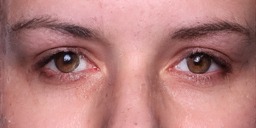} \\
\end{tabular}
\caption{Box inpainting with $\sigma_{\rvy}=0.01$, FFHQ, SD3.5-medium prior. Below each row, the region boxed on $\rvx$ in every column. Panels are from the runs of Table~\ref{tab:ffhq-models-full}.}
\label{fig:appendix-qual-ffhq-sd35-inpaint}
\end{figure}

\begin{figure}[p]
\centering
\setlength{\aqw}{\dimexpr(\linewidth-6pt)/7\relax}
\setlength{\tabcolsep}{0pt}
\renewcommand{\arraystretch}{0}
\begin{tabular}{@{}c@{\hspace{1pt}}c@{\hspace{1pt}}c@{\hspace{1pt}}c@{\hspace{1pt}}c@{\hspace{1pt}}c@{\hspace{1pt}}c@{}}
\footnotesize $\rvx$ & \footnotesize $\rvy$ & \footnotesize \paseo & \footnotesize \textsc{SILO} & \footnotesize \textsc{ReSample} & \footnotesize \textsc{GML} & \footnotesize \textsc{LDPS} \\[3pt]
\aqbox{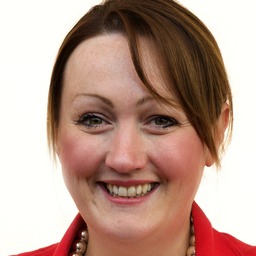}{128}{176}{384}{304} &
\includegraphics[width=\aqw]{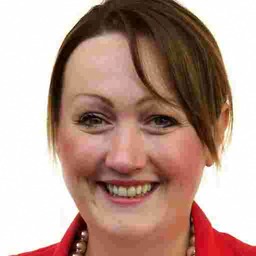} &
\includegraphics[width=\aqw]{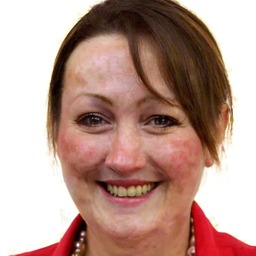} &
\includegraphics[width=\aqw]{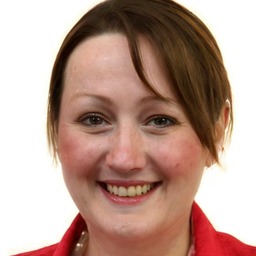} &
\includegraphics[width=\aqw]{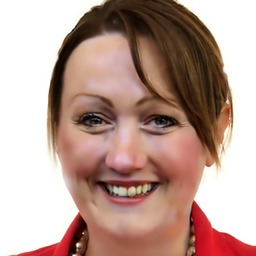} &
\includegraphics[width=\aqw]{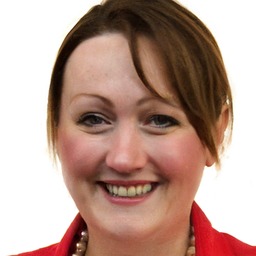} &
\includegraphics[width=\aqw]{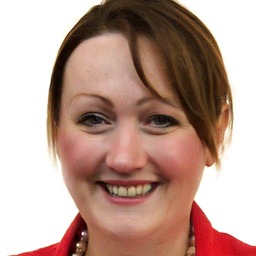} \\[1pt]
\includegraphics[width=\aqw]{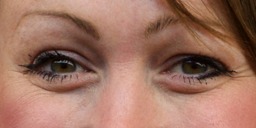} &
\includegraphics[width=\aqw]{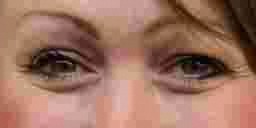} &
\includegraphics[width=\aqw]{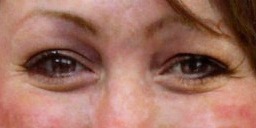} &
\includegraphics[width=\aqw]{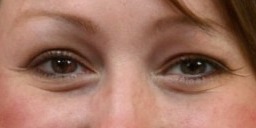} &
\includegraphics[width=\aqw]{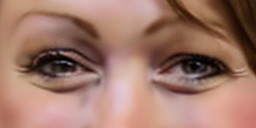} &
\includegraphics[width=\aqw]{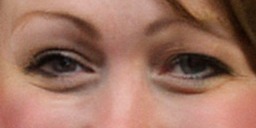} &
\includegraphics[width=\aqw]{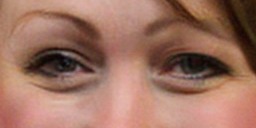} \\
\end{tabular}
\caption{JPEG ($q{=}10$) with $\sigma_{\rvy}=0.01$, FFHQ, RV-v5.1 prior. Below each row, the region boxed on $\rvx$ in every column. \paseo\ is from the runs of Table~\ref{tab:ffhq-models-full}. The other baselines' rows there quote SILO's published numbers, so their panels come from our own runs (Appendix~\ref{sec:appendix-qualitative}).}
\label{fig:appendix-qual-ffhq-rv-jpeg}
\end{figure}

\begin{figure}[p]
\centering
\setlength{\aqw}{\dimexpr(\linewidth-6pt)/7\relax}
\setlength{\tabcolsep}{0pt}
\renewcommand{\arraystretch}{0}
\begin{tabular}{@{}c@{\hspace{1pt}}c@{\hspace{1pt}}c@{\hspace{1pt}}c@{\hspace{1pt}}c@{\hspace{1pt}}c@{\hspace{1pt}}c@{}}
\footnotesize $\rvx$ & \footnotesize $\rvy$ & \footnotesize \paseo & \footnotesize \textsc{SILO} & \footnotesize \textsc{ReSample} & \footnotesize \textsc{GML} & \footnotesize \textsc{LDPS} \\[3pt]
\aqbox{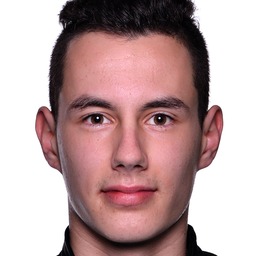}{128}{176}{384}{304} &
\includegraphics[width=\aqw]{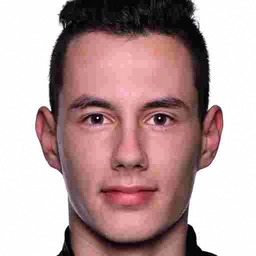} &
\includegraphics[width=\aqw]{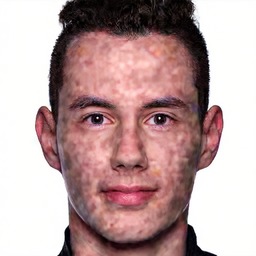} &
\includegraphics[width=\aqw]{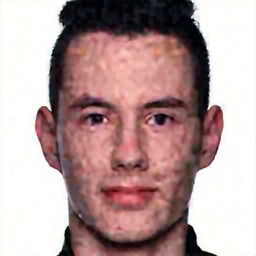} &
\includegraphics[width=\aqw]{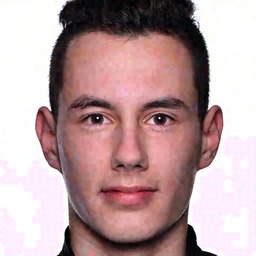} &
\includegraphics[width=\aqw]{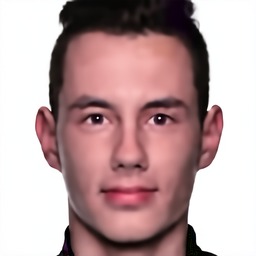} &
\includegraphics[width=\aqw]{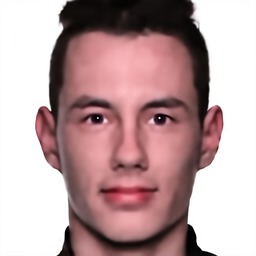} \\[1pt]
\includegraphics[width=\aqw]{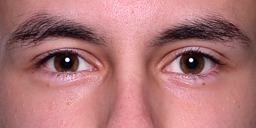} &
\includegraphics[width=\aqw]{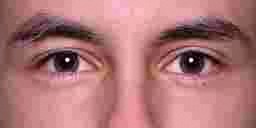} &
\includegraphics[width=\aqw]{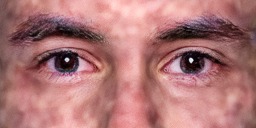} &
\includegraphics[width=\aqw]{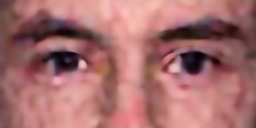} &
\includegraphics[width=\aqw]{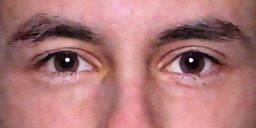} &
\includegraphics[width=\aqw]{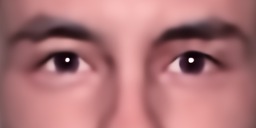} &
\includegraphics[width=\aqw]{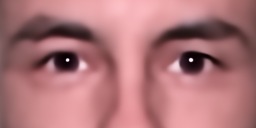} \\
\end{tabular}
\caption{JPEG ($q{=}10$) with $\sigma_{\rvy}=0.01$, FFHQ, SD3.5-medium prior. Below each row, the region boxed on $\rvx$ in every column. Panels are from the runs of Table~\ref{tab:ffhq-models-full}.}
\label{fig:appendix-qual-ffhq-sd35-jpeg}
\end{figure}

\begin{figure}[p]
\centering
\setlength{\aqw}{\dimexpr(\linewidth-9pt)/10\relax}
\setlength{\tabcolsep}{0pt}
\renewcommand{\arraystretch}{0}
\begin{tabular}{@{}c@{\hspace{1pt}}c@{\hspace{1pt}}c@{\hspace{1pt}}c@{\hspace{1pt}}c@{\hspace{1pt}}c@{\hspace{1pt}}c@{\hspace{1pt}}c@{\hspace{1pt}}c@{\hspace{1pt}}c@{}}
\scriptsize $\rvx$ & \scriptsize $\rvy$ & \scriptsize \paseo & \scriptsize \textsc{ReSample} & \scriptsize \textsc{DAPS} & \scriptsize \textsc{RedDiff} & \scriptsize \textsc{FlowDPS} & \scriptsize \textsc{FlowChef} & \scriptsize \textsc{PnP-Flow} & \scriptsize \textsc{PSLD} \\[3pt]
\aqbox{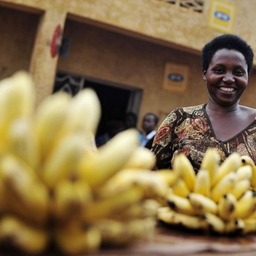}{256}{208}{512}{336} &
\includegraphics[width=\aqw]{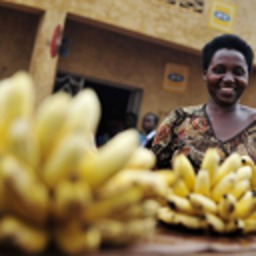} &
\includegraphics[width=\aqw]{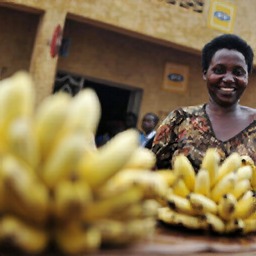} &
\includegraphics[width=\aqw]{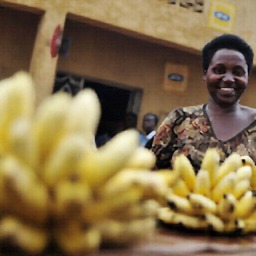} &
\includegraphics[width=\aqw]{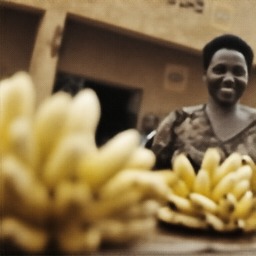} &
\includegraphics[width=\aqw]{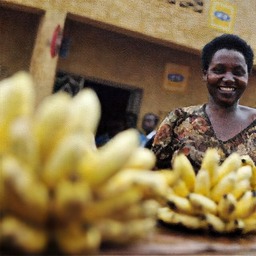} &
\includegraphics[width=\aqw]{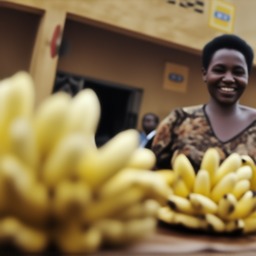} &
\includegraphics[width=\aqw]{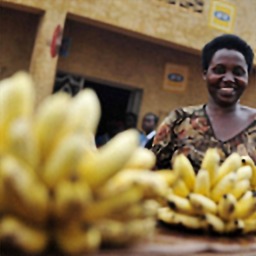} &
\includegraphics[width=\aqw]{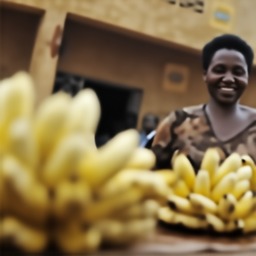} &
\includegraphics[width=\aqw]{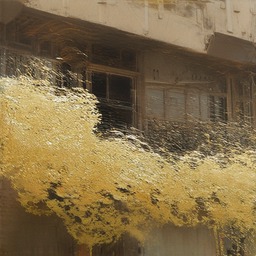} \\
\end{tabular}
\caption{SR $\times4$ with $\sigma_{\rvy}=0.01$, COCO, SD3.5-medium prior. Below each row, the region boxed on $\rvx$ in every column. Panels are from the runs of Table~\ref{tab:coco-models-arc-full}.}
\label{fig:appendix-qual-coco-sd35-sr4}
\end{figure}

\begin{figure}[p]
\centering
\setlength{\aqw}{\dimexpr(\linewidth-9pt)/10\relax}
\setlength{\tabcolsep}{0pt}
\renewcommand{\arraystretch}{0}
\begin{tabular}{@{}c@{\hspace{1pt}}c@{\hspace{1pt}}c@{\hspace{1pt}}c@{\hspace{1pt}}c@{\hspace{1pt}}c@{\hspace{1pt}}c@{\hspace{1pt}}c@{\hspace{1pt}}c@{\hspace{1pt}}c@{}}
\scriptsize $\rvx$ & \scriptsize $\rvy$ & \scriptsize \paseo & \scriptsize \textsc{ReSample} & \scriptsize \textsc{DAPS} & \scriptsize \textsc{RedDiff} & \scriptsize \textsc{FlowDPS} & \scriptsize \textsc{FlowChef} & \scriptsize \textsc{PnP-Flow} & \scriptsize \textsc{PSLD} \\[3pt]
\aqbox{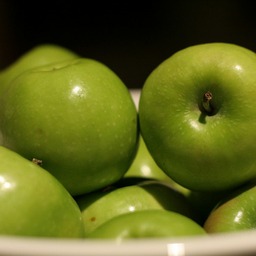}{256}{352}{512}{480} &
\includegraphics[width=\aqw]{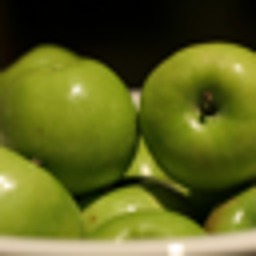} &
\includegraphics[width=\aqw]{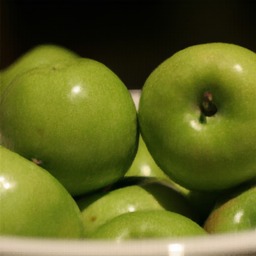} &
\includegraphics[width=\aqw]{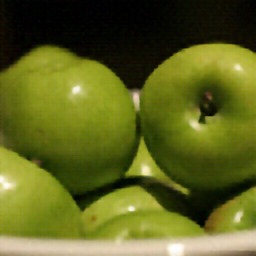} &
\includegraphics[width=\aqw]{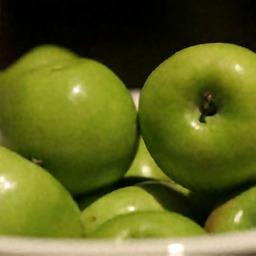} &
\includegraphics[width=\aqw]{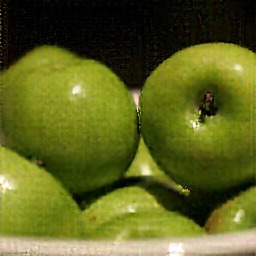} &
\includegraphics[width=\aqw]{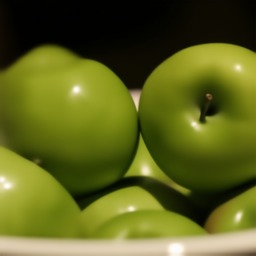} &
\includegraphics[width=\aqw]{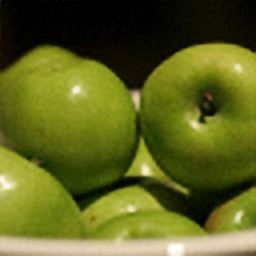} &
\includegraphics[width=\aqw]{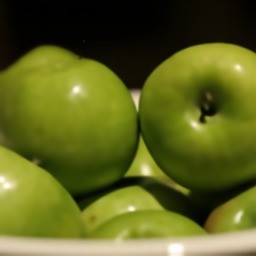} &
\includegraphics[width=\aqw]{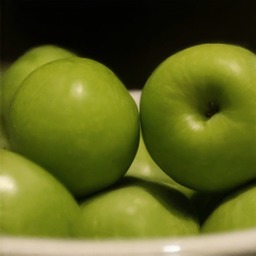} \\
\end{tabular}
\caption{SR $\times8$ with $\sigma_{\rvy}=0.01$, COCO, SD3.5-medium prior. Below each row, the region boxed on $\rvx$ in every column. Panels are from the runs of Table~\ref{tab:coco-models-arc-full}.}
\label{fig:appendix-qual-coco-sd35-sr8}
\end{figure}

\begin{figure}[p]
\centering
\setlength{\aqw}{\dimexpr(\linewidth-9pt)/10\relax}
\setlength{\tabcolsep}{0pt}
\renewcommand{\arraystretch}{0}
\begin{tabular}{@{}c@{\hspace{1pt}}c@{\hspace{1pt}}c@{\hspace{1pt}}c@{\hspace{1pt}}c@{\hspace{1pt}}c@{\hspace{1pt}}c@{\hspace{1pt}}c@{\hspace{1pt}}c@{\hspace{1pt}}c@{}}
\scriptsize $\rvx$ & \scriptsize $\rvy$ & \scriptsize \paseo & \scriptsize \textsc{ReSample} & \scriptsize \textsc{DAPS} & \scriptsize \textsc{RedDiff} & \scriptsize \textsc{FlowDPS} & \scriptsize \textsc{FlowChef} & \scriptsize \textsc{PnP-Flow} & \scriptsize \textsc{PSLD} \\[3pt]
\aqbox{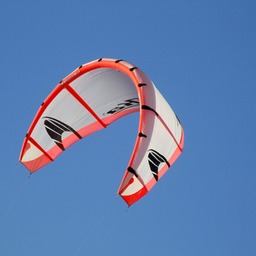}{64}{192}{320}{320} &
\includegraphics[width=\aqw]{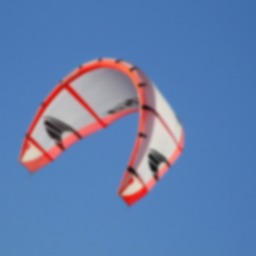} &
\includegraphics[width=\aqw]{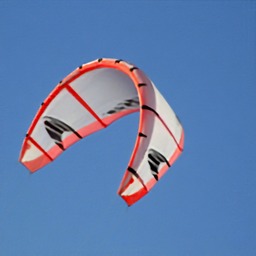} &
\includegraphics[width=\aqw]{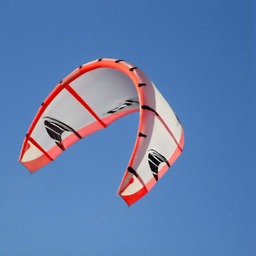} &
\includegraphics[width=\aqw]{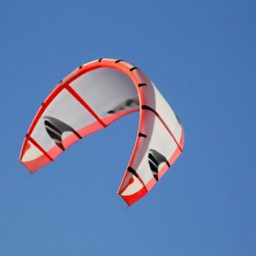} &
\includegraphics[width=\aqw]{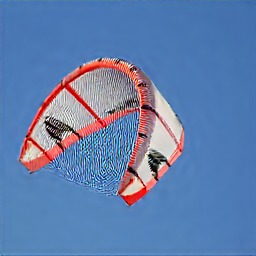} &
\includegraphics[width=\aqw]{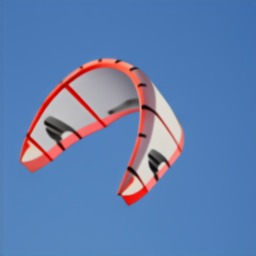} &
\includegraphics[width=\aqw]{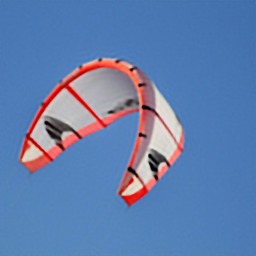} &
\includegraphics[width=\aqw]{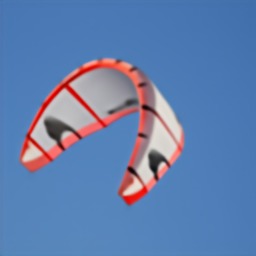} &
\includegraphics[width=\aqw]{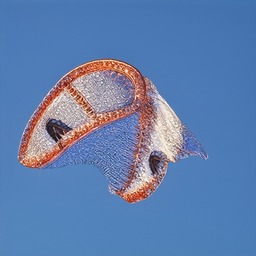} \\
\end{tabular}
\caption{Gaussian blur with $\sigma_{\rvy}=0.01$, COCO, SD3.5-medium prior. Below each row, the region boxed on $\rvx$ in every column. Panels are from the runs of Table~\ref{tab:coco-models-arc-full}.}
\label{fig:appendix-qual-coco-sd35-blur}
\end{figure}

\begin{figure}[p]
\centering
\setlength{\aqw}{\dimexpr(\linewidth-9pt)/10\relax}
\setlength{\tabcolsep}{0pt}
\renewcommand{\arraystretch}{0}
\begin{tabular}{@{}c@{\hspace{1pt}}c@{\hspace{1pt}}c@{\hspace{1pt}}c@{\hspace{1pt}}c@{\hspace{1pt}}c@{\hspace{1pt}}c@{\hspace{1pt}}c@{\hspace{1pt}}c@{\hspace{1pt}}c@{}}
\scriptsize $\rvx$ & \scriptsize $\rvy$ & \scriptsize \paseo & \scriptsize \textsc{ReSample} & \scriptsize \textsc{DAPS} & \scriptsize \textsc{RedDiff} & \scriptsize \textsc{FlowDPS} & \scriptsize \textsc{FlowChef} & \scriptsize \textsc{PnP-Flow} & \scriptsize \textsc{PSLD} \\[3pt]
\aqbox{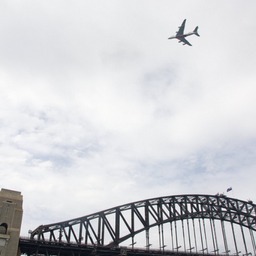}{128}{192}{384}{320} &
\includegraphics[width=\aqw]{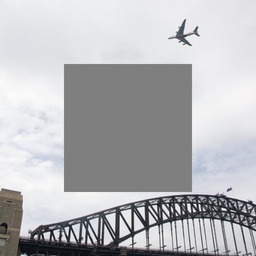} &
\includegraphics[width=\aqw]{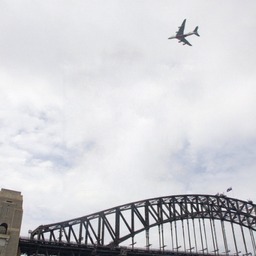} &
\includegraphics[width=\aqw]{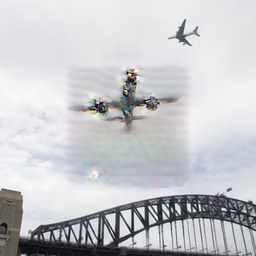} &
\includegraphics[width=\aqw]{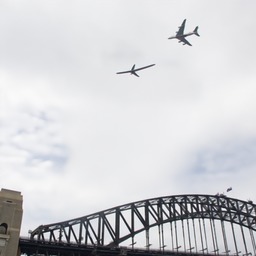} &
\includegraphics[width=\aqw]{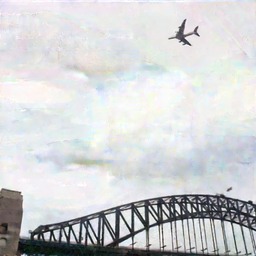} &
\includegraphics[width=\aqw]{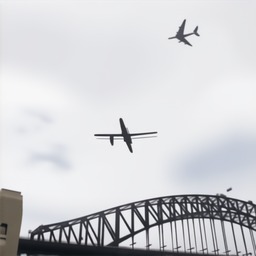} &
\includegraphics[width=\aqw]{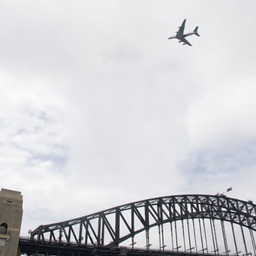} &
\includegraphics[width=\aqw]{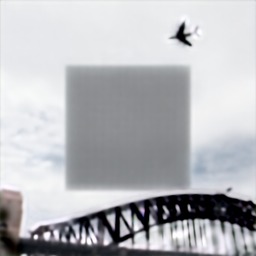} &
\includegraphics[width=\aqw]{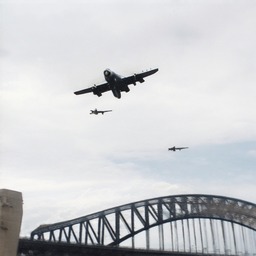} \\
\end{tabular}
\caption{Box inpainting with $\sigma_{\rvy}=0.01$, COCO, SD3.5-medium prior. Below each row, the region boxed on $\rvx$ in every column. Panels are from the runs of Table~\ref{tab:coco-models-arc-full}.}
\label{fig:appendix-qual-coco-sd35-inpaint}
\end{figure}

\begin{figure}[p]
\centering
\setlength{\aqw}{\dimexpr(\linewidth-8pt)/9\relax}
\setlength{\tabcolsep}{0pt}
\renewcommand{\arraystretch}{0}
\begin{tabular}{@{}c@{\hspace{1pt}}c@{\hspace{1pt}}c@{\hspace{1pt}}c@{\hspace{1pt}}c@{\hspace{1pt}}c@{\hspace{1pt}}c@{\hspace{1pt}}c@{\hspace{1pt}}c@{}}
\scriptsize $\rvx$ & \scriptsize $\rvy$ & \scriptsize \paseo & \scriptsize \textsc{ReSample} & \scriptsize \textsc{DAPS} & \scriptsize \textsc{RedDiff} & \scriptsize \textsc{FlowDPS} & \scriptsize \textsc{FlowChef} & \scriptsize \textsc{PnP-Flow} \\[3pt]
\aqbox{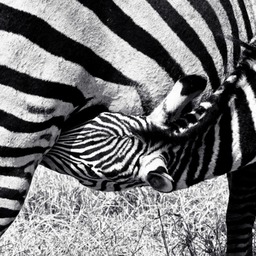}{192}{384}{448}{512} &
\includegraphics[width=\aqw]{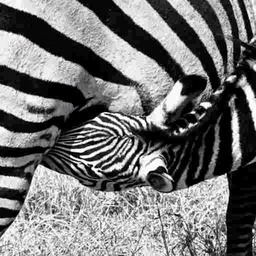} &
\includegraphics[width=\aqw]{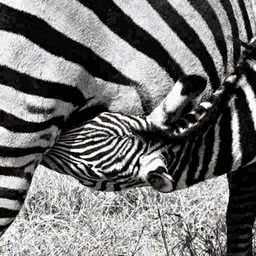} &
\includegraphics[width=\aqw]{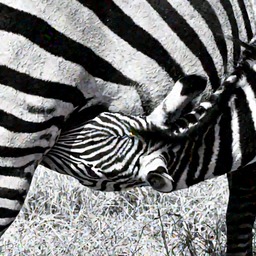} &
\includegraphics[width=\aqw]{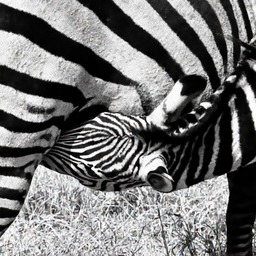} &
\includegraphics[width=\aqw]{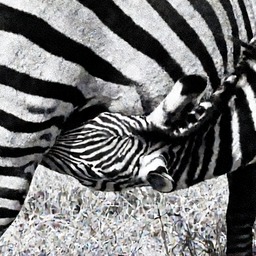} &
\includegraphics[width=\aqw]{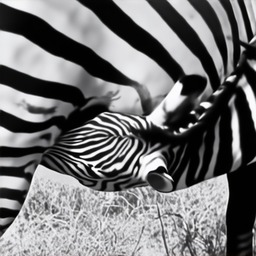} &
\includegraphics[width=\aqw]{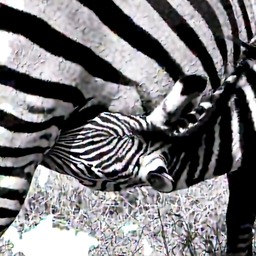} &
\includegraphics[width=\aqw]{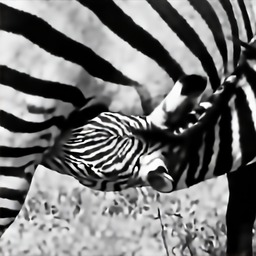} \\
\end{tabular}
\caption{JPEG ($q{=}10$) with $\sigma_{\rvy}=0.01$, COCO, SD3.5-medium prior. Below each row, the region boxed on $\rvx$ in every column. Panels are from the runs of Table~\ref{tab:coco-models-arc-full}.}
\label{fig:appendix-qual-coco-sd35-jpeg}
\end{figure}

\end{document}